\documentclass[journal]{IEEEtran}

\pdfoutput=1
\usepackage[pdftex]{graphicx}
\DeclareGraphicsExtensions{.pdf,.jpeg,.png,.bmp,.ai}

\ifCLASSOPTIONcompsoc
  \usepackage[caption=false,font=normalsize,labelfont=sf,textfont=sf]{subfig}
\else
  \usepackage[caption=false,font=footnotesize]{subfig}
\fi
\newcommand{\etal}{\textit{et al}. }
\newcommand{\ie}{\textit{i}.\textit{e}. }
\newcommand{\eg}{\textit{e}.\textit{g}. }

\usepackage{multirow}

\begin{document}
%
\title{Estimation and Restoration of\\Compositional Degradation Using\\Convolutional Neural Networks}
%
%
%

\author{Kazutaka~Uchida,
        Masayuki~Tanaka,~\IEEEmembership{Member,~IEEE,}
        and~Masatoshi~Okutomi,~\IEEEmembership{Member,~IEEE,}
\thanks{K. Uchida, M.Tanaka, and M.Okutomi are with Tokyo Institute of Technology, Tokyo Japan.}
\thanks{K. Uchida is also with Kadinche Corporation, Tokyo, Japan.}
\thanks{M. Tanaka is also with National Institute of Advanced Industrial Science and Technology, Tokyo, Japan.}}

%
%

\ifCLASSOPTIONpeerreview
\markboth{Journal of \LaTeX\ Class Files,~Vol.~14, No.~8, August~2015}%
{Shell \MakeLowercase{\textit{et al.}}: Bare Demo of IEEEtran.cls for IEEE Journals}
\fi

%



\maketitle


\begin{abstract}
Image restoration from a single image degradation type, such as blurring, hazing, random noise, and compression has been investigated for decades.
However, image degradations in practice are often a mixture of several types of degradation.
Such compositional degradations complicate restoration because they require the differentiation of different degradation types and levels.
In this paper, we propose a convolutional neural network (CNN) model for estimating the degradation properties of a given degraded image.
Furthermore, we introduce an image restoration CNN model that adopts the estimated degradation properties as its input.
Experimental results show that the proposed degradation estimation model can successfully infer the degradation properties of compositionally degraded images.
The proposed restoration model can restore degraded images by exploiting the estimated degradation properties and can achieve both blind and nonblind image restorations.
\end{abstract}


%
\IEEEpeerreviewmaketitle

\section{Introduction}


\IEEEPARstart{D}{egradation} estimation is an important process for image restoration.
The precise identification of the degradation type and level facilitates the subsequent restoration process because this approach allows the restoration problem to be considered a nonblind problem rather than a blind problem.


Degradation estimation methods for a specific degradation type such as Gaussian noise \cite{chen2015efficient, liu2012noise,liu2013estimation,liu2013single}, blurring \cite{shi2015just,park2017unified} and JPEG compression \cite{li2015statistical,fan2003identification,fridrich2008detection,neelamani2006jpeg,doan2016quality} have been proposed for decades.
For example, Liu \etal \cite{liu2012noise,liu2013estimation,liu2013single} proposed a model-based approach for the estimation of Gaussian noise level on a given image.
Shi \etal \cite{shi2015just} proposed a blurring detection and level estimation method that outputs a blur map of the given image.
These methods perform well for a single type of degradation.


Along with degradation estimation, image restoration from degraded images has been a hot topic during the past several decades \cite{buades2005review}.
Many restoration methods from degradation, such as Gaussian noise \cite{bosco2005fast, liu2008automatic}, blurring \cite{zhao2016single,xu2010two}, image compression \cite{xiong1997deblocking}, and hazing\cite{fattal2008single} have been proposed.
These methods include both blind and nonblind restoration methods.
Nonblind restoration methods estimate the input degradation parameters by using one of the degradation estimation methods, whereas most blind restoration algorithms infer the degradation parameters internally and implicitly or explicitly for the restoration.


Several image restoration methods that utilize convolutional neural networks (CNNs) have recently been introduced with the rise of deep neural networks \cite{uchida2018coupled,tanaka2018weighted}.
In particular, learning-based methods are useful because of the simplicity of the training process.
These methods map a degraded image directly on the restored image, thus implicitly estimating the degradation parameters.
On the contrary, model-based restoration methods remain popular because of the mathematical clarity in describing degradation and restoration problems.
Some of these models require degradation parameters on an input image for nonblind restoration to initialize the internal restoration model.




In practice, images are degraded during imaging process.
The light wave that becomes diffused in the air causes hazing degradation, and blurring occurs when the image projected on an image sensor is unfocussed.
Random noises are added by the image sensor.
Furthermore, quantization and image compression cause image degradation.

Image degradation effects may form a cascade in captured images, and this phenomenon is called compositional degradation.
For example, blurring is usually followed by random noise and image compression.


Restoration from compositional degradation is challenging problem.
Zhang \etal \cite{zhang2017beyond} proposed a CNN-based image restoration method called DnCNN, which applies denoising, deblocking, or superresolution to a degraded input image depending on the  degradation type of that image.
Although the model can handle different types of degradation, it can apply only one restoration operation at a time. Therefore, the model is not directly applicable to images with compositional degradation.
When we downsample a degraded image with additive white Gaussian noise (AWGN) by using DnCNN, the restored image is a super-resolved image because the model recognizes the input as a low resolution image.
If DnCNN is applied again to this super-resolved image to reduce noise, the restored image will not be denoised because the super-resolved AWGN has different properties from that of the AWGN samples used in the training phase.

A solution to this problem could be the multiple application of a restoration method to restore an image in a step-by-step manner.
This type of step-by-step restoration strategy does not work normally because the initial restoration produces an image with a degradation property that is different from that expected by the second restoration algorithm, as illustrated in Section \ref{sec:compositional}.




In this paper, we introduce a method for the estimation of degradation types and levels for compositional degradation.
We also propose an image restoration network using the estimated degradation type and level to perform blind resotration for compositional degradation.
This paper is an extended version of our previous work published in \cite{uchida2018nonblind}.
We extend the nonblind restoration in \cite{uchida2018nonblind} to the blind restoration by adding a degradation estimation network for compositional degradation.
 We first describe compositional degradation in Section \ref{sec:compositional} and then propose our method in Section \ref{sec:proposed}. Experiments on degradation estimation and restoration are proposed in Section \ref{sec:experiments}, and we provide our conclusions in Section \ref{sec:conclusion}.



\section{Degradation Estimation for Compositional Degradation}

\label{sec:compositional}

\subsection{Compositional Degradation}

\begin{figure}[!t]
\centering
\includegraphics[width=0.9\linewidth]{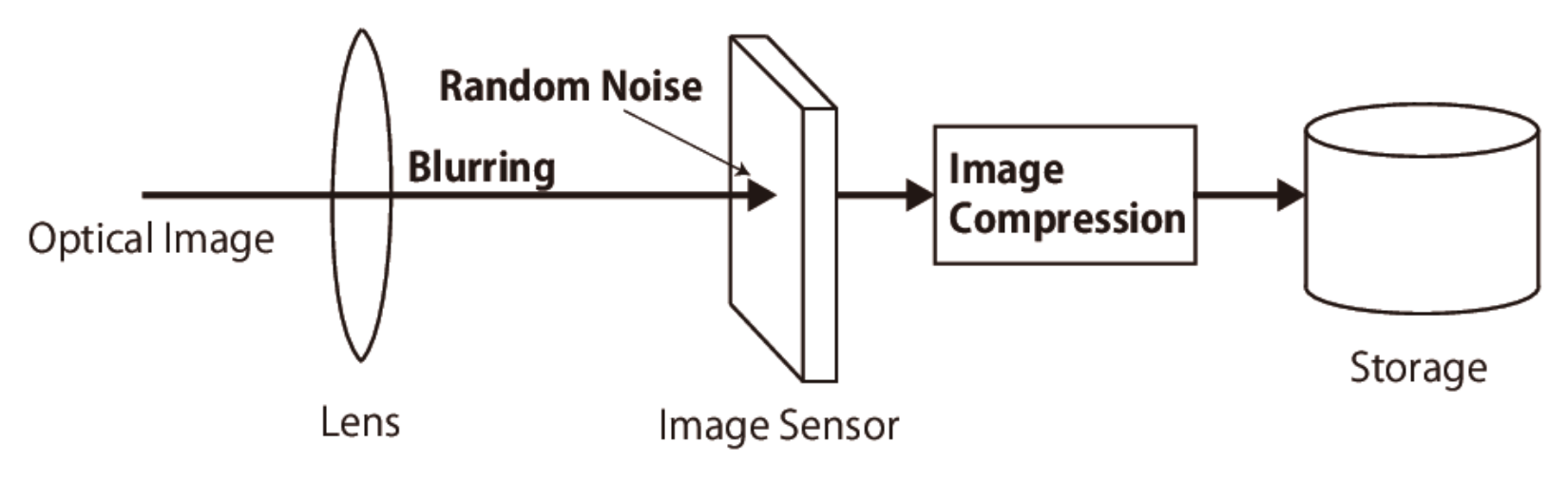}
\caption{Compositional degradation model}
\label{fig:composite_degradation_model}
\end{figure}

Image capturing involves a chain of degradations.
Figure \ref{fig:composite_degradation_model} shows a typical model of degradations during the image capturing process.
First, the optical image is blurred owing to the low focus of the lens and the finite pixel size of the image sensor.
Thereafter, the image is contaminated by random noise with the sensor.
The captured image is quantized into a certain resolution and compressed by an image compression algorithm for storage.
The stored image is then resized to simplify usage, \eg attachment to an email, which is another example of degradation.

In this way, most captured images are degraded with different types of degradation.
We call these consecutive degradations of different types as compositional degradation.

Although compositional degradation is common, most image restoration algorithms focus on a single degradation type and not on compositional degradation as an ensemble.
Some studies concentrated on compositional degradation to achieve robustness against perturbation, such as random noise \cite{fan2003identification,fridrich2008detection,neelamani2006jpeg,xu2010two}.
However, these previous studies treated compositional degradation as an end-to-end system.

\subsection{Degradation Estimation for Compositional Degradation}

Degradation level estimation for a single type is achieved well for single type of degraded images with targeted degradation.
However, current estimation methods perform poorly on compositional degradation images.

Liu \etal \cite{liu2013single} proposed a noise level estimation method for AWGN. As shown in Section \ref{sec:experiments}, noise level estimation is accurate for images that are only degraded with AWGN in double-precise intensity resolution.
On the contrary, degradation estimation shows low accuracy when applied to degraded images with AWGN followed by JPEG compression even with 100\% quality factor (AWGN+JPEG).

Another example is blurring level estimation.
As discussed in Section \ref{sec:experiments},
the estimation for the normal blurred image is accurate. However, many regions are wrongly marked as ``forcused'' for the degraded image with AWGN (blur+AWGN).

These observations imply that existing degradation estimation methods are powerful when applied to single type of degraded images with a single-target degradation type but are not robust for compositional degradations, \ie, AWGN+JPEG compression for noise detection and blur+AWGN for blur-level estimation.

As we stated previously, the images captured by a sensor are degraded with consecutive degradations of different types. Therefore, practical degradation estimation should be designed to simultaneously detect multiple degradations and their levels.

\subsection{Restoration from Compositional Degradation}

\begin{figure*}[t!]
  \captionsetup{farskip=0pt}
    \begin{center}
  \subfloat[Original]{
    \includegraphics[width=0.19\linewidth]{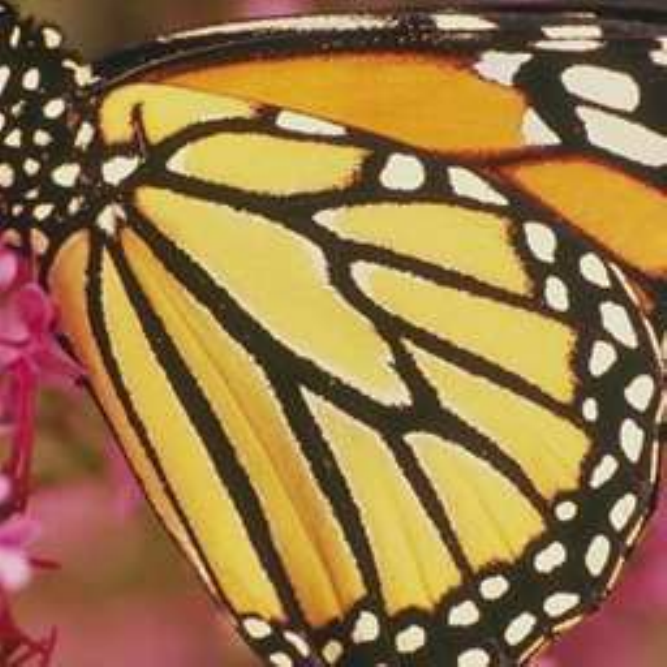}
    \label{step_by_step_restoration_a}\hfill
  }
  \subfloat[Degraded with blur and AWGN]{
     \includegraphics[width=0.19\linewidth]{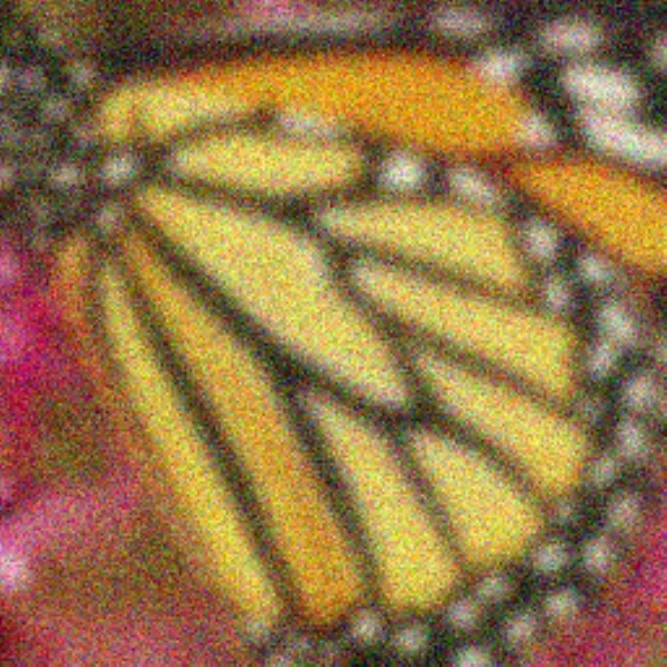}
    \label{step_by_step_restoration_b}
  }
  \subfloat[Denoised by \cite{dabov2007image}]{%
     \includegraphics[width=0.19\linewidth]{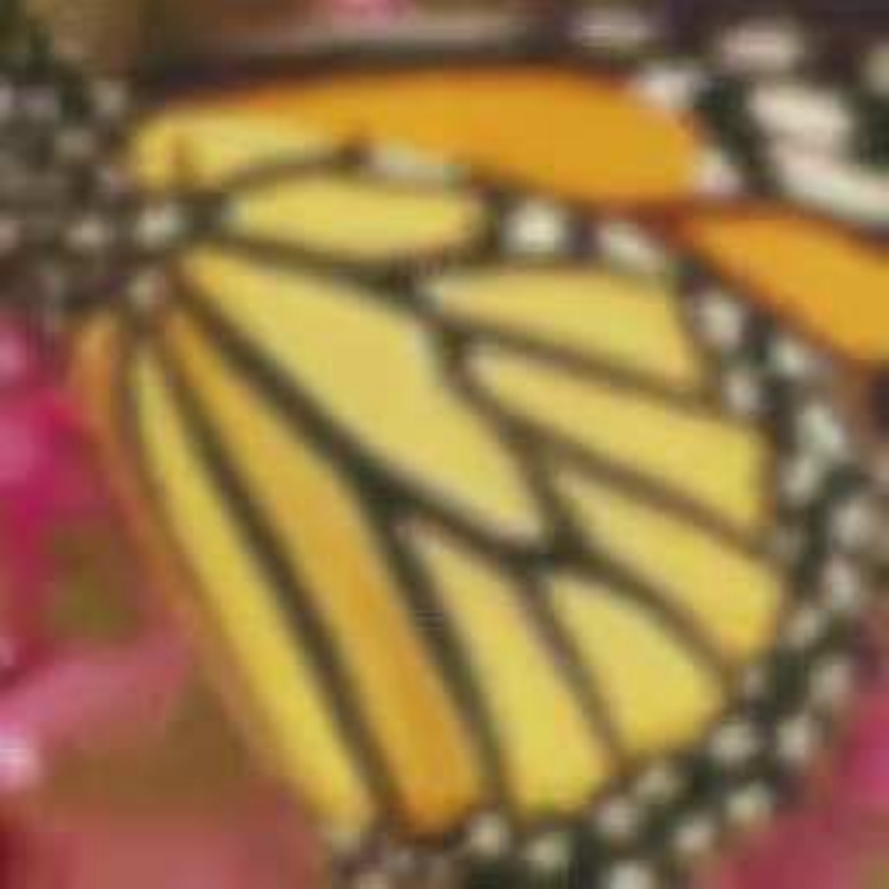}
    \label{step_by_step_restoration_c}\hfill
  }
  \subfloat[Deblurred by \cite{xu2010two}]{%
     \includegraphics[width=0.19\linewidth]{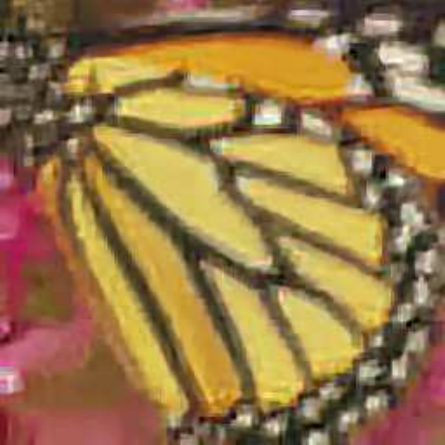}
    \label{step_by_step_restoration_d}
  }
  \subfloat[Directly deblurred by \cite{xu2010two}]{%
     \includegraphics[width=0.19\linewidth]{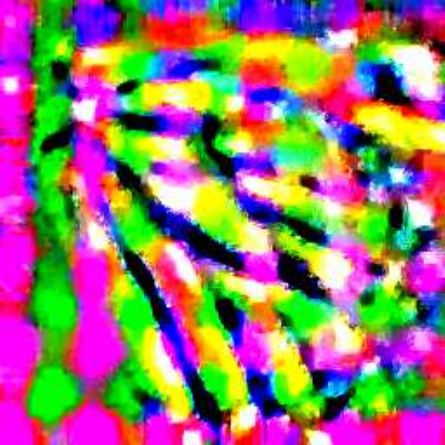}
    \label{step_by_step_restoration_e}
  }
  \end{center}
  \caption{Step-by-step restoration from compositional degradation}
  \label{fig:step_by_step_restoration}
\end{figure*}

Image restoration from compositional degradation is a more complex problem than that from single-type degradation.
Figure \ref{fig:step_by_step_restoration} shows an example restoration of a blur+AWGN degraded image (Fig. \ref{step_by_step_restoration_b}) that consecutively applies denoising \cite{dabov2007image} and a deblurring algorithm \cite{xu2010two} to restore degraded images in the inverse order. Denoising successfully reduces random noise (Fig. \ref{step_by_step_restoration_c}), but the subsequent deblurring fails to restore the image (Fig. \ref{step_by_step_restoration_d}) because the denoising is not perfect, \ie, the purely blurred image is not restored from the noised blurred image. Furthermore, there is a gap between the purely blurred image and the denoised blurred image. Most deblurring algorithms assume that an input image is purely blurred; therefore, the algorithms do not work well for such restored images.
Directly deblurring the blur+AWGN image by \cite{xu2010two} totally fails (Fig. \ref{step_by_step_restoration_e}) because the algorithm does not assume such random noise in an input image.

Restoration from compositional degradation is difficult in such a step-by-step restoration strategy, which applies a restoration method for each degradation type consecutively.
An end-to-end restoration algorithm, which directly converts an input image with compositional degradation to a restored image, can be used to solve this problem.

A learning-based end-to-end restoration algorithm is suitable for this problem, however, there are two drawbacks.
First, internally detected degradation types and levels are not observable. Therefore, a user cannot identify how the algorithm inferred the input degradation, \ie, the selection of restoration strategy for the input image.
Second, there is no control point to adjust the restoration strategy and strength because only the degraded image is the input and no measure is provided to change the output.
To solve the drawbacks, a new network structure is proposed in the next section.

\section{Proposed Method}

\label{sec:proposed}

To realize degradation estimation and restoration from compositional degradation, we propose a CNN-based network model that comprises two subnetworks: a degradation estimation network and a restoration network.

\subsection{Overview}

\begin{figure*}[!t]
\centering
\includegraphics[width=1.0\linewidth]{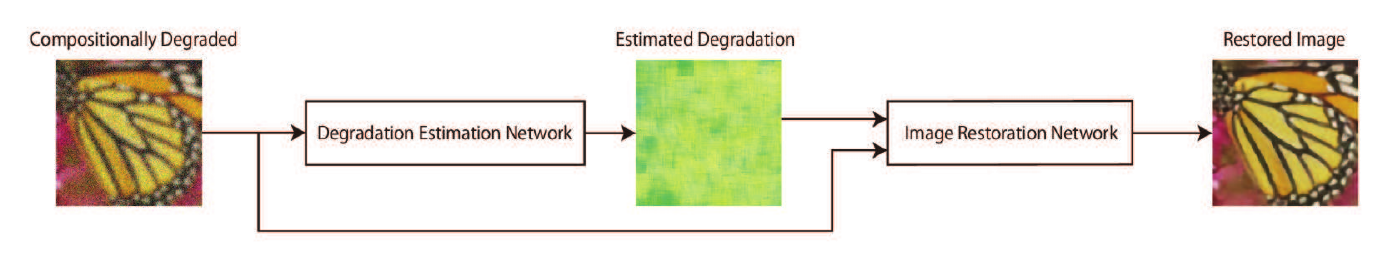}
\caption{Flow of compositional degradation estimation and restoration}
\label{fig:flow_of_compositional_degradation}
\end{figure*}

Figure \ref{fig:flow_of_compositional_degradation} shows an overview of the proposed model.

The degradation estimation network infers the degradation attributes, \ie, the degradation types and their strengths (or levels), from an input image with compositional degradation.
The network assumes that there are $N$ degradation types and the input image is degraded in a certain order.

The restoration network predicts a clean image of the input image with the inferred degradation levels.
Therefore, the restoration network itself is considered a nonblind restoration processor.

There are two reasons why the network is divided into two subnetworks.
First, restoration strategies can be controlled by the degradation parameter input to the restoration network unlike DnCNN \cite{zhang2017beyond}, which directly maps a degraded image to the restored image.
One of the advantages of this restoration is that it can be performed in both nonblind and blind modes, thus allowing users to interactively adjust the restoration strength by changing the degradation parameter in the restoration network.
Second, the estimated degradation types and levels are visible from the output of the degradation estimation network.
This is an advantage over other networks, such as DnCNN, which implicitly infers degradation and wherein the estimated degradation is not visible. Visibility helps users interpret the restored image because restoration is often unsuccessful owing to failures in degradation estimation.

\subsection{Degradation Estimation Network}

\begin{figure}[!t]
\centering
\subfloat[Degradation estimation network] {
  \includegraphics[width=0.9\linewidth]{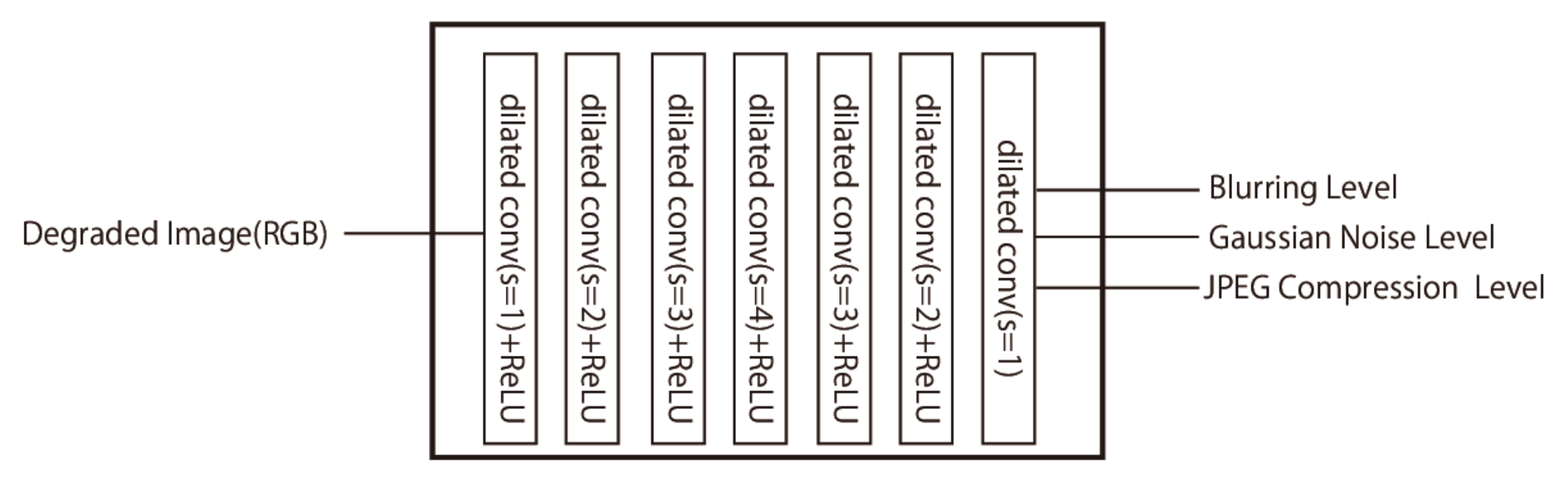}
  \label{fig:network_estimation}
}
\caption{Network structures}
\end{figure}

Figure \ref{fig:network_estimation} shows the structure of the degradation estimation network that consists of seven dilated convolutional layers \cite{yu2015multi} with dilated rates of one, two, three, four, three, two, and one. The number of input channels is three for RGB-degraded images.
Zhang \etal \cite{zhang2017learning} introduced a similar network structure.
The outputs of all layers have the same size as the input image. The filter size is $3 \times 3$ and the number of output channels is 64 for the first 6 layers and $N$ for the last layer, where $N$ is the number of degradation level maps to be inferred. The activation function is ReLU \cite{nair2010rectified} for all convolutional layers except for the last layer.
In the current study, $N$ is set to three to correspond to the blur, AWGN, and JPEG degradation types.

The values in each output channel are the estimated degradation strength of the corresponding degradation type at each position in the image.
The degradation strength value is normalized between zero and one, where a value of zero means not degraded (\ie, clean), and a value of one indicates fully degraded.

The network is trained so that the degraded parameters, \ie, the types and levels, can be correctly inferred from a compositionally degraded image.
During training, input images are generated with random degradation types and parameters from clean image patches and the corresponding output feature maps are generated with the true degradation parameters.

\subsection{Image Restoration Network}

The image restoration network is a nonblind restoration processor that follows the concept proposed in the previous work \cite{uchida2018nonblind}.
One of the advantages of a nonblind network is robustness against perturbation on a degradation model.
Furthermore, the controllability of the restoration strategy enables stable image restoration.

The network structure proposed here is based on the network structure of \cite{uchida2018nonblind}, but our degradation estimation network has seven dilated convolutional layers.
The input has $(3+N)$ channels: three for degraded images (RGB) and $N$ for the degradation attribute channels. The output has three channels for the restored image (RGB) with the same size as the input.
The input image is added to the output of the final convolution layer with the skip connection.
The values in the input channels of the degradation parameters are assumed to be normalized within $[0,1]$ in accordance with the output of the degradation estimation network.

During training, the network is trained so that it can infer the clean image from the degraded input image and the degradation parameters. The training dataset is created in the same manner as the degradation estimation network. Randomly degraded images and the true degradation parameters are the input, and the original images are the output.
Optimization is executed to minimize the mean-squared loss function of the restored images and the true images by using SDG or Adam algorithm \cite{kingma2014adam}.

\section{Experiments}

\label{sec:experiments}

The proposed model is trained, and the performance of the degradation estimation and restoration is evaluated.

\subsection{Training}

To train the models, image datasets containing 291 images are used: 91 images from \cite{yang2010image} are used, and 200 images from the Berkeley segmentation dataset \cite{MartinFTM01} are used in a similar manner as that in \cite{Kim_2016_VDSR}.

Images from the training datasets are randomly cropped into patches with a size of $60 \times 60$. The patches are then increased eight times with data augmentation by rotating and mirroring, thus resulting in 561 k patches in total.

Compositional degradation consisting of subsequent blurring, AWGN, and JPEG compression with random degradation levels is applied to each patch.
Blurring is applied by filtering with a Gaussian kernel with a standard deviation of $\sigma$, where $\sigma$ ranges from 0 to 3.5.
AWGN is generated with noise level $\lambda$ ranging from 0 to 55.
JPEG compression is applied with quality factor $q$ ranging from 5\% to 100\%.
Ten percent of the patches are not JPEG compressed.
The random degradations are applied on the fly to generate training mini-batches that contain 128 patches each.

The values for degradation parameter channels are calculated as follows:
\begin{eqnarray}
v_b(x,y) &=& V_b(\sigma) \\
v_n(x,y) &=& V_n(\lambda) \\
v_c(x,y) &=& V_c(q),
\end{eqnarray}
where $v_b(x,y)$, $v_n(x,y)$, and $v_c(x,y)$ are the degradation parameters at position $(x, y)$ for blurring, AWGN, and JPEG compression channel, respectively.
$V_b(\cdot)$, $V_n(\cdot)$, and $V_c(\cdot)$ are functions so that
\begin{eqnarray}
V_b(\sigma) &=& \sigma / 3.5\\
V_n(\lambda) &=& \lambda / 55\\
V_c(q) &=&
  \left\{
    \begin{array}{ll}
      0.9 (100 - q) / 100 + 0.1 & ({\rm JPEG comp.})\\
      0 & ({\rm uncompressed}).
    \end{array}
  \right.
\end{eqnarray}.

To train the degradation estimation model, the loss is calculated as the mean-squared error between the model output and the true degradation properties of the degraded input patches.
The optimization of the model parameters is performed in 80 epochs by using Adam algorithm \cite{kingma2014adam}.

To train the restoration model, degraded patches (on the first to the third channels) and their true degradation properties (on the fourth to the sixth channels) are entered into the model. The loss is calculated as the mean-squared error between the model output and the corresponding clean images to the input.
The optimization is performed with Adam algorithm in 80 epochs.
The implementation is written with the Keras framework \cite{chollet2015keras} and performed on a PC with Nvidia TITAN X GPUs.

\subsection{Performance Evaluation}

\subsubsection{Degradation Estimation}

\begin{figure*}[t!]
    \footnotesize
    \begin{center}
\hspace*{-10mm}
      \begin{tabular}{|l|l|ccc|ccc|ccc|} \hline
      &  & \multicolumn{3}{|c|}{$\sigma=0$} & \multicolumn{3}{|c|}{$\sigma=1.5$} & \multicolumn{3}{|c|}{$\sigma=3.0$} \\ \hline
      & & $q=100$ & $q=50$ & $q=10$ & $q=100$ & $q=50$ & $q=10$ & $q=100$ & $q=50$ & $q=10$ \\ \hline
\multirow{3}{*}{$\lambda=0$} &
Degraded  &
       \includegraphics[width=0.08\linewidth]{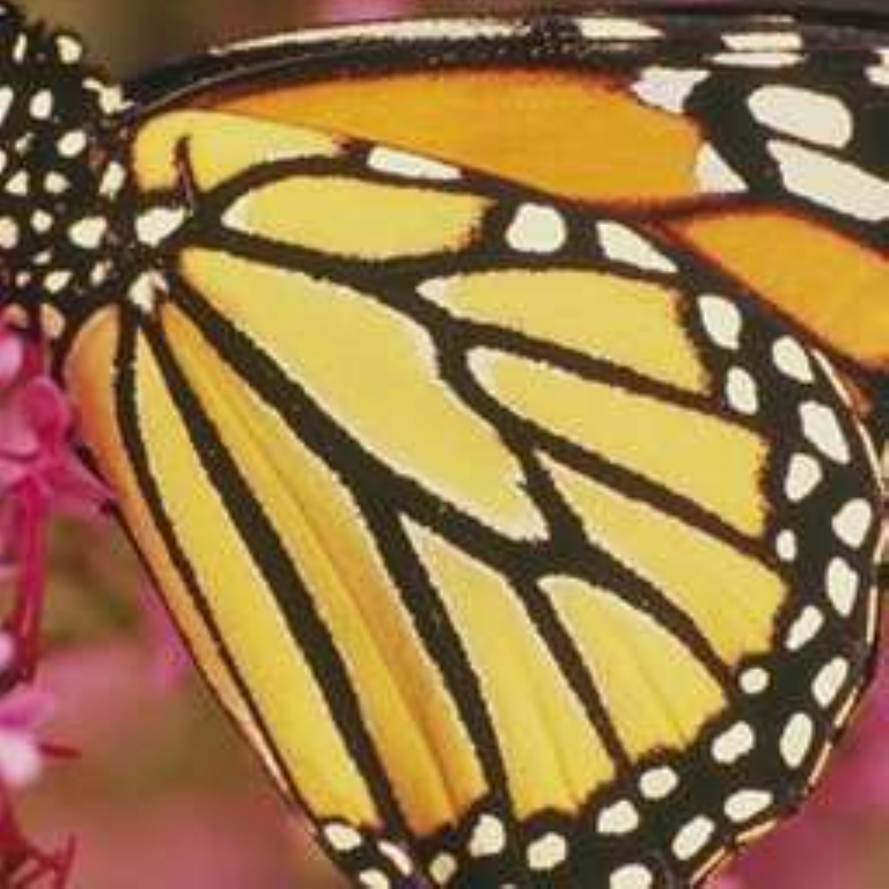} &
       \includegraphics[width=0.08\linewidth]{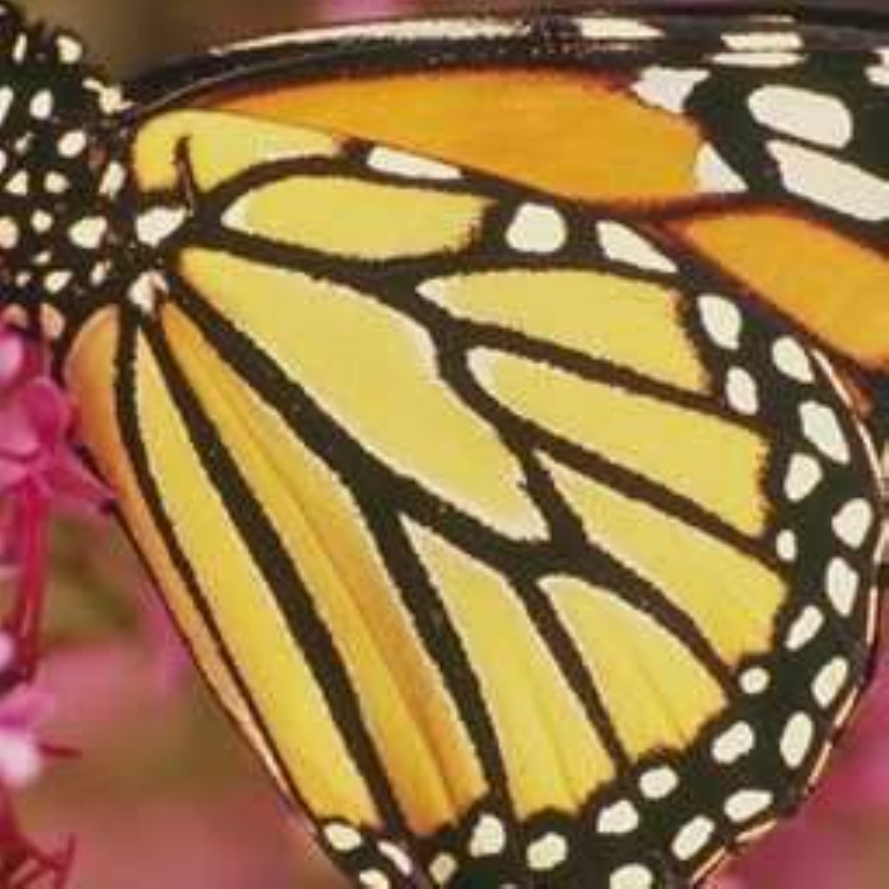} &
       \includegraphics[width=0.08\linewidth]{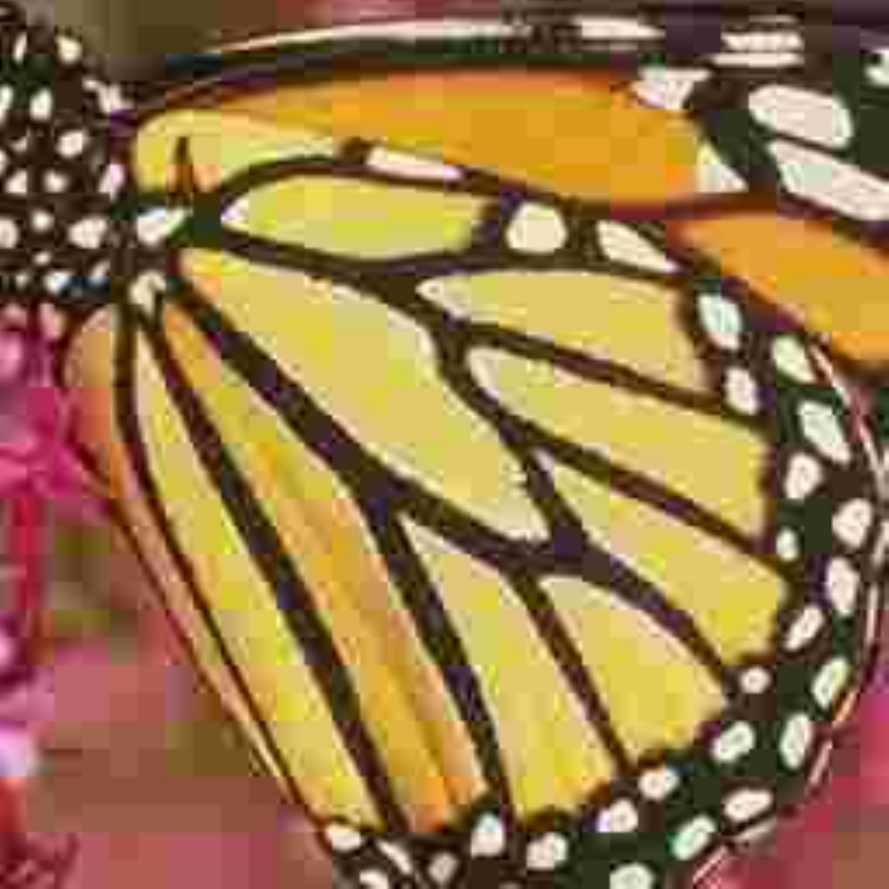} &
       \includegraphics[width=0.08\linewidth]{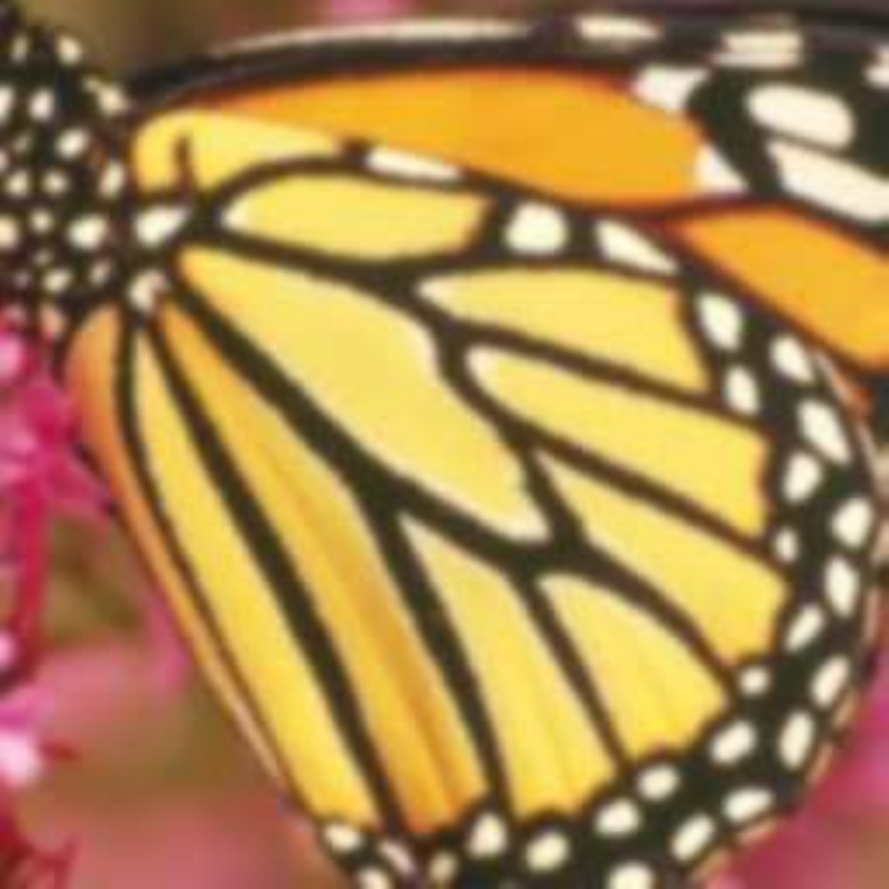} &
       \includegraphics[width=0.08\linewidth]{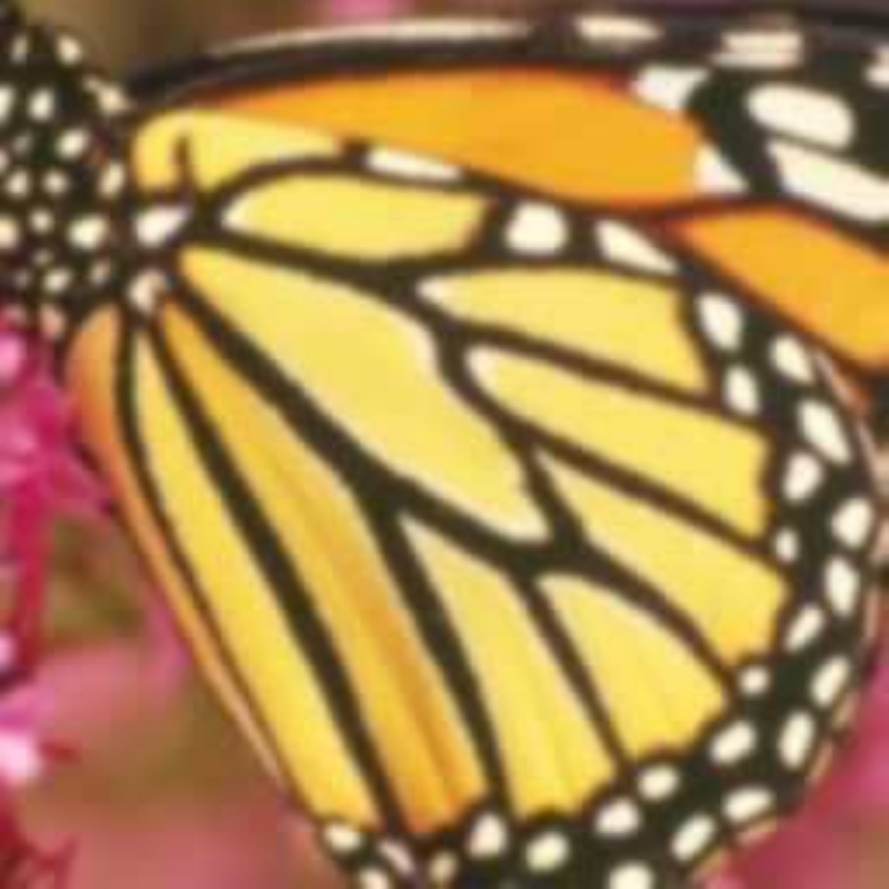} &
       \includegraphics[width=0.08\linewidth]{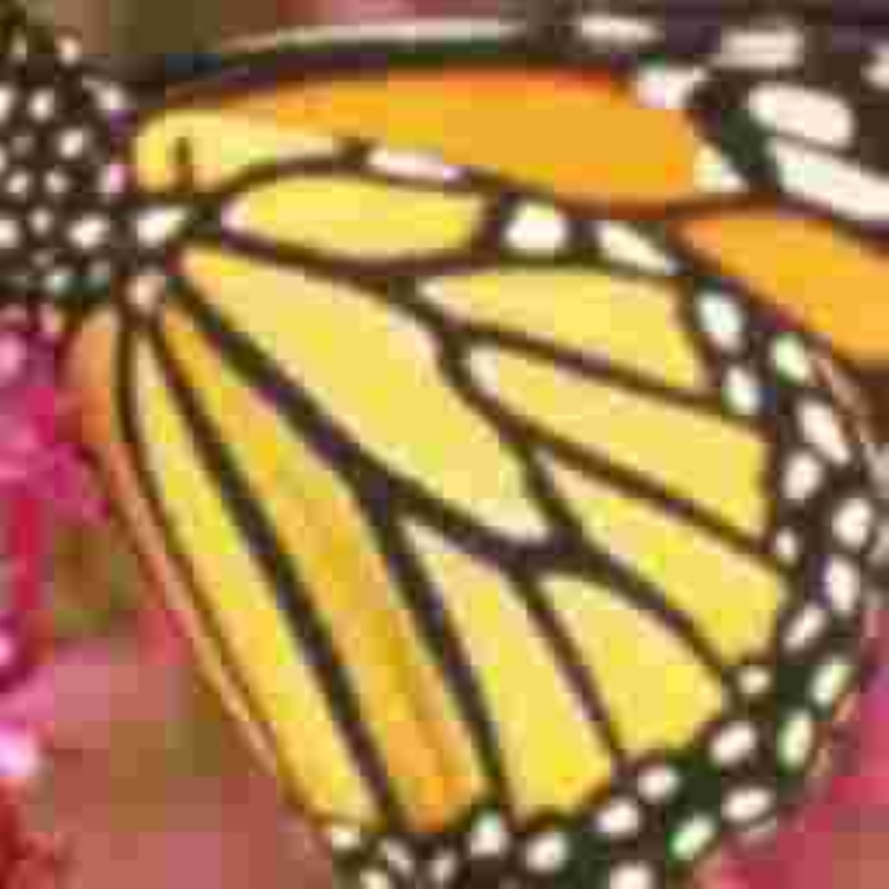} &
       \includegraphics[width=0.08\linewidth]{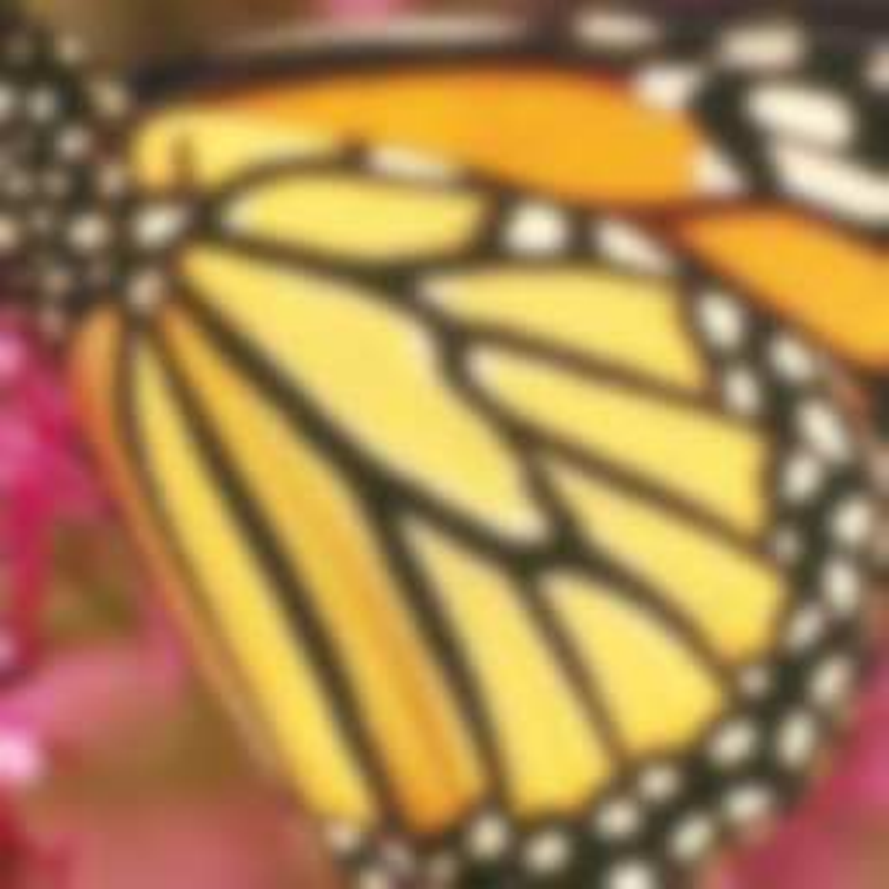} &
       \includegraphics[width=0.08\linewidth]{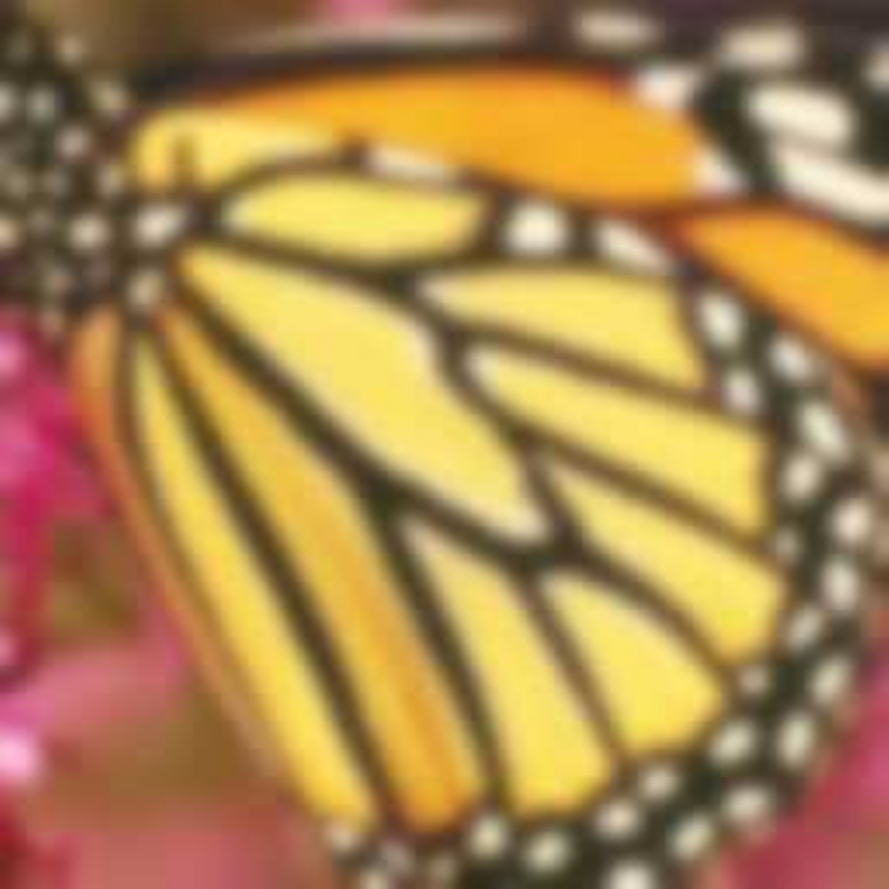} &
       \includegraphics[width=0.08\linewidth]{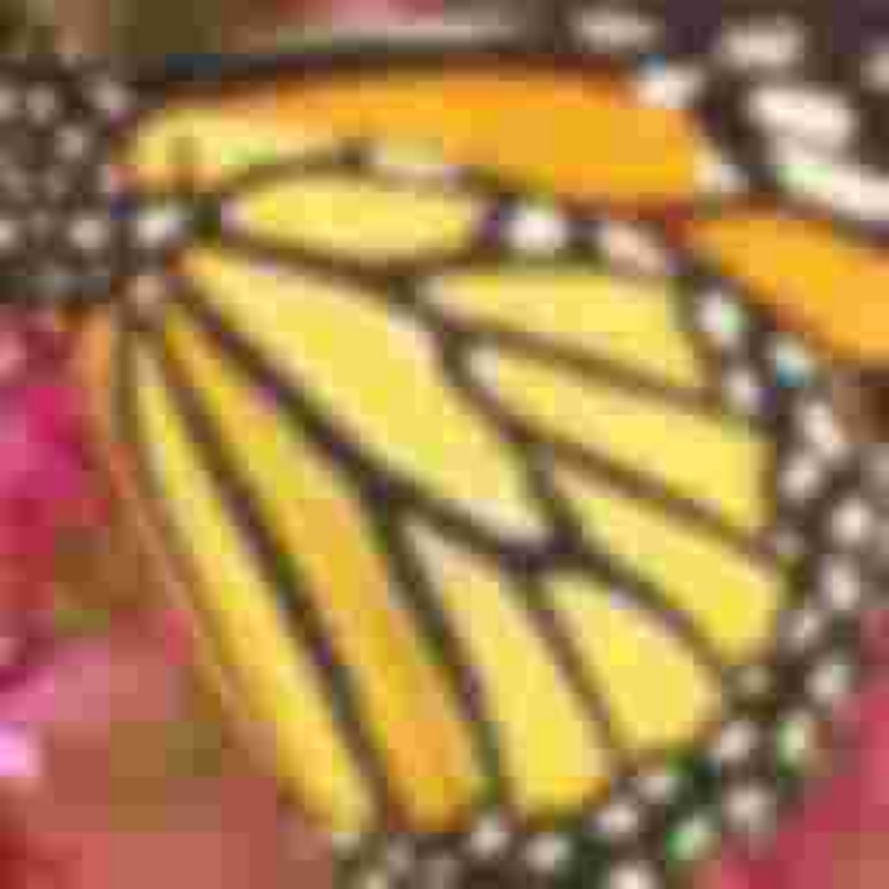} \\
&Grand Truth       &
       \includegraphics[width=0.08\linewidth]{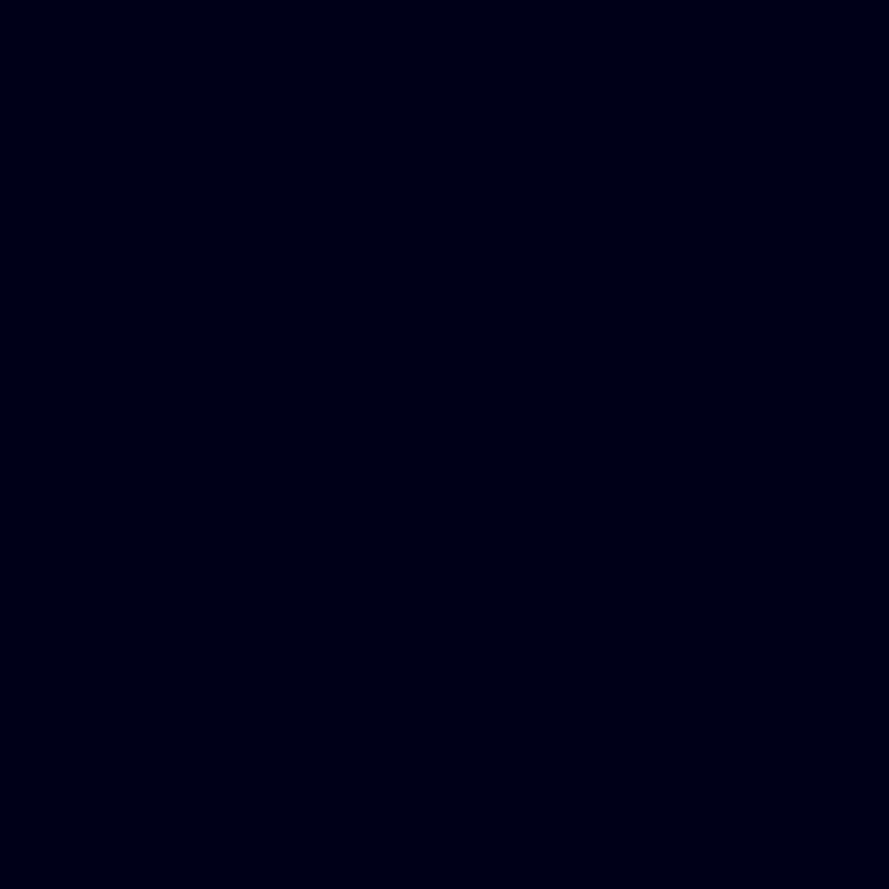} &
       \includegraphics[width=0.08\linewidth]{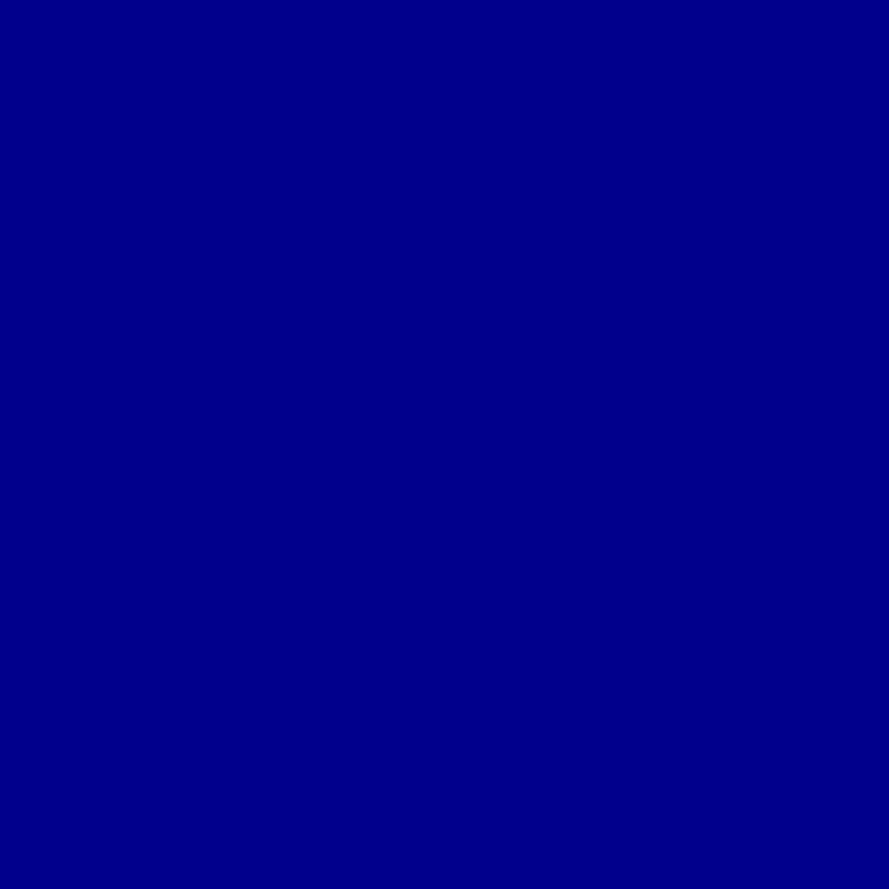} &
       \includegraphics[width=0.08\linewidth]{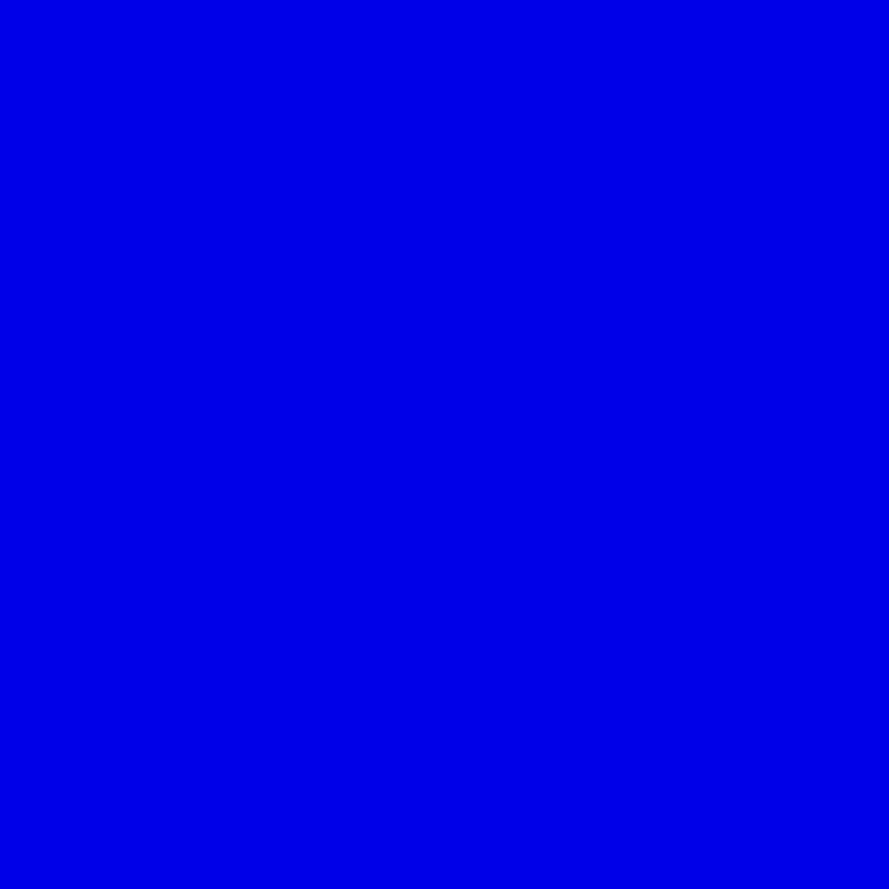} &
       \includegraphics[width=0.08\linewidth]{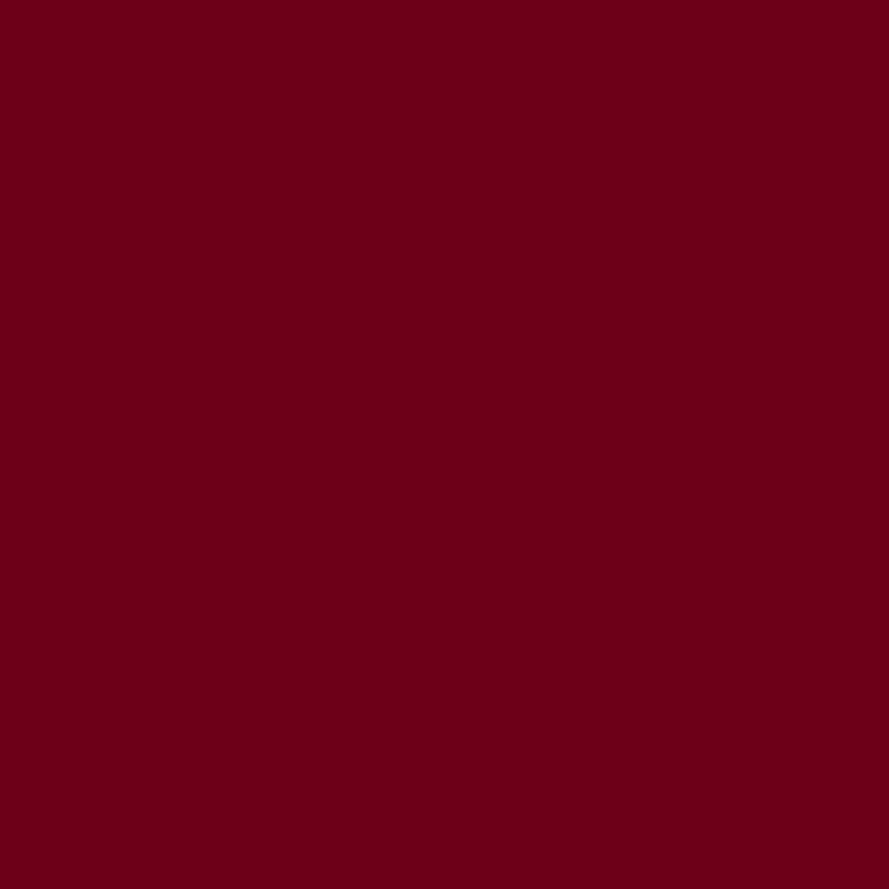} &
       \includegraphics[width=0.08\linewidth]{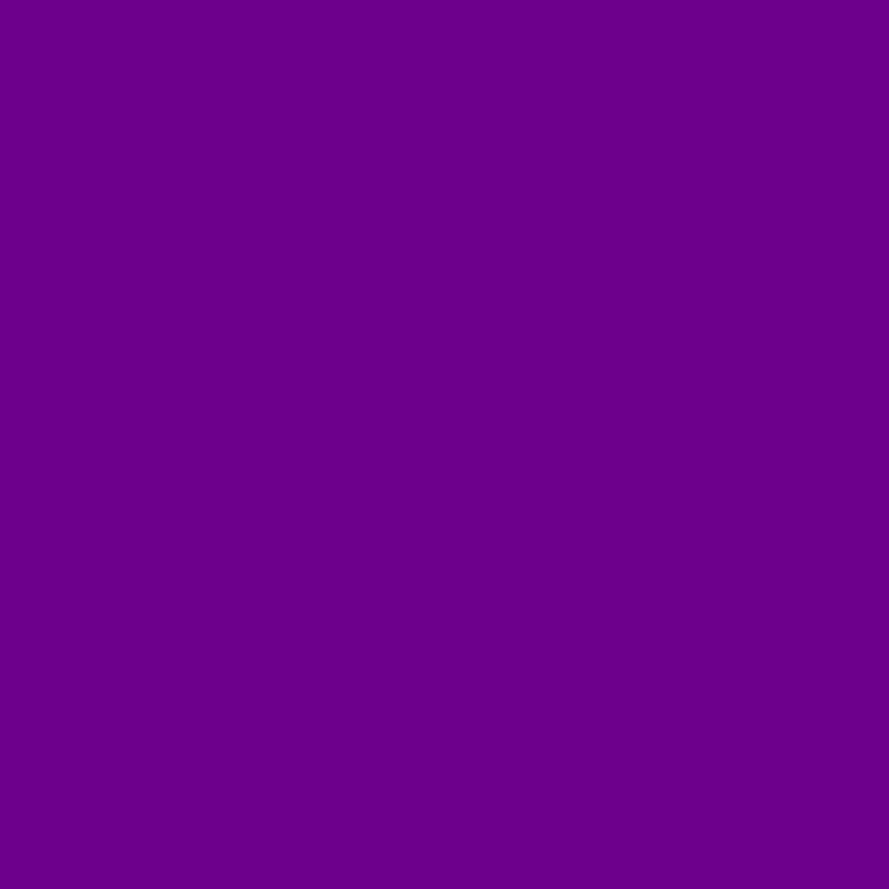} &
       \includegraphics[width=0.08\linewidth]{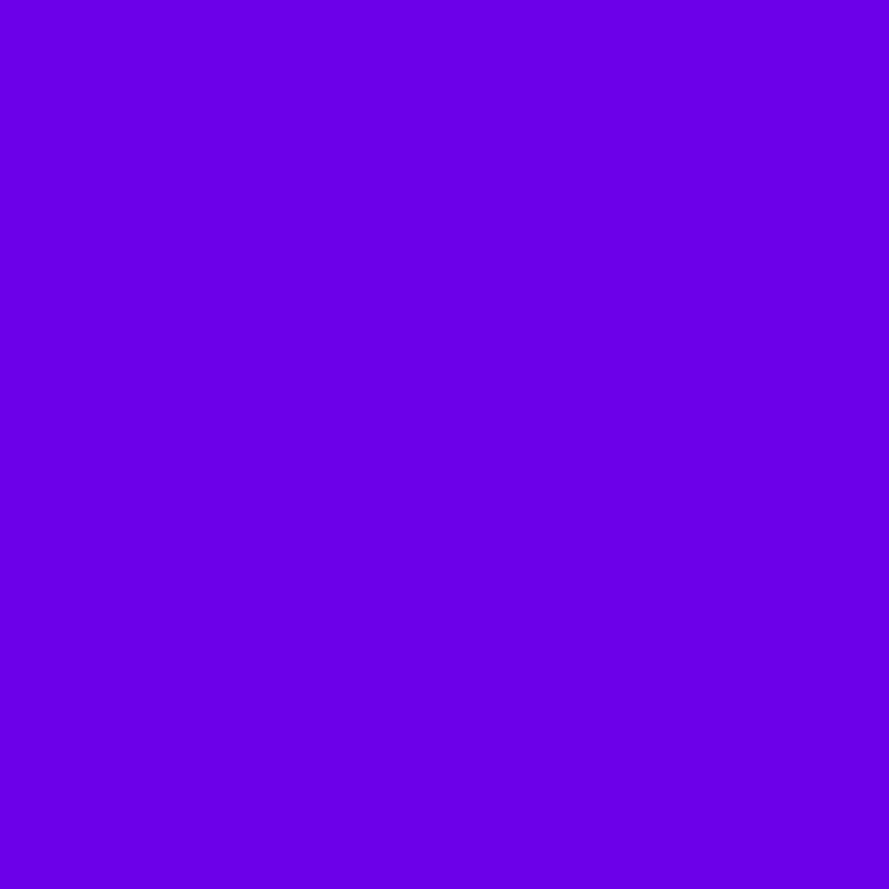} &
       \includegraphics[width=0.08\linewidth]{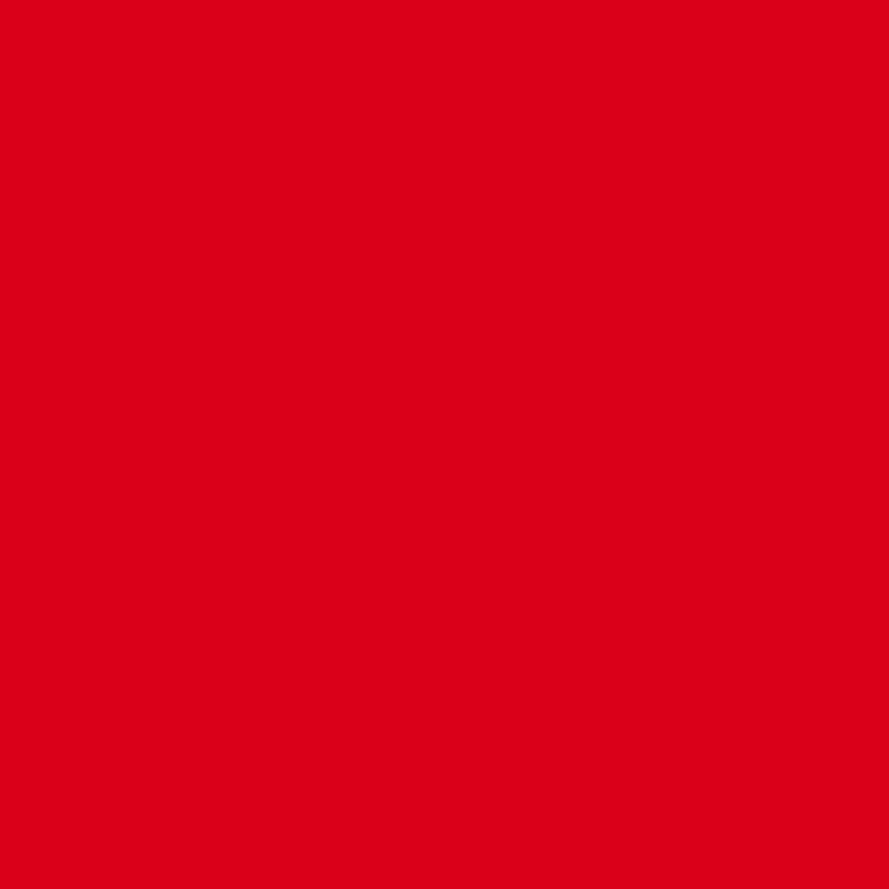} &
       \includegraphics[width=0.08\linewidth]{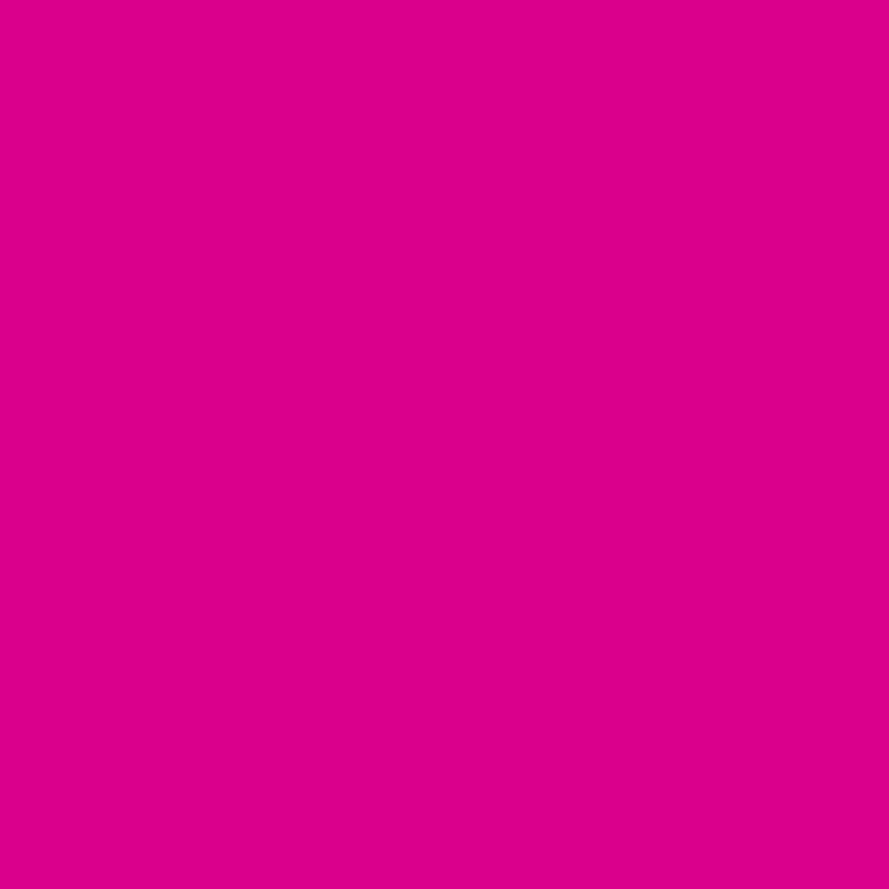} &
       \includegraphics[width=0.08\linewidth]{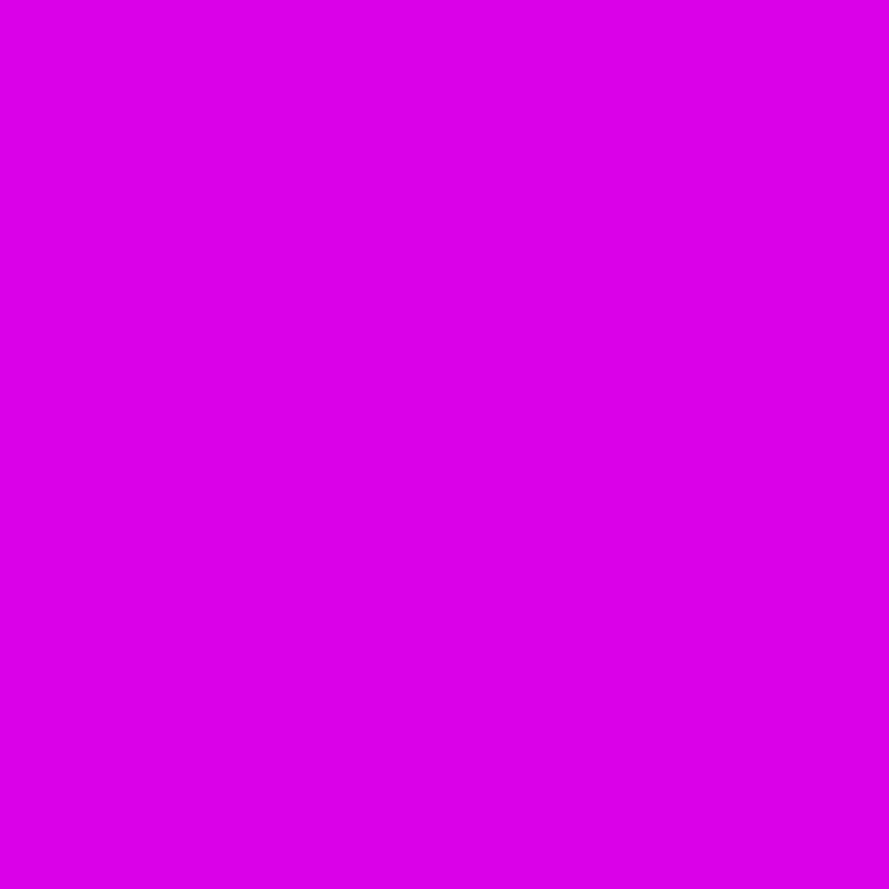} \\
&Estimated       &
        \includegraphics[width=0.08\linewidth]{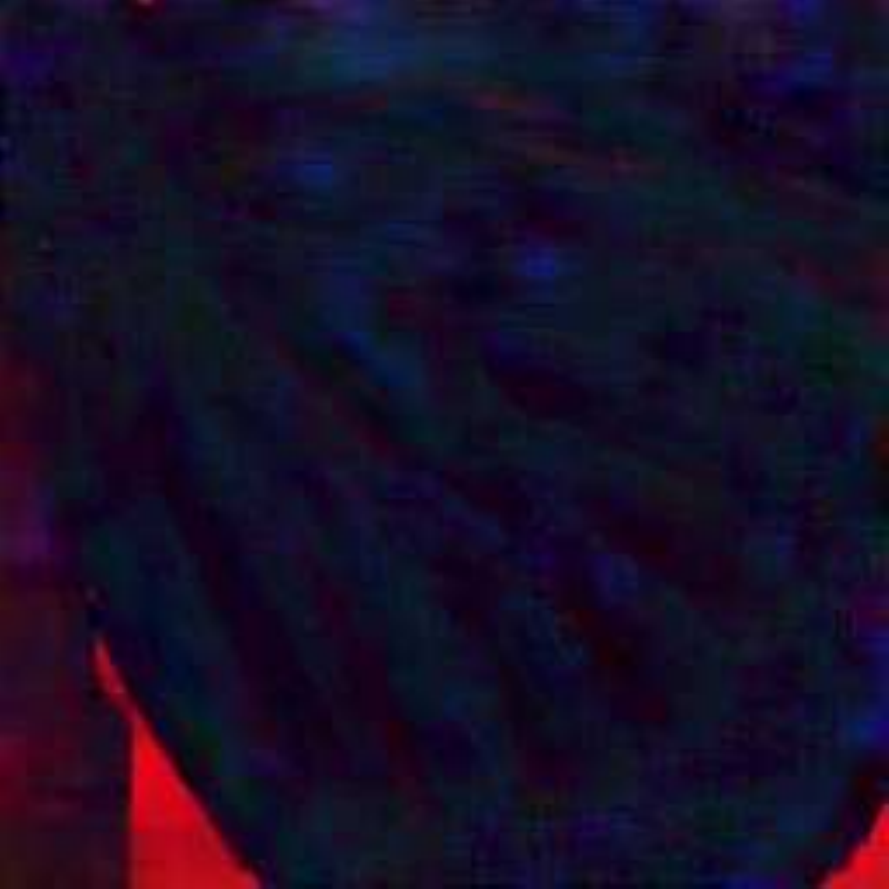} &
        \includegraphics[width=0.08\linewidth]{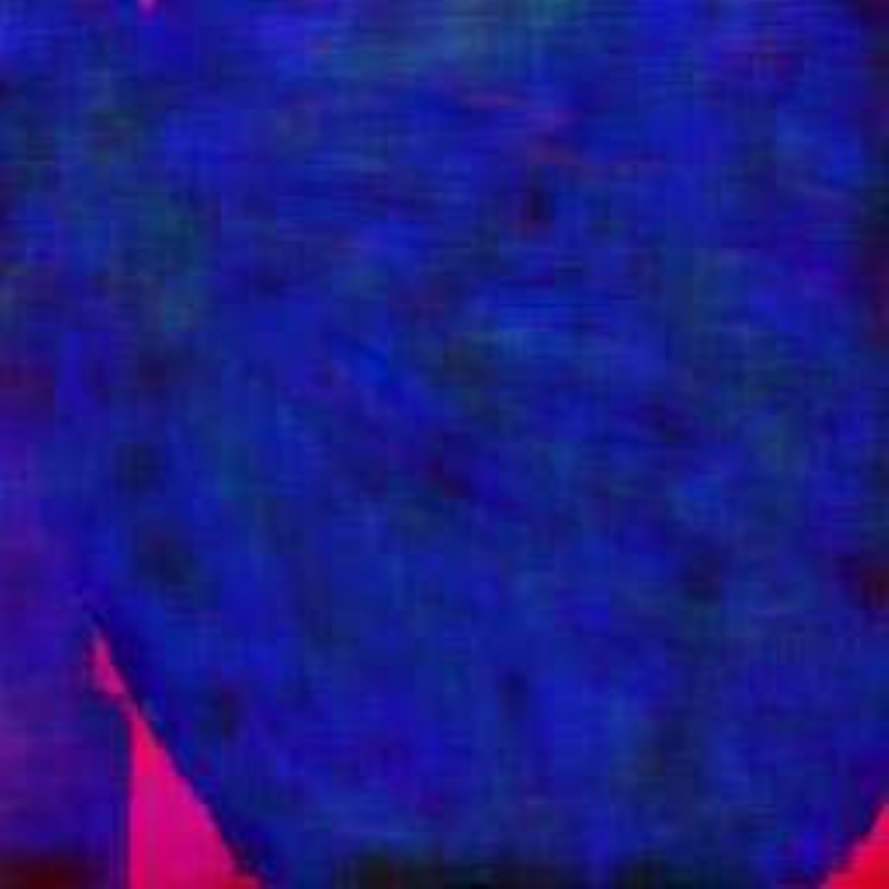} &
        \includegraphics[width=0.08\linewidth]{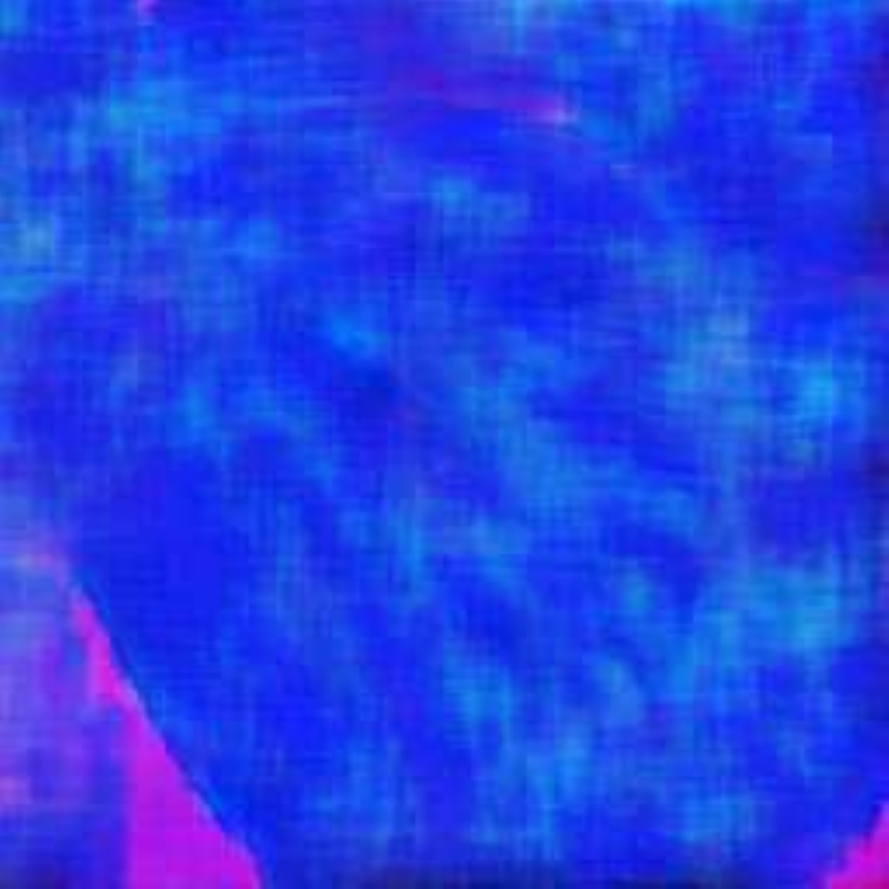} &
        \includegraphics[width=0.08\linewidth]{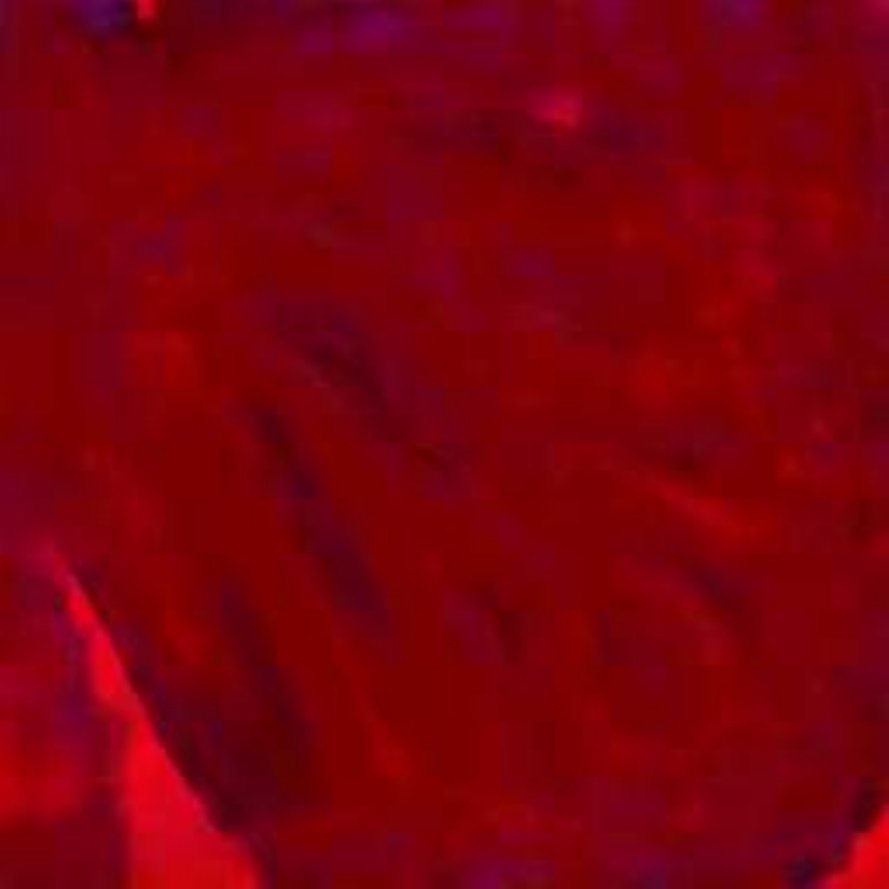} &
        \includegraphics[width=0.08\linewidth]{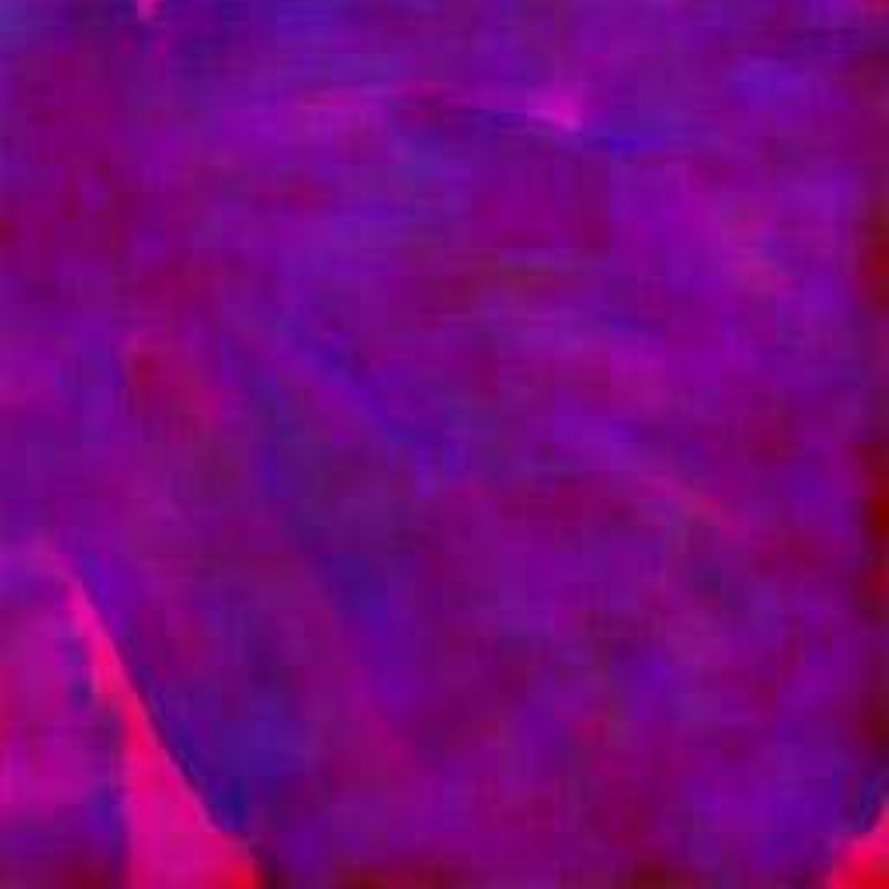} &
        \includegraphics[width=0.08\linewidth]{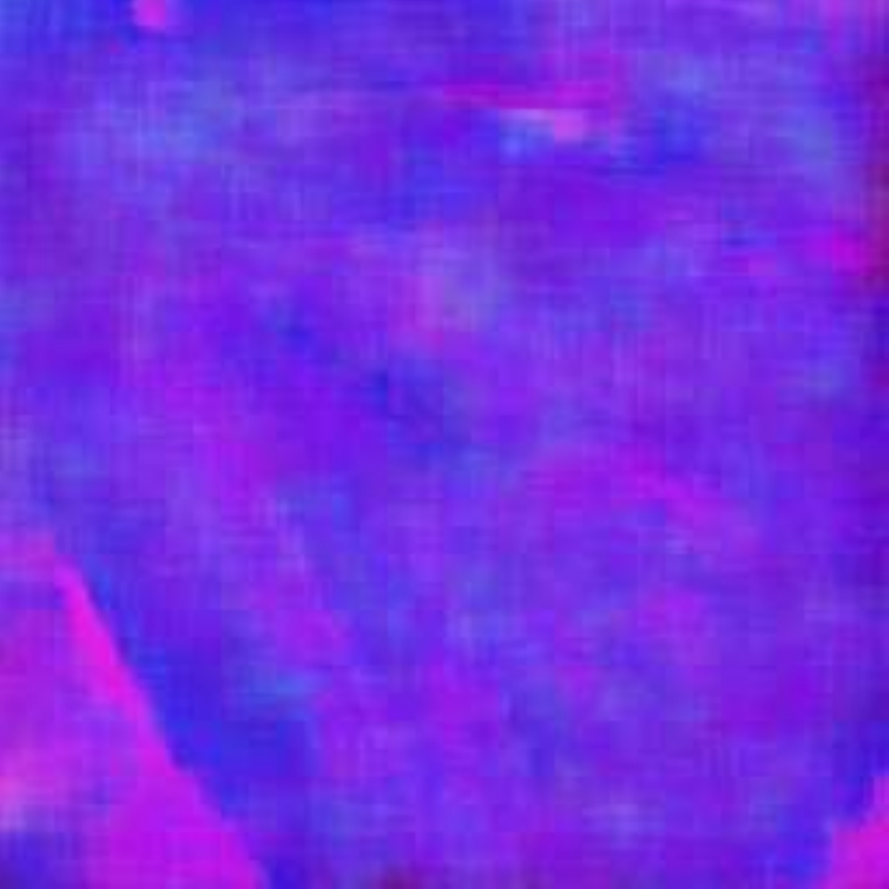} &
        \includegraphics[width=0.08\linewidth]{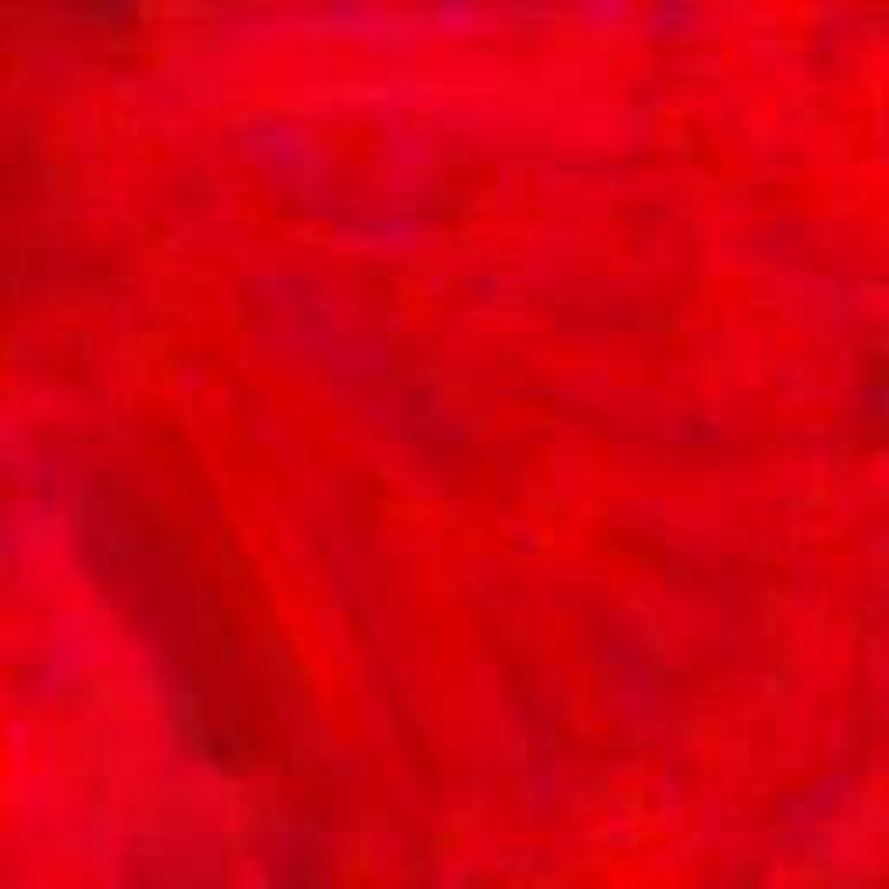} &
        \includegraphics[width=0.08\linewidth]{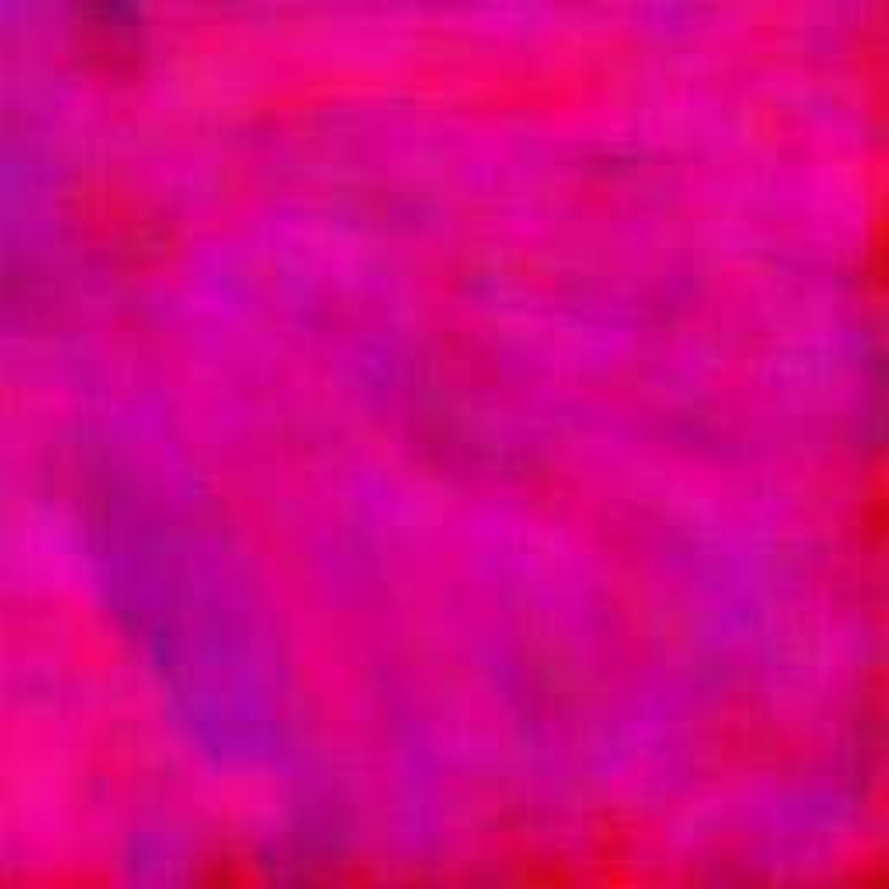} &
        \includegraphics[width=0.08\linewidth]{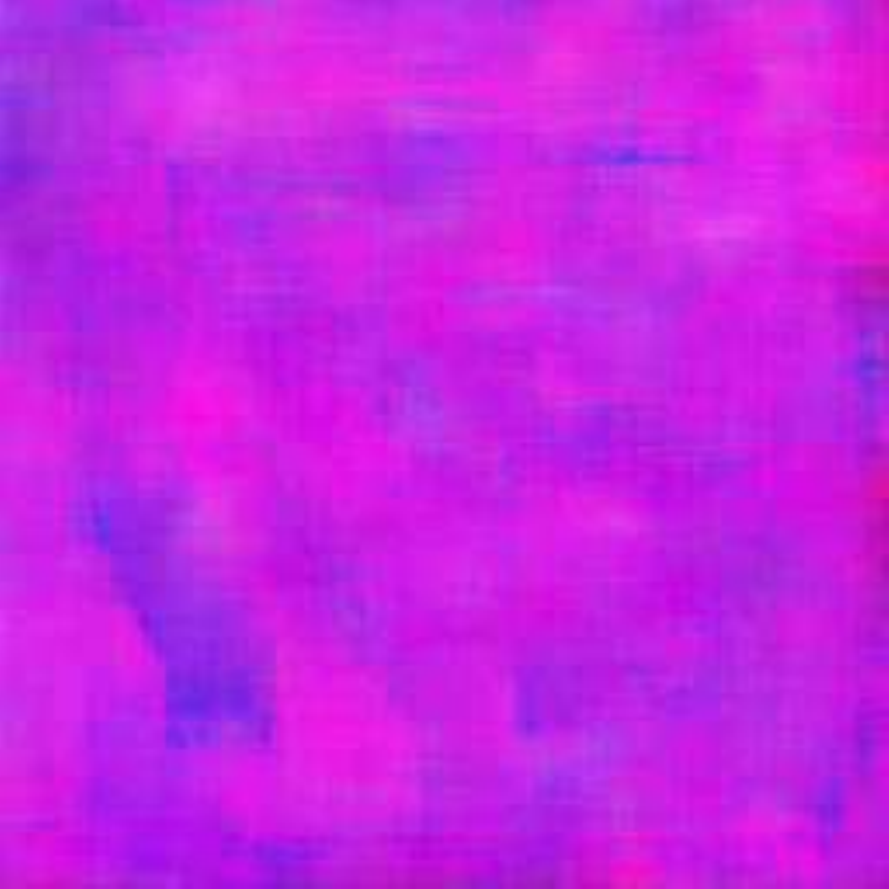} \\ \hline
\multirow{3}{*}{$\lambda=25$} &
Degraded  &
       \includegraphics[width=0.08\linewidth]{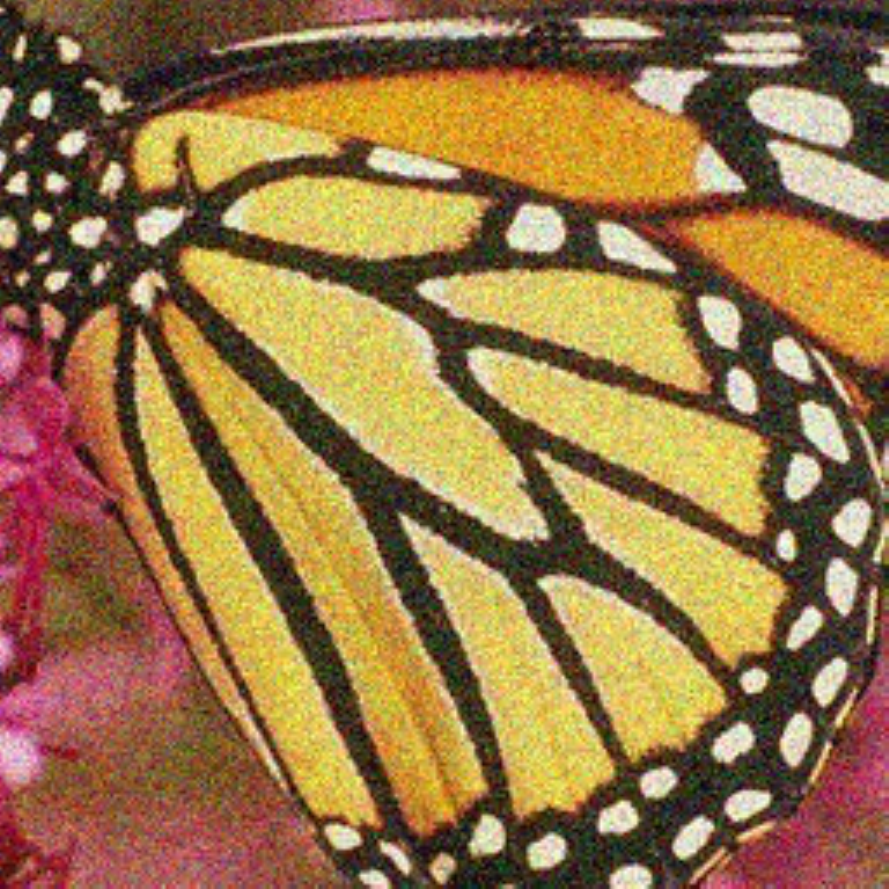} &
       \includegraphics[width=0.08\linewidth]{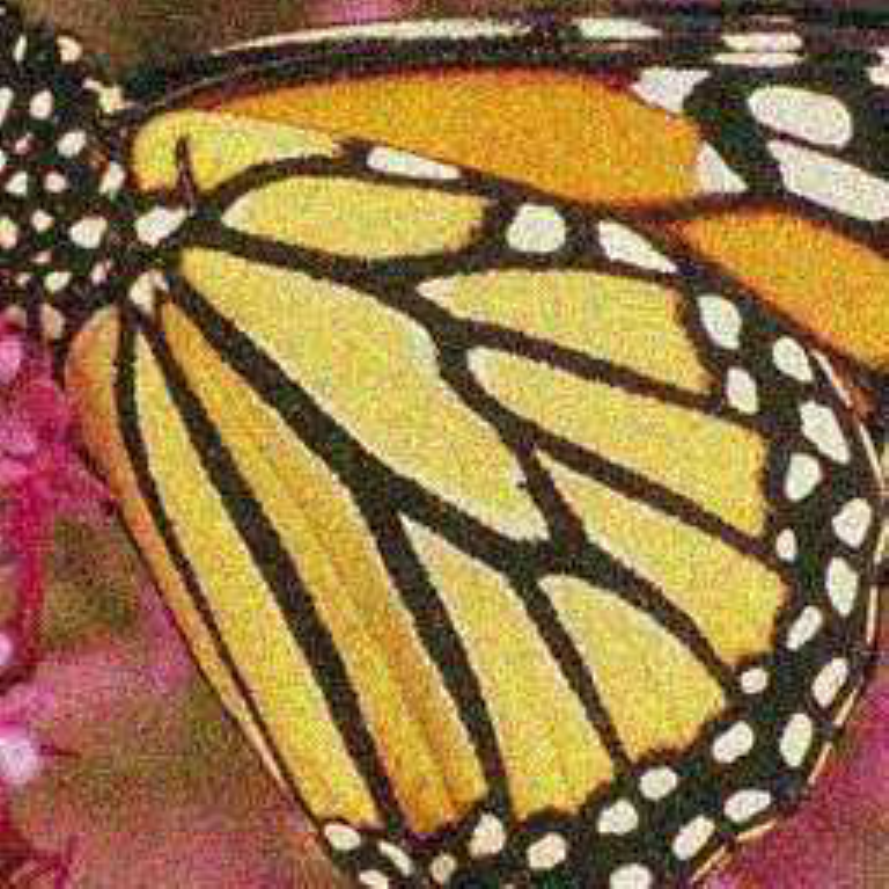} &
       \includegraphics[width=0.08\linewidth]{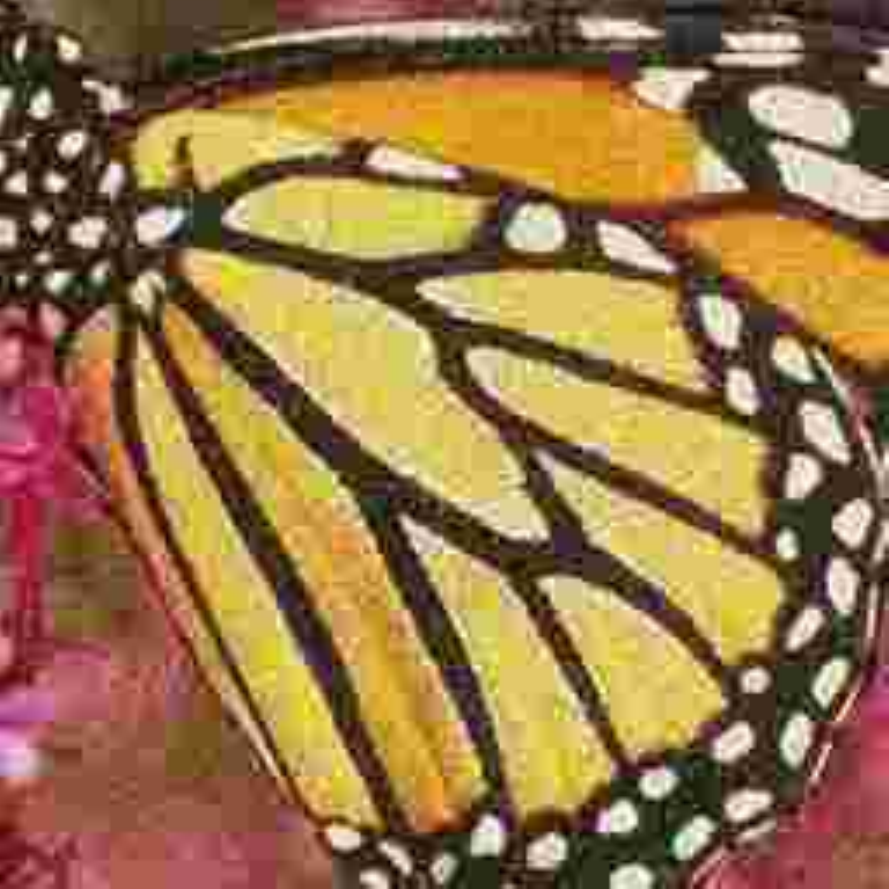} &
       \includegraphics[width=0.08\linewidth]{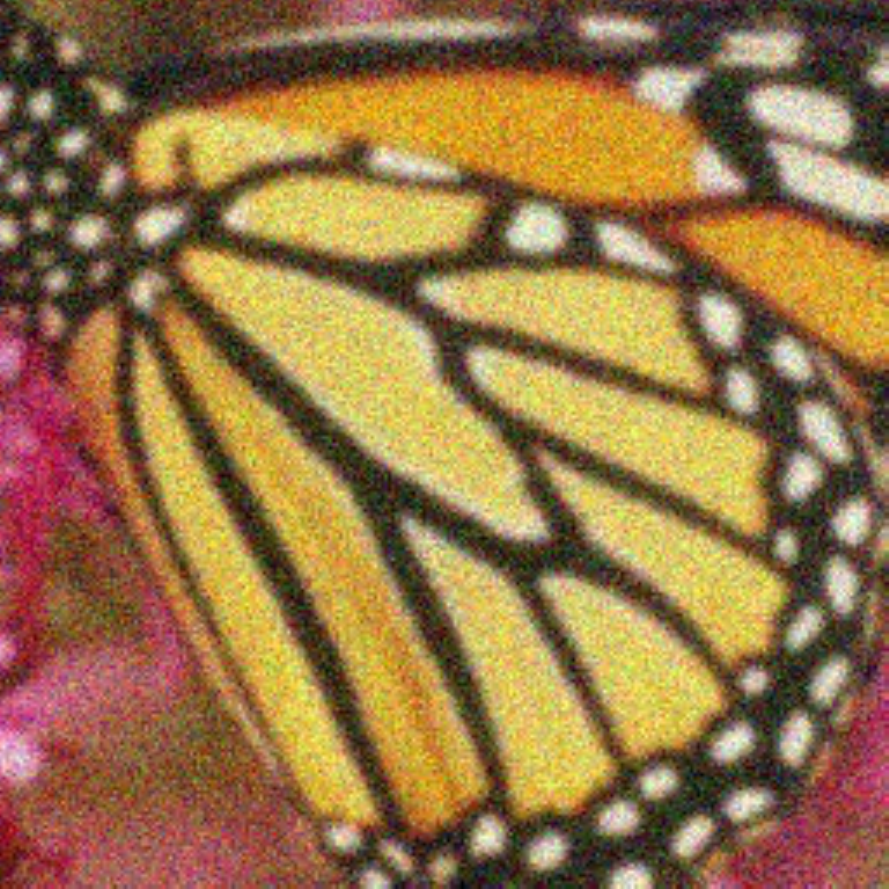} &
       \includegraphics[width=0.08\linewidth]{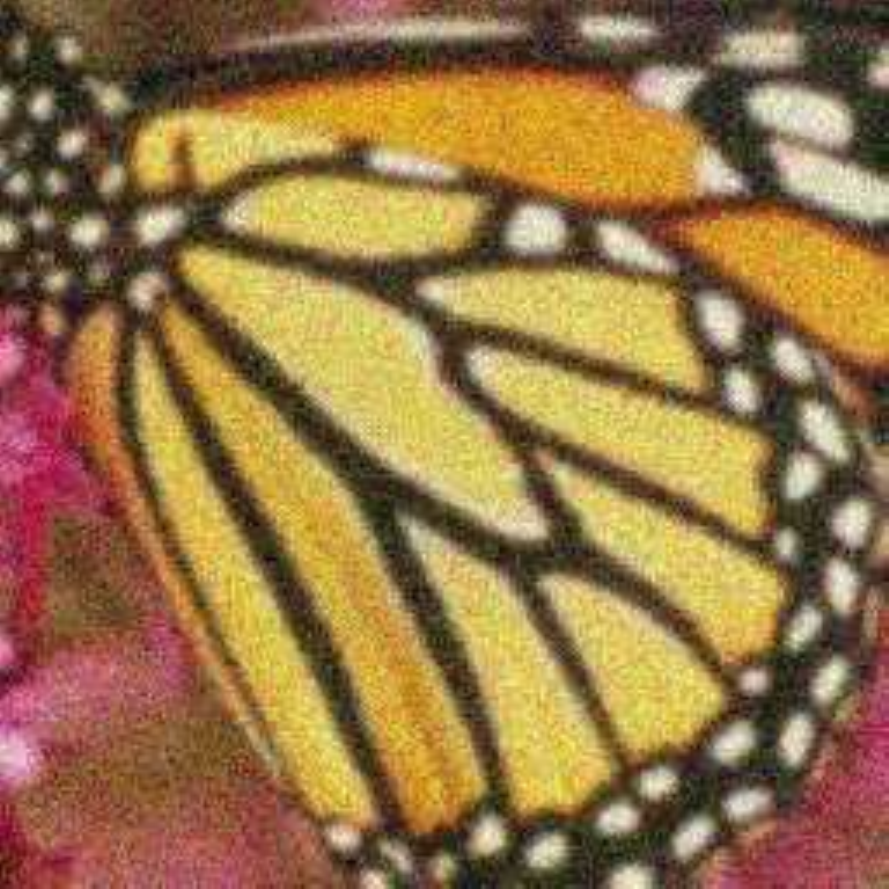} &
       \includegraphics[width=0.08\linewidth]{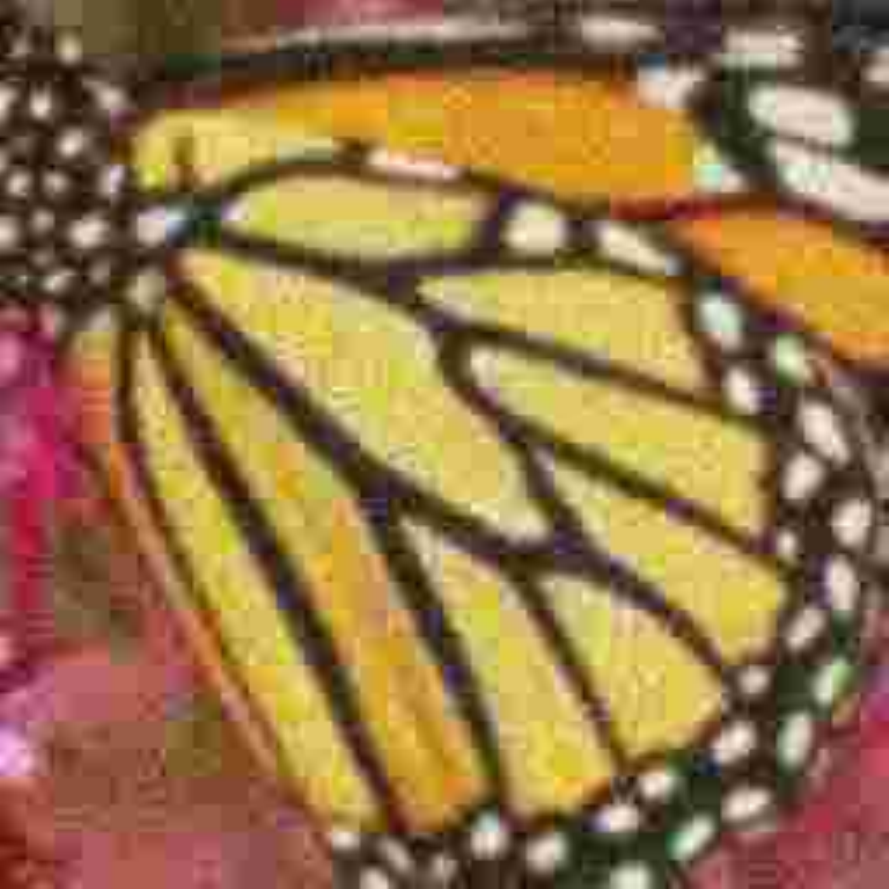} &
       \includegraphics[width=0.08\linewidth]{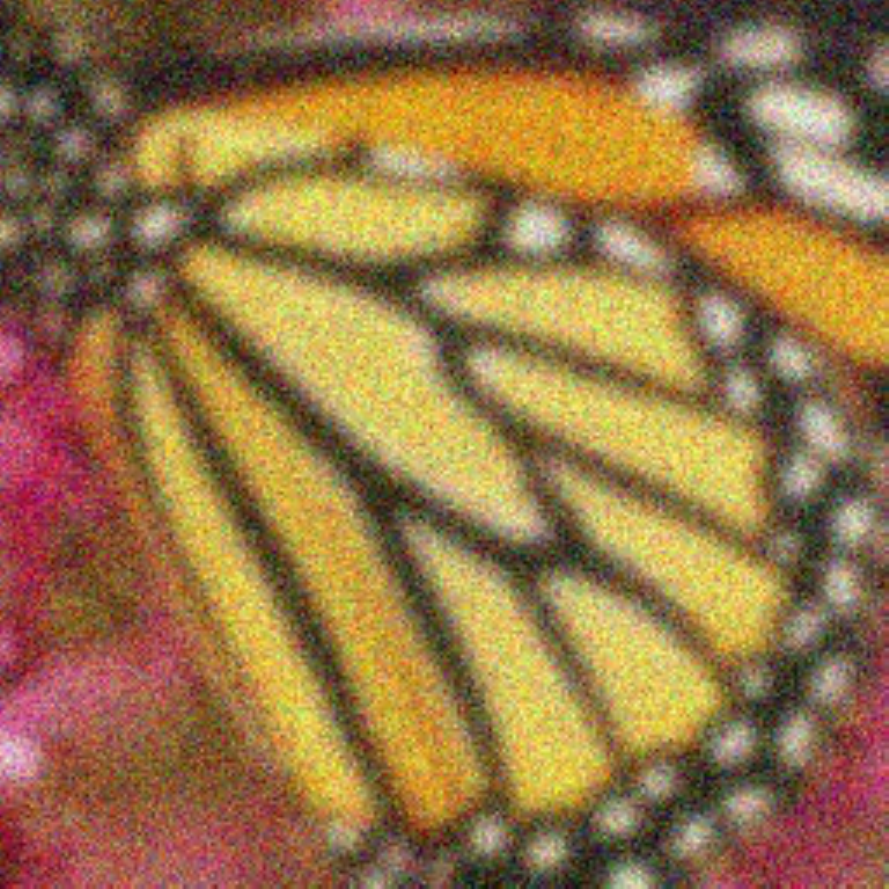} &
       \includegraphics[width=0.08\linewidth]{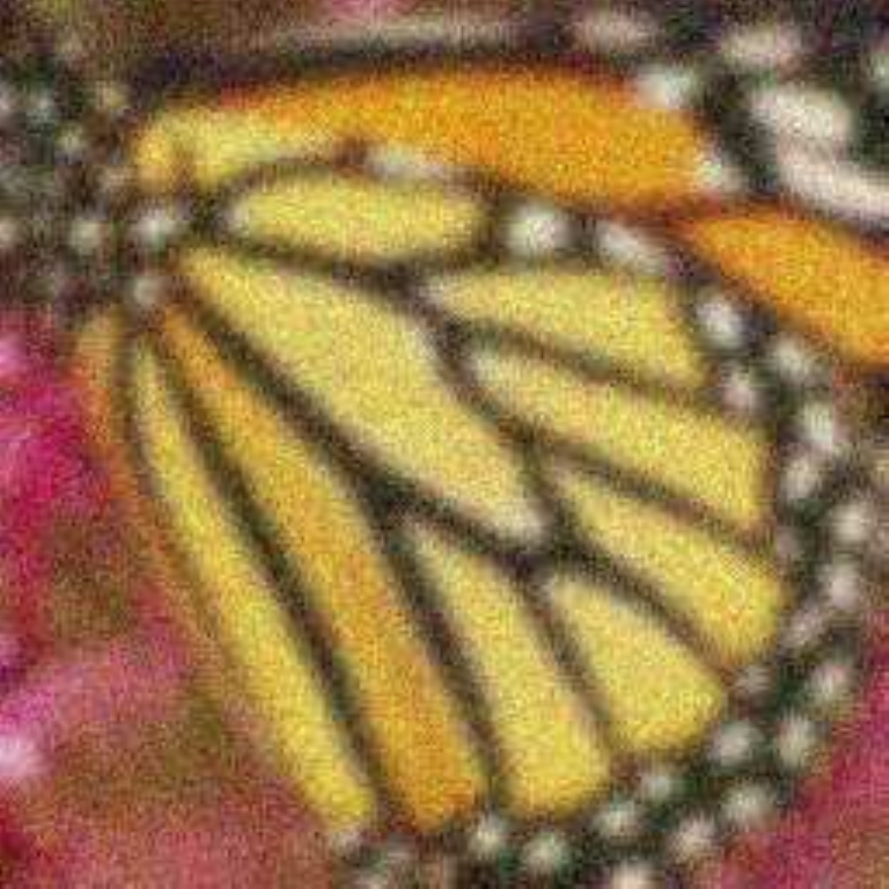} &
       \includegraphics[width=0.08\linewidth]{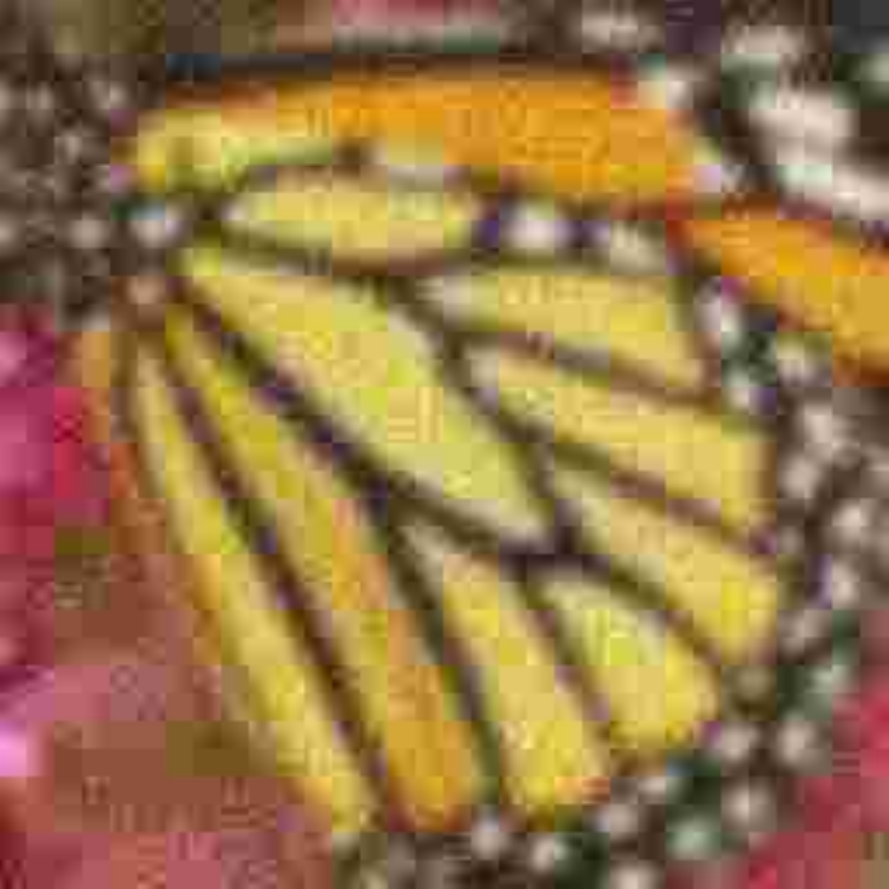} \\
&Grand Truth       &
       \includegraphics[width=0.08\linewidth]{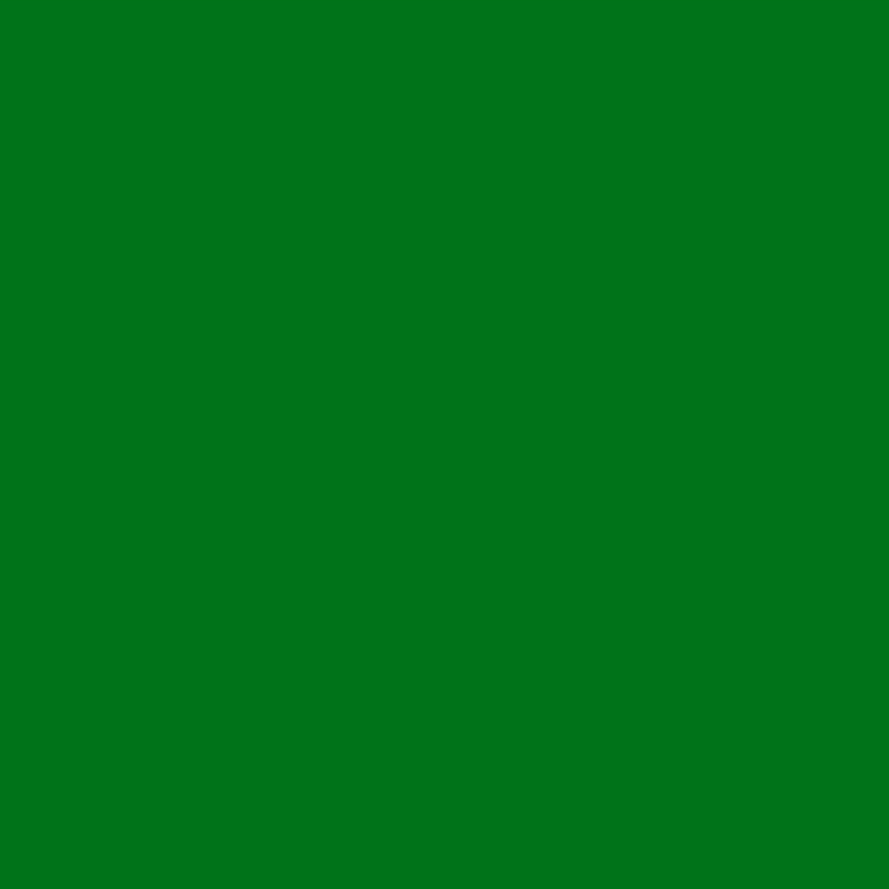} &
       \includegraphics[width=0.08\linewidth]{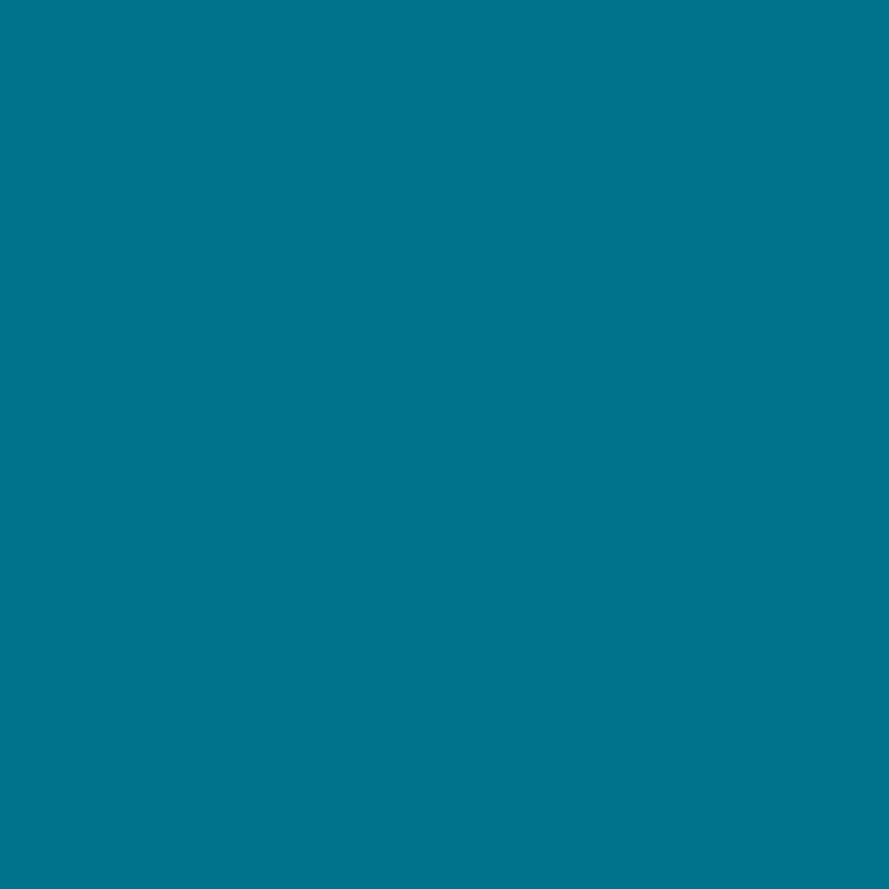} &
       \includegraphics[width=0.08\linewidth]{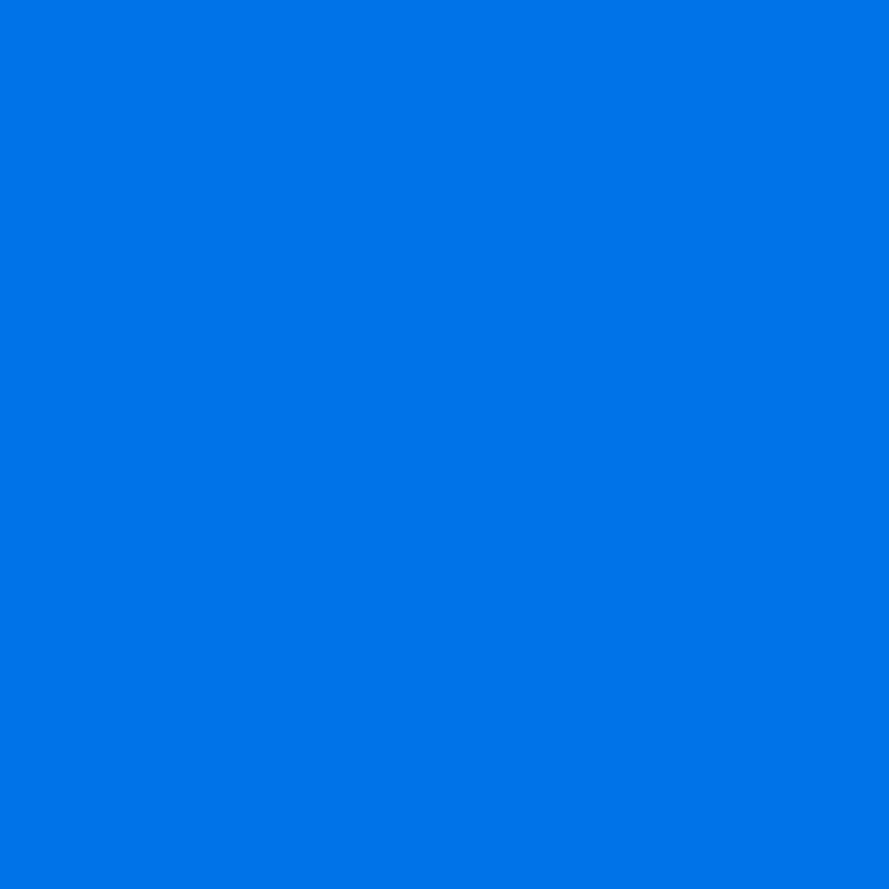} &
       \includegraphics[width=0.08\linewidth]{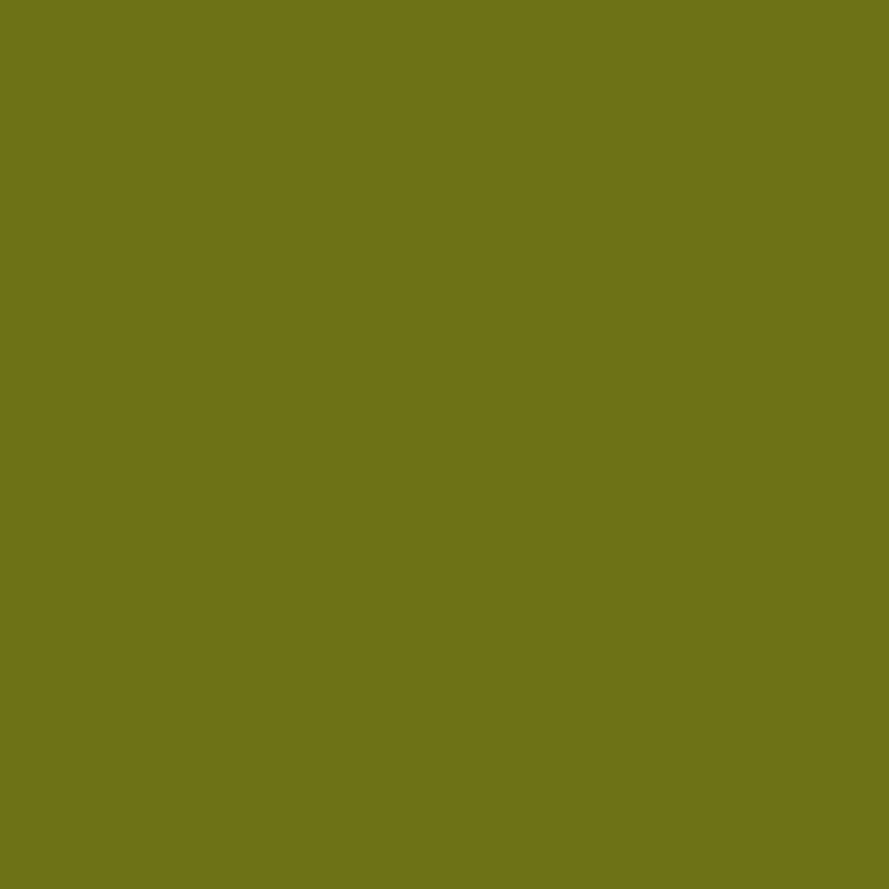} &
       \includegraphics[width=0.08\linewidth]{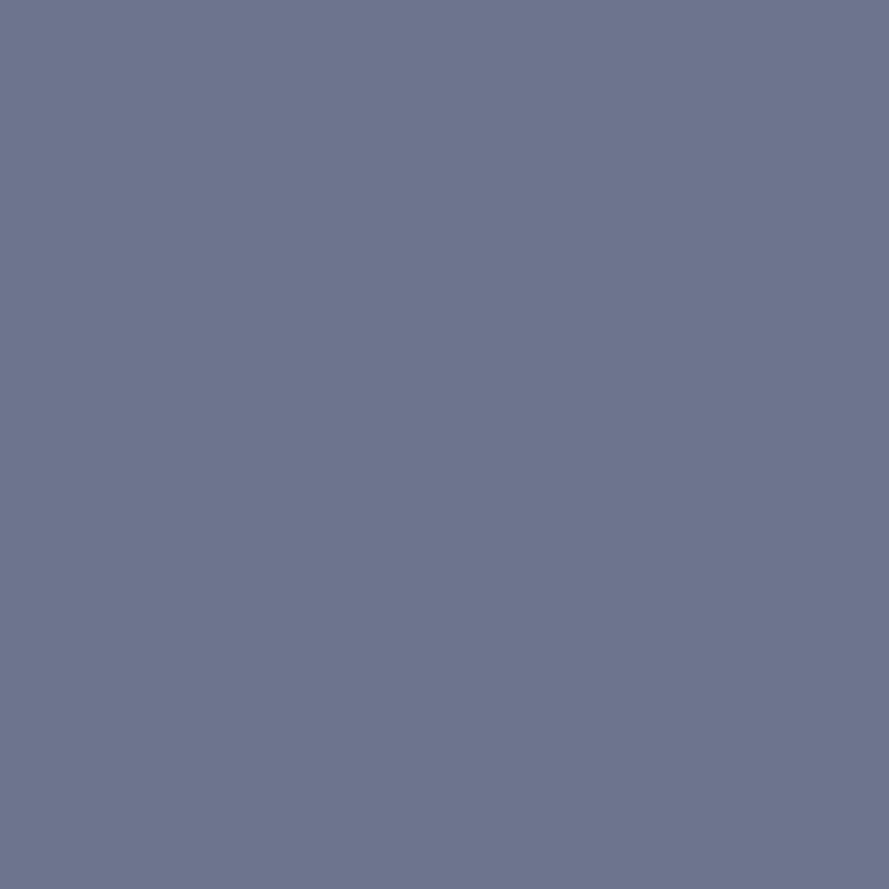} &
       \includegraphics[width=0.08\linewidth]{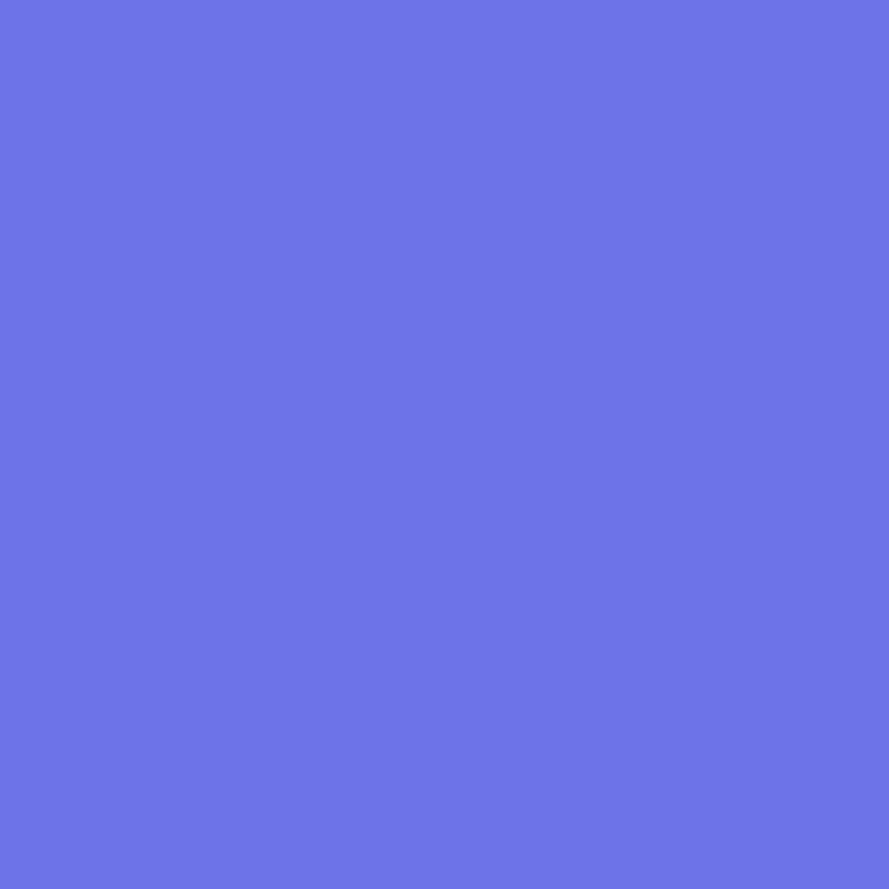} &
       \includegraphics[width=0.08\linewidth]{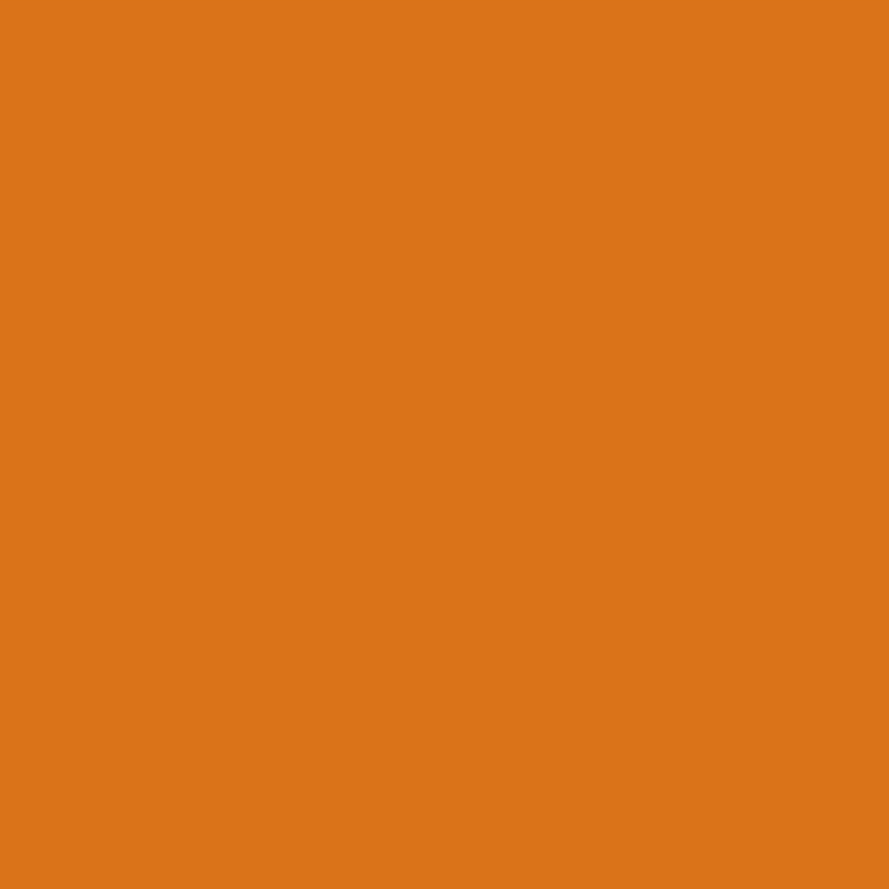} &
       \includegraphics[width=0.08\linewidth]{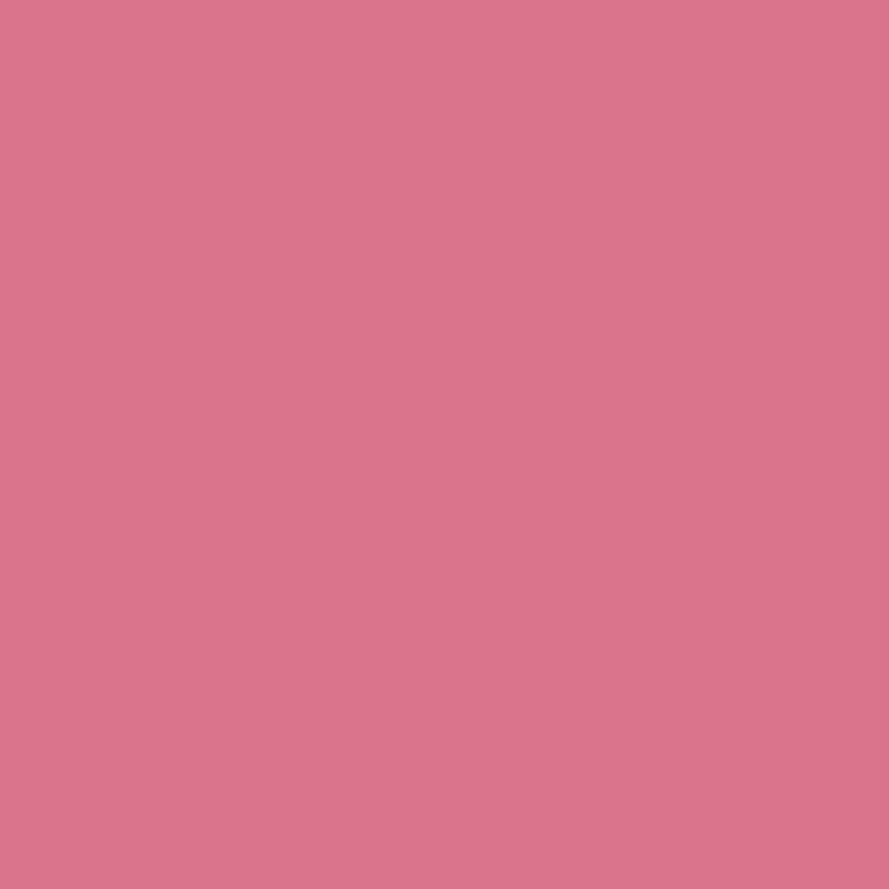} &
       \includegraphics[width=0.08\linewidth]{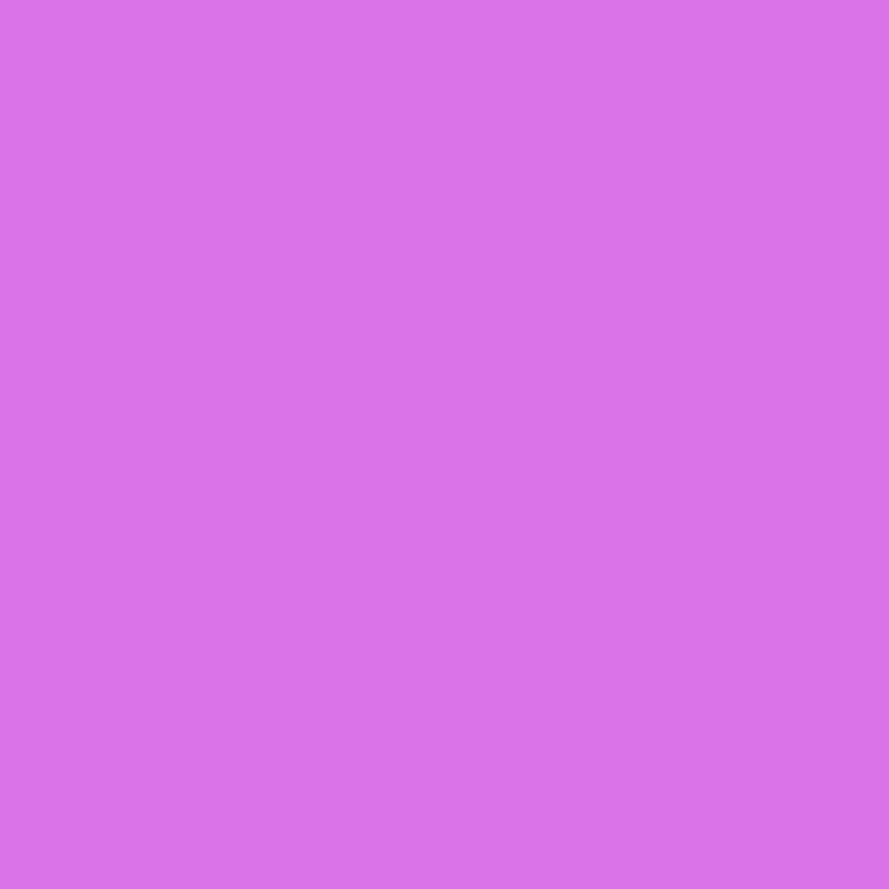} \\
&Estimated       &
        \includegraphics[width=0.08\linewidth]{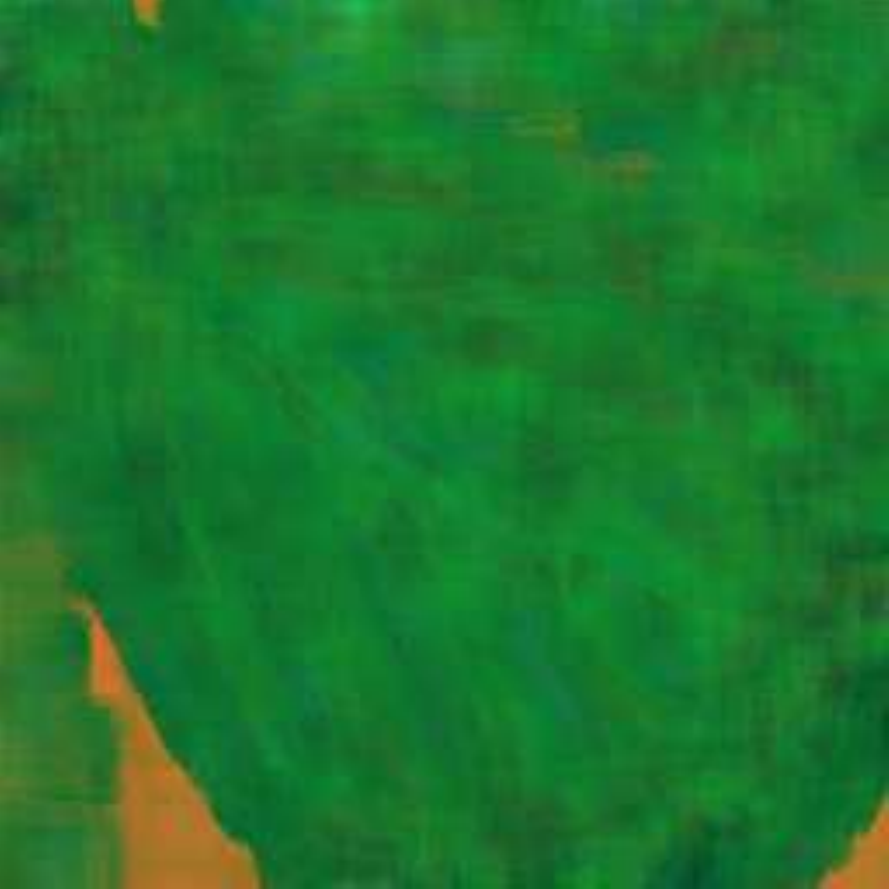} &
        \includegraphics[width=0.08\linewidth]{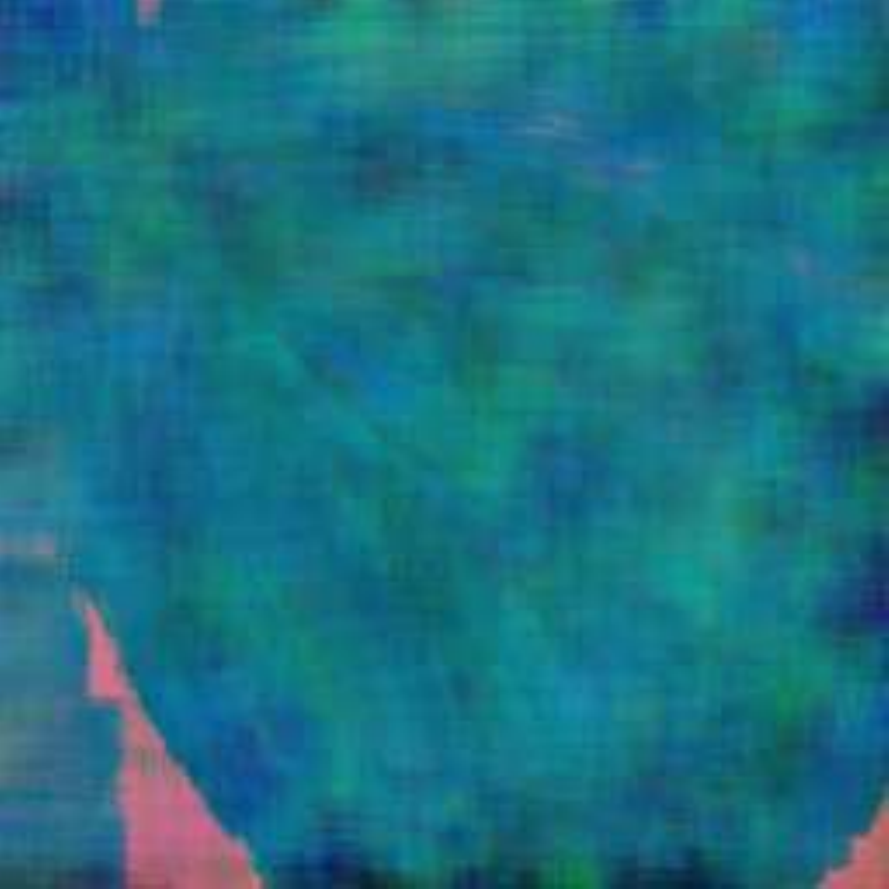} &
        \includegraphics[width=0.08\linewidth]{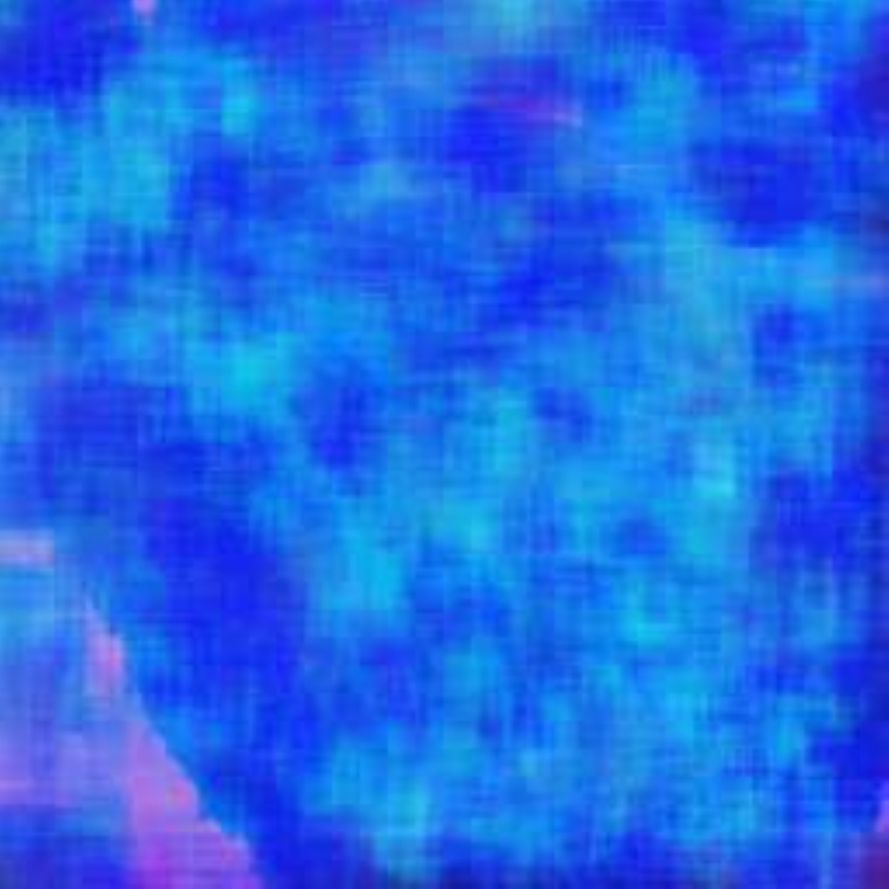} &
        \includegraphics[width=0.08\linewidth]{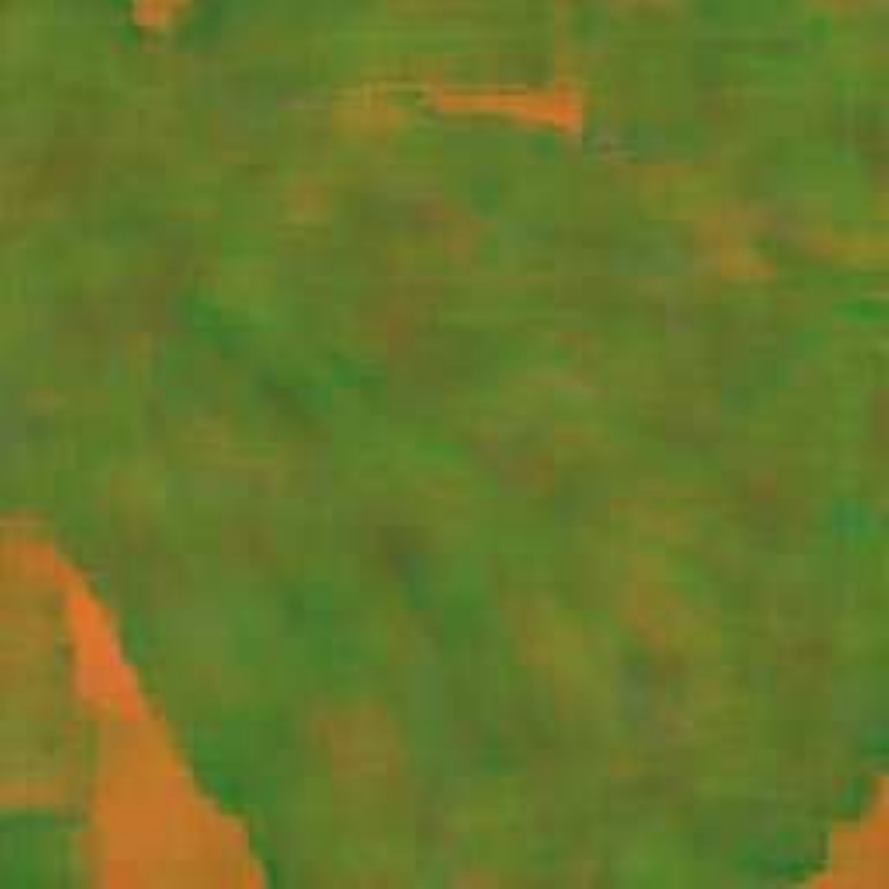} &
        \includegraphics[width=0.08\linewidth]{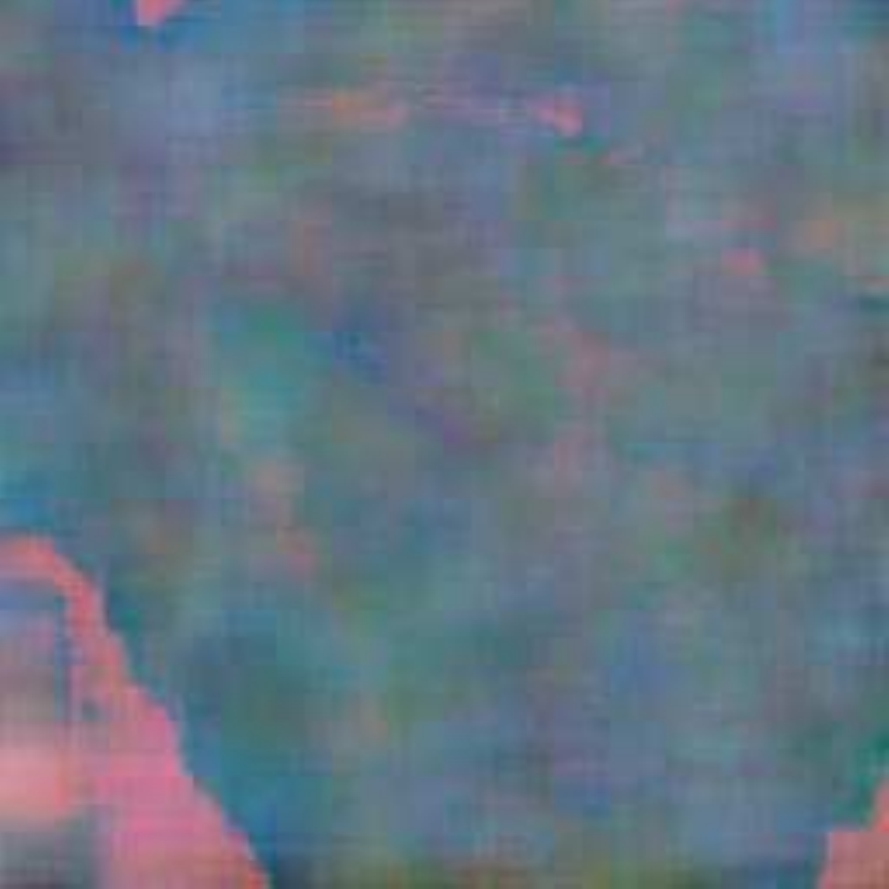} &
        \includegraphics[width=0.08\linewidth]{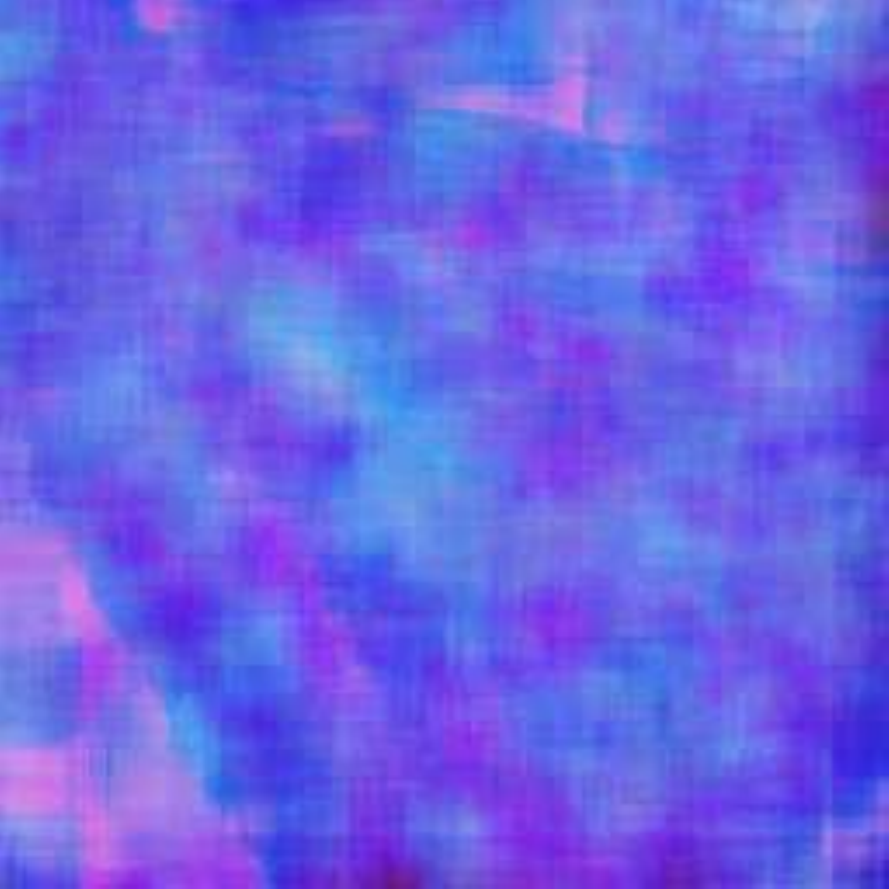} &
        \includegraphics[width=0.08\linewidth]{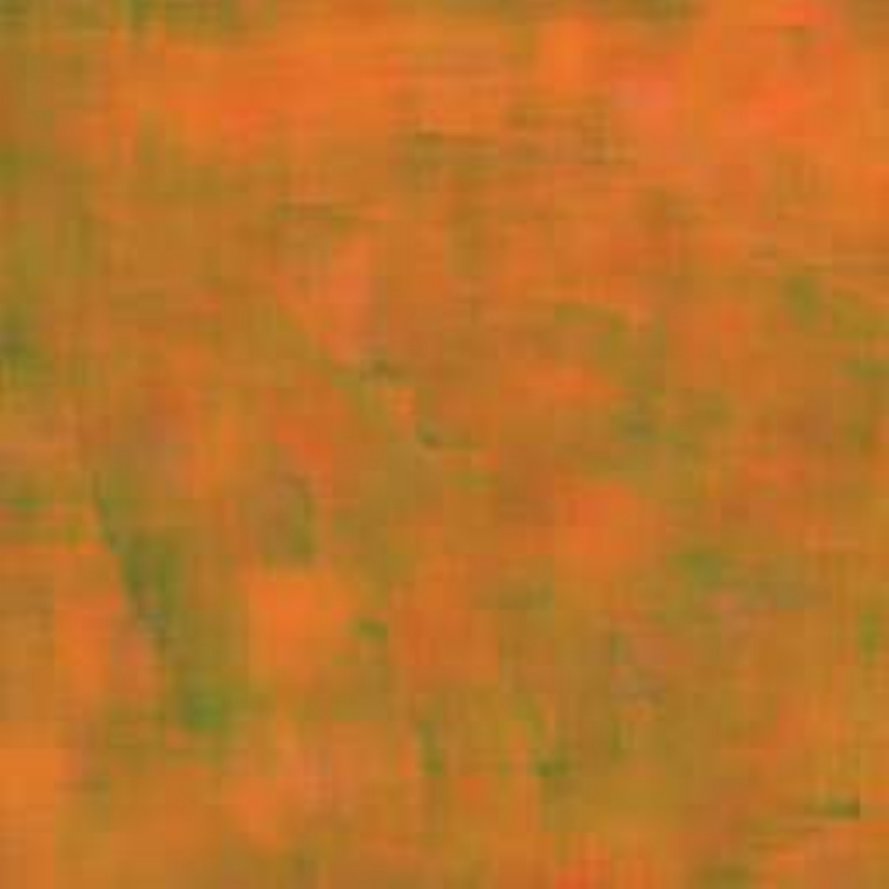} &
        \includegraphics[width=0.08\linewidth]{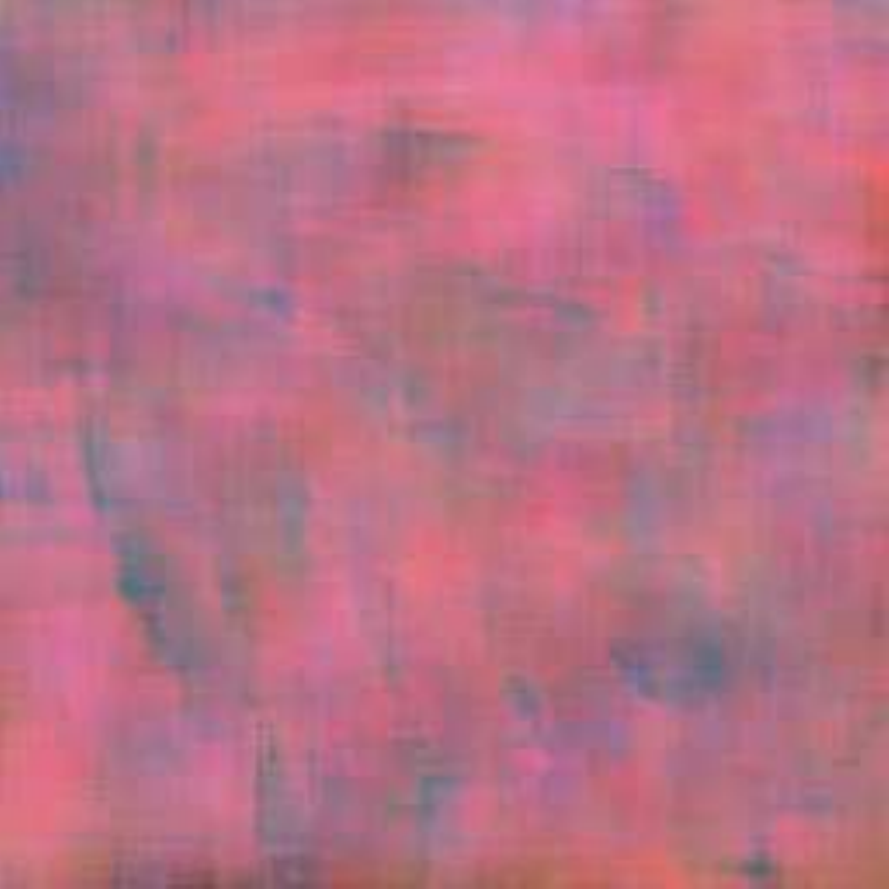} &
        \includegraphics[width=0.08\linewidth]{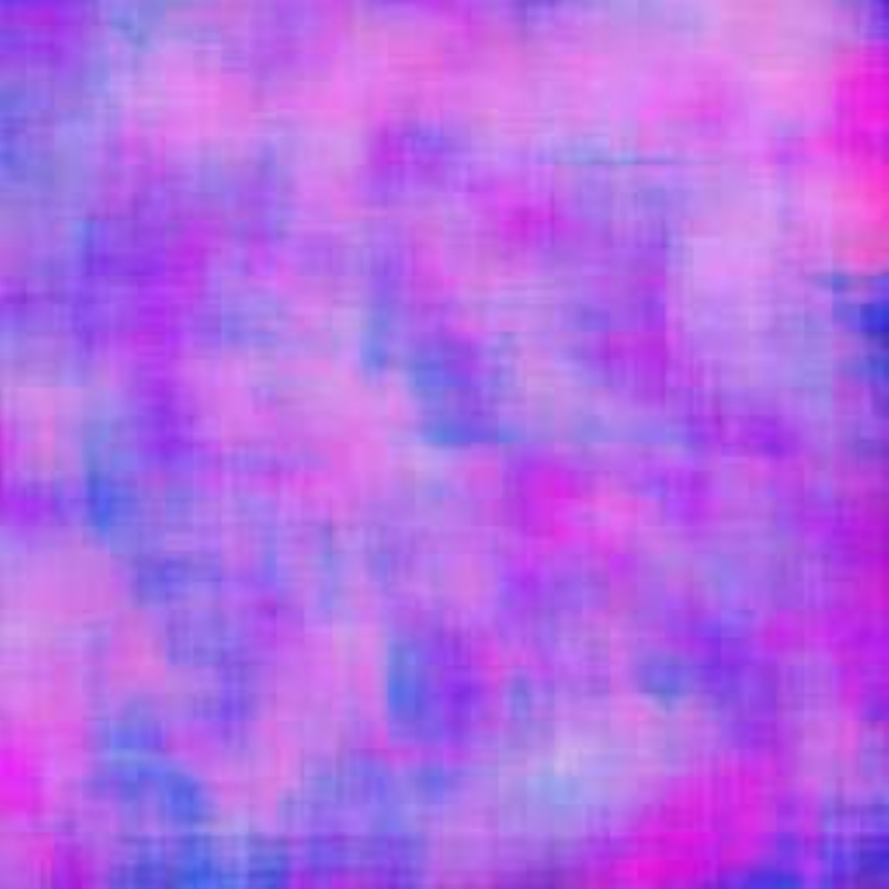} \\ \hline
\multirow{3}{*}{$\lambda=55$} &
Degraded  &
       \includegraphics[width=0.08\linewidth]{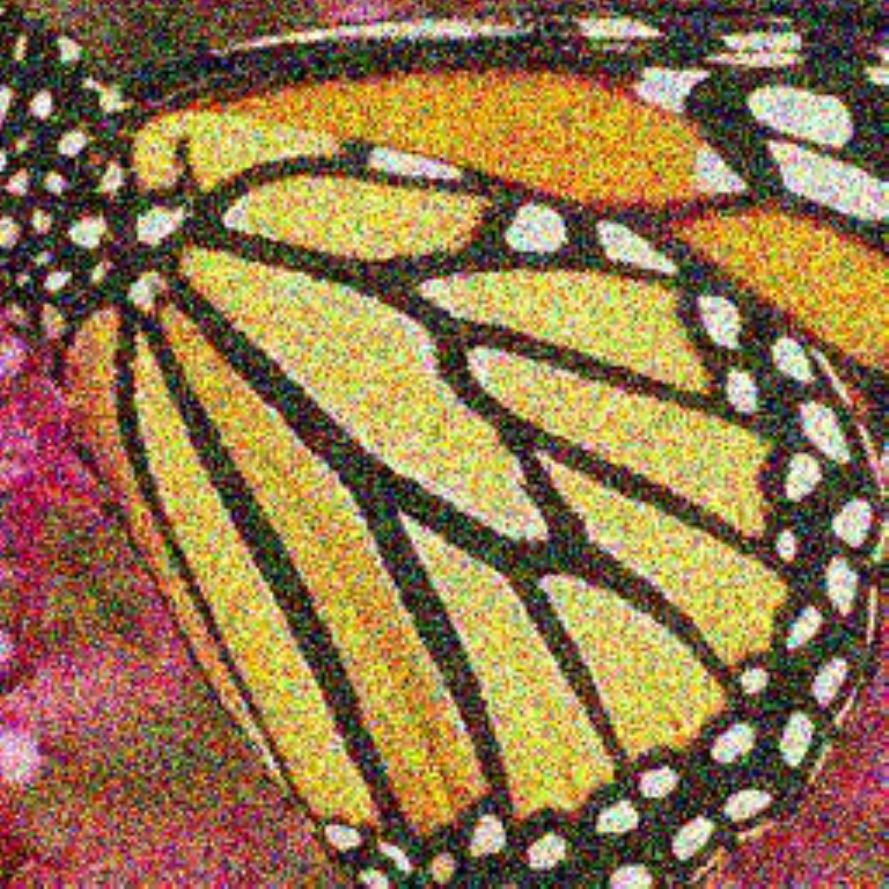} &
       \includegraphics[width=0.08\linewidth]{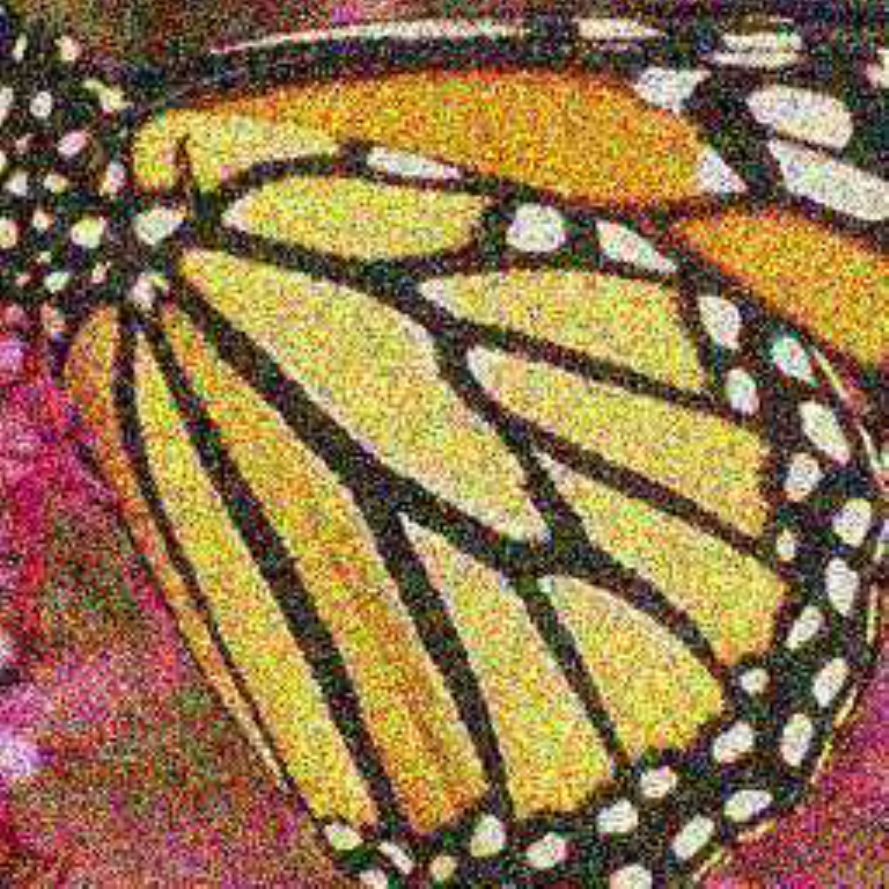} &
       \includegraphics[width=0.08\linewidth]{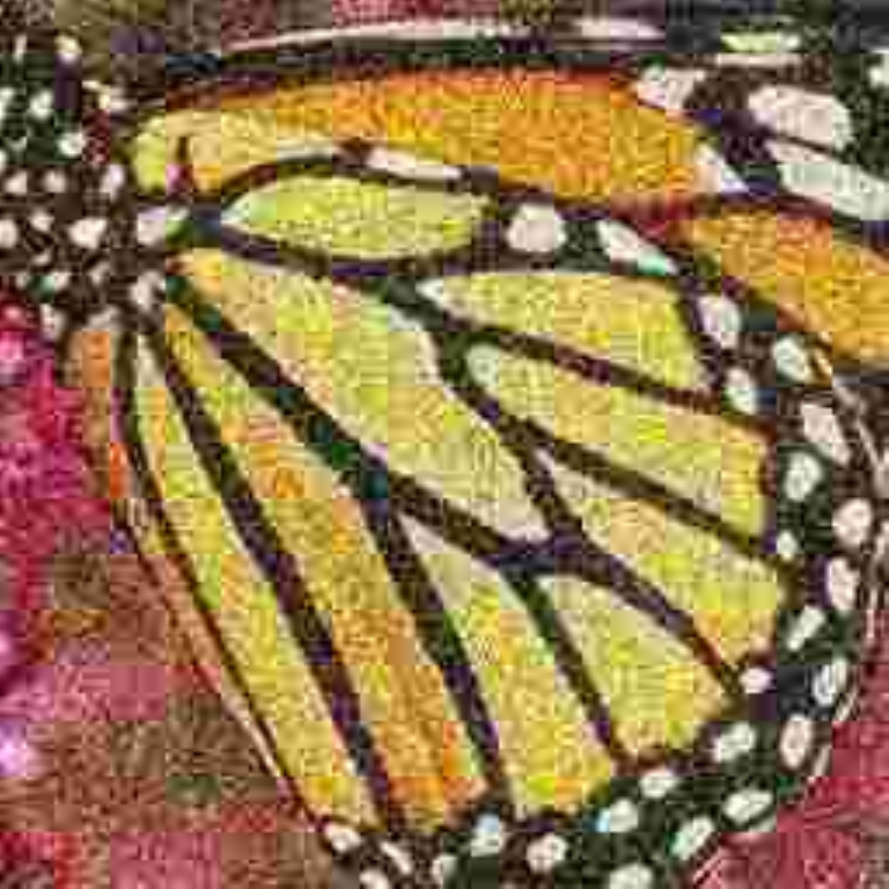} &
       \includegraphics[width=0.08\linewidth]{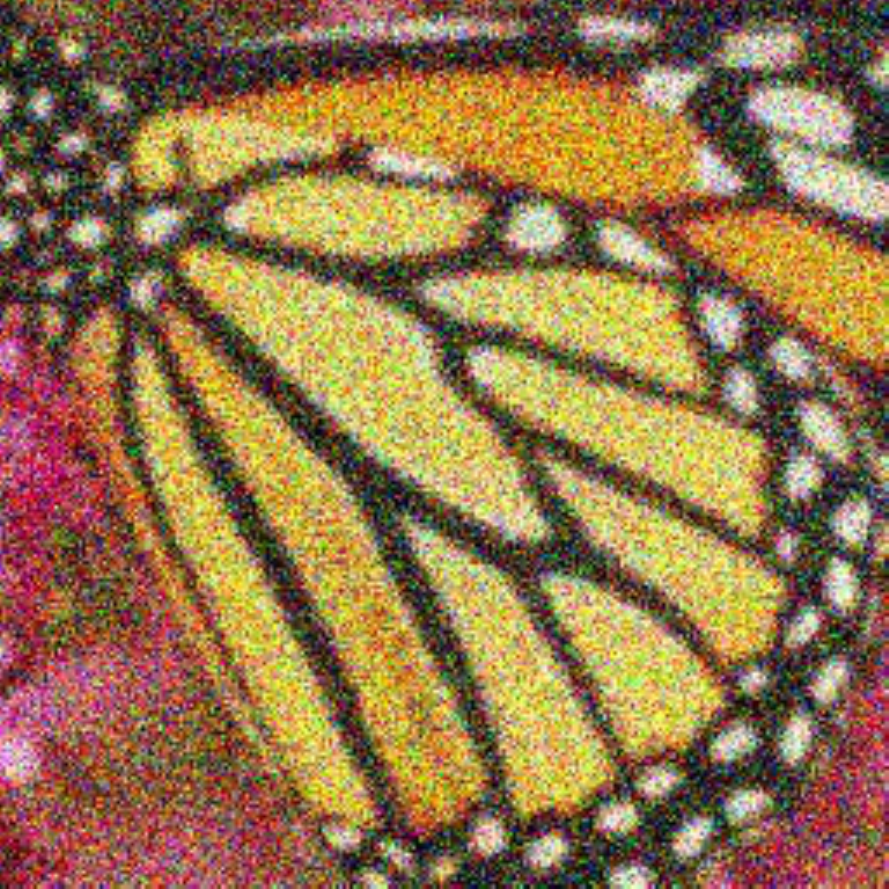} &
       \includegraphics[width=0.08\linewidth]{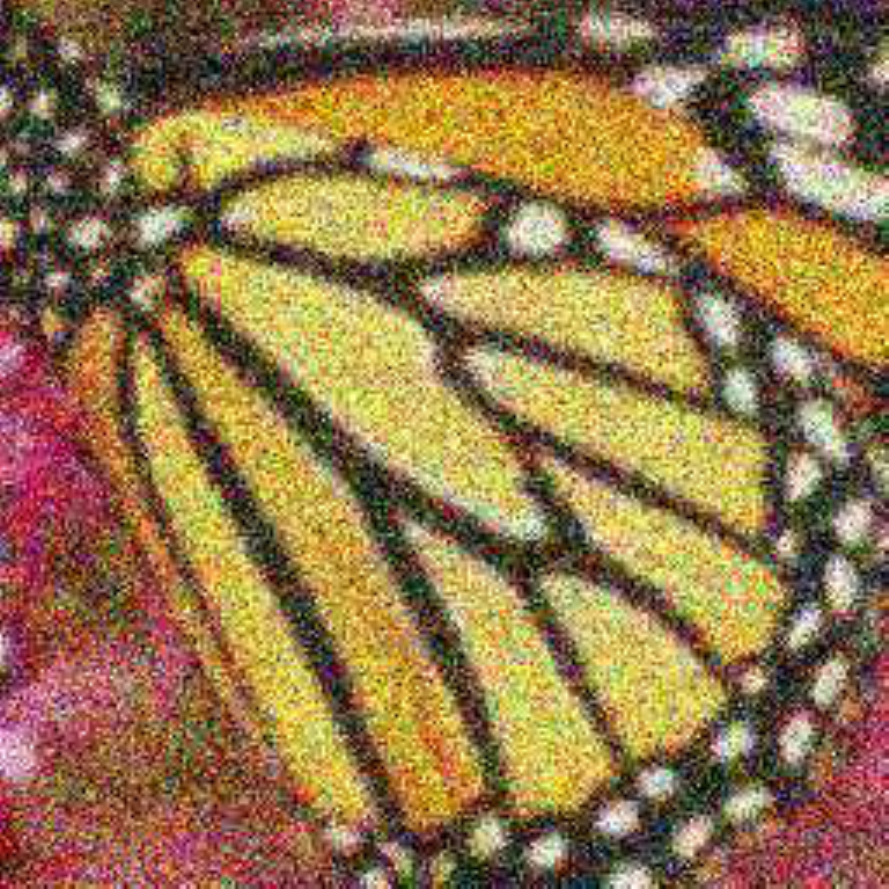} &
       \includegraphics[width=0.08\linewidth]{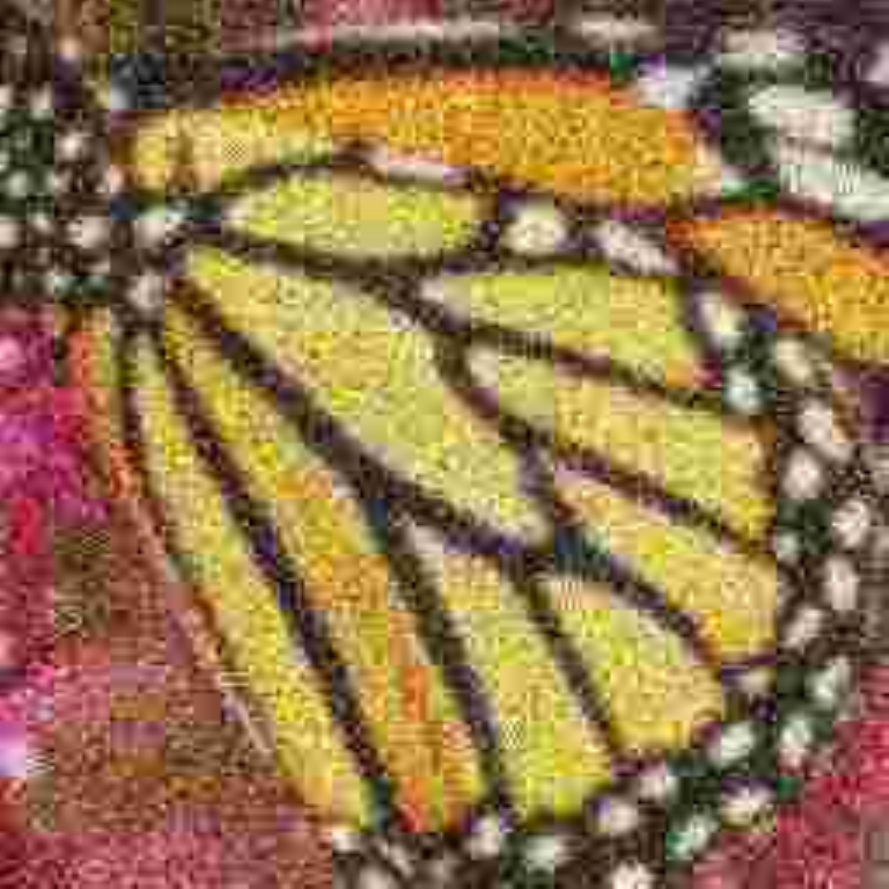} &
       \includegraphics[width=0.08\linewidth]{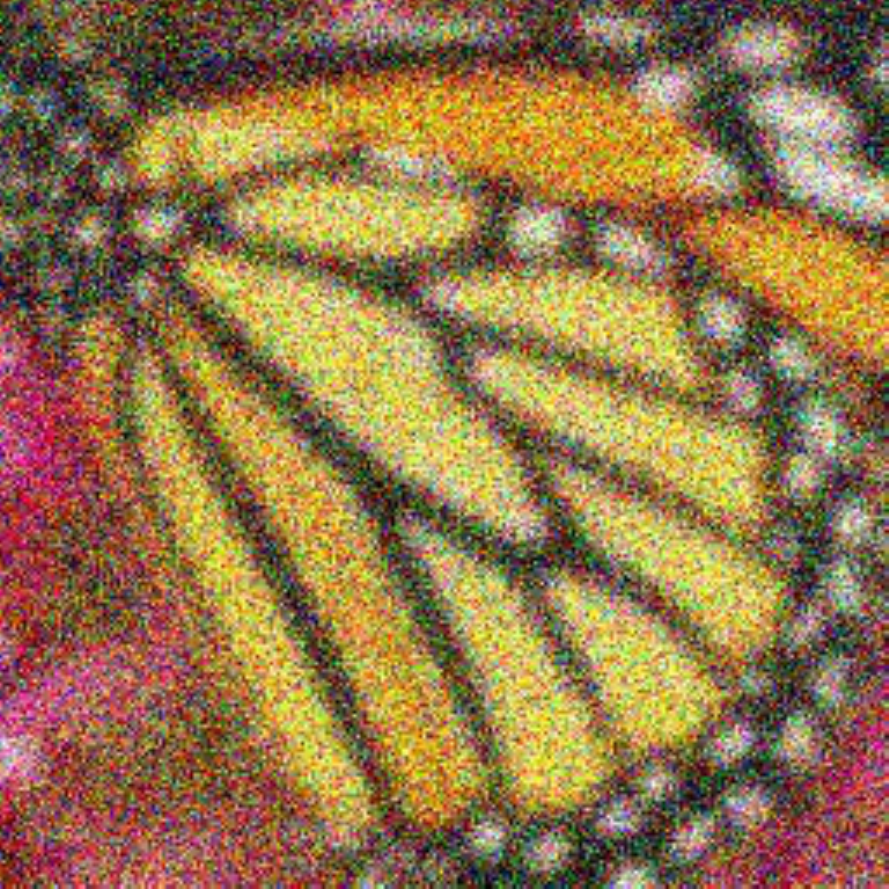} &
       \includegraphics[width=0.08\linewidth]{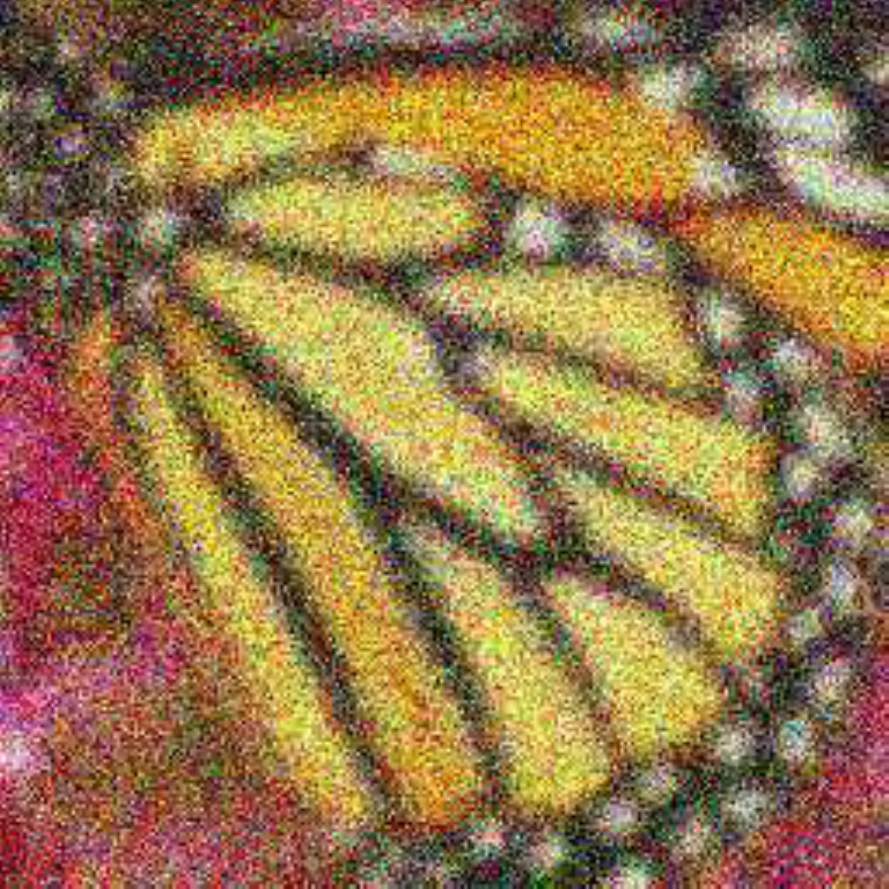} &
       \includegraphics[width=0.08\linewidth]{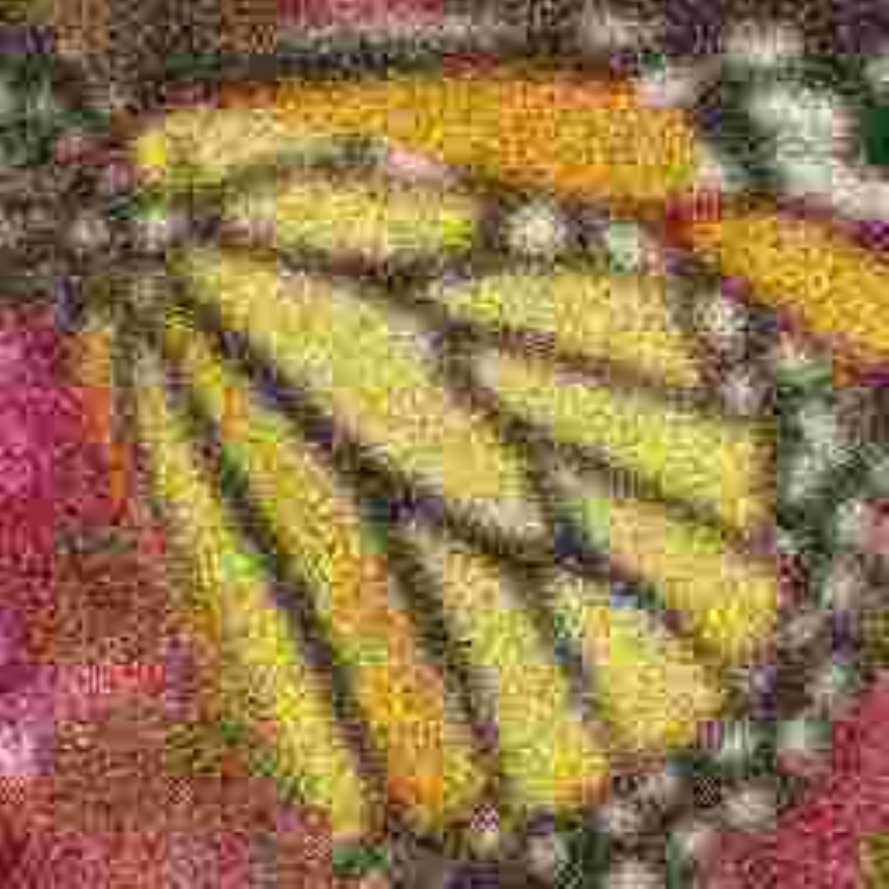} \\
&Grand Truth       &
       \includegraphics[width=0.08\linewidth]{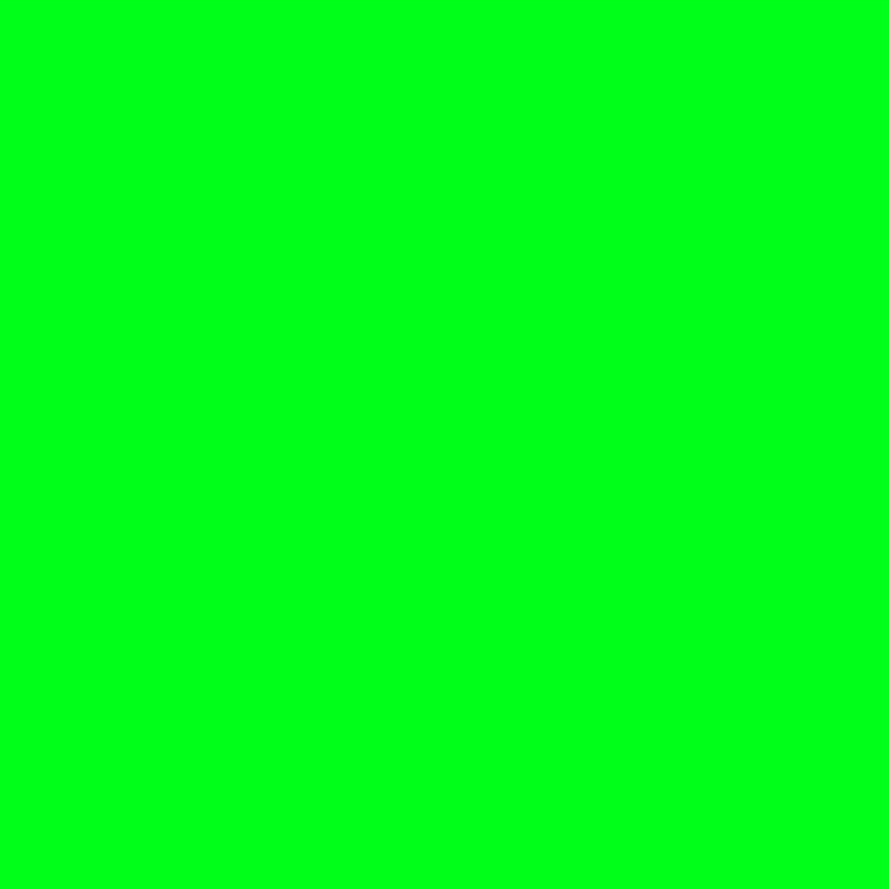} &
       \includegraphics[width=0.08\linewidth]{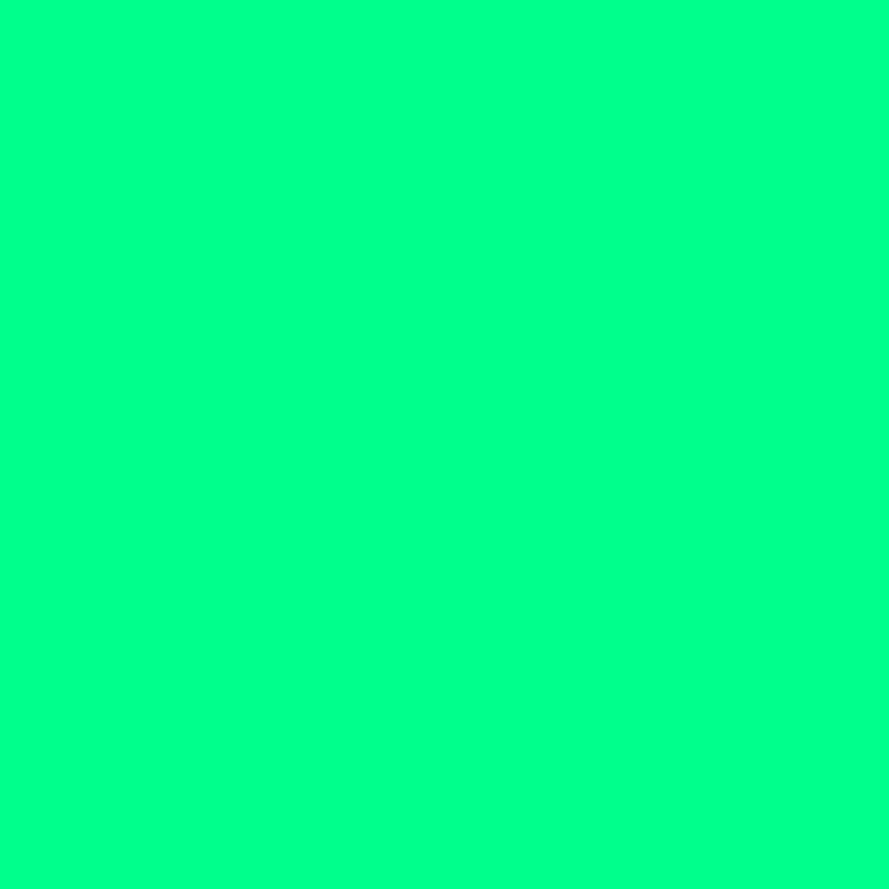} &
       \includegraphics[width=0.08\linewidth]{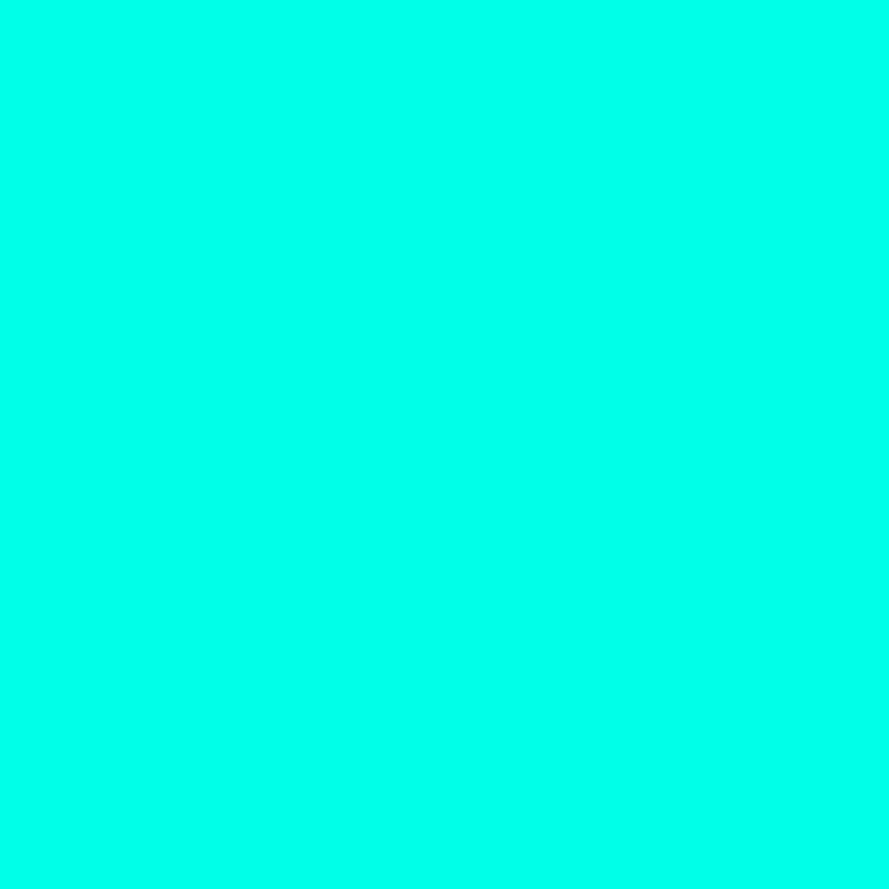} &
       \includegraphics[width=0.08\linewidth]{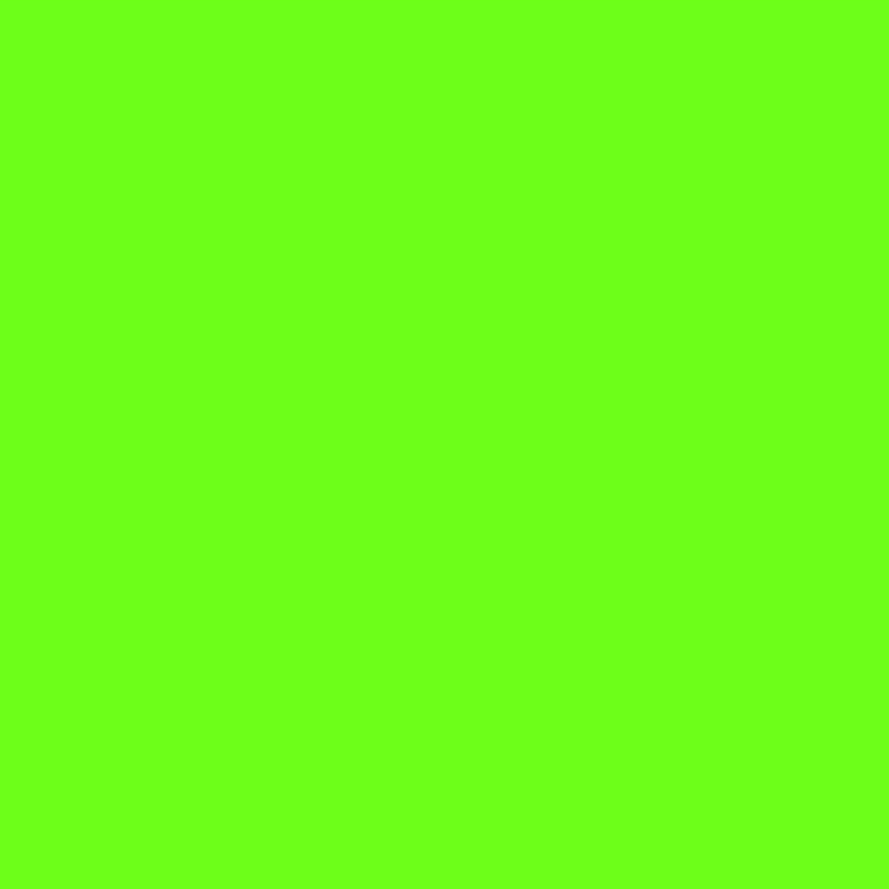} &
       \includegraphics[width=0.08\linewidth]{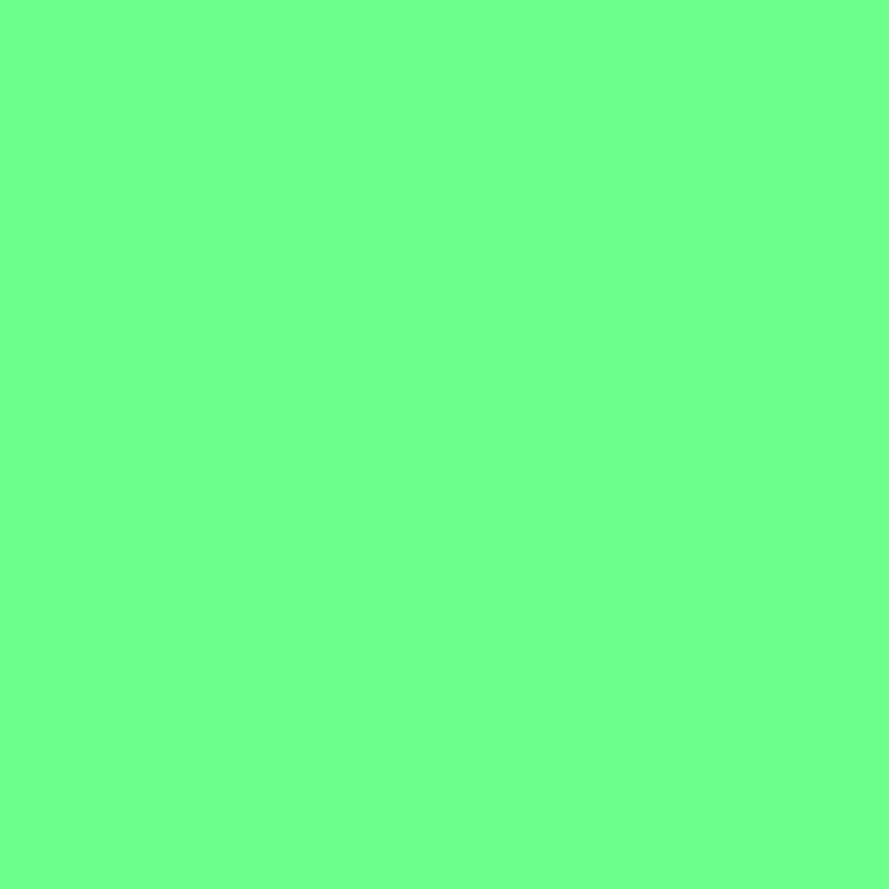} &
       \includegraphics[width=0.08\linewidth]{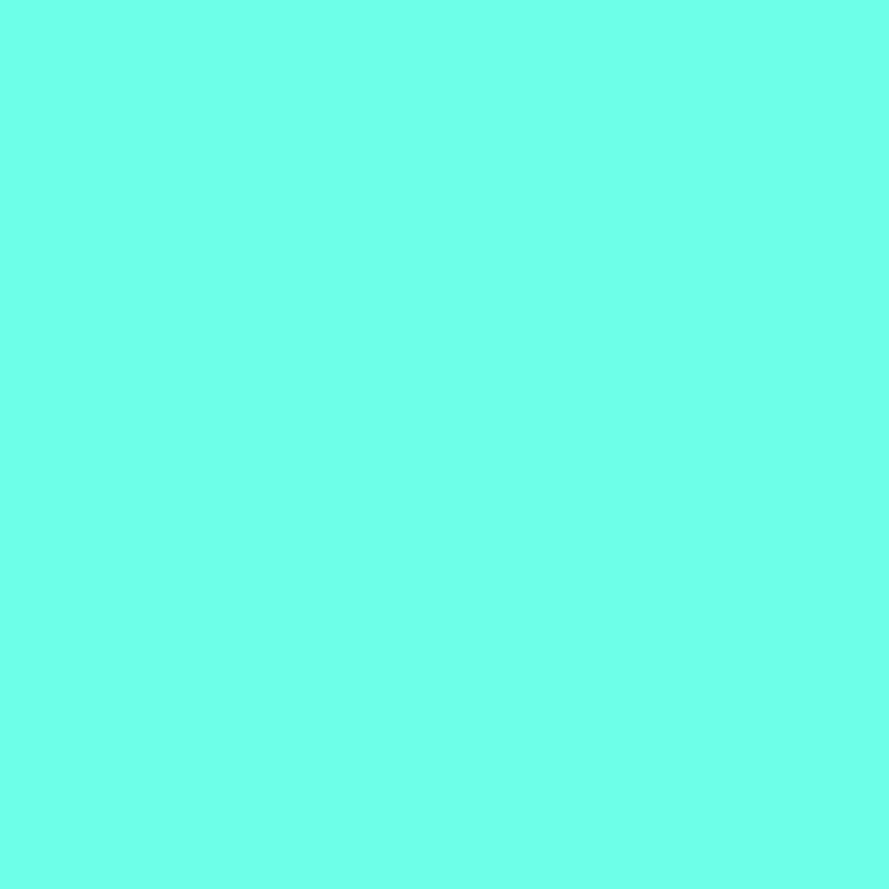} &
       \includegraphics[width=0.08\linewidth]{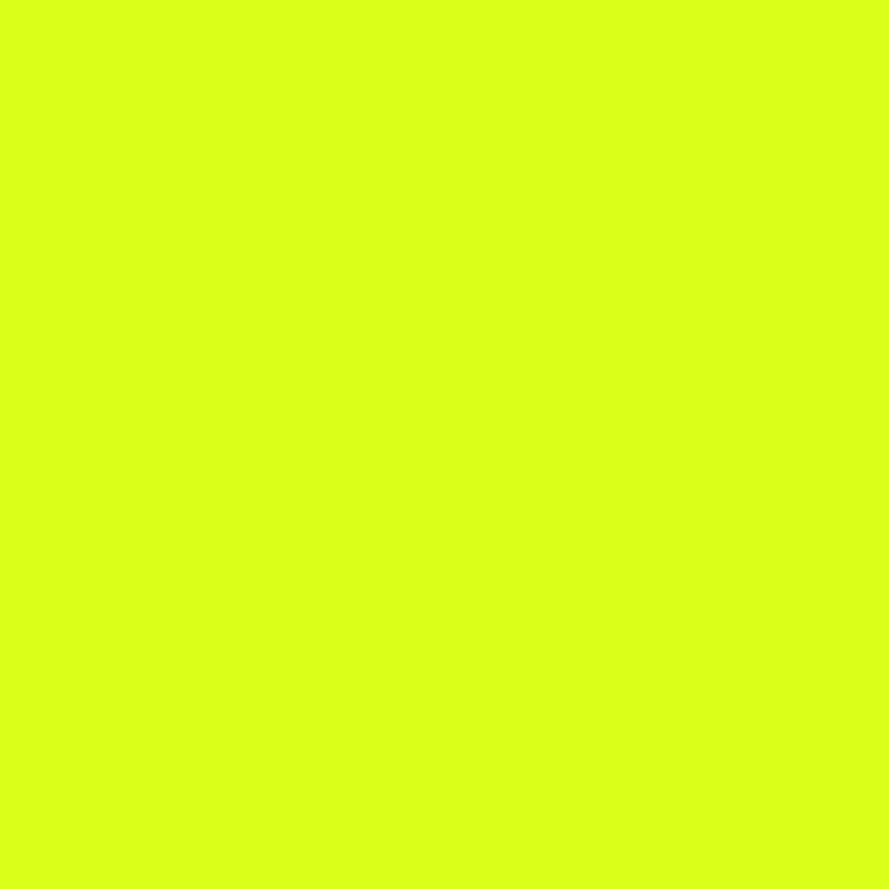} &
       \includegraphics[width=0.08\linewidth]{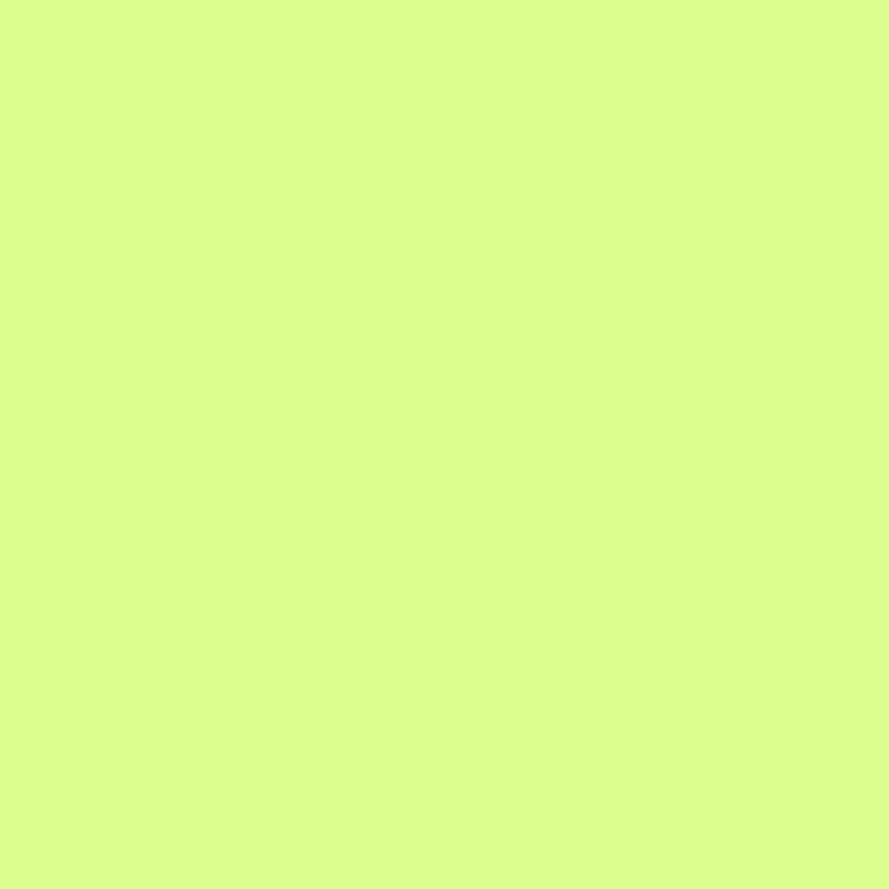} &
       \includegraphics[width=0.08\linewidth]{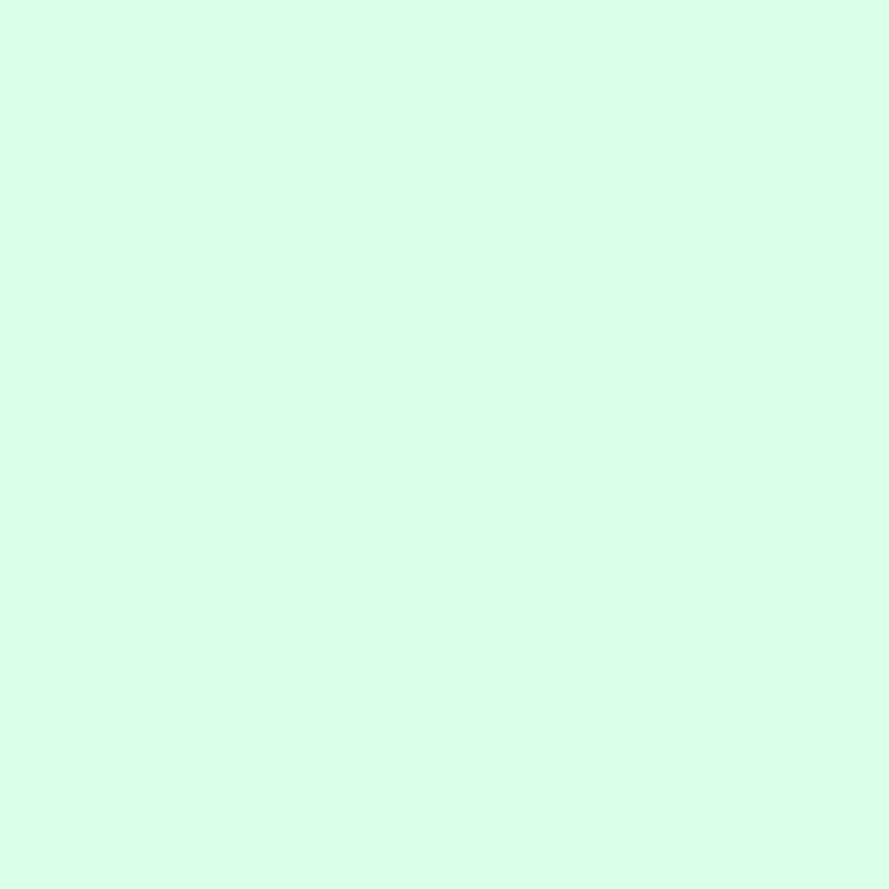} \\
&Estimated       &
        \includegraphics[width=0.08\linewidth]{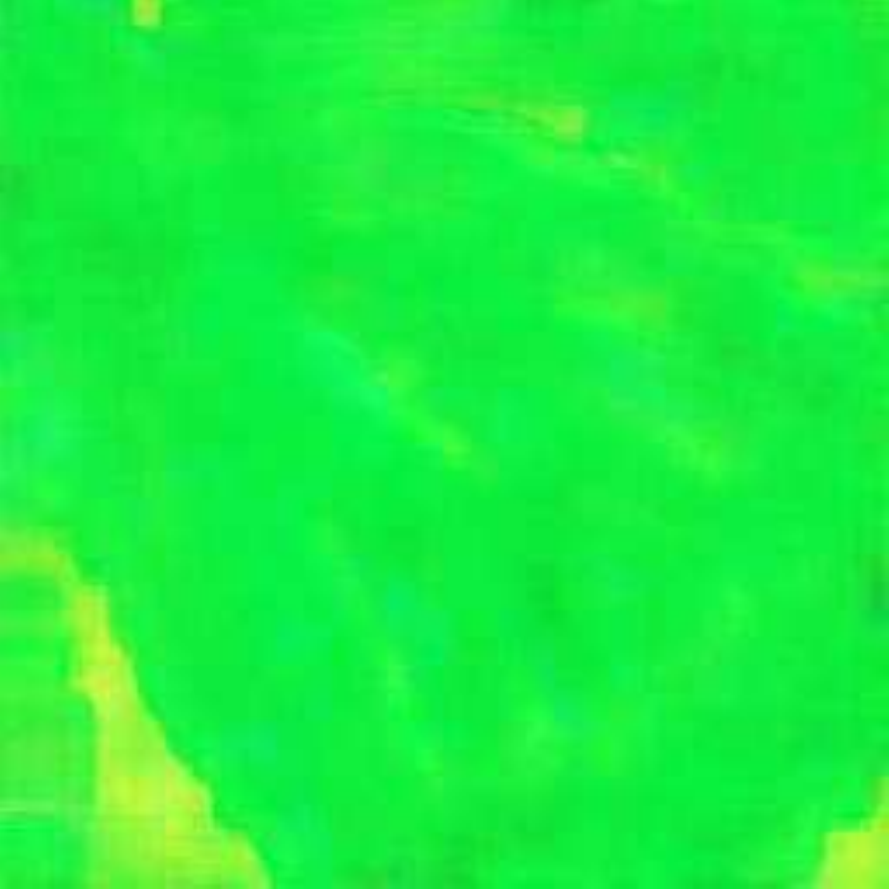} &
        \includegraphics[width=0.08\linewidth]{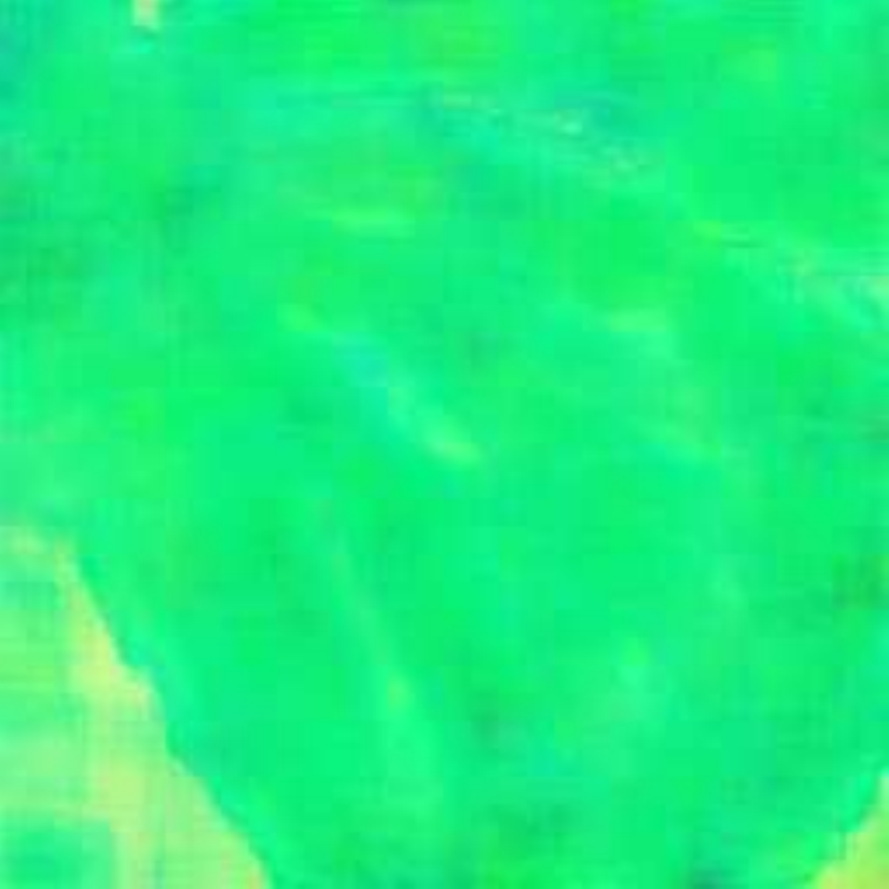} &
        \includegraphics[width=0.08\linewidth]{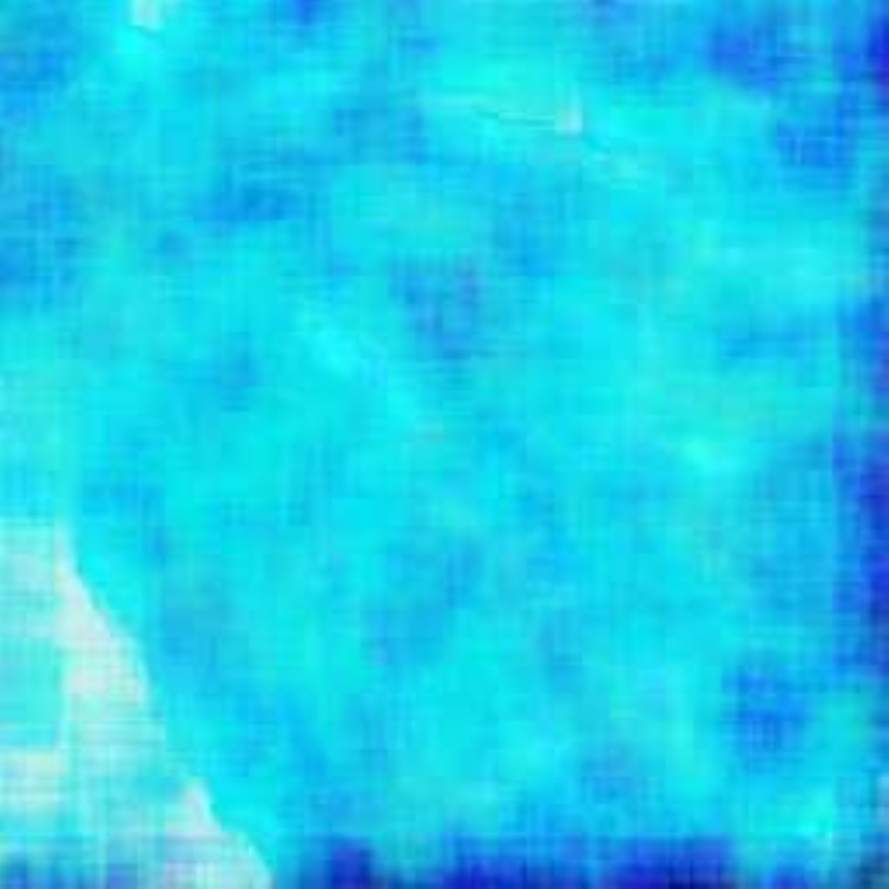} &
        \includegraphics[width=0.08\linewidth]{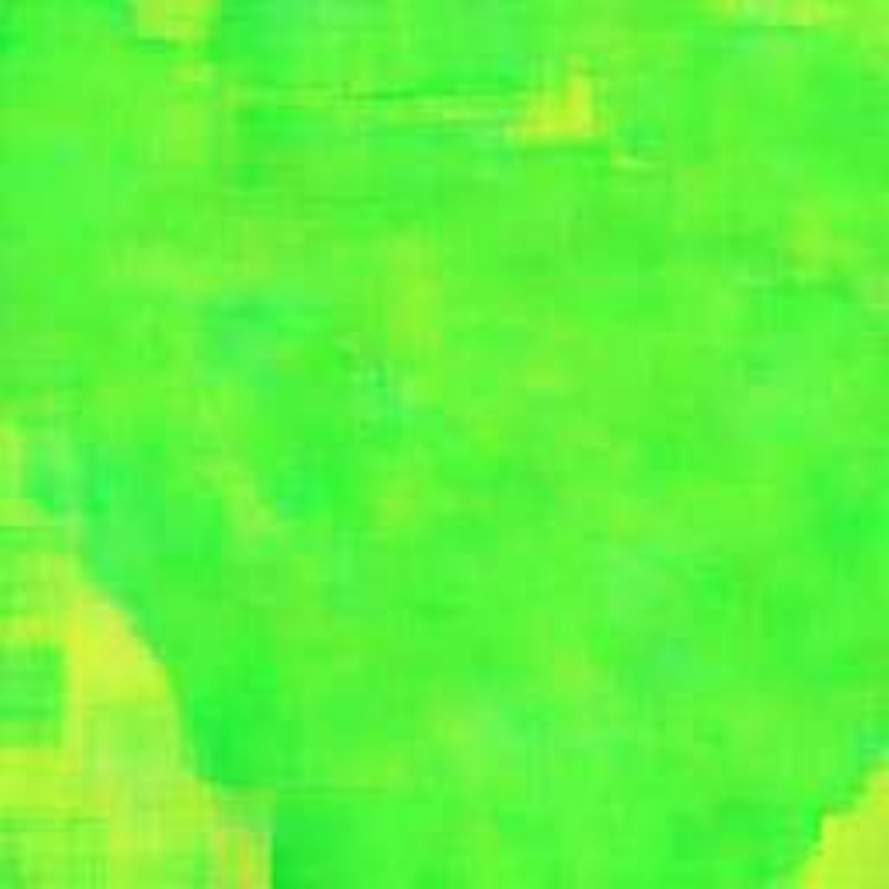} &
        \includegraphics[width=0.08\linewidth]{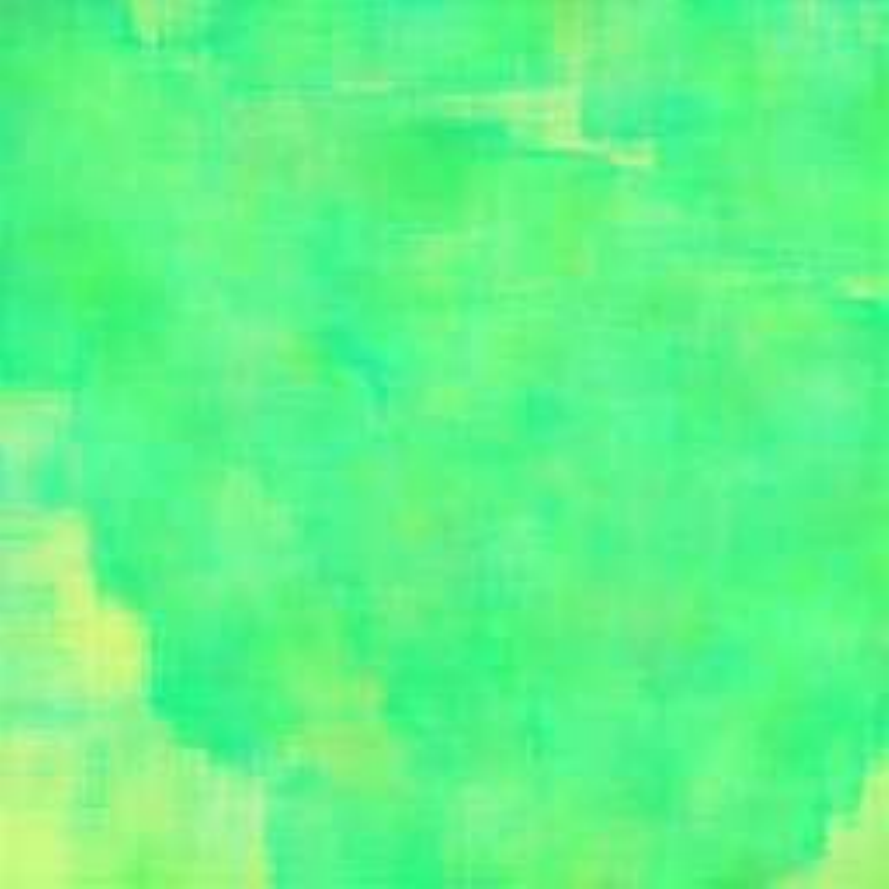} &
        \includegraphics[width=0.08\linewidth]{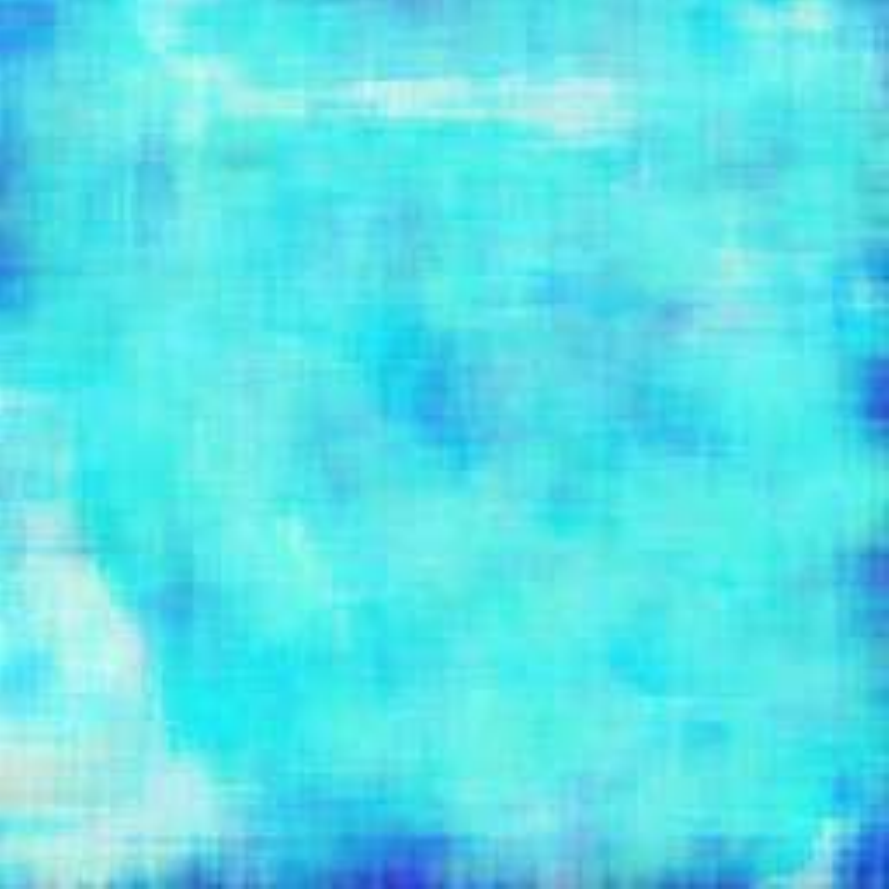} &
        \includegraphics[width=0.08\linewidth]{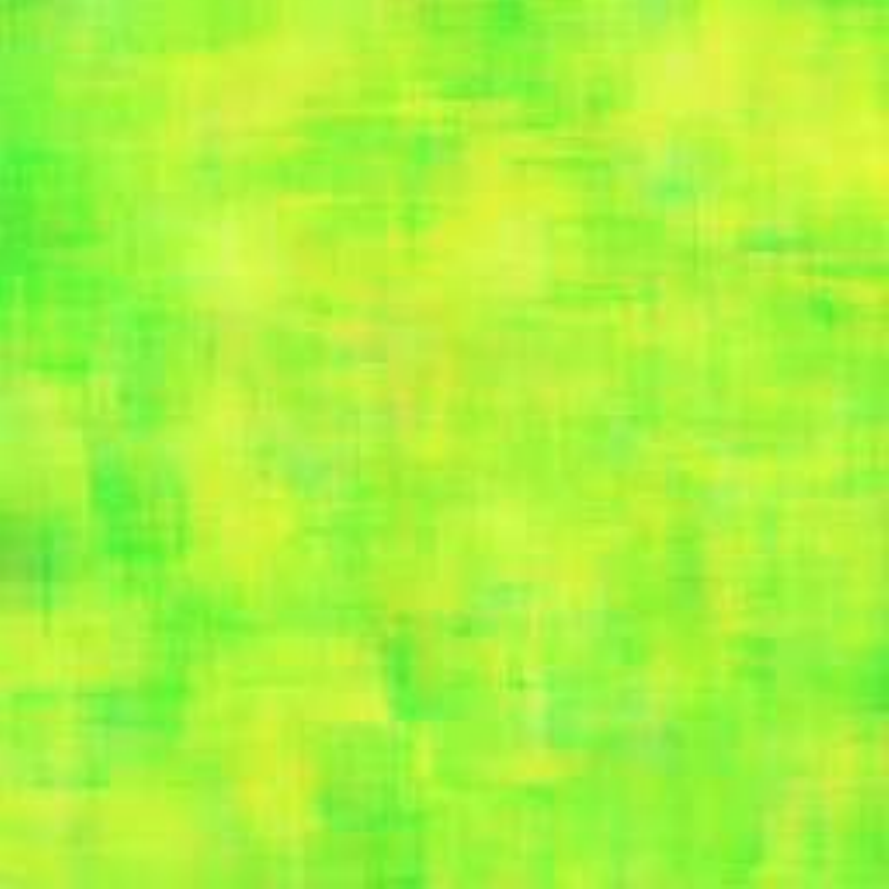} &
        \includegraphics[width=0.08\linewidth]{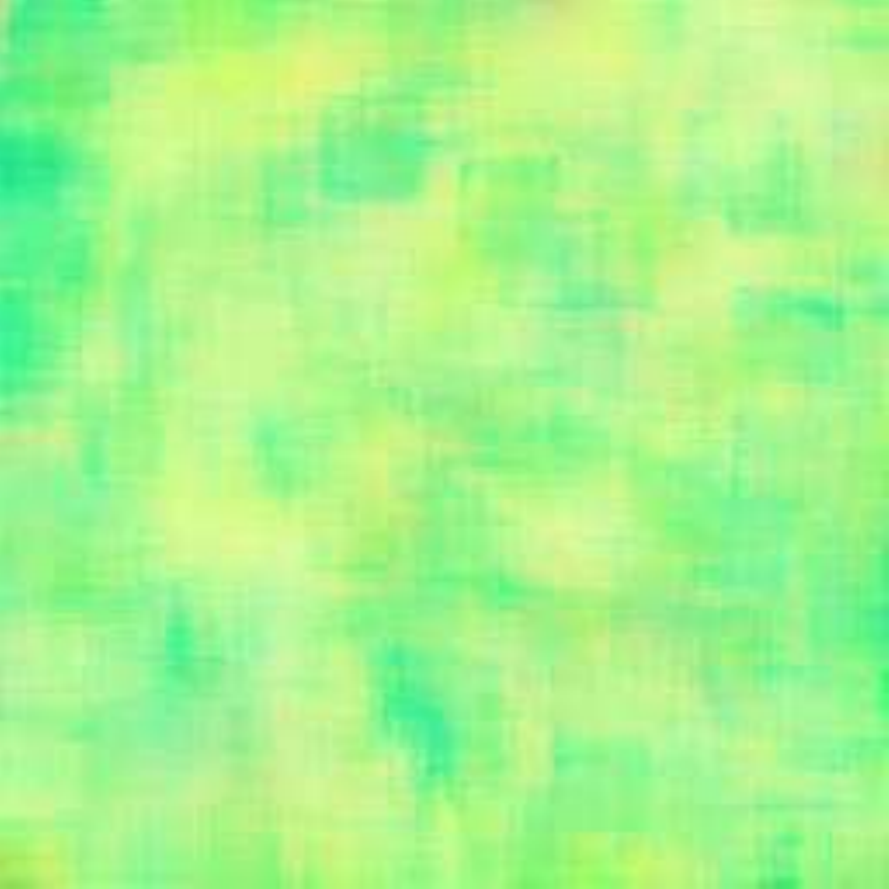} &
        \includegraphics[width=0.08\linewidth]{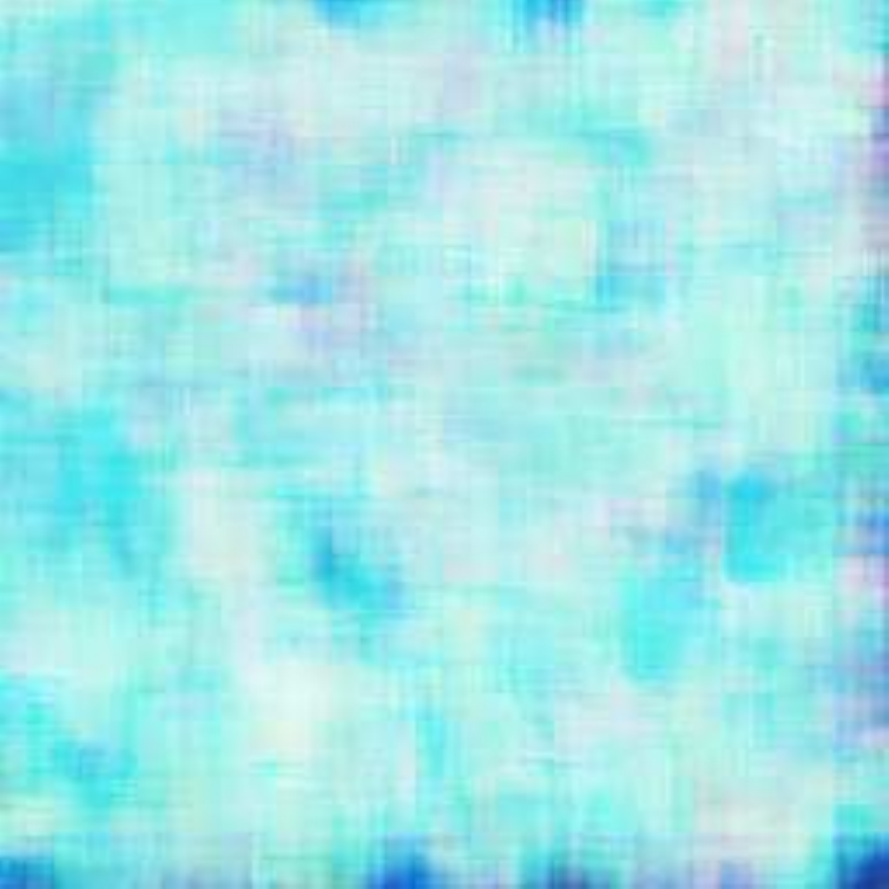} \\ \hline
     \end{tabular}
    \end{center}
  \caption{Examples of estimated degradation attributes for different degradation parameters.}
  \label{fig:degradations}
\end{figure*}

First, the accuracy of the degradation property estimation is evaluated using the Set5 dataset.
Figure \ref{fig:degradations} shows some examples of degradation estimation.
The degradation properties are represented as RGB components, wherein the red component indicates the blurring level, the green component shows  the AWGN level, and the blue component corresponds to the JPEG compression degradation.
Overall, the degradation parameters are estimated well.

Table \ref{tab:degradation_metrics} shows the root mean-squared error (RMSE) of the estimated degradation attributes for various compositional degradation parameters. Each cell shows the RMSE of the blurring level, noise level, and JPEG blocking level from top to bottom.
The noise levels and JPEG qualities are estimated well with errors of less than 0.2.
The blurring levels are estimated with errors of less than 0.3.

\newcommand{\specialcell}[2][c]{%
  \begin{tabular}[#1]{@{}c@{}}#2\end{tabular}}

\begin{table*}[t!]
  \caption{RMSE of Estimated Degradation Attributes for Different Degradation Parameters (Set5).}
  \label{tab:degradation_metrics}
    \begin{center}
      \begin{tabular}{|c|c|c|c|c|c|c|c|c|c|} \hline
          & \multicolumn{3}{|c|}{$\sigma=0$} & \multicolumn{3}{|c|}{$\sigma=1.5$} & \multicolumn{3}{|c|}{$\sigma=3.0$} \\ \hline
          & $q=100$ & $q=50$ & $q=10$ & $q=100$ & $q=50$ & $q=10$ & $q=100$ & $q=50$ & $q=10$ \\ \hline
      $\lambda=0$ &
      \specialcell{$0.240$\\$0.073$\\$0.131$} &
      \specialcell{$0.279$\\$0.120$\\$0.085$} &
      \specialcell{$0.348$\\$0.248$\\$0.041$} &
      \specialcell{$0.142$\\$0.009$\\$0.081$} &
      \specialcell{$0.161$\\$0.039$\\$0.084$} &
      \specialcell{$0.173$\\$0.180$\\$0.036$} &
      \specialcell{$0.071$\\$0.013$\\$0.077$} &
      \specialcell{$0.080$\\$0.034$\\$0.083$} &
      \specialcell{$0.129$\\$0.147$\\$0.032$}

      \\ \hline
      $\lambda=25$ &
      \specialcell{$0.291$\\$0.033$\\$0.081$} &
      \specialcell{$0.298$\\$0.050$\\$0.050$} &
      \specialcell{$0.323$\\$0.123$\\$0.028$} &
      \specialcell{$0.136$\\$0.022$\\$0.074$} &
      \specialcell{$0.145$\\$0.029$\\$0.044$} &
      \specialcell{$0.144$\\$0.114$\\$0.022$} &
      \specialcell{$0.161$\\$0.019$\\$0.073$} &
      \specialcell{$0.164$\\$0.024$\\$0.039$} &
      \specialcell{$0.206$\\$0.099$\\$0.019$}

      \\ \hline
      $\lambda=55$ &
      \specialcell{$0.330$\\$0.052$\\$0.152$} &
      \specialcell{$0.340$\\$0.057$\\$0.042$} &
      \specialcell{$0.346$\\$0.161$\\$0.024$} &
      \specialcell{$0.156$\\$0.047$\\$0.145$} &
      \specialcell{$0.160$\\$0.048$\\$0.035$} &
      \specialcell{$0.152$\\$0.126$\\$0.020$} &
      \specialcell{$0.222$\\$0.045$\\$0.150$} &
      \specialcell{$0.237$\\$0.046$\\$0.035$} &
      \specialcell{$0.274$\\$0.108$\\$0.021$}
      \\ \hline
      \end{tabular}
    \end{center}
\end{table*}

\begin{figure}[!t]
\centering
\includegraphics[width=1.0\linewidth]{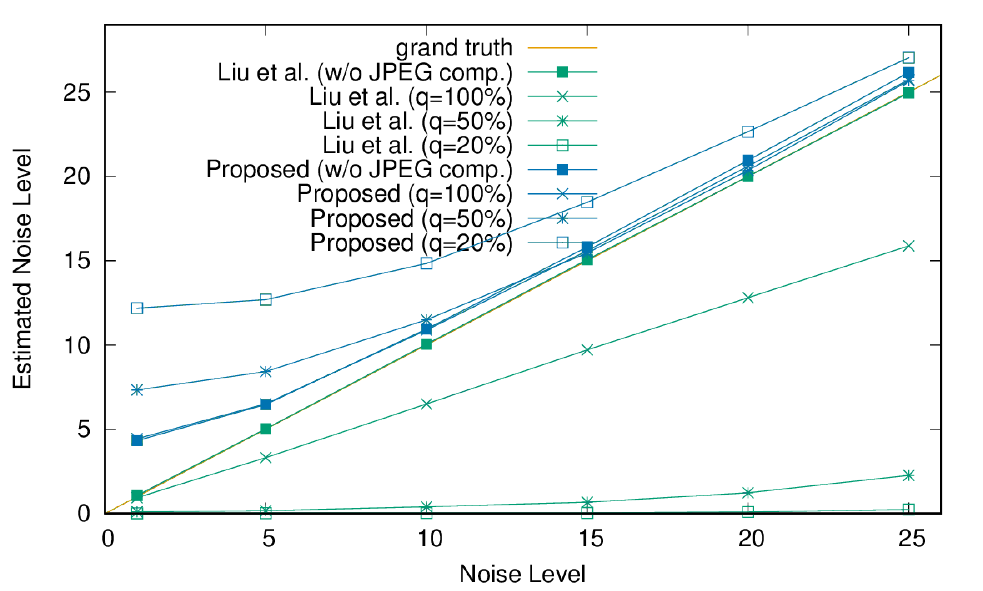}
\caption{Noise level estimation with a method of \cite{liu2013single} on compositional degradation images}
\label{fig:liu_uchida}
\end{figure}

Figure \ref{fig:liu_uchida} shows a comparison of the estimation performance of the method of Liu \etal \cite{liu2013single}. Their method precisely estimated the noise level of a noised image without JPEG compression. However, it fails to estimate the noise level of noised images with JPEG compression ($q=100$).
On the contrary, our method is capable of estimating the noise level with JPEG compression ($q=100$).
Note that JPEG compression with $q=100$ does not mean an uncompressed. Even if $q=100$, the JPEG image included compression distortion.

\begin{figure}[!t]
\centering
\includegraphics[width=1.0\linewidth]{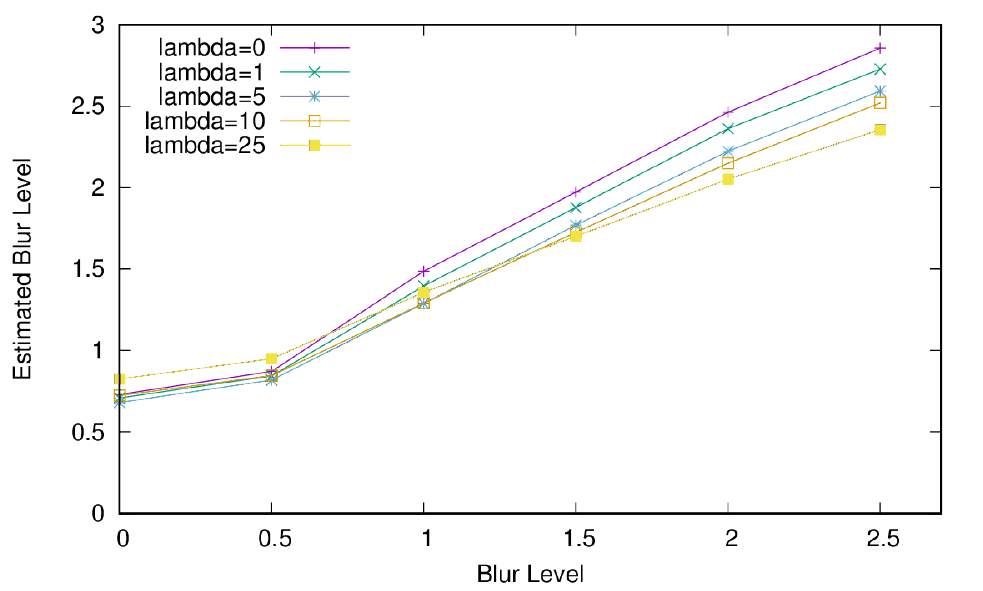}
\caption{Blur-level estimation for each noise level.}
\label{fig:uchida_blur}
\end{figure}

Figure \ref{fig:uchida_blur} demonstrates the performances on blur-level estimation under various noise levels.
Blur levels are successfully estimated regardless of the additive noises, except for the zero blur level.

\begin{figure}[!t]
\centering
\includegraphics[width=1.0\linewidth]{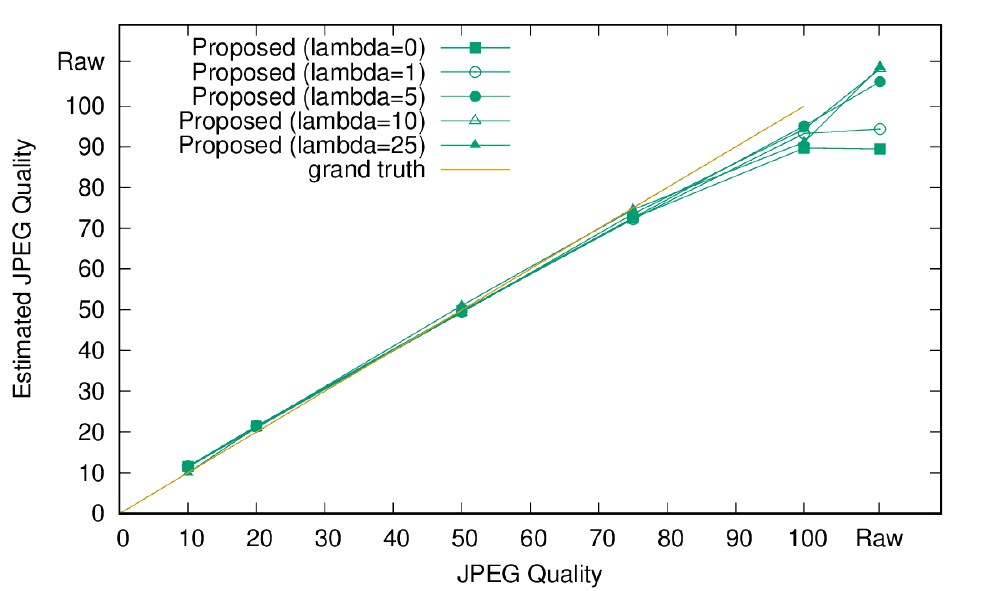}
\caption{JPEG quality factor estimation for each noise level.}
\label{fig:uchida_jpg}
\end{figure}

Figure \ref{fig:uchida_jpg} shows the results on JPEG quality estimation under various noise levels by the proposed algorithm.
The performance is almost constant for the noise levels because the noises are added before the JPEG compression.

\begin{figure*}[t!]
  \captionsetup{farskip=0pt}
    \begin{center}
  \subfloat[Original]{
       \includegraphics[width=0.24\linewidth]{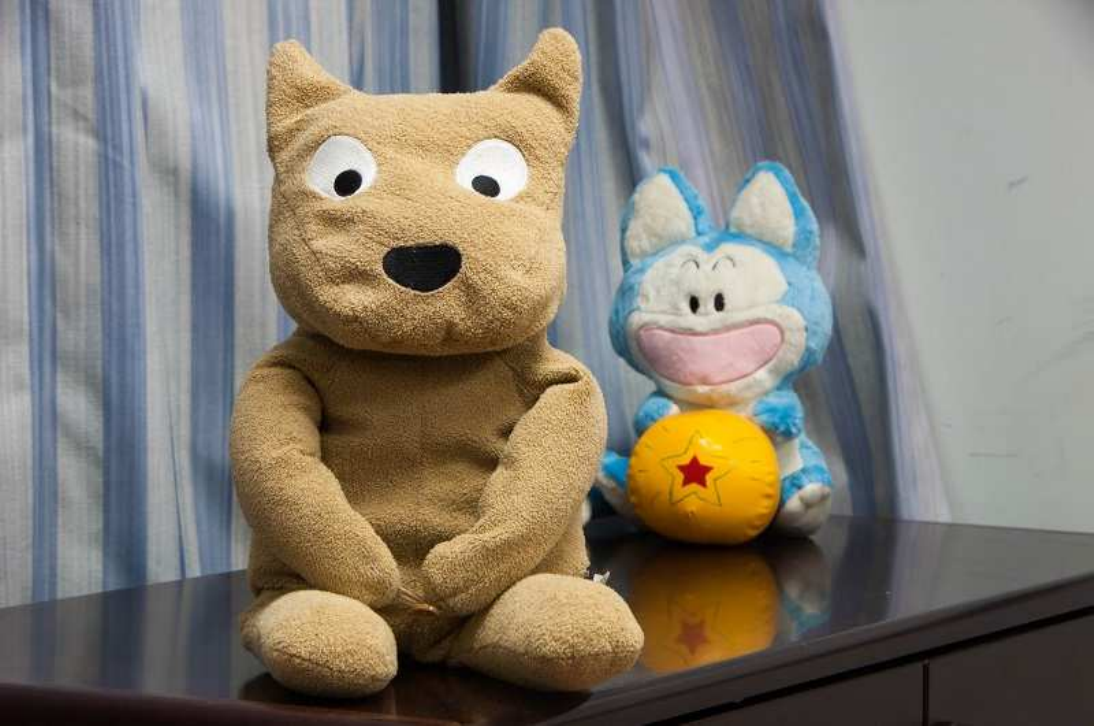}
    \label{shi_a}\hfill
  }
  \subfloat[Degraded with AWGN $\lambda=15$]{%
       \includegraphics[width=0.24\linewidth]{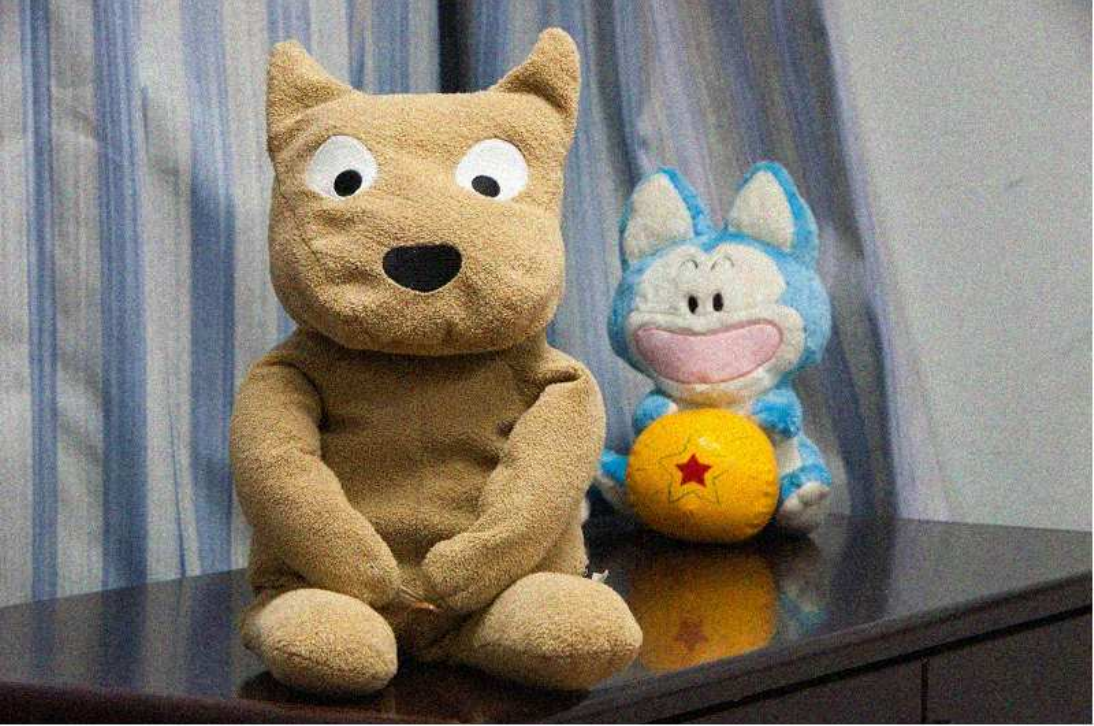}
    \label{shi_b}
    }
  \subfloat[Blur Map for Original (Shi \etal \cite{shi2015just})]{%
       \includegraphics[width=0.24\linewidth]{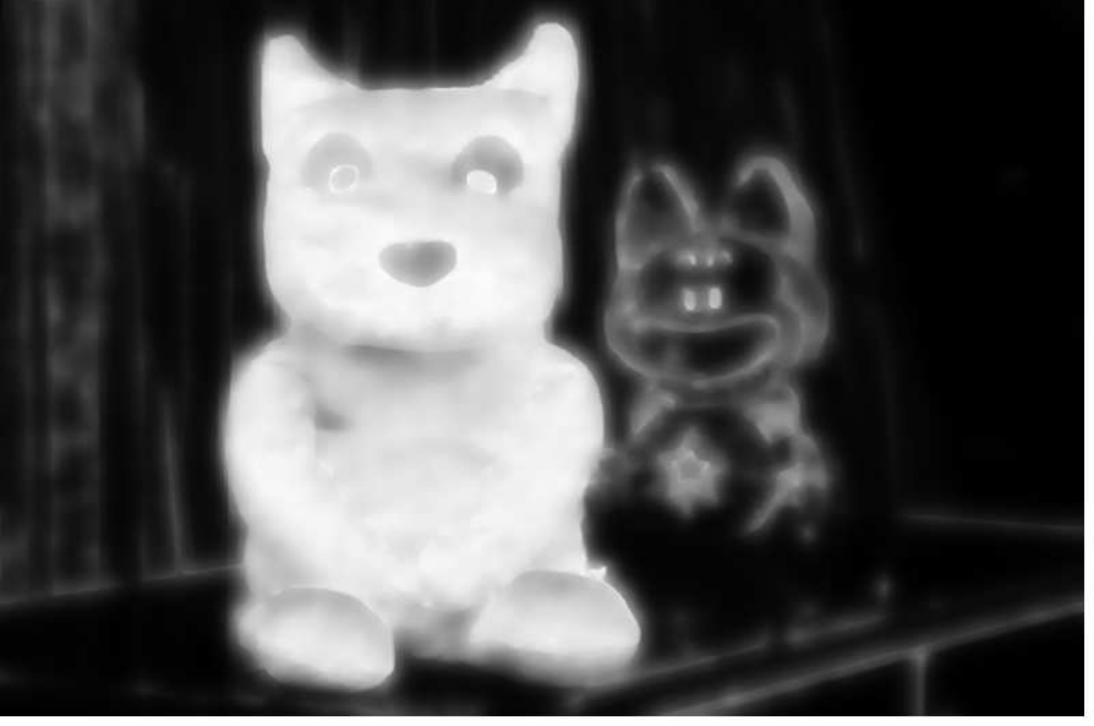}
    \label{shi_c}\hfill
    }
  \subfloat[Blur Map for the Degraded Image (Shi \etal \cite{shi2015just})]{%
       \includegraphics[width=0.24\linewidth]{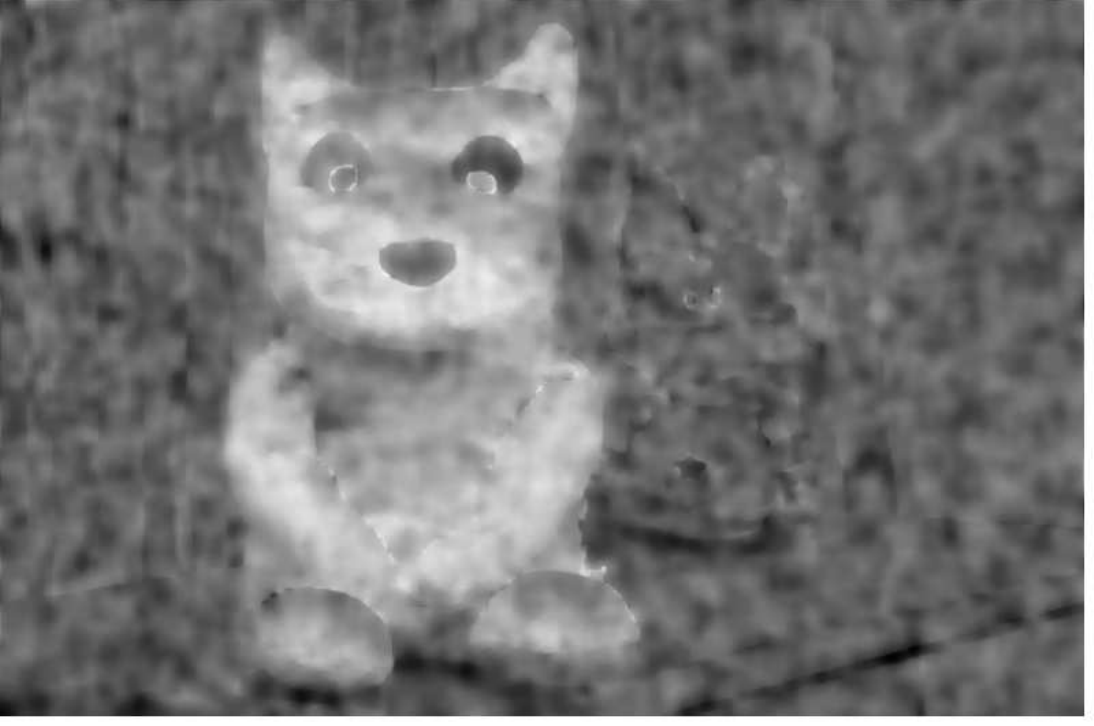}
    \label{shi_d}
    }\\
    \subfloat[Estimated Degradation for Original (proposed)]{%
         \includegraphics[width=0.24\linewidth]{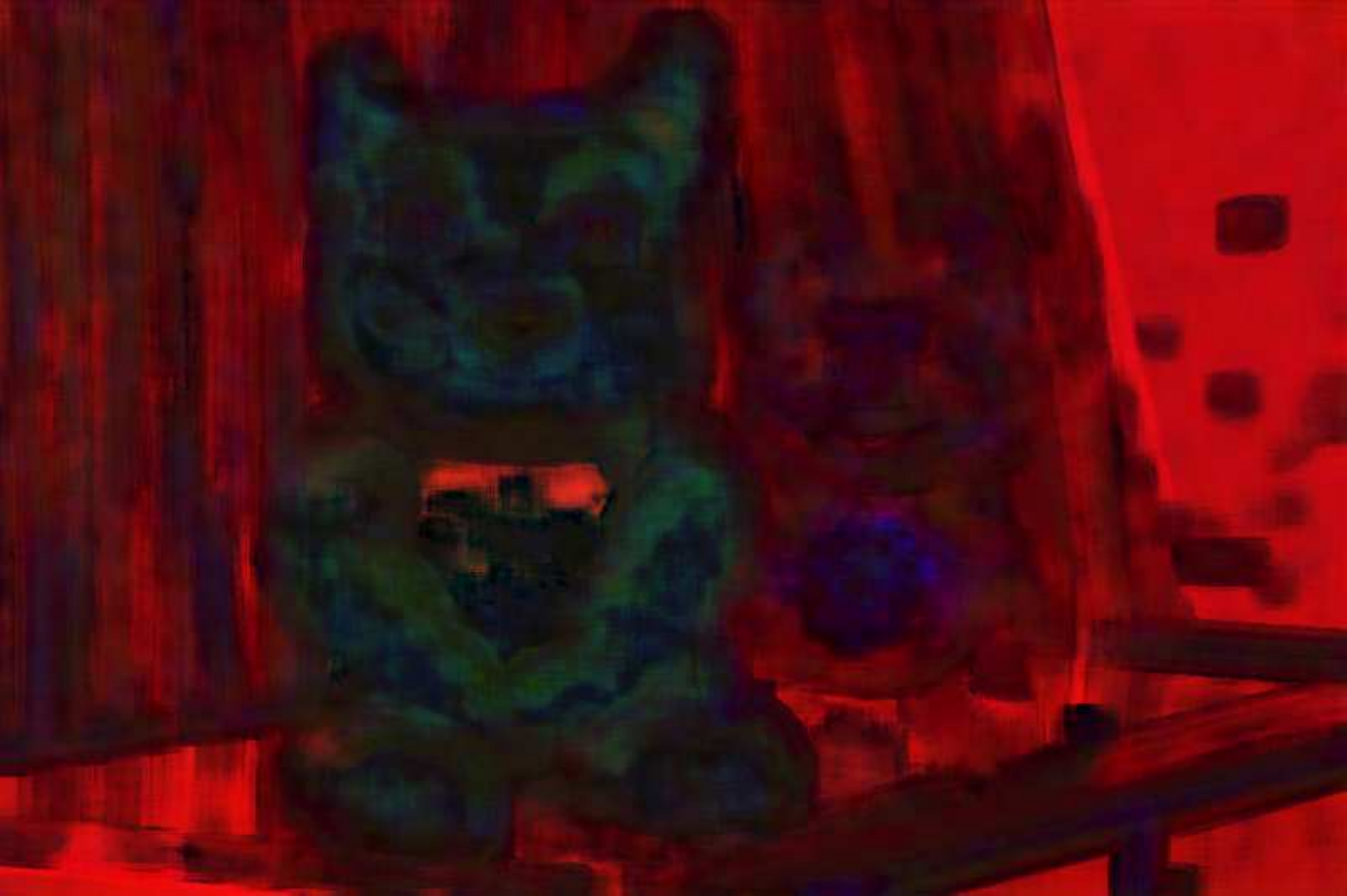}
      \label{shi_g}\hfill
      }
    \subfloat[Estimated Degradation for the Degraded Image (proposed)]{%
         \includegraphics[width=0.24\linewidth]{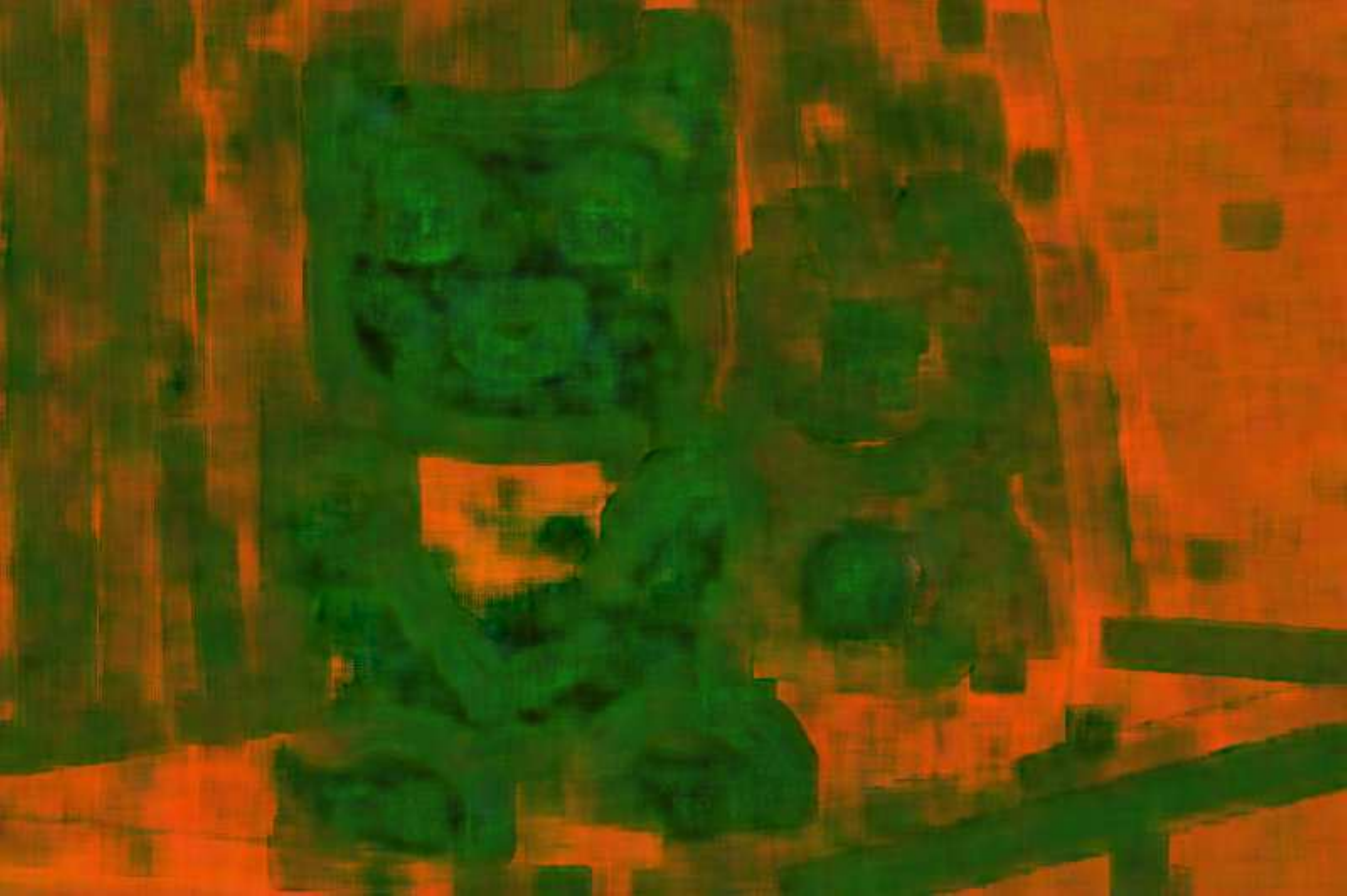}
      \label{shi_h}
      }
  \subfloat[Blur Map for Original (proposed)]{%
       \includegraphics[width=0.24\linewidth]{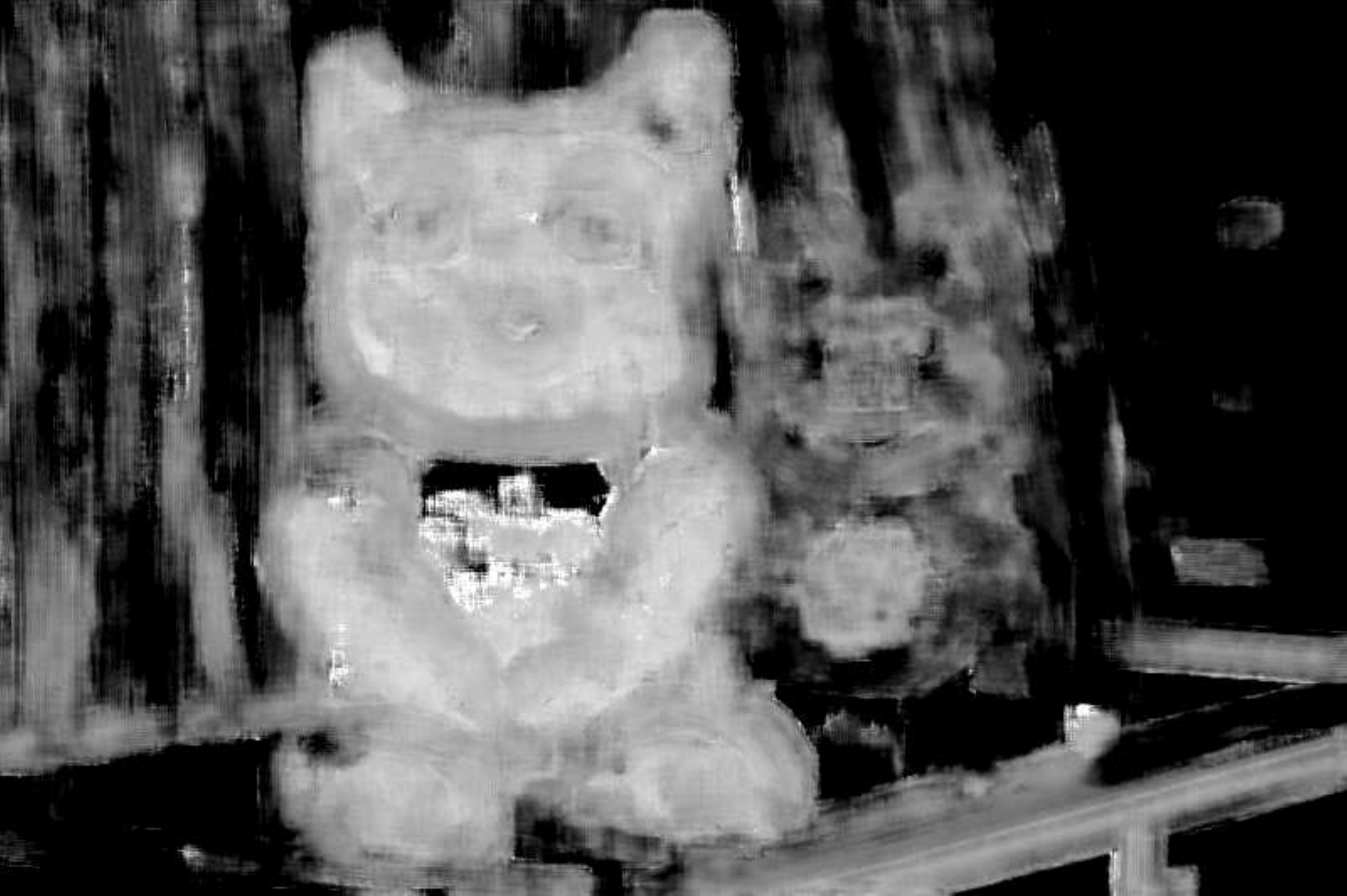}
    \label{shi_e}\hfill
    }
  \subfloat[Blur Map for the Degraded Image (proposed)]{%
       \includegraphics[width=0.24\linewidth]{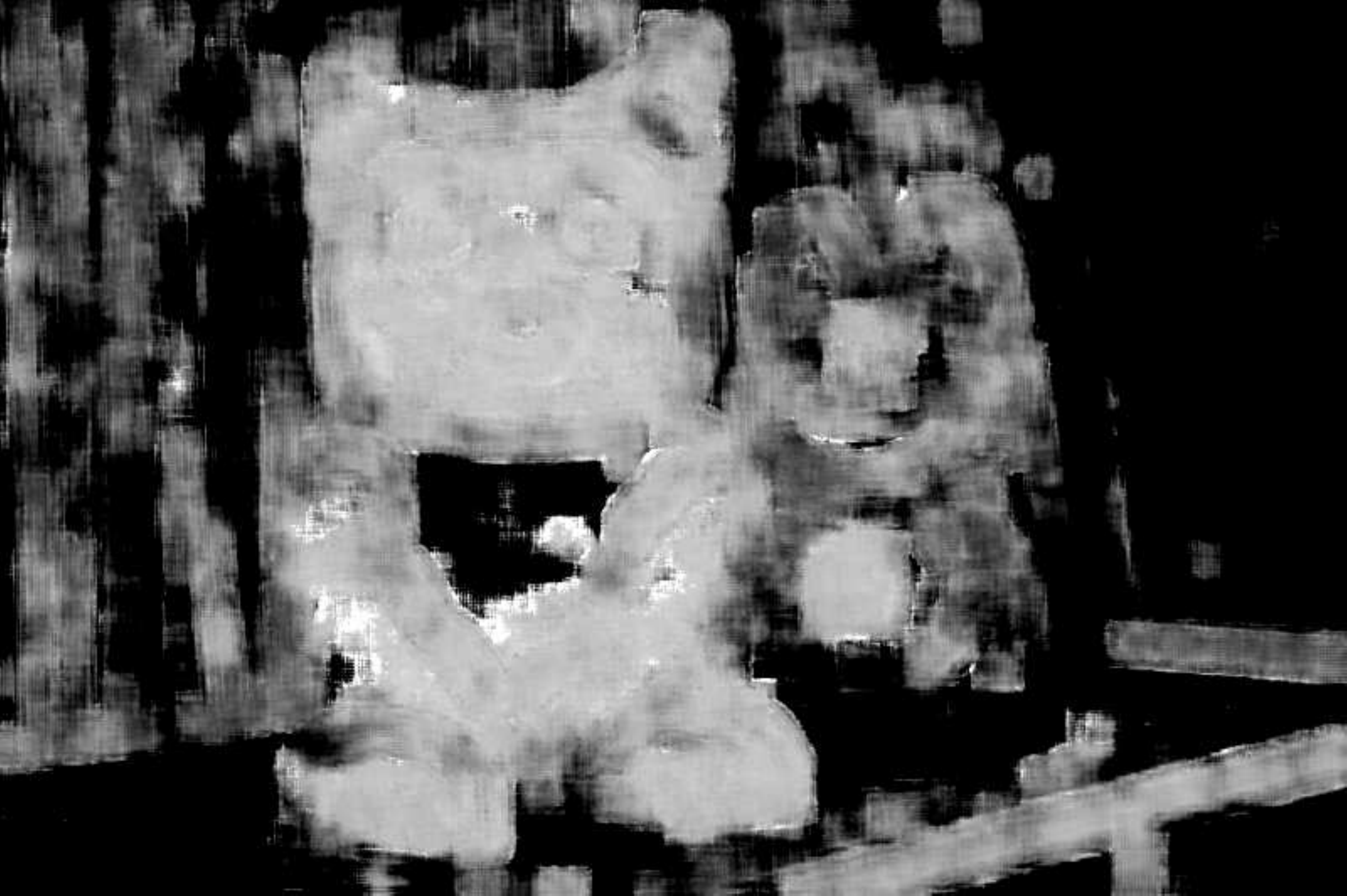}
    \label{shi_f}
    }
  \end{center}
    \caption{Estimated blur map for compositional degradation}
    \label{fig:shi}
\end{figure*}

Figure \ref{fig:shi} shows a comparison of blur-level estimation with the proposed algorithm and Shi \etal \cite{shi2015just}.
As shown in Fig. \ref{shi_c}, their method successfully estimated the blur level of the original image (Fig. \ref{shi_a}).
However, it failed to estimate the blur level of the noised image with $\lambda=15$ (Fig. \ref{shi_b}), as shown in Fig. \ref{shi_d}.
On the contrary, our method estimates the blur level of both the clean and noised images (Fig. \ref{shi_e} and \ref{shi_f}).
Figure \ref{shi_e} and \ref{shi_f} illustrate blur map channel of the estimated degradations.
These blur maps are adequately constant independently from the noise.

\subsubsection{Blind Restoration}

Restoration performance is measured using the integrated model presented in Fig. \ref{fig:flow_of_compositional_degradation}.
Table \ref{tab:restoration_metrics} shows the average PSNR of the restored images from the degraded Set5 dataset for each degradation parameter.
The restoration is performed by the proposed model and by using several existing methods\cite{zhang2017beyond, liu2013single, dabov2007image, shan2008high}.
As for the restoration by the existing methods,
first the degraded image is JPEG-deblocked by \cite{zhang2017beyond},
then nonblind restoration by \cite{dabov2007image} is applied referring to the noise level estimated by \cite{liu2013single},
and then deblurring by \cite{shan2008high} is applied.

\begin{figure*}[t!]
    \footnotesize
    \begin{center}
\hspace*{-10mm}
      \begin{tabular}{|l|l|ccc|ccc|ccc|} \hline
        &  & \multicolumn{3}{|c|}{$\sigma=0$} & \multicolumn{3}{|c|}{$\sigma=1.5$} & \multicolumn{3}{|c|}{$\sigma=3.0$} \\ \hline
        & & $q=100$ & $q=50$ & $q=10$ & $q=100$ & $q=50$ & $q=10$ & $q=100$ & $q=50$ & $q=10$ \\ \hline
  \multirow{3}{*}{$\lambda=0$} &
Degraded  &
      \includegraphics[width=0.08\linewidth]{images/butterfly/degraded_000_00_100_pdf.pdf} &
      \includegraphics[width=0.08\linewidth]{images/butterfly/degraded_000_00_50_pdf.pdf} &
      \includegraphics[width=0.08\linewidth]{images/butterfly/degraded_000_00_10_pdf.pdf} &
      \includegraphics[width=0.08\linewidth]{images/butterfly/degraded_150_00_100_pdf.pdf} &
      \includegraphics[width=0.08\linewidth]{images/butterfly/degraded_150_00_50_pdf.pdf} &
      \includegraphics[width=0.08\linewidth]{images/butterfly/degraded_150_00_10_pdf.pdf} &
      \includegraphics[width=0.08\linewidth]{images/butterfly/degraded_300_00_100_pdf.pdf} &
      \includegraphics[width=0.08\linewidth]{images/butterfly/degraded_300_00_50_pdf.pdf} &
      \includegraphics[width=0.08\linewidth]{images/butterfly/degraded_300_00_10_pdf.pdf} \\
&Restored       &
      \includegraphics[width=0.08\linewidth]{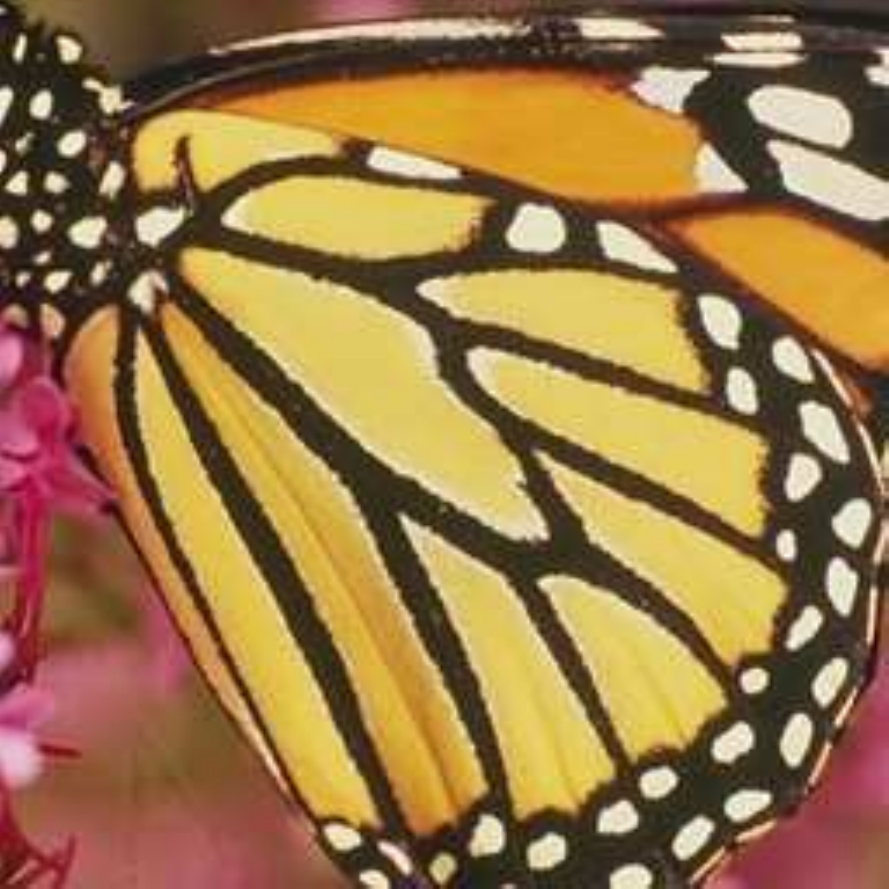} &
      \includegraphics[width=0.08\linewidth]{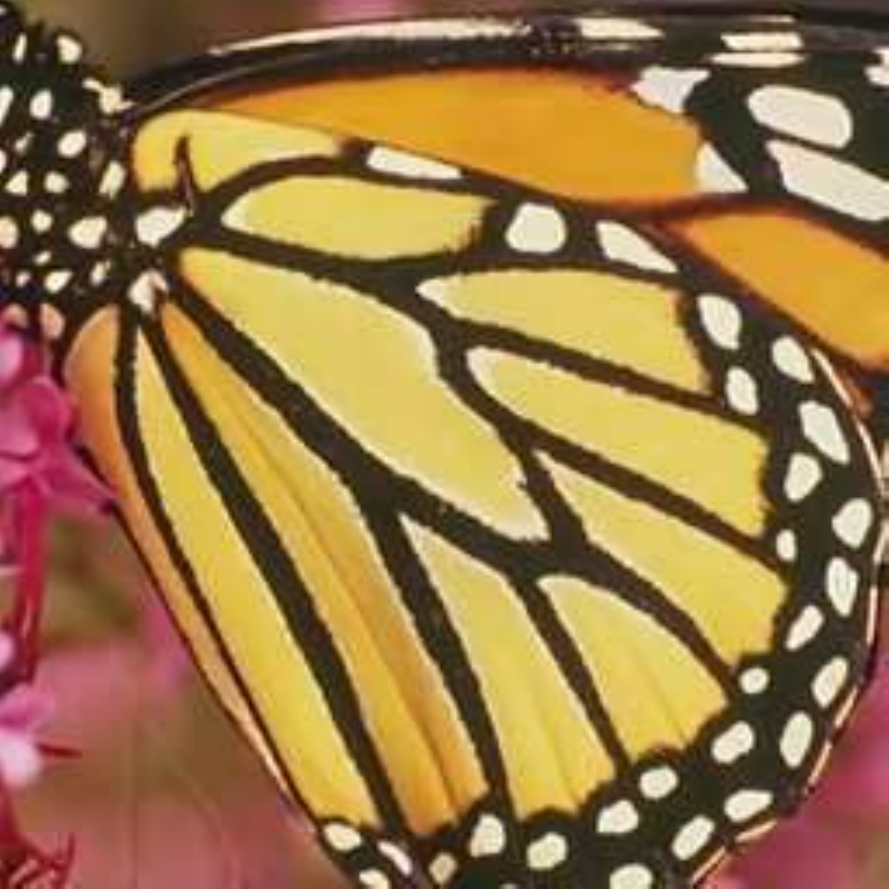} &
      \includegraphics[width=0.08\linewidth]{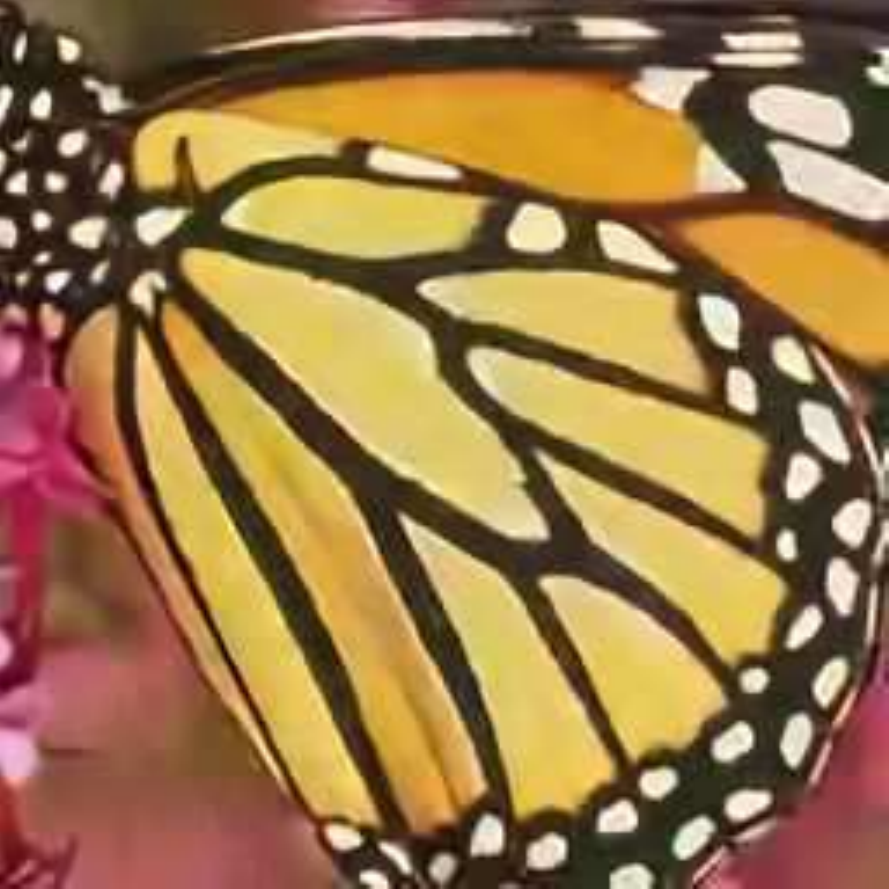} &
      \includegraphics[width=0.08\linewidth]{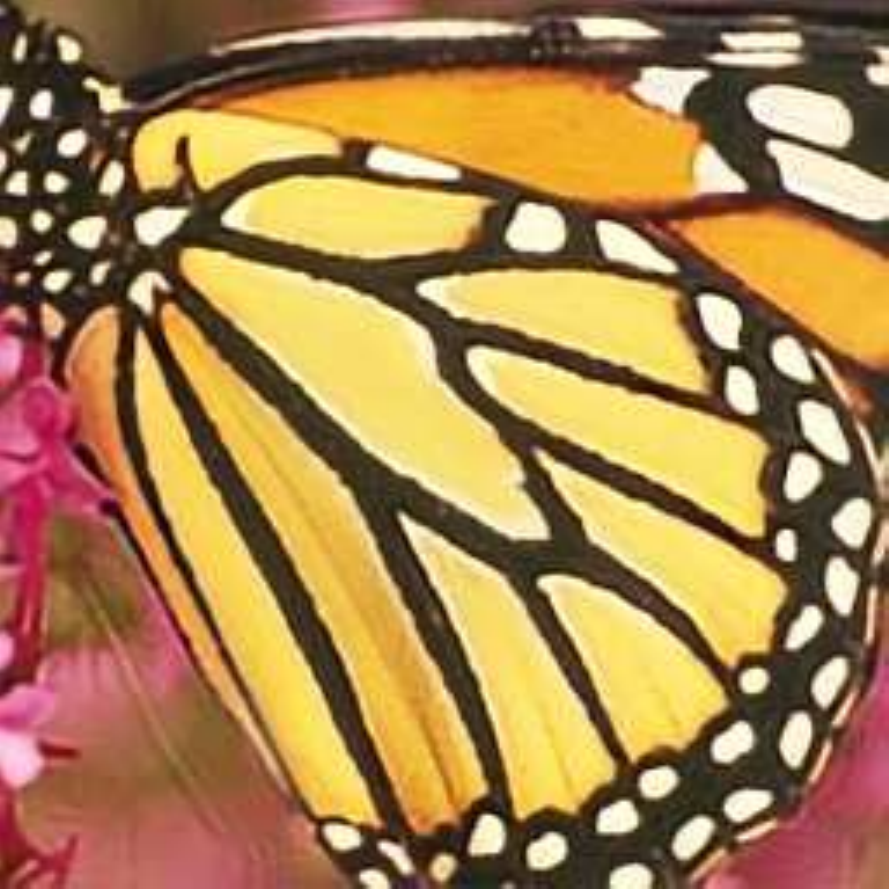} &
      \includegraphics[width=0.08\linewidth]{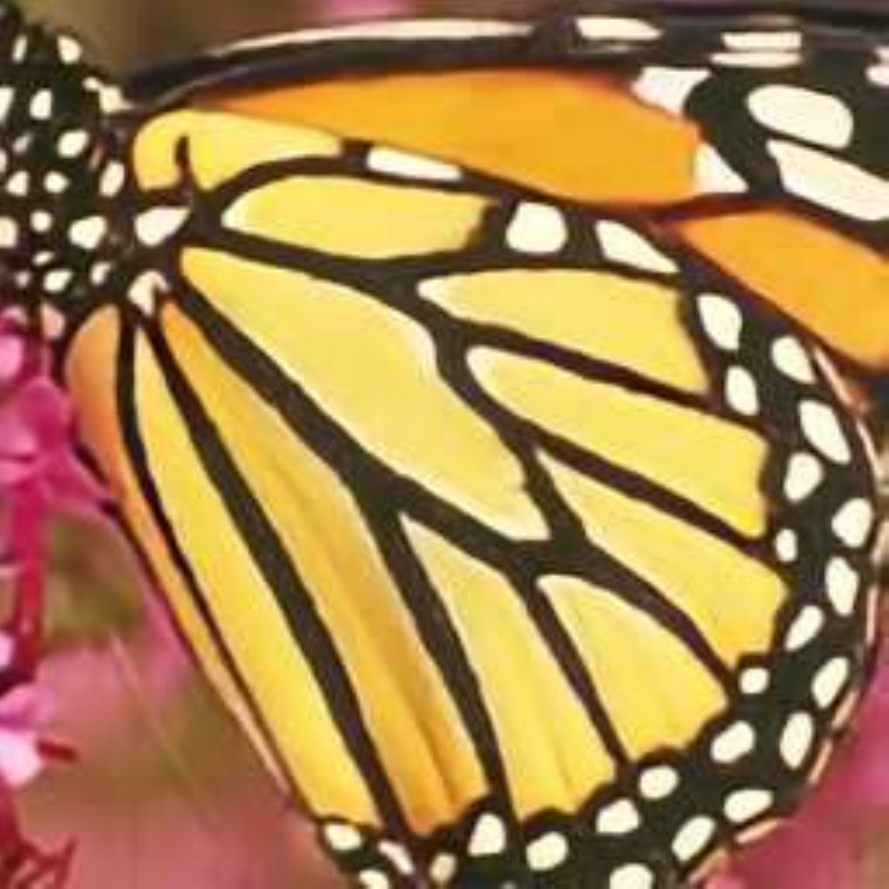} &
      \includegraphics[width=0.08\linewidth]{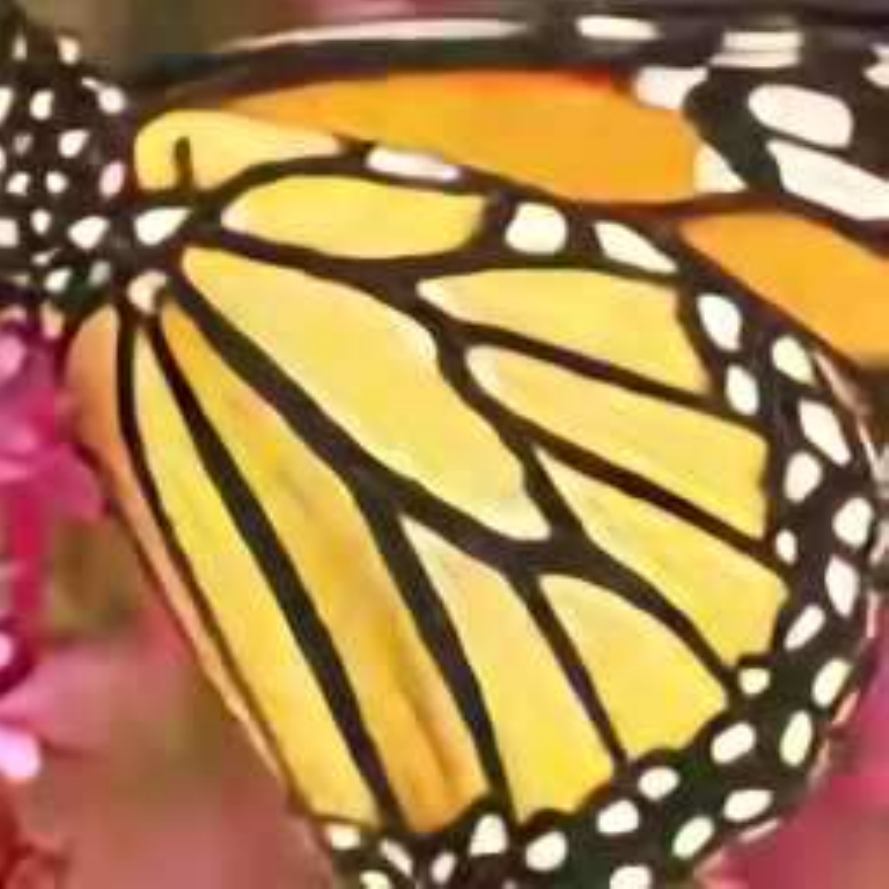} &
      \includegraphics[width=0.08\linewidth]{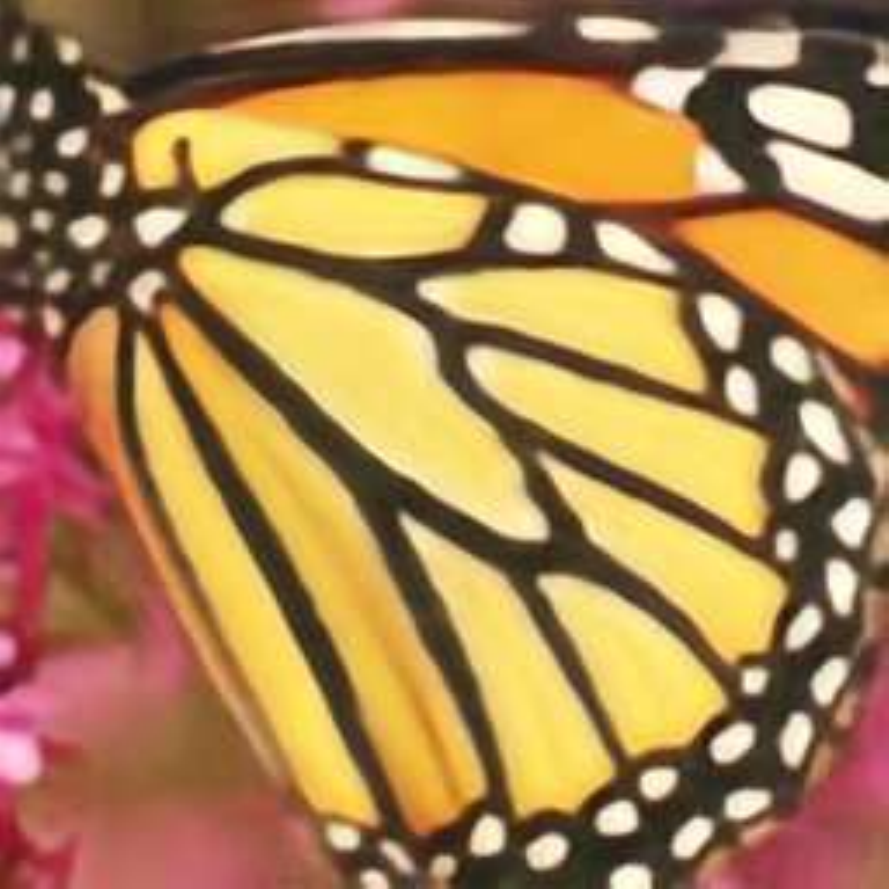} &
      \includegraphics[width=0.08\linewidth]{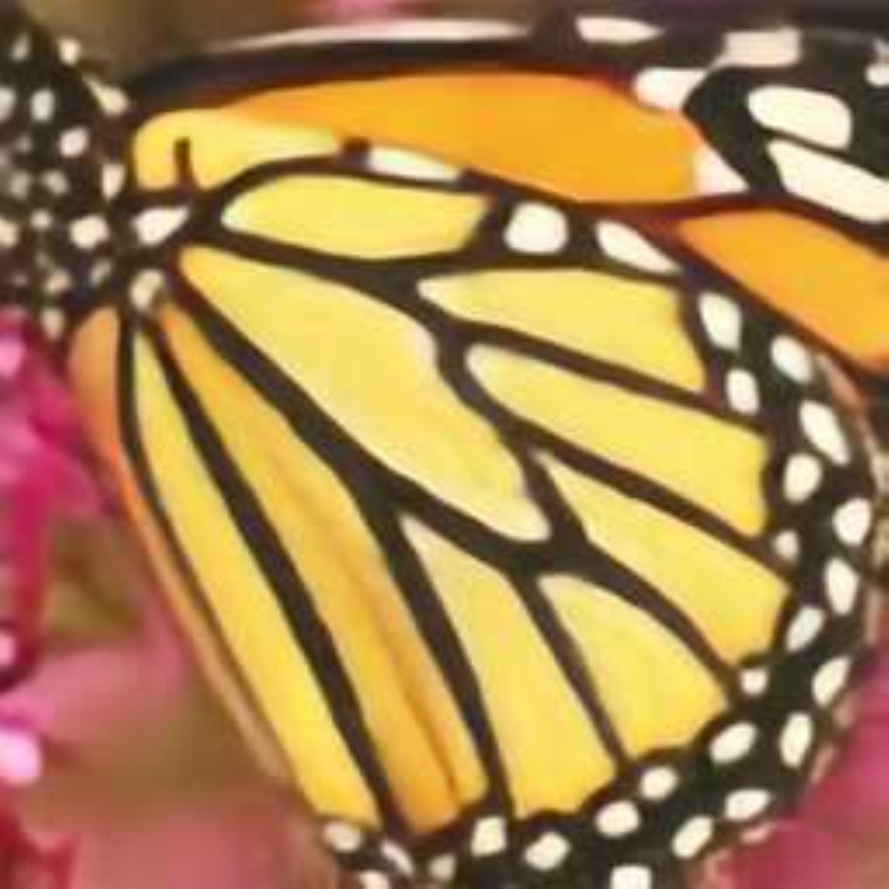} &
      \includegraphics[width=0.08\linewidth]{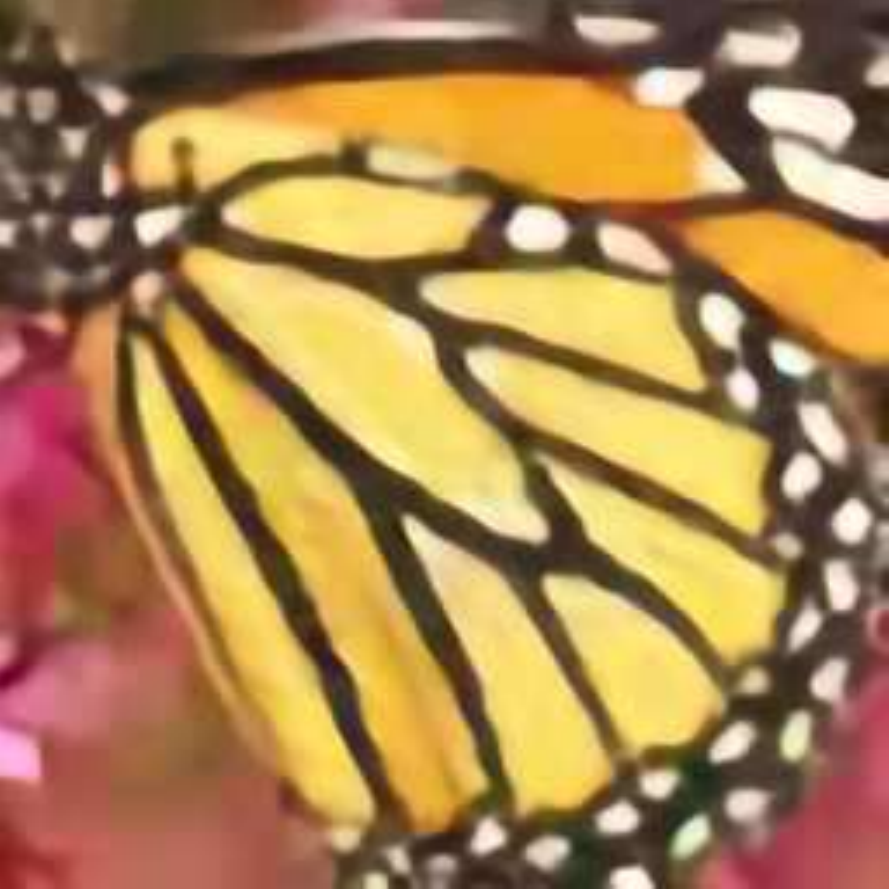} \\ \hline
  \multirow{3}{*}{$\lambda=25$} &
Degraded  &
        \includegraphics[width=0.08\linewidth]{images/butterfly/degraded_000_250_100_pdf.pdf} &
        \includegraphics[width=0.08\linewidth]{images/butterfly/degraded_000_250_50_pdf.pdf} &
        \includegraphics[width=0.08\linewidth]{images/butterfly/degraded_000_250_10_pdf.pdf} &
        \includegraphics[width=0.08\linewidth]{images/butterfly/degraded_150_250_100_pdf.pdf} &
        \includegraphics[width=0.08\linewidth]{images/butterfly/degraded_150_250_50_pdf.pdf} &
        \includegraphics[width=0.08\linewidth]{images/butterfly/degraded_150_250_10_pdf.pdf} &
        \includegraphics[width=0.08\linewidth]{images/butterfly/degraded_300_250_100_pdf.pdf} &
        \includegraphics[width=0.08\linewidth]{images/butterfly/degraded_300_250_50_pdf.pdf} &
        \includegraphics[width=0.08\linewidth]{images/butterfly/degraded_300_250_10_pdf.pdf} \\
&Restored      &
      \includegraphics[width=0.08\linewidth]{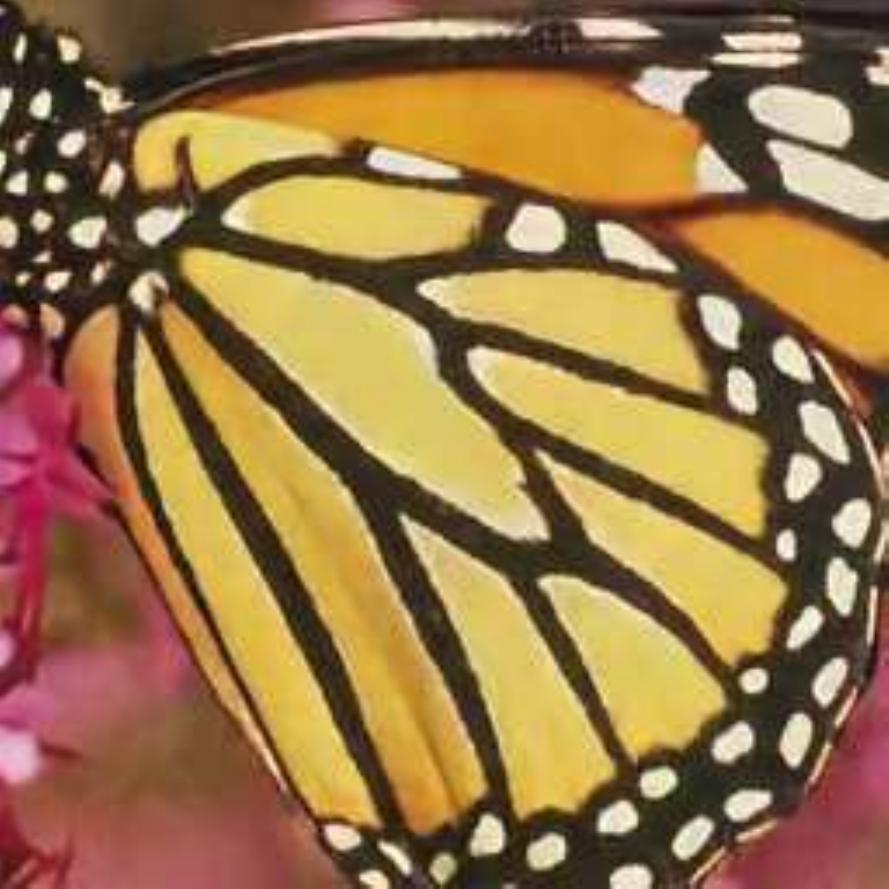} &
      \includegraphics[width=0.08\linewidth]{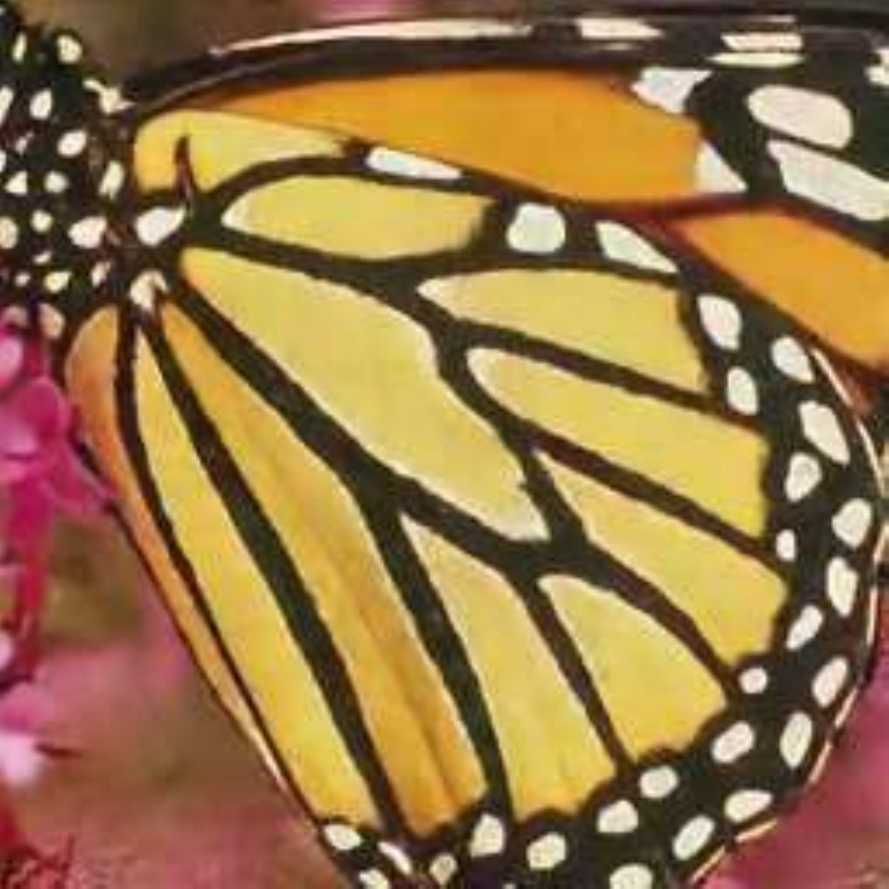} &
      \includegraphics[width=0.08\linewidth]{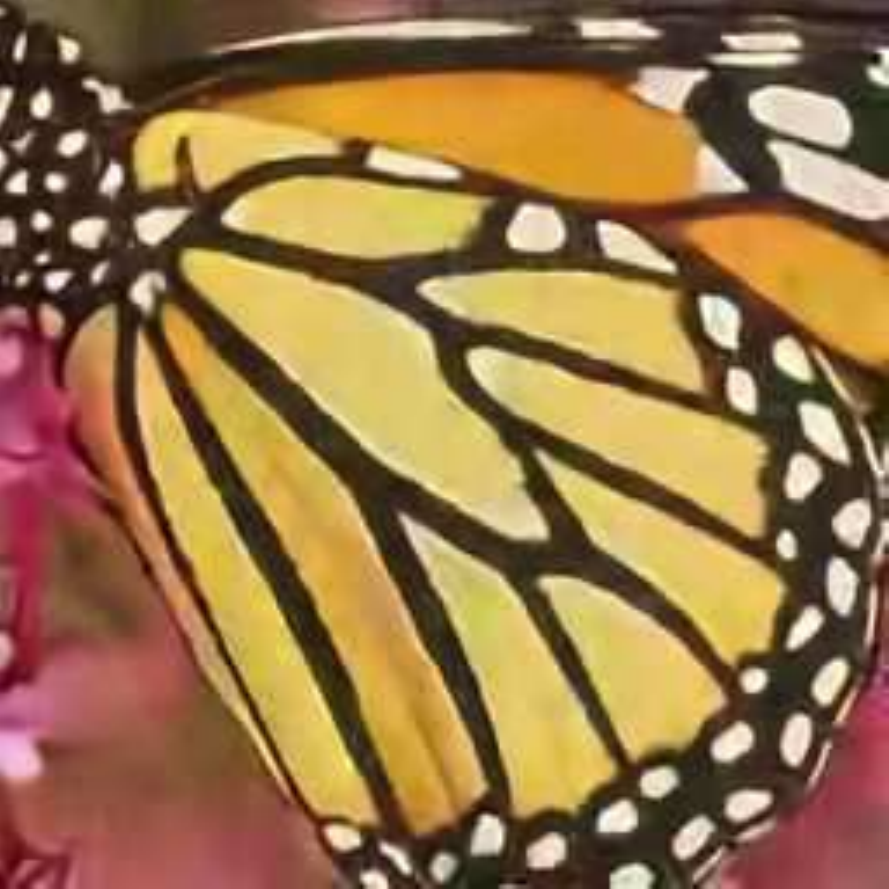} &
      \includegraphics[width=0.08\linewidth]{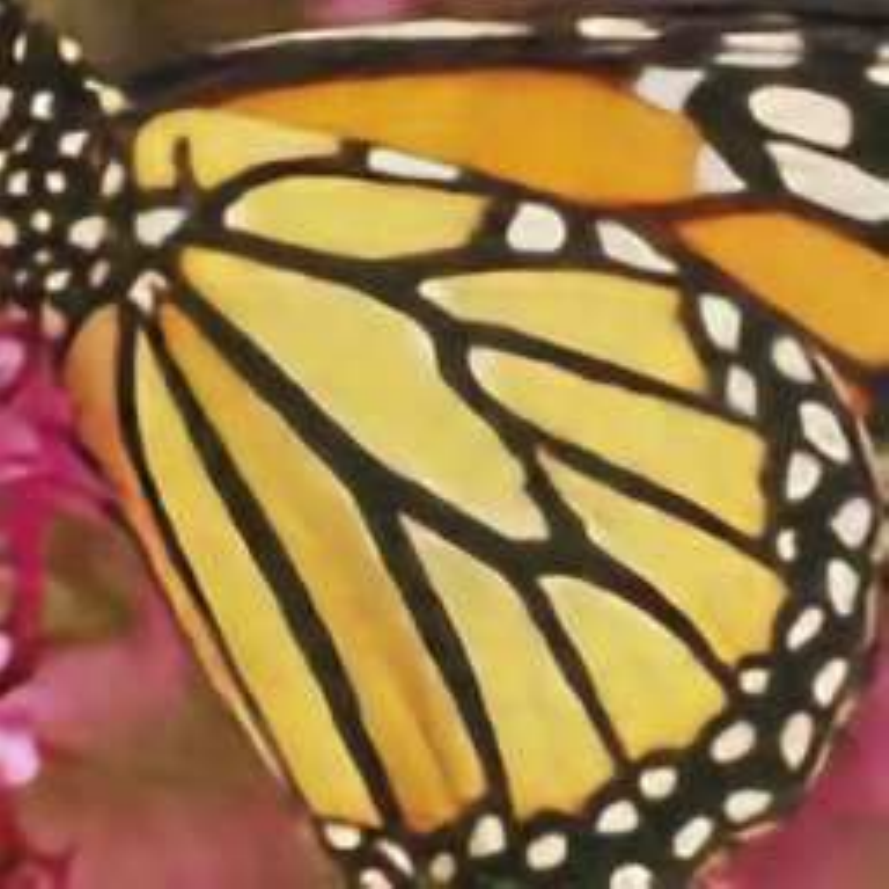} &
      \includegraphics[width=0.08\linewidth]{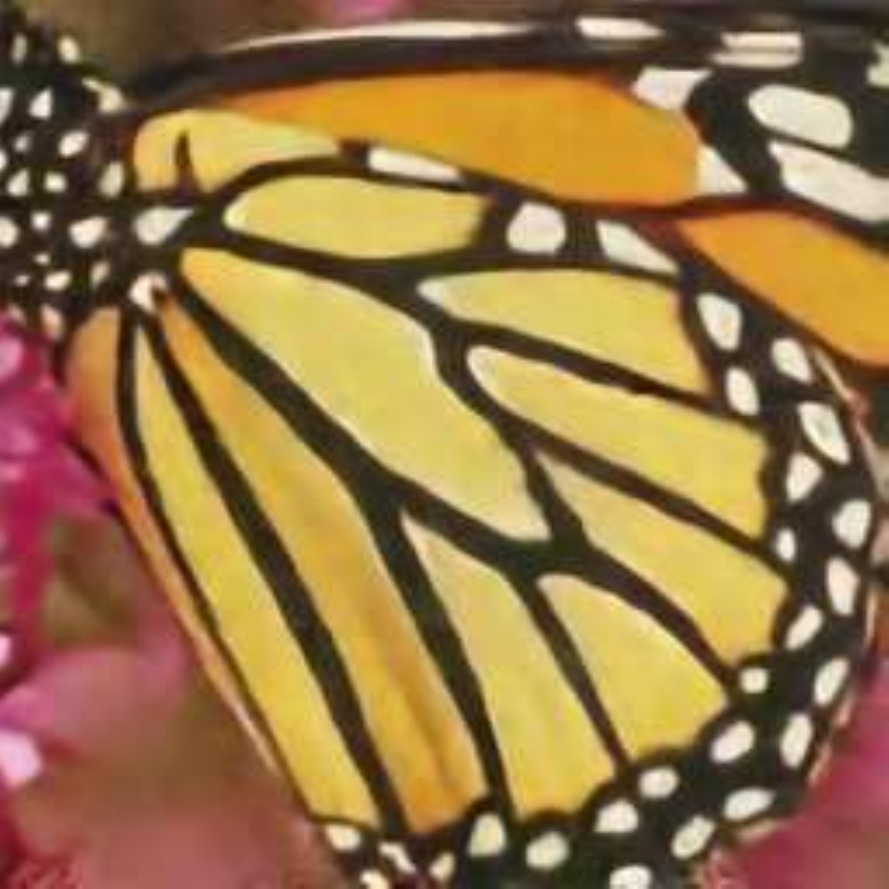} &
      \includegraphics[width=0.08\linewidth]{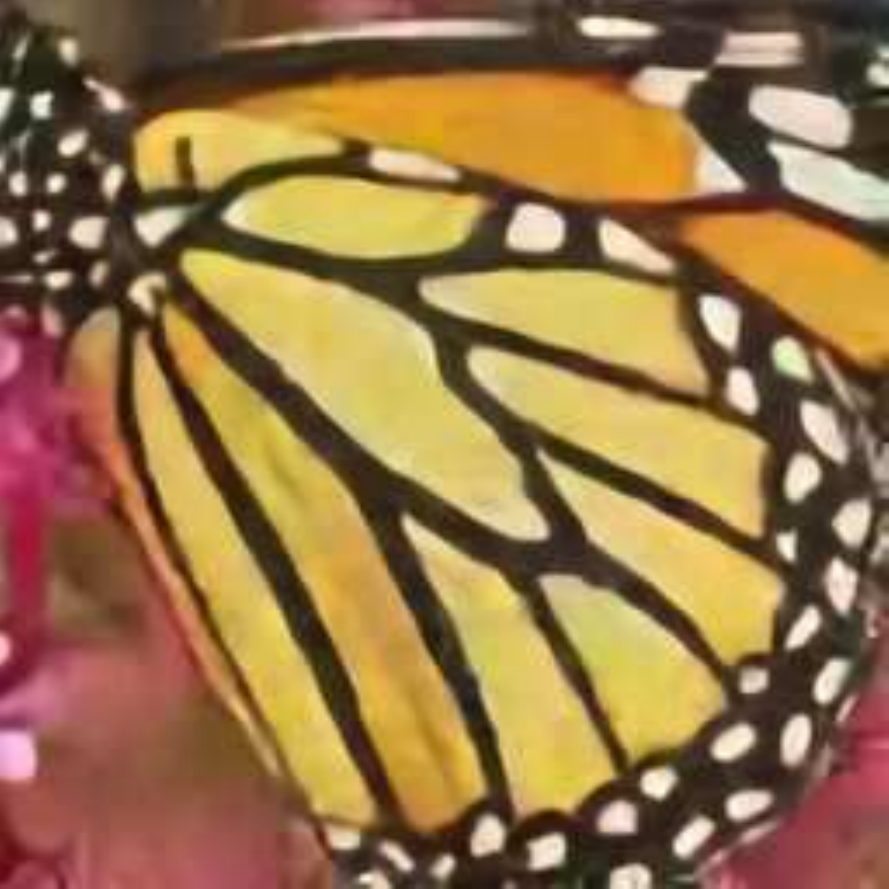} &
      \includegraphics[width=0.08\linewidth]{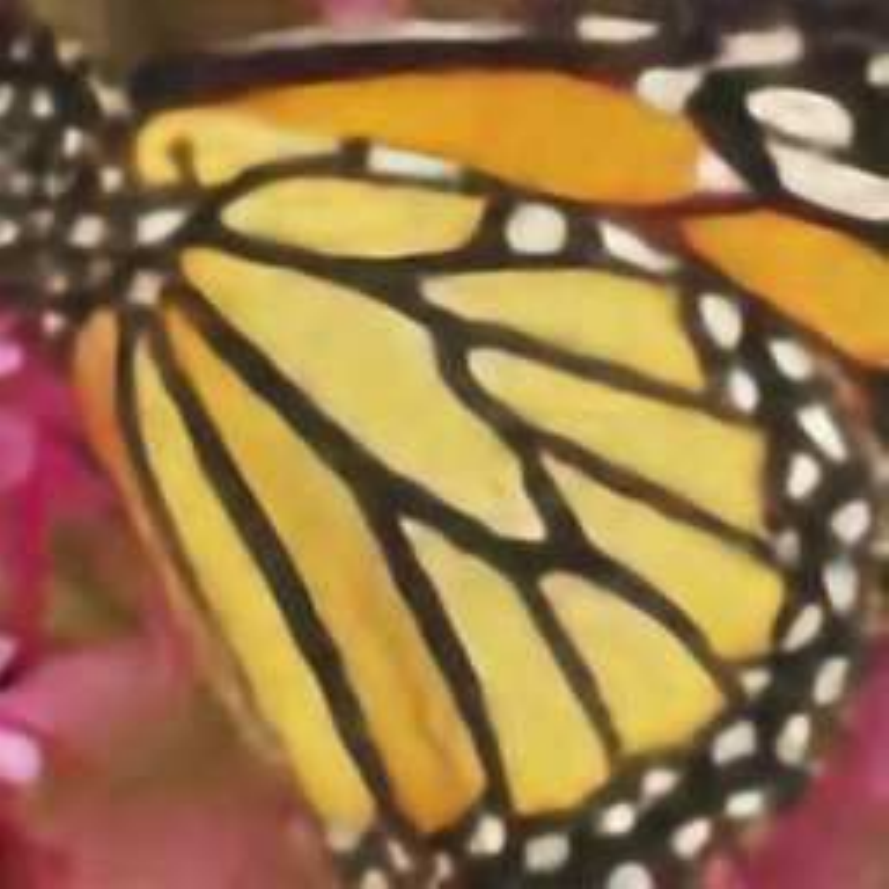} &
      \includegraphics[width=0.08\linewidth]{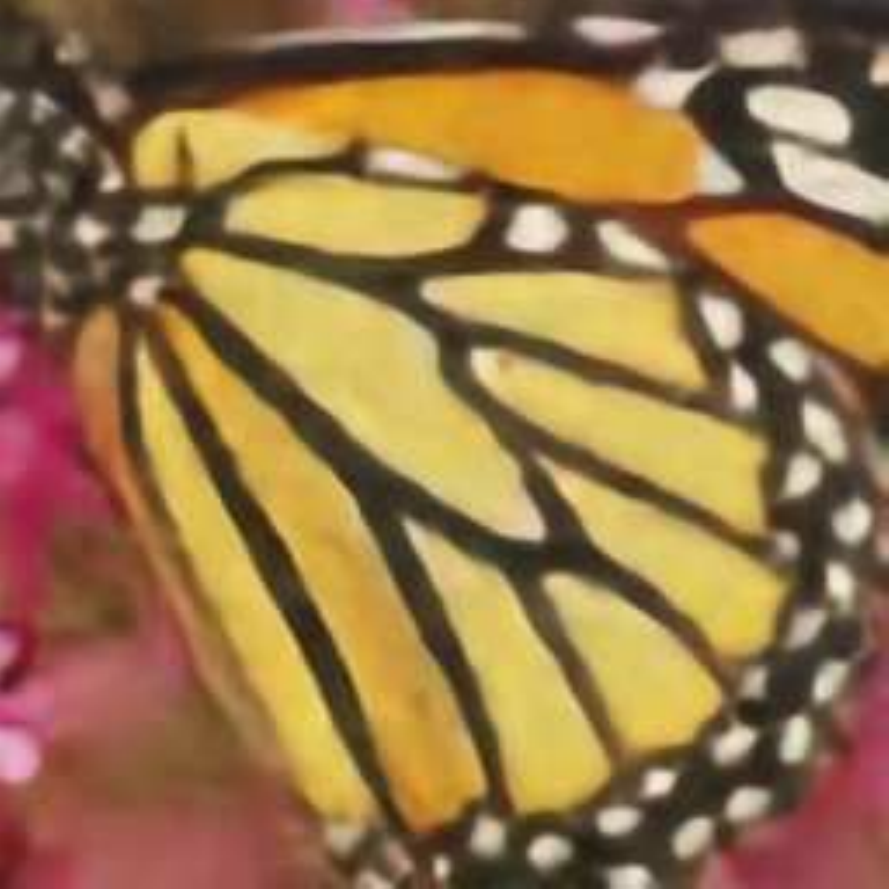} &
      \includegraphics[width=0.08\linewidth]{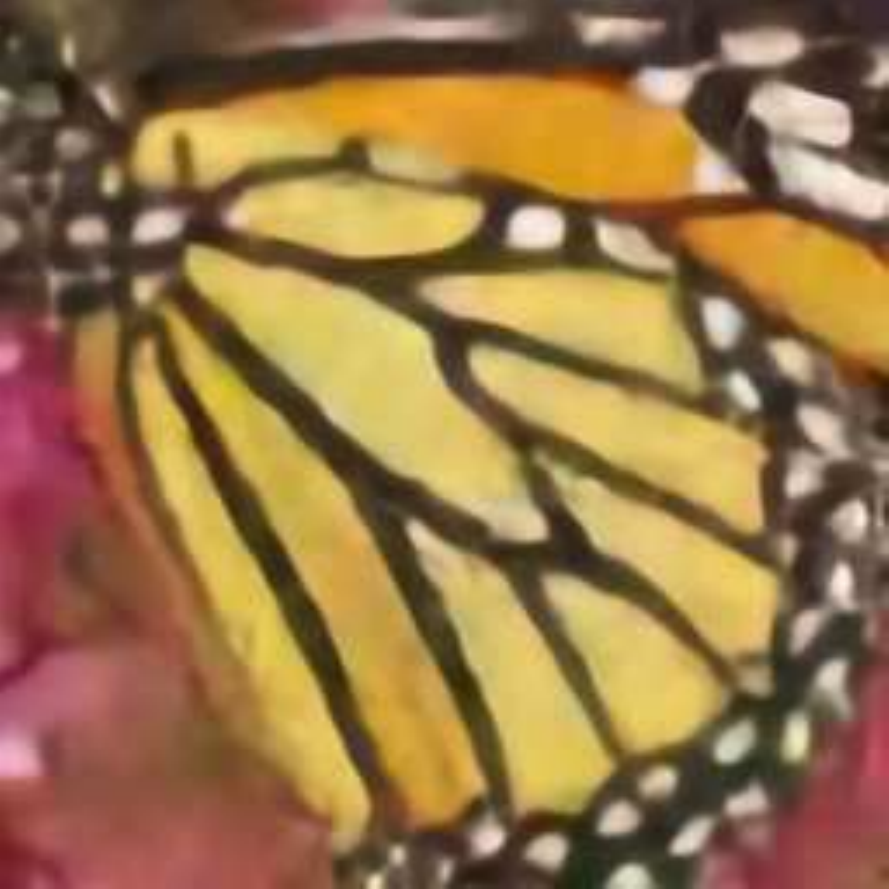} \\ \hline
  \multirow{3}{*}{$\lambda=55$} &
Degraded  &
      \includegraphics[width=0.08\linewidth]{images/butterfly/degraded_000_550_100_pdf.pdf} &
      \includegraphics[width=0.08\linewidth]{images/butterfly/degraded_000_550_50_pdf.pdf} &
      \includegraphics[width=0.08\linewidth]{images/butterfly/degraded_000_550_10_pdf.pdf} &
      \includegraphics[width=0.08\linewidth]{images/butterfly/degraded_150_550_100_pdf.pdf} &
      \includegraphics[width=0.08\linewidth]{images/butterfly/degraded_150_550_50_pdf.pdf} &
      \includegraphics[width=0.08\linewidth]{images/butterfly/degraded_150_550_10_pdf.pdf} &
      \includegraphics[width=0.08\linewidth]{images/butterfly/degraded_300_550_100_pdf.pdf} &
      \includegraphics[width=0.08\linewidth]{images/butterfly/degraded_300_550_50_pdf.pdf} &
      \includegraphics[width=0.08\linewidth]{images/butterfly/degraded_300_550_10_pdf.pdf} \\
&Restored     &
      \includegraphics[width=0.08\linewidth]{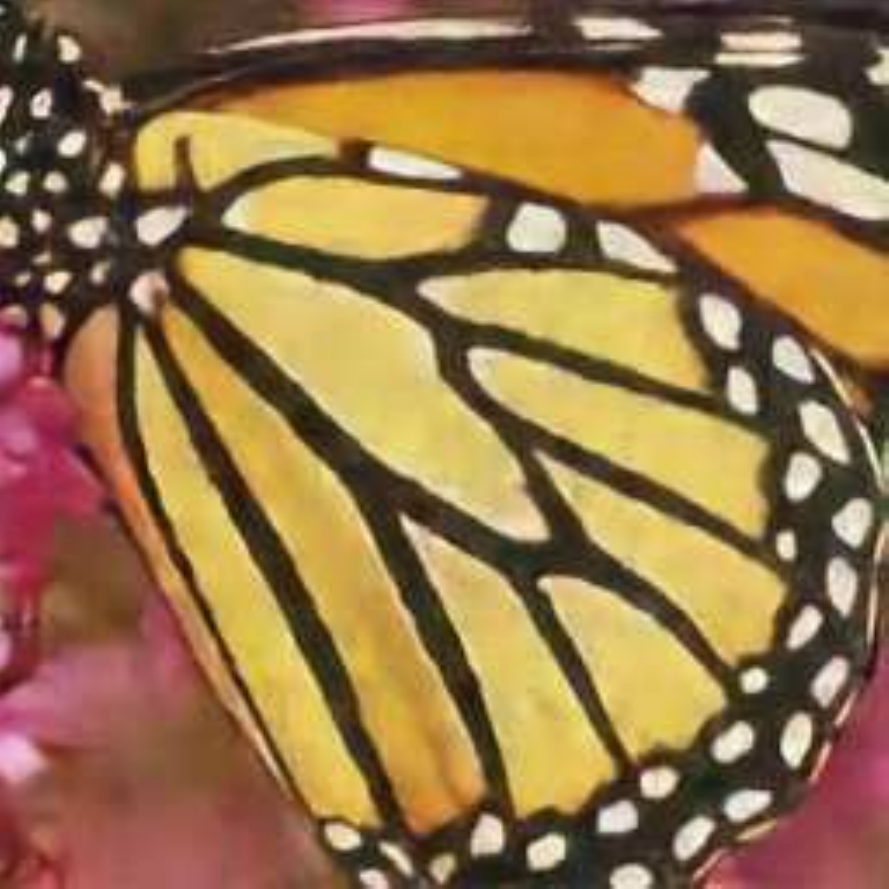} &
      \includegraphics[width=0.08\linewidth]{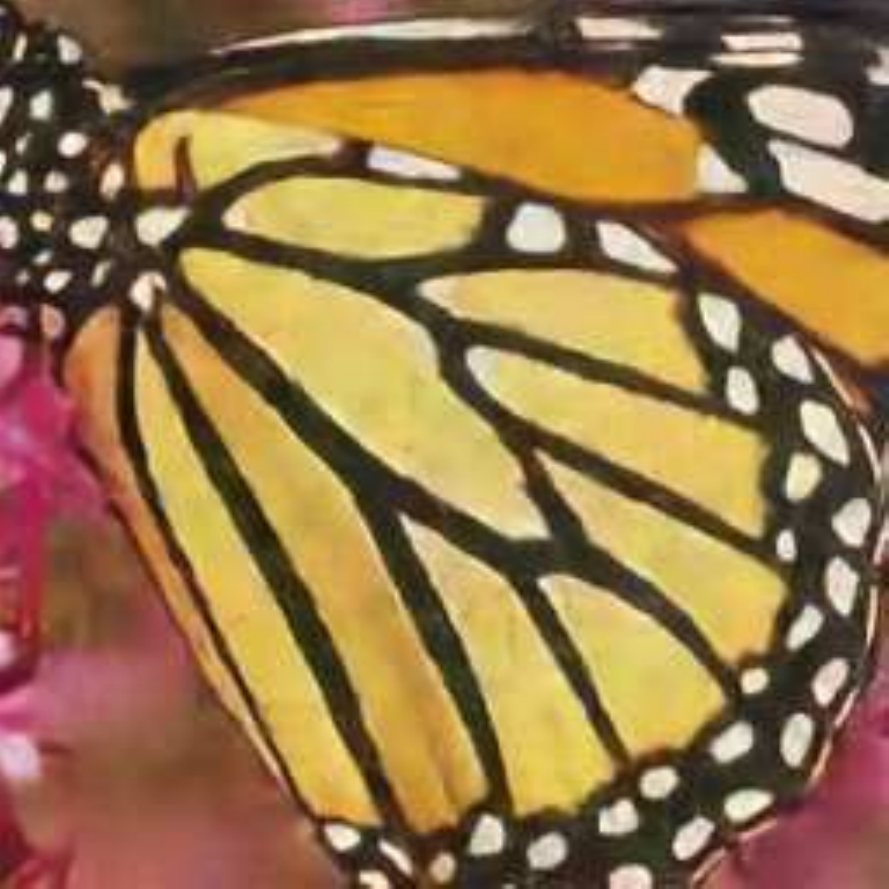} &
      \includegraphics[width=0.08\linewidth]{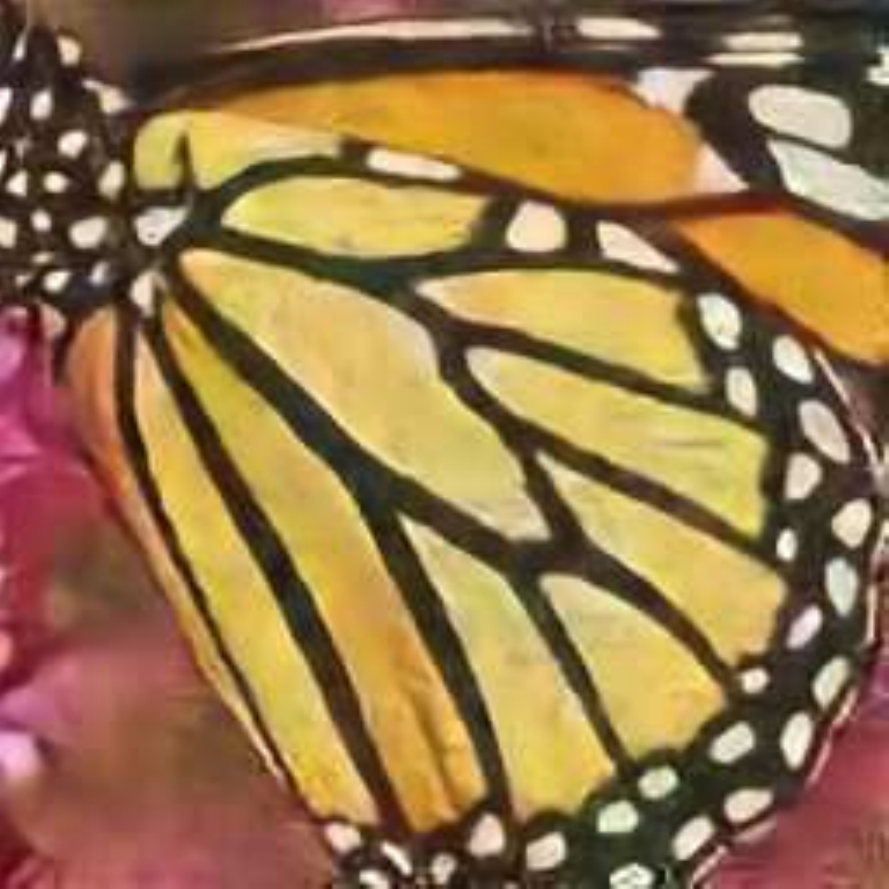} &
      \includegraphics[width=0.08\linewidth]{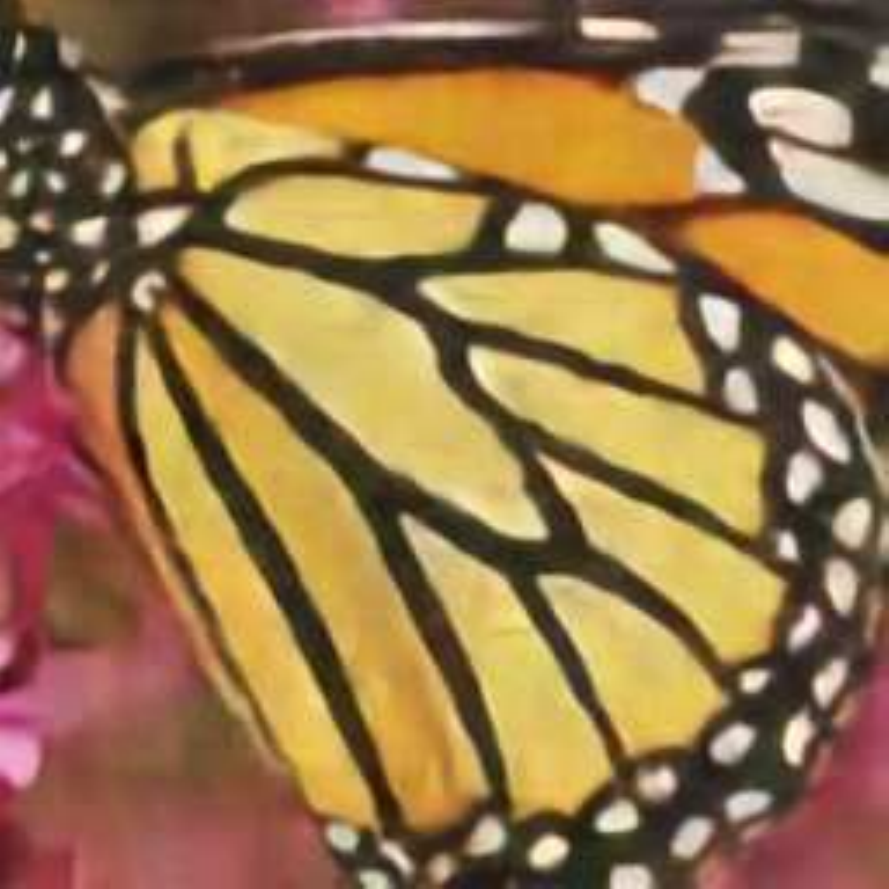} &
      \includegraphics[width=0.08\linewidth]{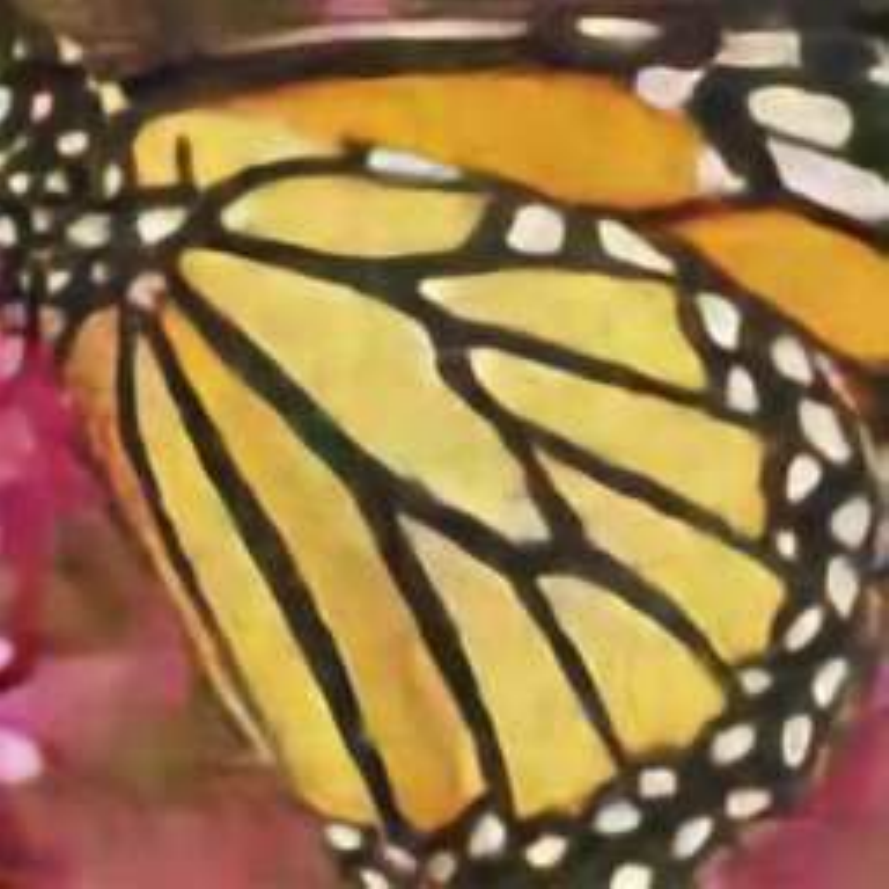} &
      \includegraphics[width=0.08\linewidth]{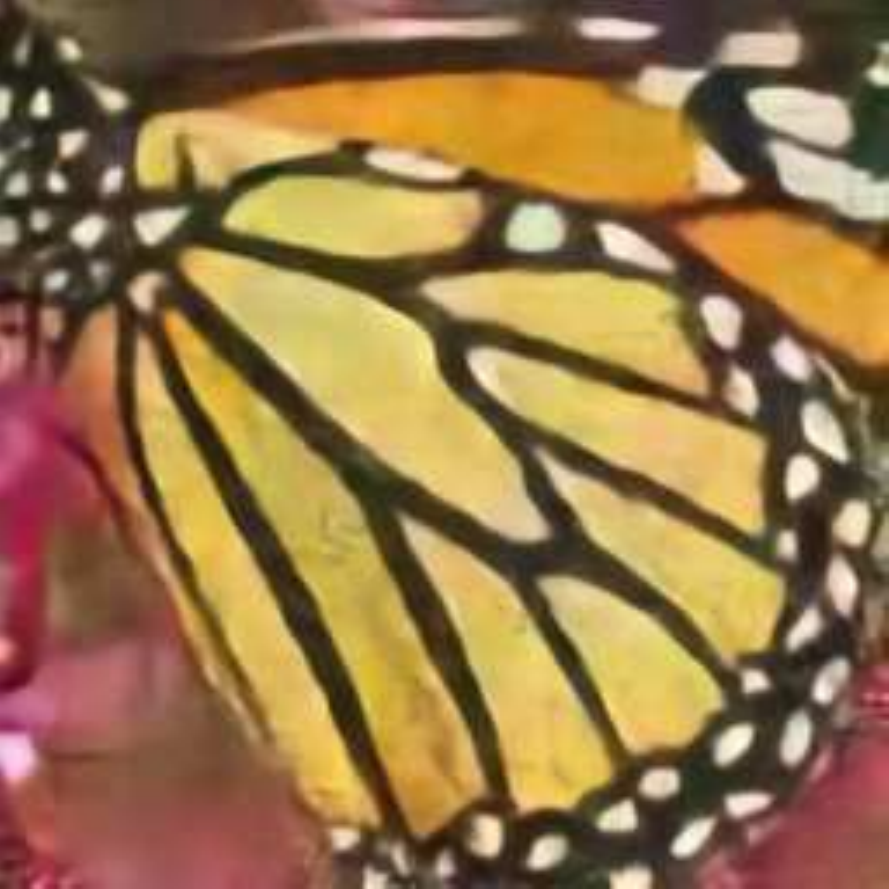} &
      \includegraphics[width=0.08\linewidth]{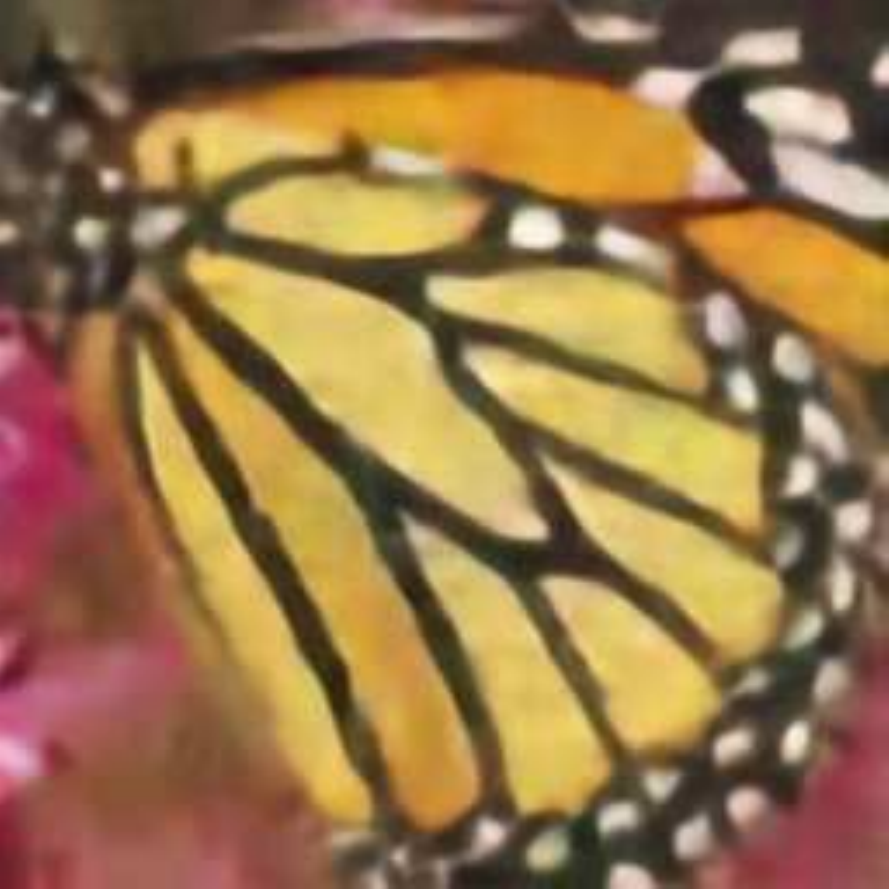} &
      \includegraphics[width=0.08\linewidth]{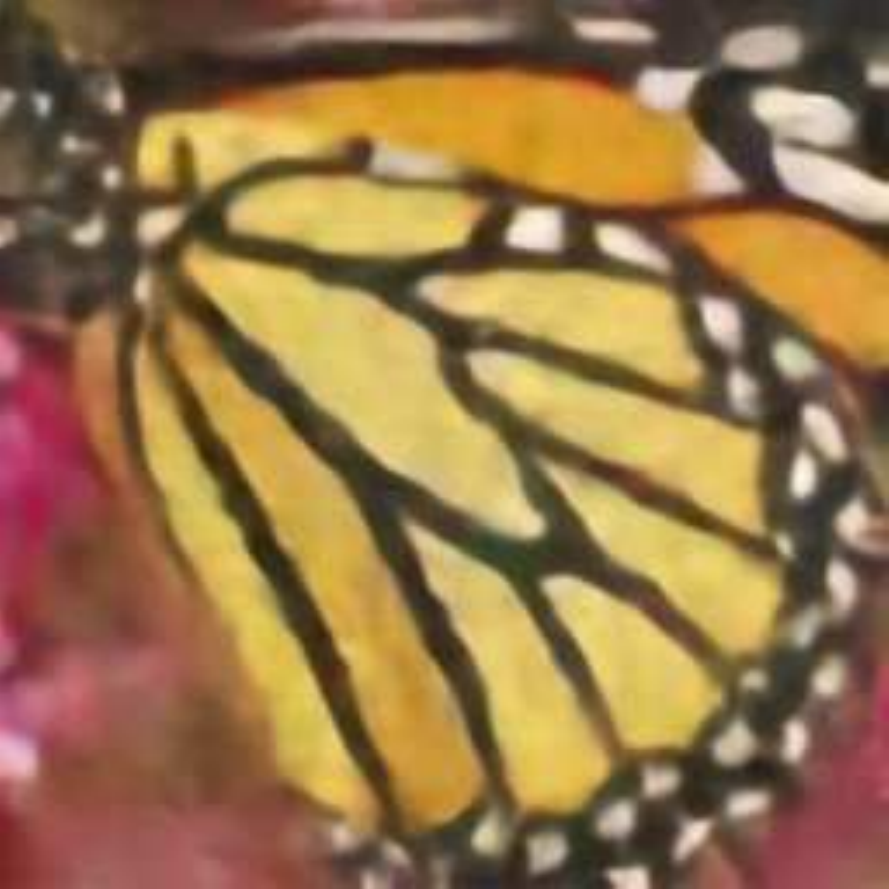} &
      \includegraphics[width=0.08\linewidth]{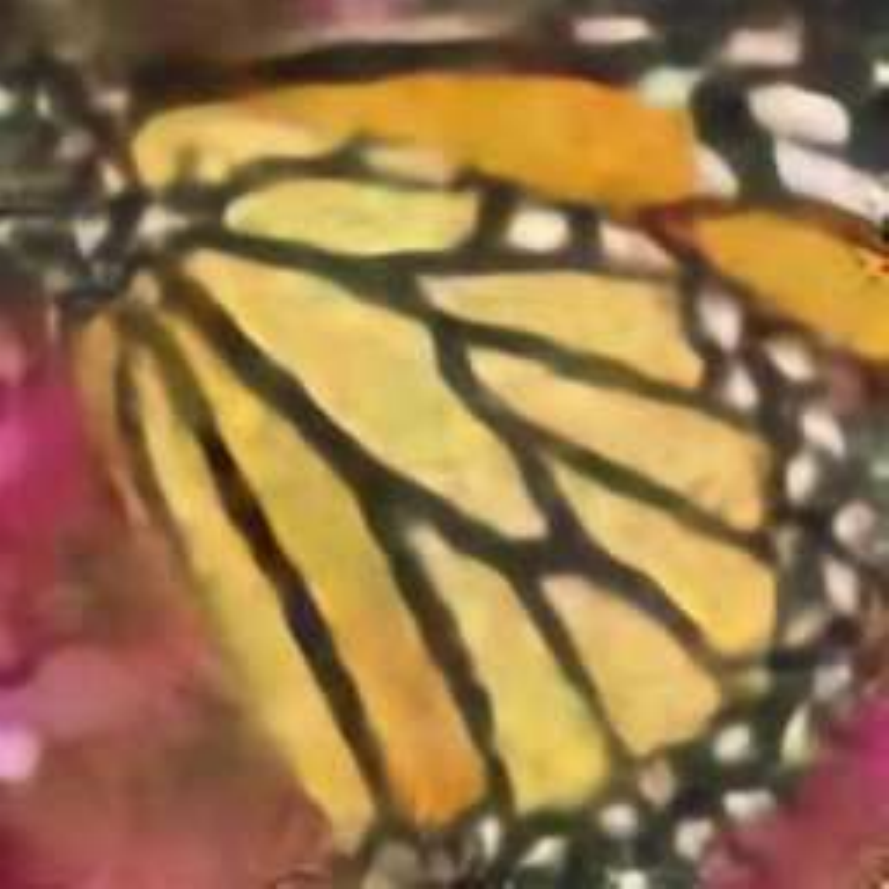} \\ \hline
     \end{tabular}
  \caption{Examples of the proposed blind restoration with different degradation parameters.}
  \label{fig:restorations}
    \end{center}
\end{figure*}

Figure \ref{fig:restorations} shows the examples of the proposed blind restoration with different degradation parameters.
A certain level of restoration from compositional degradation is achieved.

\begin{figure*}[t!]
    \footnotesize
    \begin{center}
      \begin{tabular}{|c|c|ccc|c|} \hline
      &  & \multicolumn{3}{|c|}{Existing Methods} & Ours \\ \hline
      Original & Degraded & Deblocked \cite{zhang2017beyond} & Then, Denoised \cite{liu2013single, dabov2007image}& Then, Deblurred \cite{shan2008high}& Restored\\ \hline
       \includegraphics[width=0.13\linewidth]{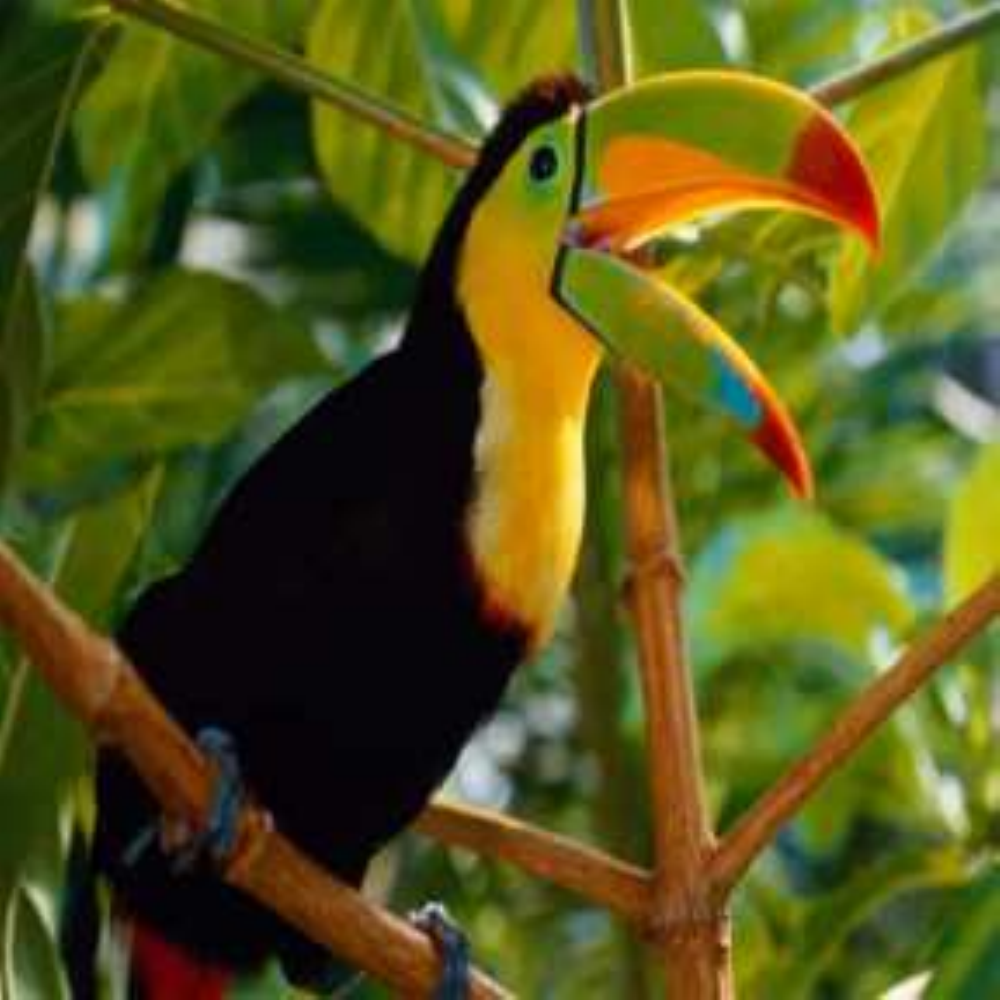} &
       \includegraphics[width=0.13\linewidth]{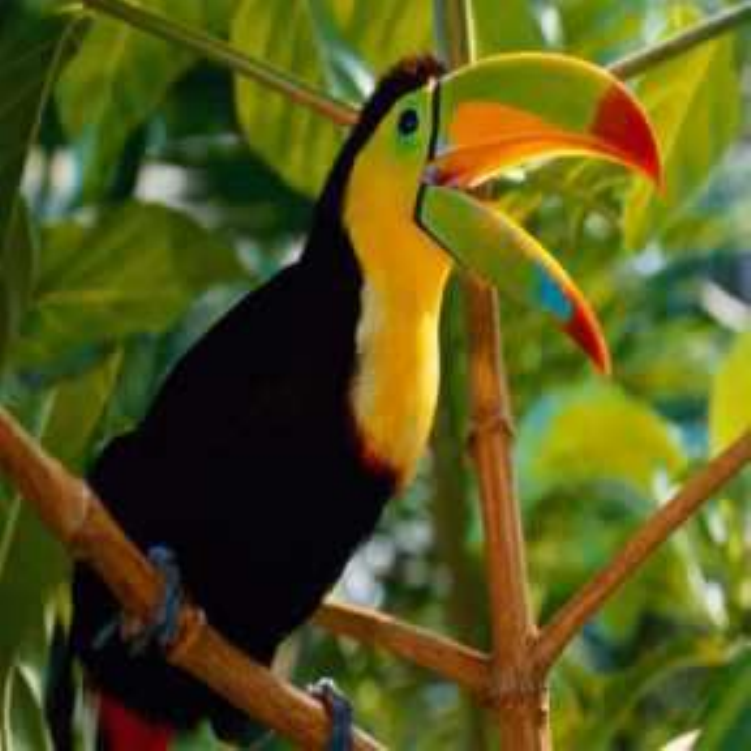} &
       \includegraphics[width=0.13\linewidth]{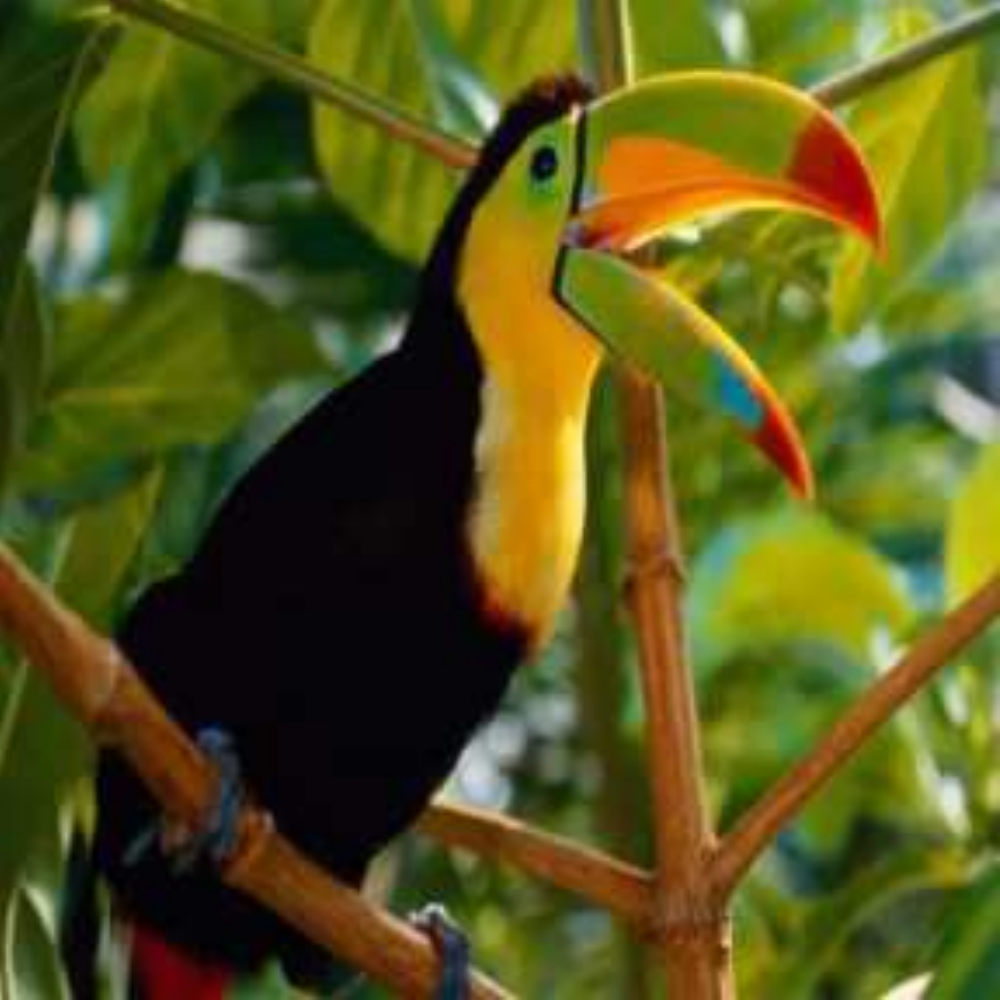} &
       \includegraphics[width=0.13\linewidth]{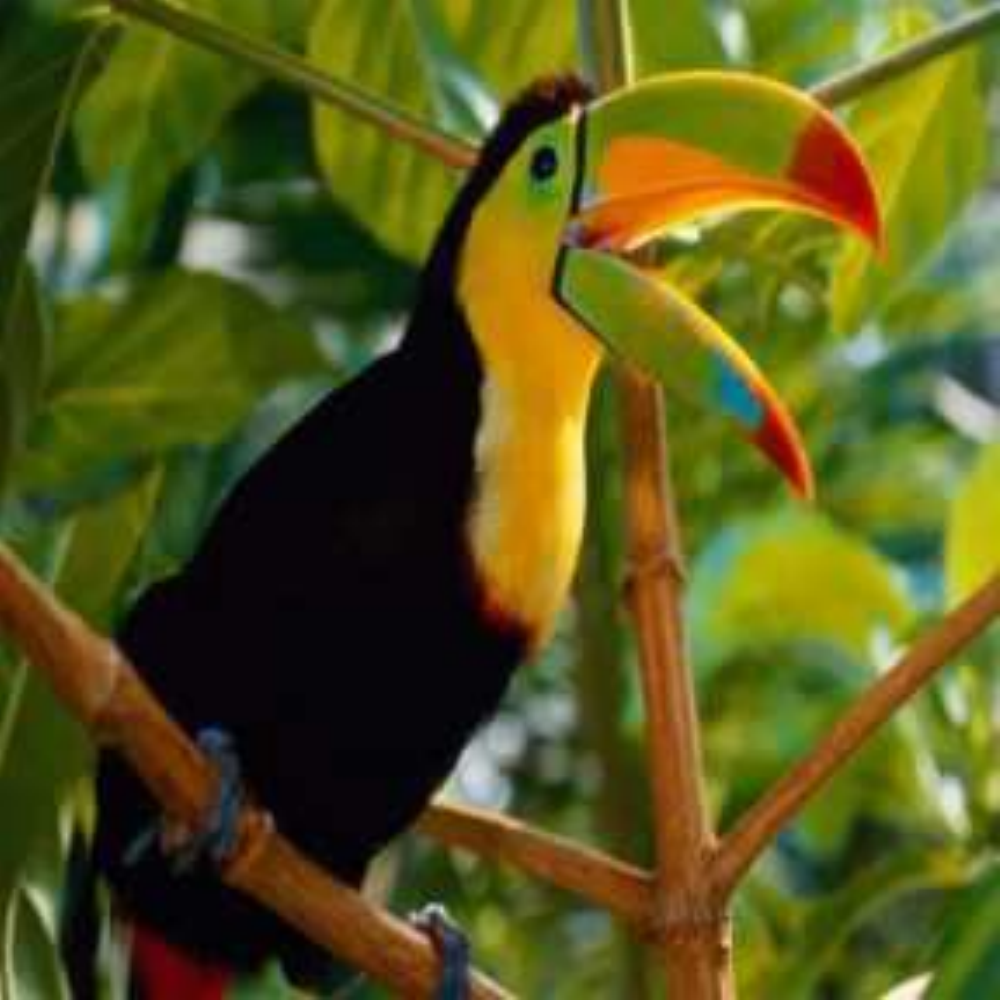} &
       \includegraphics[width=0.13\linewidth]{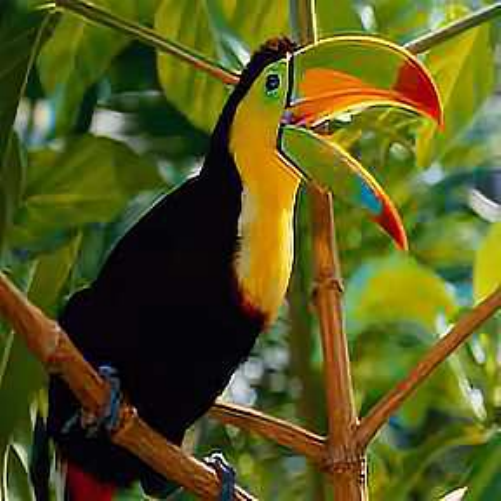} &
       \includegraphics[width=0.13\linewidth]{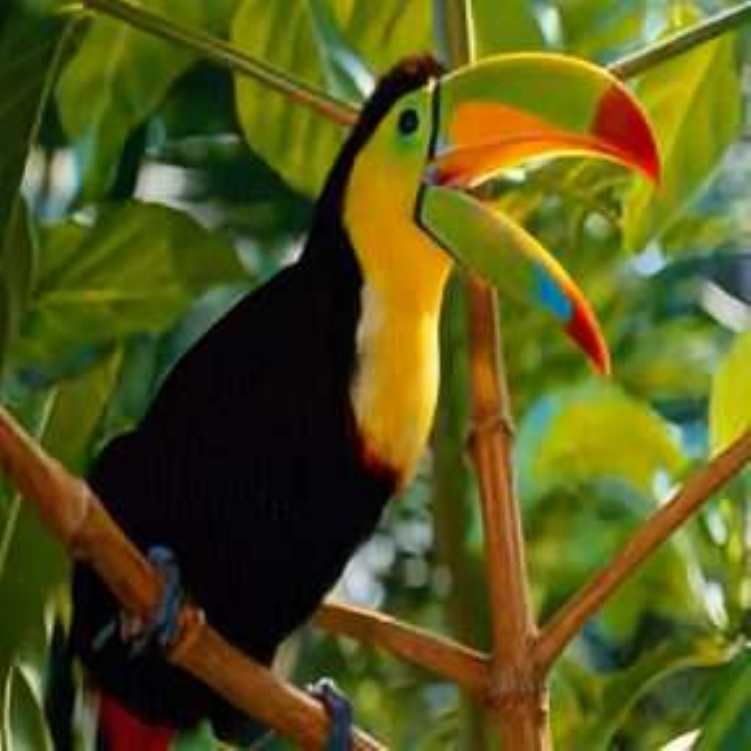} \\
       & \footnotesize{$\sigma=0, \lambda=0, q=50$} & & & & \\ \hline

       \includegraphics[width=0.13\linewidth]{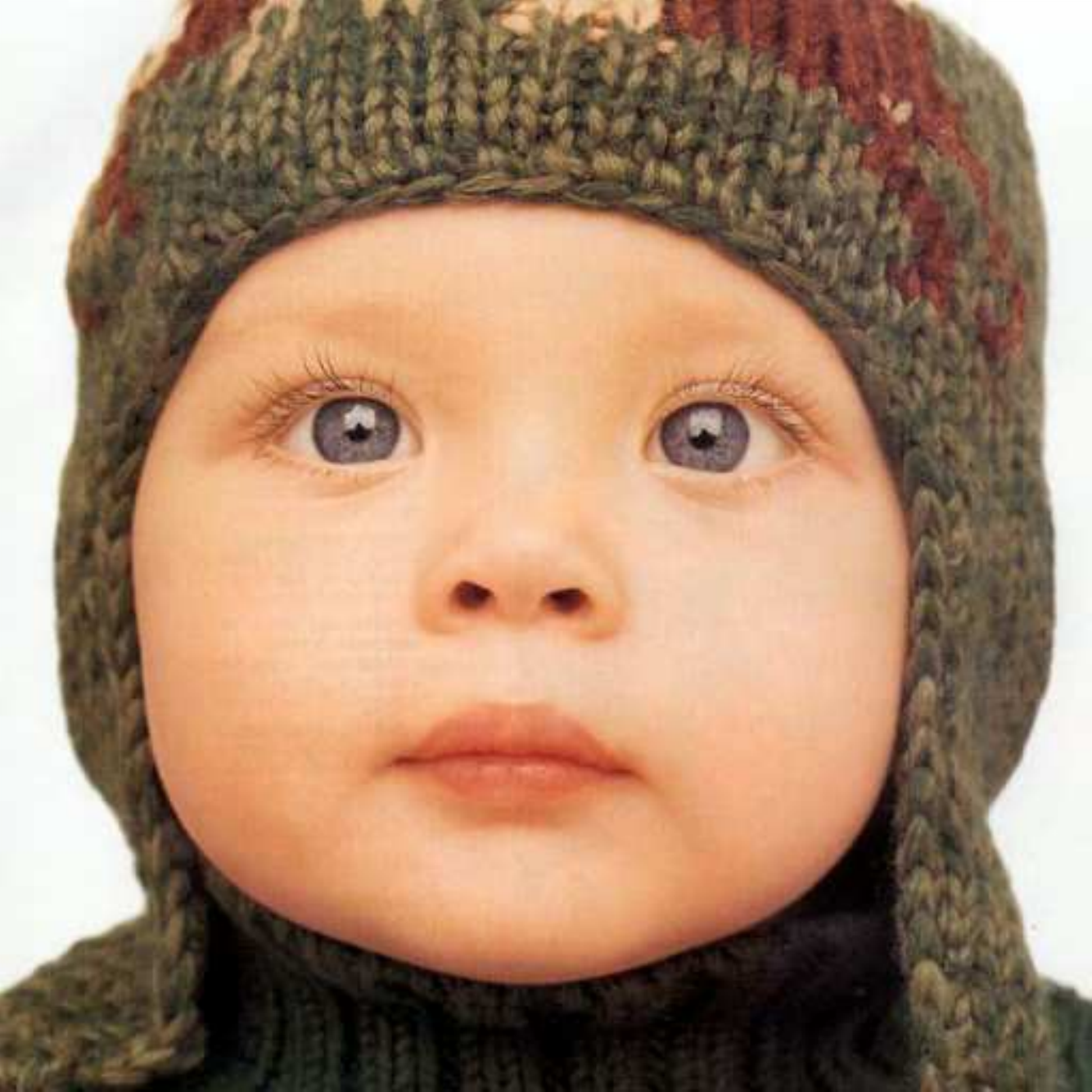} &
       \includegraphics[width=0.13\linewidth]{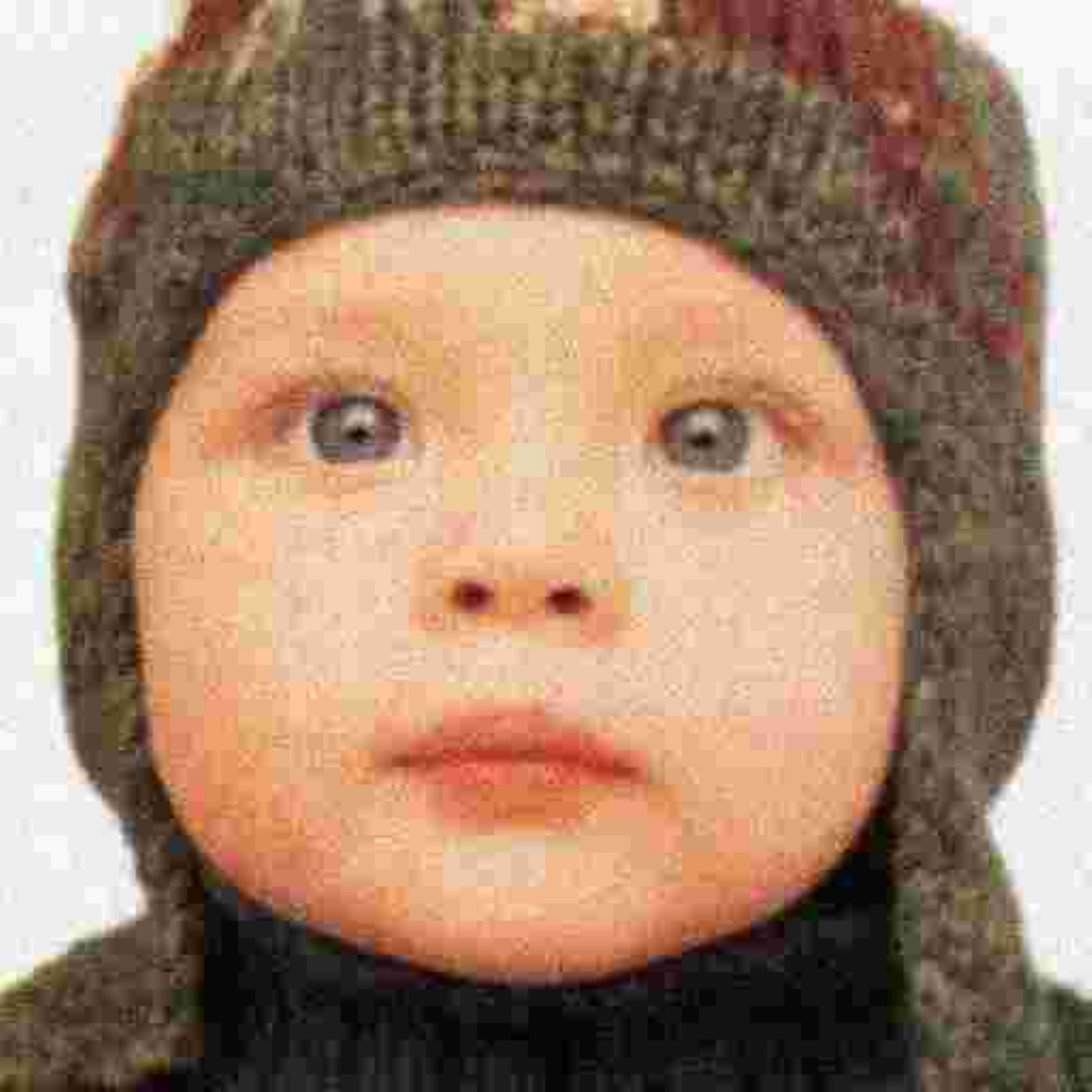} &
       \includegraphics[width=0.13\linewidth]{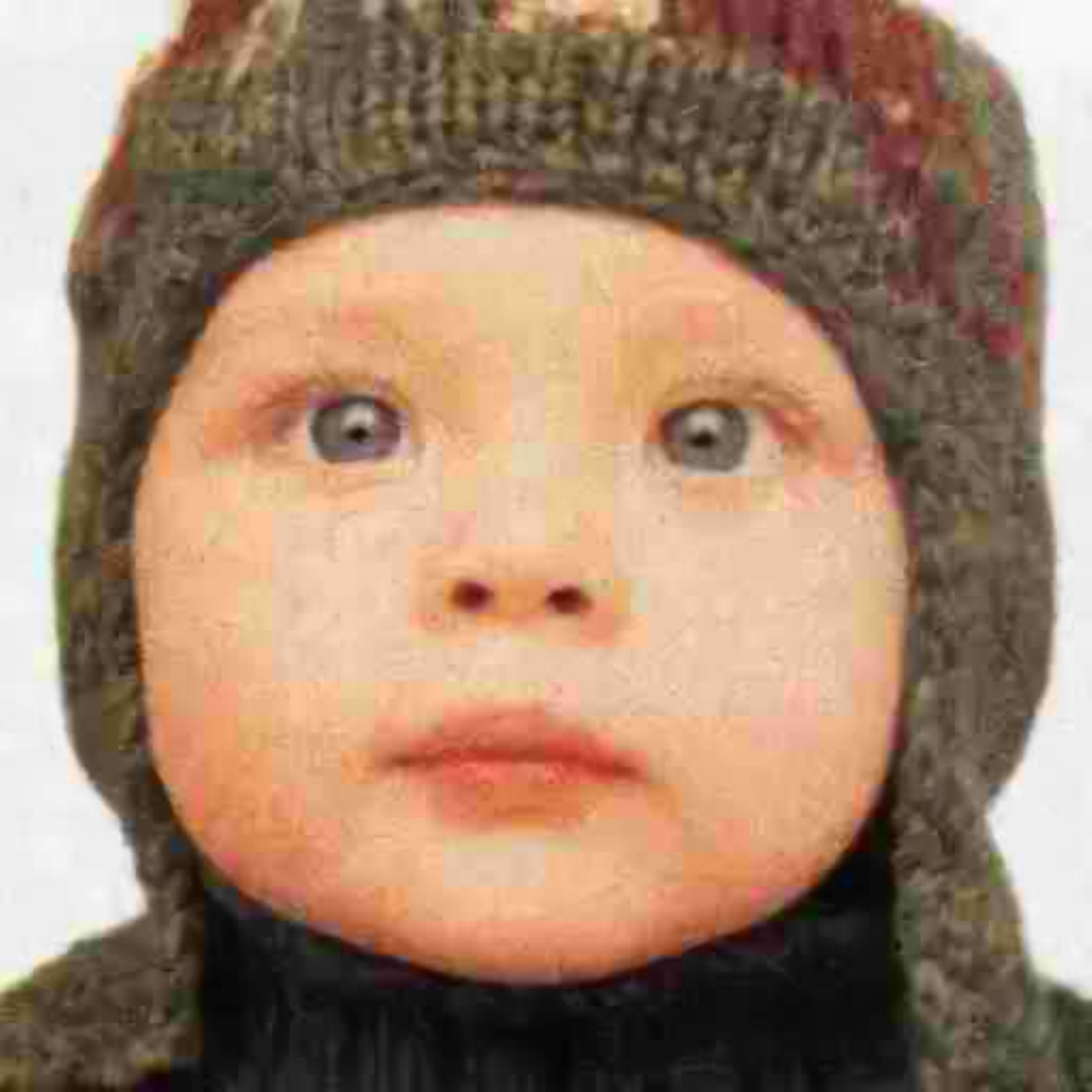} &
       \includegraphics[width=0.13\linewidth]{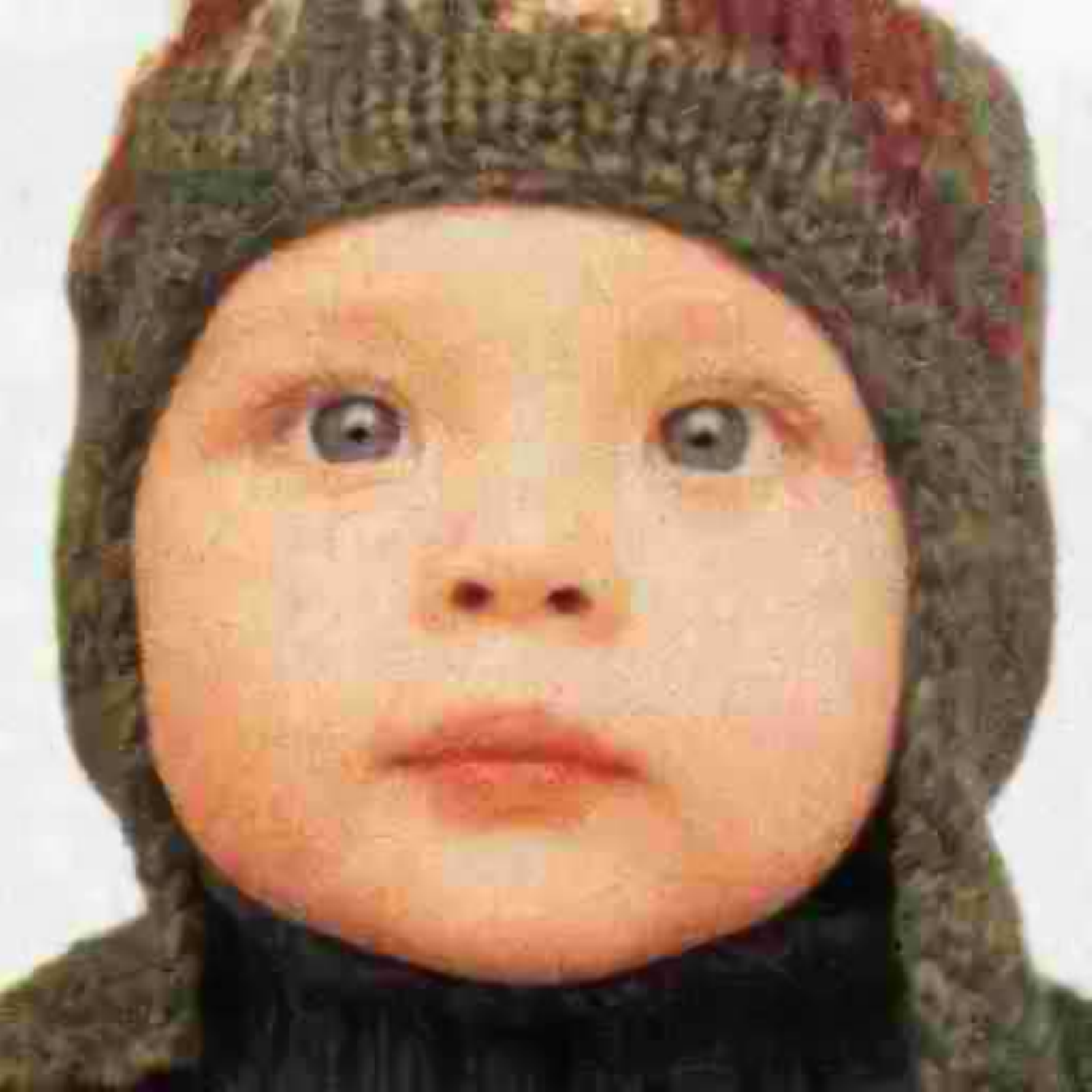} &
       \includegraphics[width=0.13\linewidth]{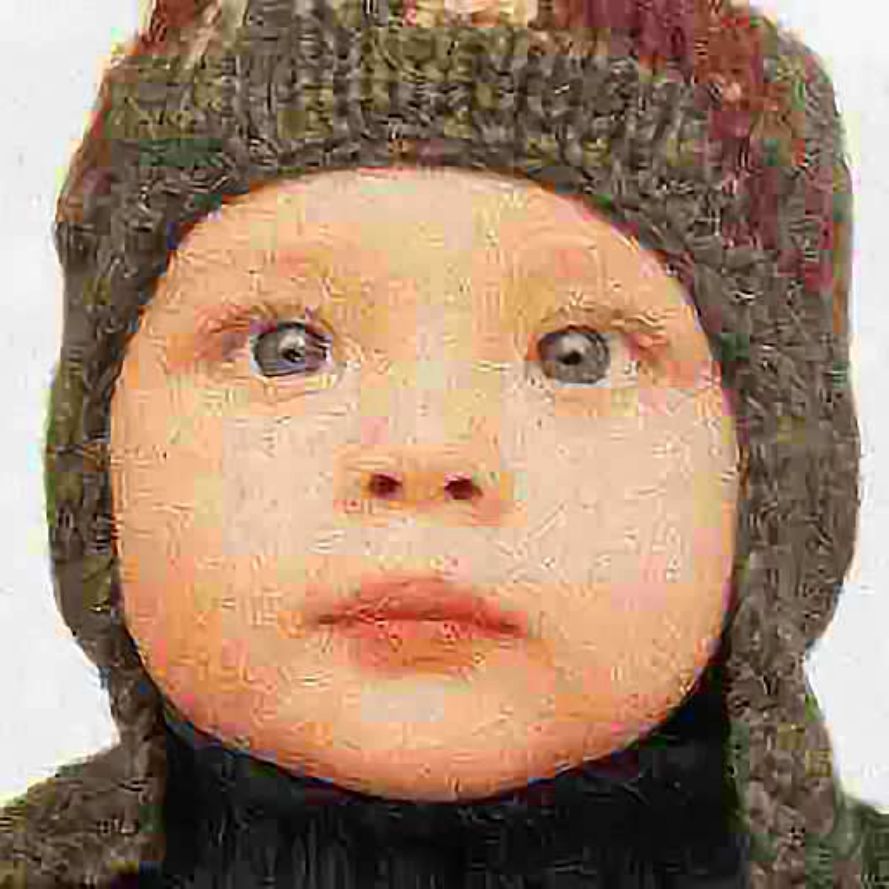} &
       \includegraphics[width=0.13\linewidth]{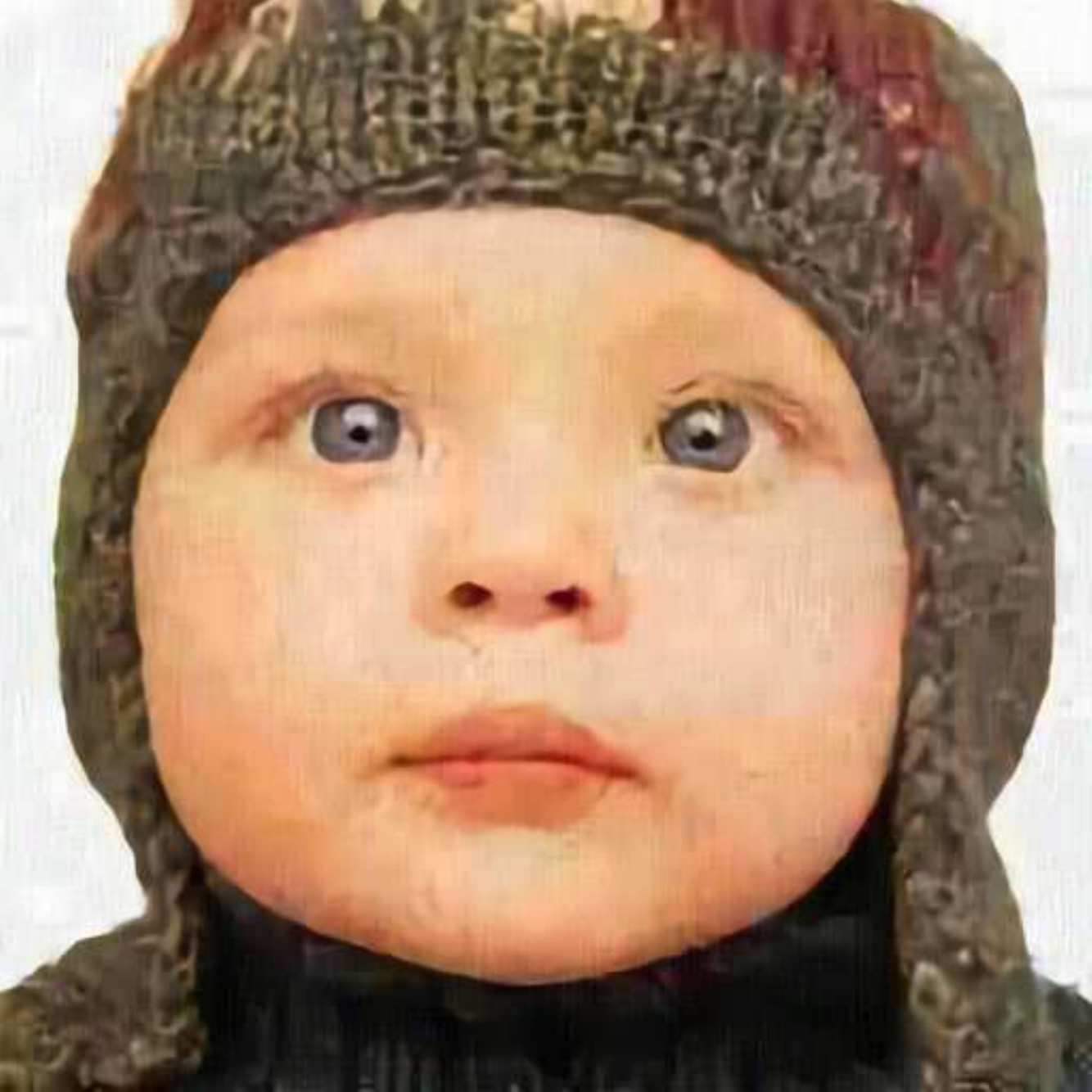} \\
       & \footnotesize{$\sigma=1.5, \lambda=25, q=10$} & & & & \\ \hline

       \includegraphics[width=0.13\linewidth]{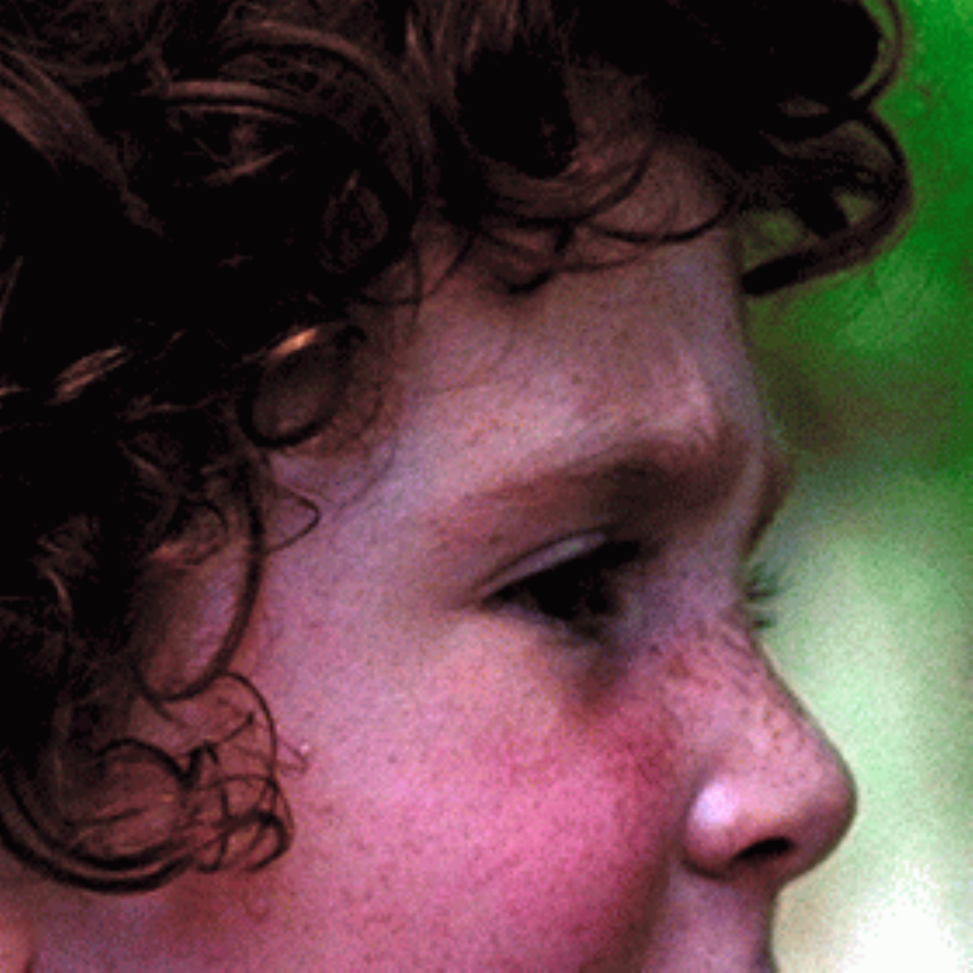} &
       \includegraphics[width=0.13\linewidth]{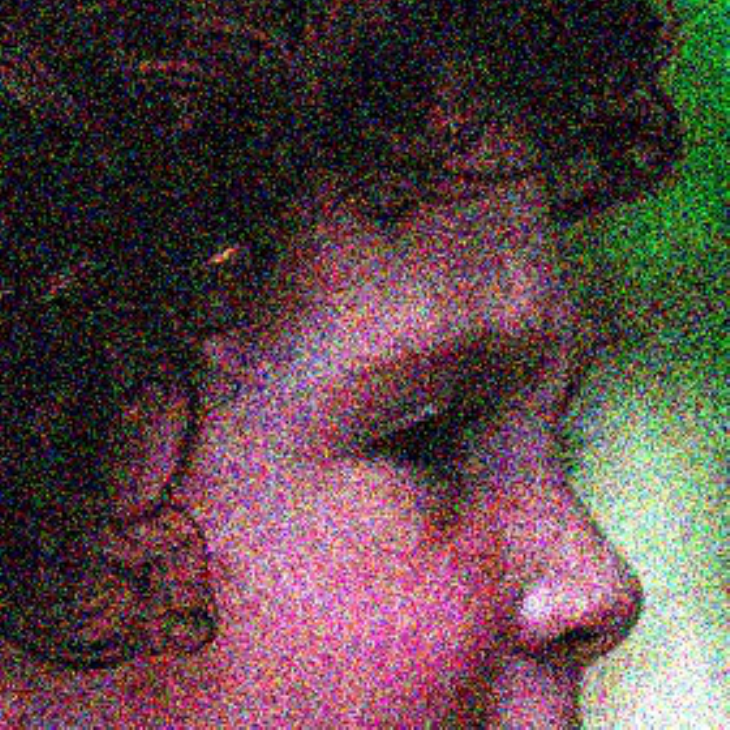} &
       \includegraphics[width=0.13\linewidth]{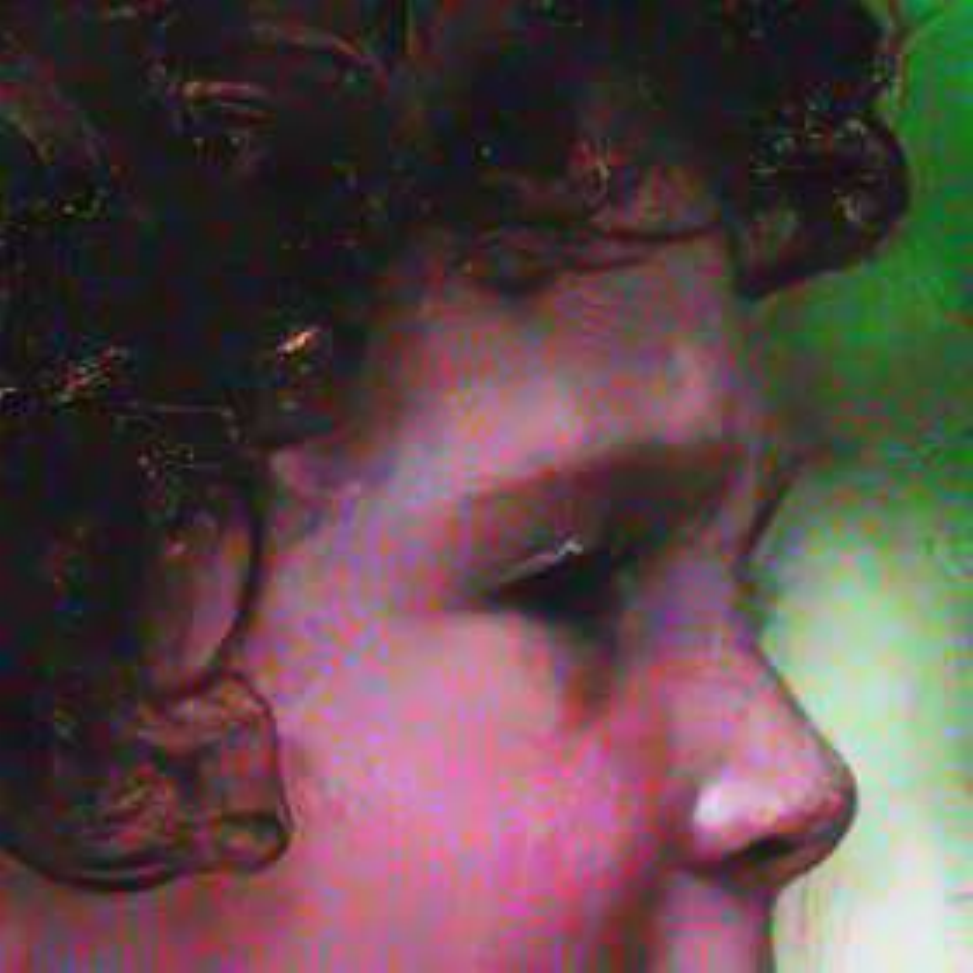} &
       \includegraphics[width=0.13\linewidth]{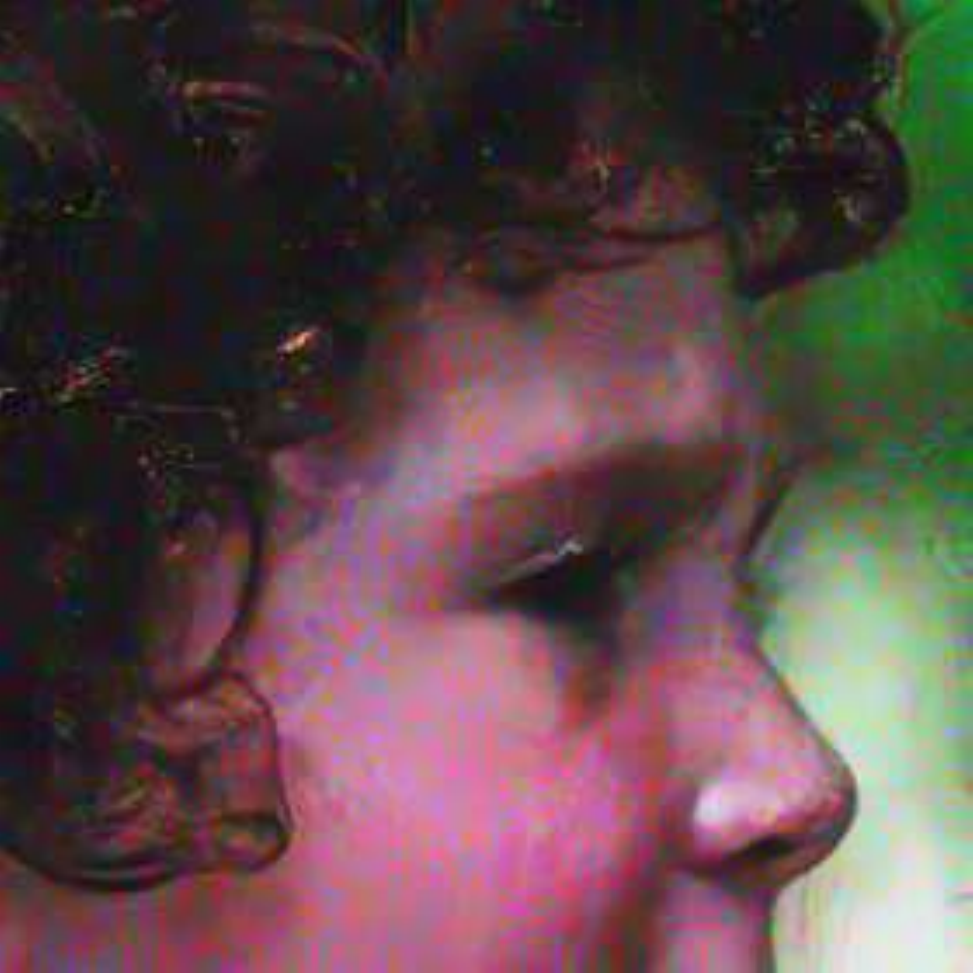} &
       \includegraphics[width=0.13\linewidth]{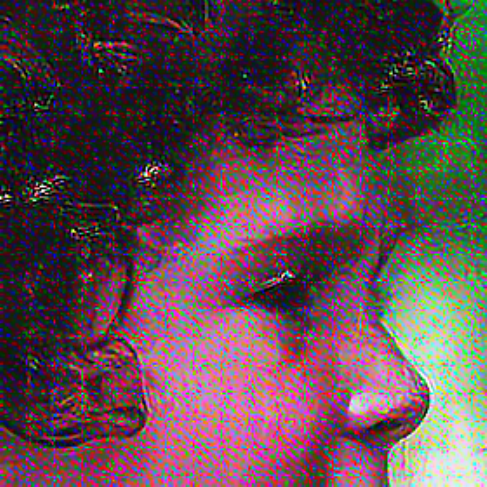} &
       \includegraphics[width=0.13\linewidth]{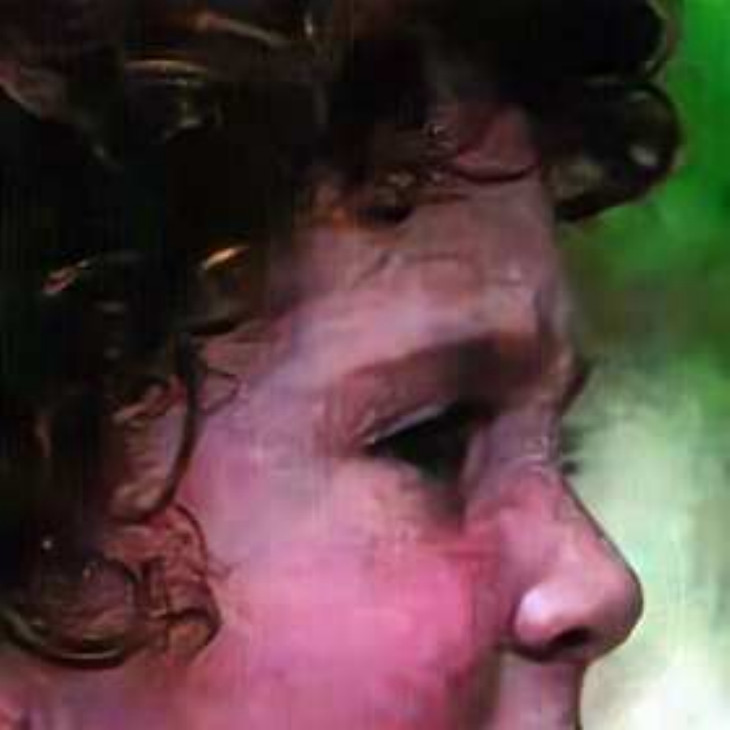} \\
       & \footnotesize{$\sigma=0, \lambda=55, q=100$} & & & & \\ \hline

       \includegraphics[width=0.13\linewidth]{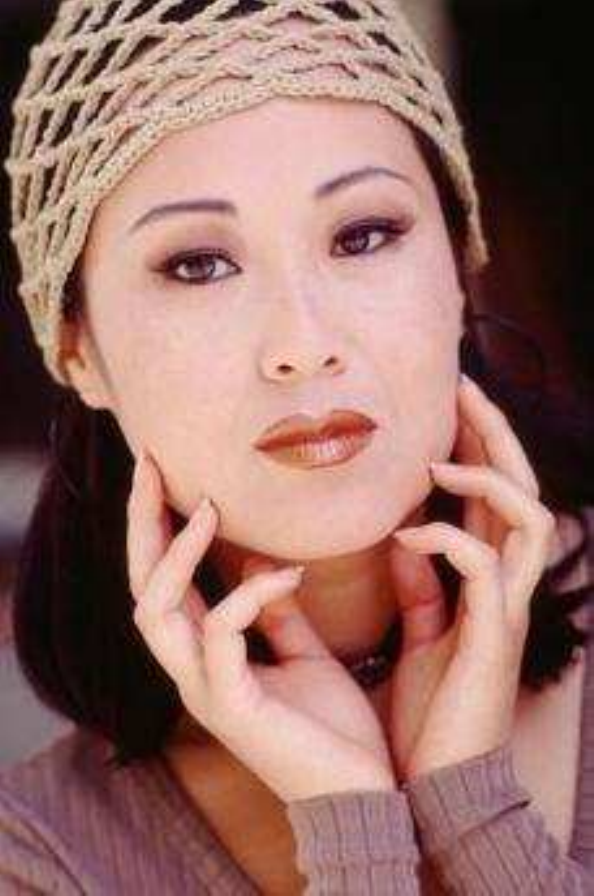} &
       \includegraphics[width=0.13\linewidth]{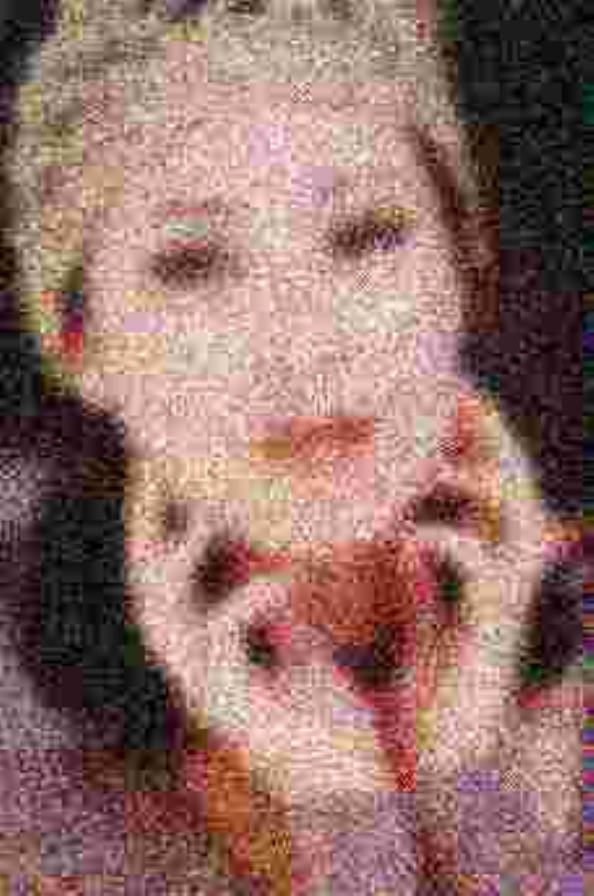} &
       \includegraphics[width=0.13\linewidth]{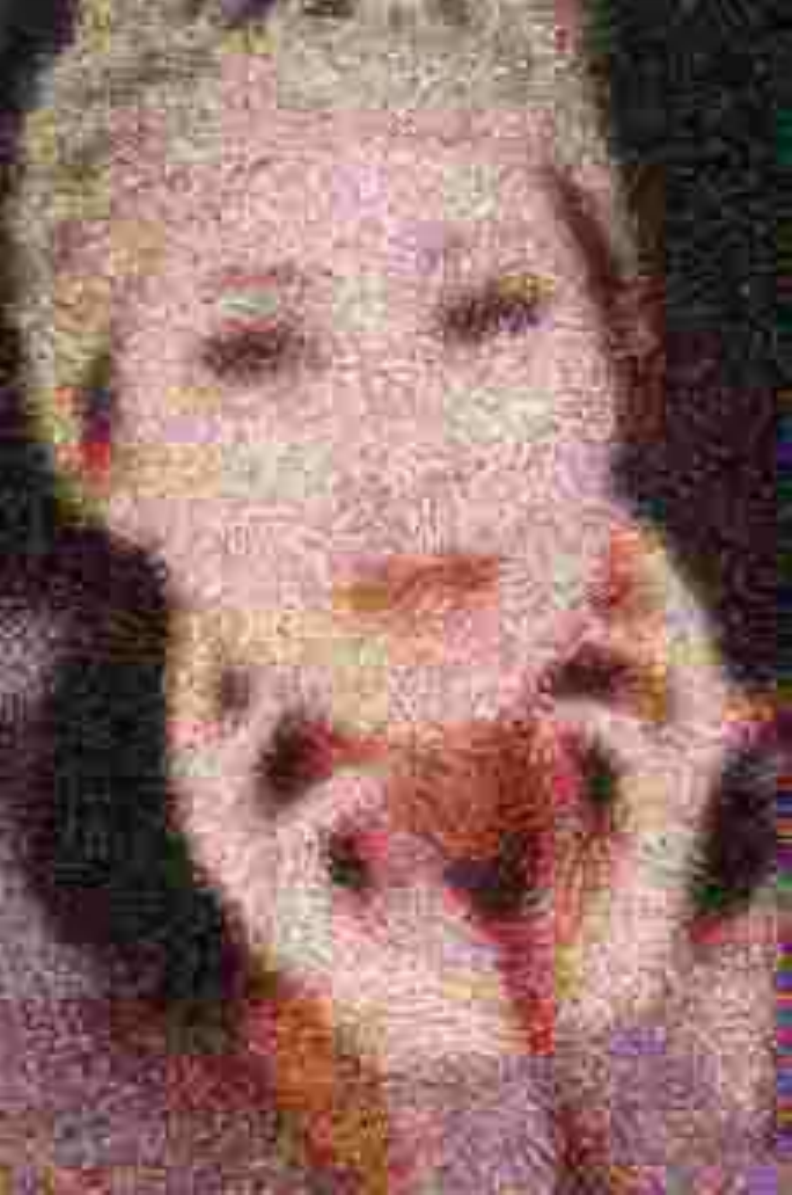} &
       \includegraphics[width=0.13\linewidth]{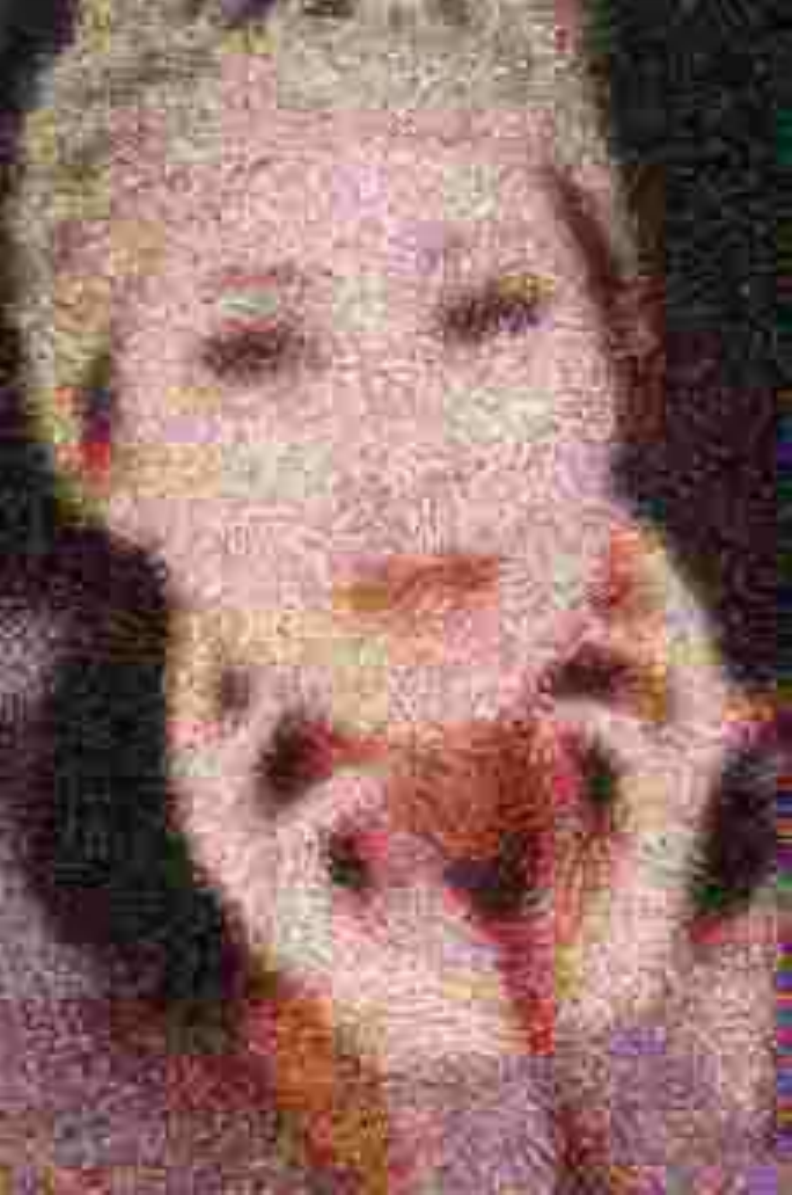} &
       \includegraphics[width=0.13\linewidth]{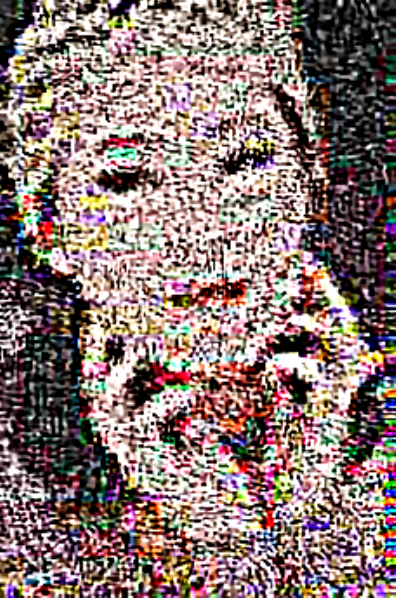} &
       \includegraphics[width=0.13\linewidth]{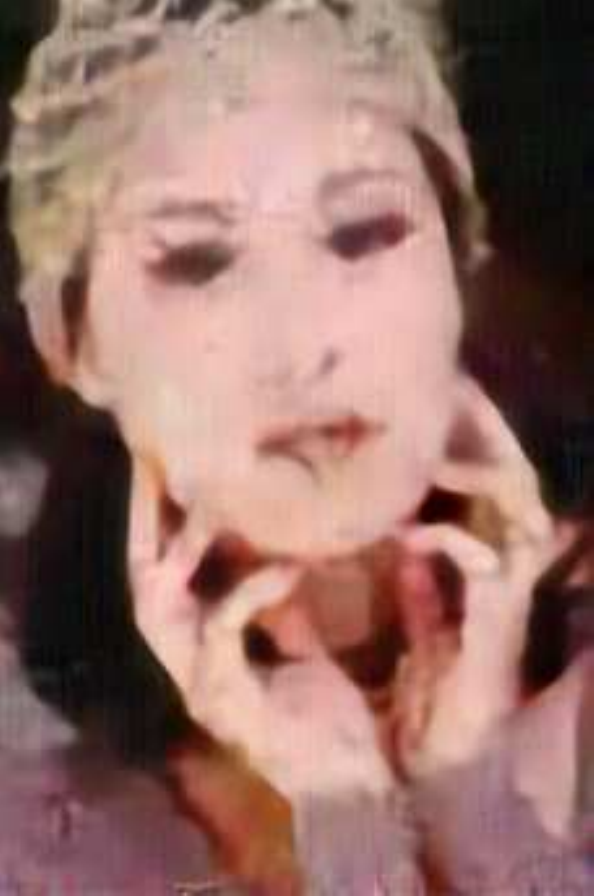} \\
       & \footnotesize{$\sigma=3.0, \lambda=55, q=10$} & & & & \\ \hline

     \end{tabular}
  \caption{Blind restoration images by the proposed method and existing methods.}
  \label{fig:existing_restorations}
    \end{center}
\end{figure*}

Figure \ref{fig:existing_restorations} demonstrates some examples of blind image restoration using existing methods and the proposed method.
The restoration result using existing methods has low performance because more restoration processes are applied and because each restoration method is not robust against degradation perturbation.
Conversely, the proposed method has the advantage of a restoration process that involves a simple one-to-one mapping from a degraded image to the restored image.

\begin{table*}[t!]
  \caption{Restoration Performance (PSNR) for Different Degradation Parameters (Set5).}
  \label{tab:restoration_metrics}
    \begin{center}
      \begin{tabular}{|c|c|c|c|c|c|c|c|c|c|c|} \hline
  \multicolumn{2}{|c|}{}  & \multicolumn{3}{|c|}{$\sigma=0$} & \multicolumn{3}{|c|}{$\sigma=1.5$} & \multicolumn{3}{|c|}{$\sigma=3.0$} \\ \cline{3-11}
    \multicolumn{2}{|c|}{}    & $q=100$ & $q=50$ & $q=10$ & $q=100$ & $q=50$ & $q = 10$ & $q=100$ & $q=50$ & $q=10$ \\ \hline
    $\lambda=0$ &
    Existing \cite{zhang2017beyond, liu2013single, dabov2007image, shan2008high} &

    22.24 &
    22.05 &
    22.09 &
    \bf{28.55} &
    26.30 &
    22.70 &
    23.98 &
    23.02 &
    19.31\\

    & Ours (Blind) &

    \bf{33.86} &
    \bf{31.46} &
    \bf{27.75} &
    26.93 &
    \bf{27.43} &
    \bf{25.71} &
    \bf{27.03} &
    \bf{26.22} &
    \bf{24.20} \\

    & Ours (Nonblind) &

    40.52 &
    32.74 &
    28.01 &
    30.85 &
    29.10 &
    26.18 &
    27.52 &
    26.53 &
    24.32 \\ \hline

          $\lambda=25$  &
          Existing &

          17.68 &
          13.48 &
          21.00 &
          14.27 &
          11.74 &
          21.64 &
          11.05 &
          9.48 &
          19.12\\

          & Ours (Blind) &
          \bf{30.52} &
          \bf{29.85} &
          \bf{27.19} &
          \bf{27.99} &
          \bf{27.61} &
          \bf{25.87} &
          \bf{25.35} &
          \bf{25.10} &
          \bf{23.90} \\

          & Ours (Nonblind) &

          31.09 &
          30.33 &
          27.65 &
          28.42 &
          27.99 &
          26.22 &
          25.71 &
          25.49 &
          24.23 \\ \hline

          $\lambda=55$  &
          Existing &
          12.80 &
          14.59 &
          15.40 &
          10.35 &
          13.93 &
          12.88 &
          8.90 &
          12.13 &
          10.49 \\

          & Ours (Blind) &
          \bf{27.64} &
          \bf{27.32} &
          \bf{25.83} &
          \bf{25.86} &
          \bf{25.60} &
          \bf{24.54} &
          \bf{23.86} &
          \bf{23.71} &
          \bf{22.90} \\

      & Ours (Nonblind) &

      27.78 &
      27.50 &
      26.14 &
      26.05 &
      25.86 &
      24.95 &
      24.05 &
      23.94 &
      23.24 \\ \hline

      \end{tabular}
    \end{center}
\end{table*}

\subsection{Restoration of Actual Images}

Actual images have different degradations in different regions.
Blur levels differ depending on the distance to objects, and noise levels vary with regard to intensity.
Therefore, estimated degradation often changes by region, and different restoration strategies are applied accordingly.

\begin{figure*}[t!]
  \captionsetup{farskip=0pt}
    \begin{center}
  \subfloat[Original]{
       \includegraphics[width=0.32\linewidth]{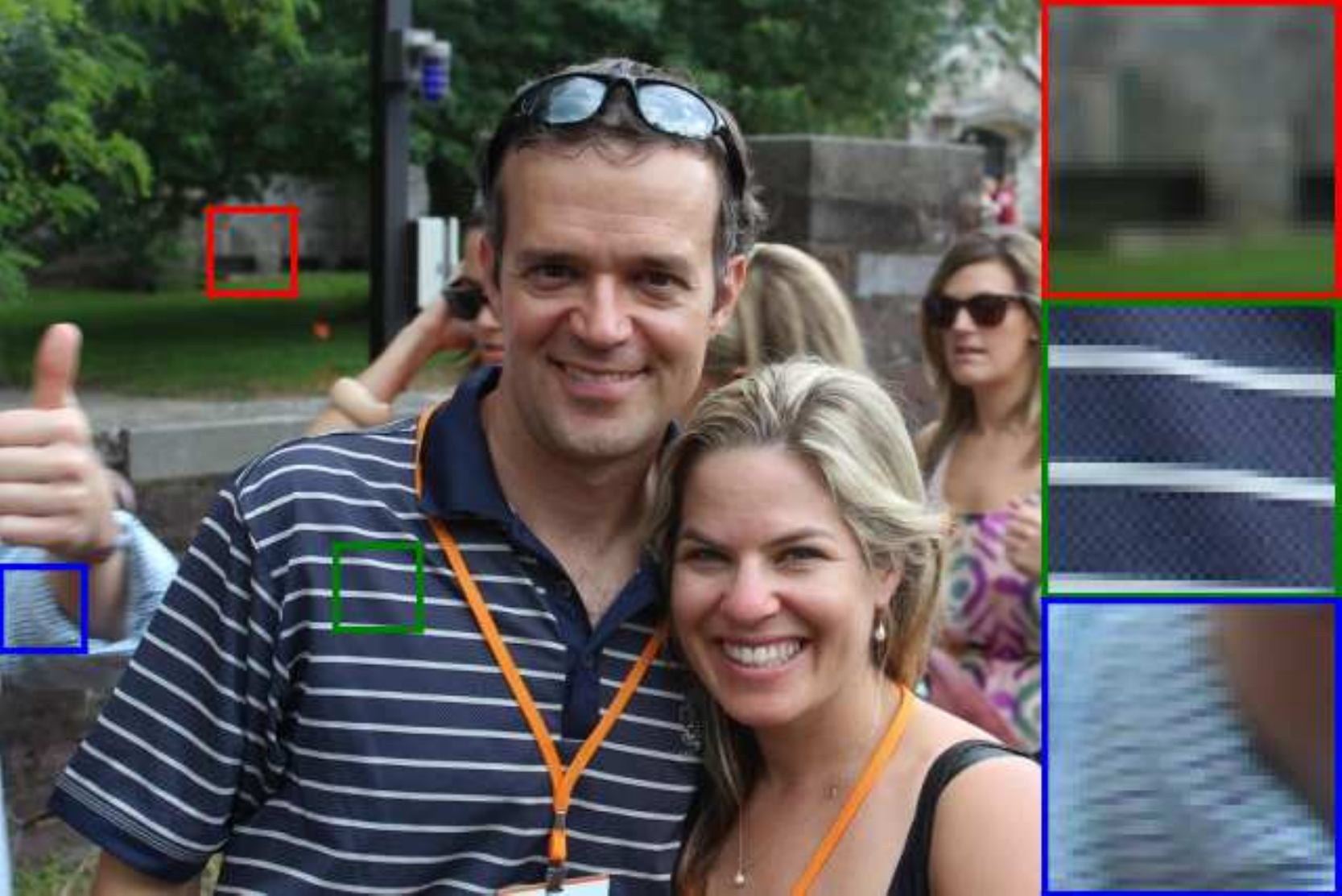}
    \label{shi_6a}\hfill
  }
  \subfloat[Estimated Degradation]{%
       \includegraphics[width=0.32\linewidth]{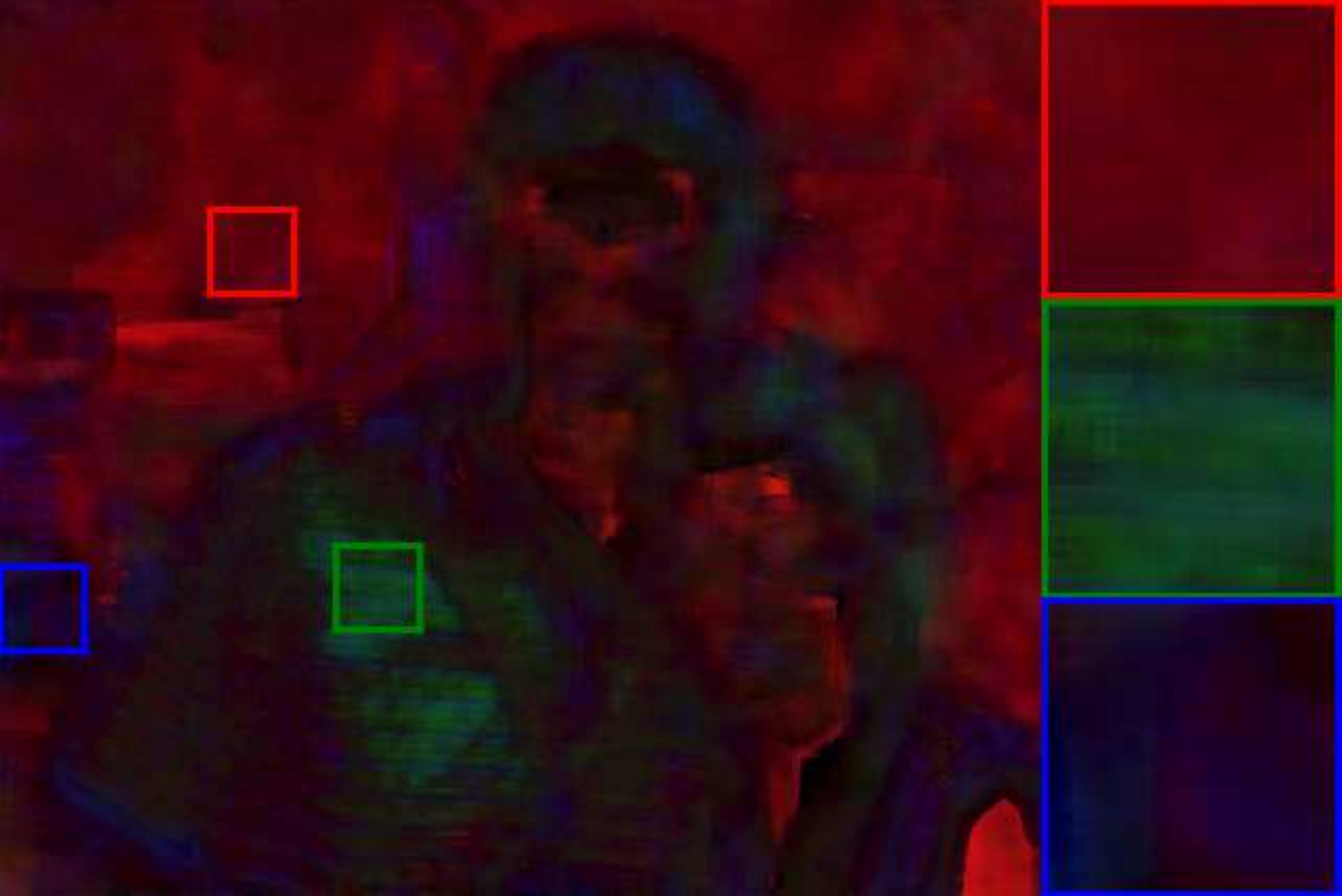}
    \label{shi_6b}
    }
  \subfloat[Restored]{%
       \includegraphics[width=0.32\linewidth]{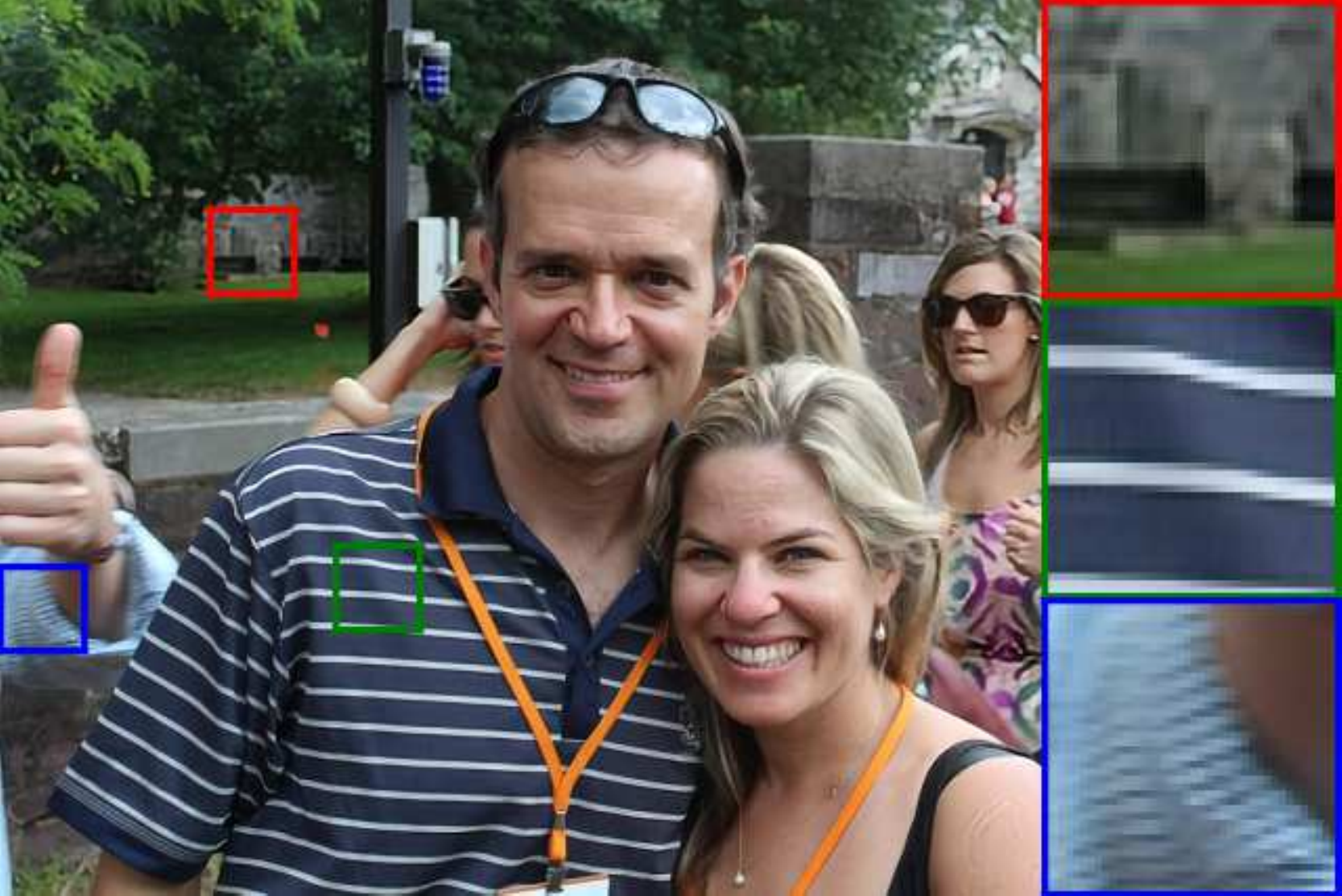}
    \label{shi_6c}\hfill
    }
  \end{center}
    \caption{Restoration from normal photo image; (a) is the original image. The degradation estimation network detects different degradations by region as shown in (b). (c) is the restored image which is deblurred and denoised according to the estimated degradation types and strength.}
    \label{fig:shi_6}
\end{figure*}

Figure \ref{fig:shi_6} shows an example of the degradation estimation and restoration of an actual image.
For instance, a region of a man’s head is estimated as noised (Fig. \ref{shi_6b}; represented by the green component), and the noise has been removed in the output image in Fig. \ref{shi_6c}.
A region estimated as blurred (\ie, represented by red) is also deblurred.

\begin{figure*}[t!]
  \captionsetup{farskip=0pt}
    \begin{center}
  \subfloat[Original]{
       \includegraphics[width=0.32\linewidth]{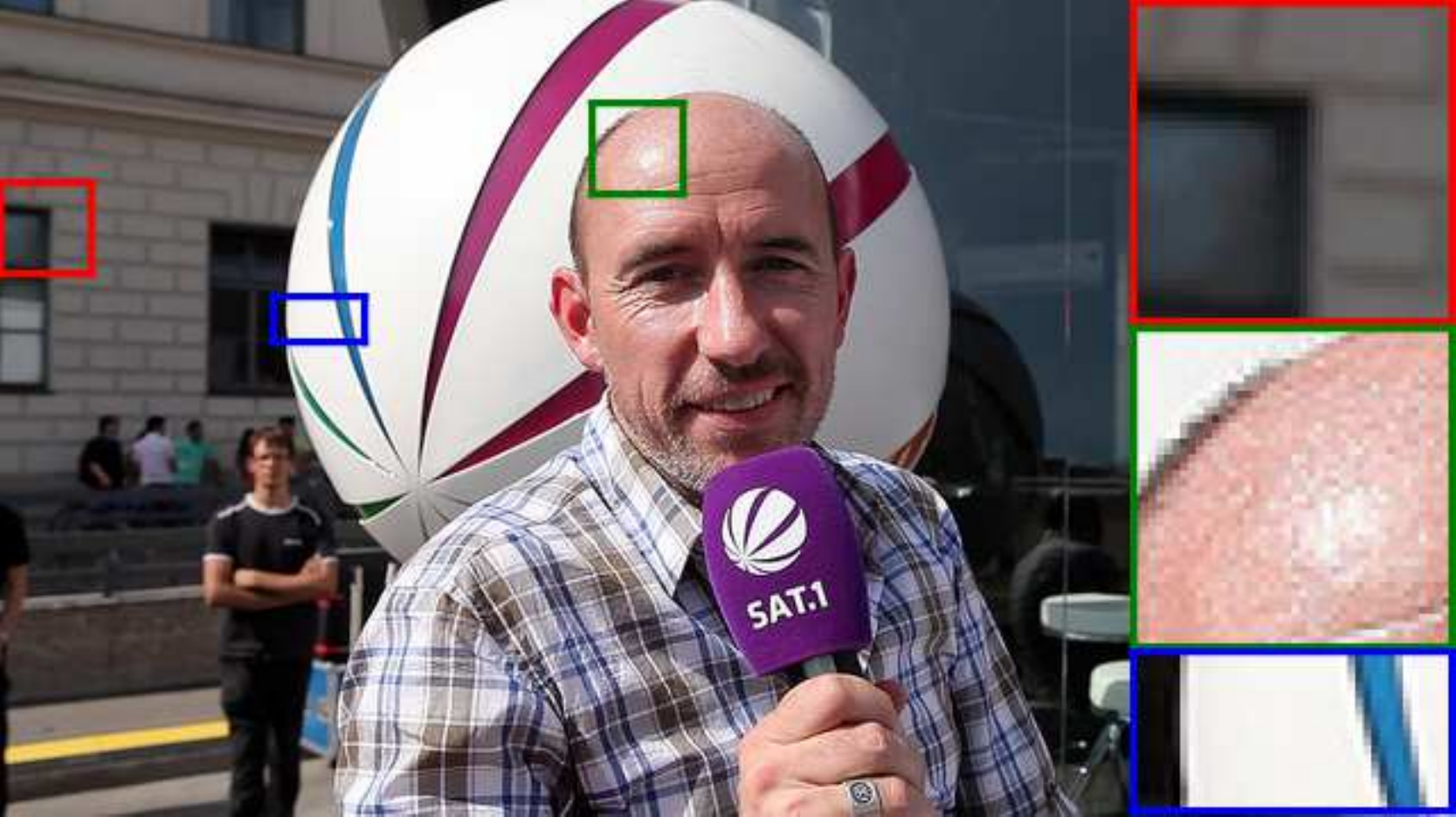}
    \label{shi_7a}\hfill
  }
  \subfloat[Estimated Degradation]{%
       \includegraphics[width=0.32\linewidth]{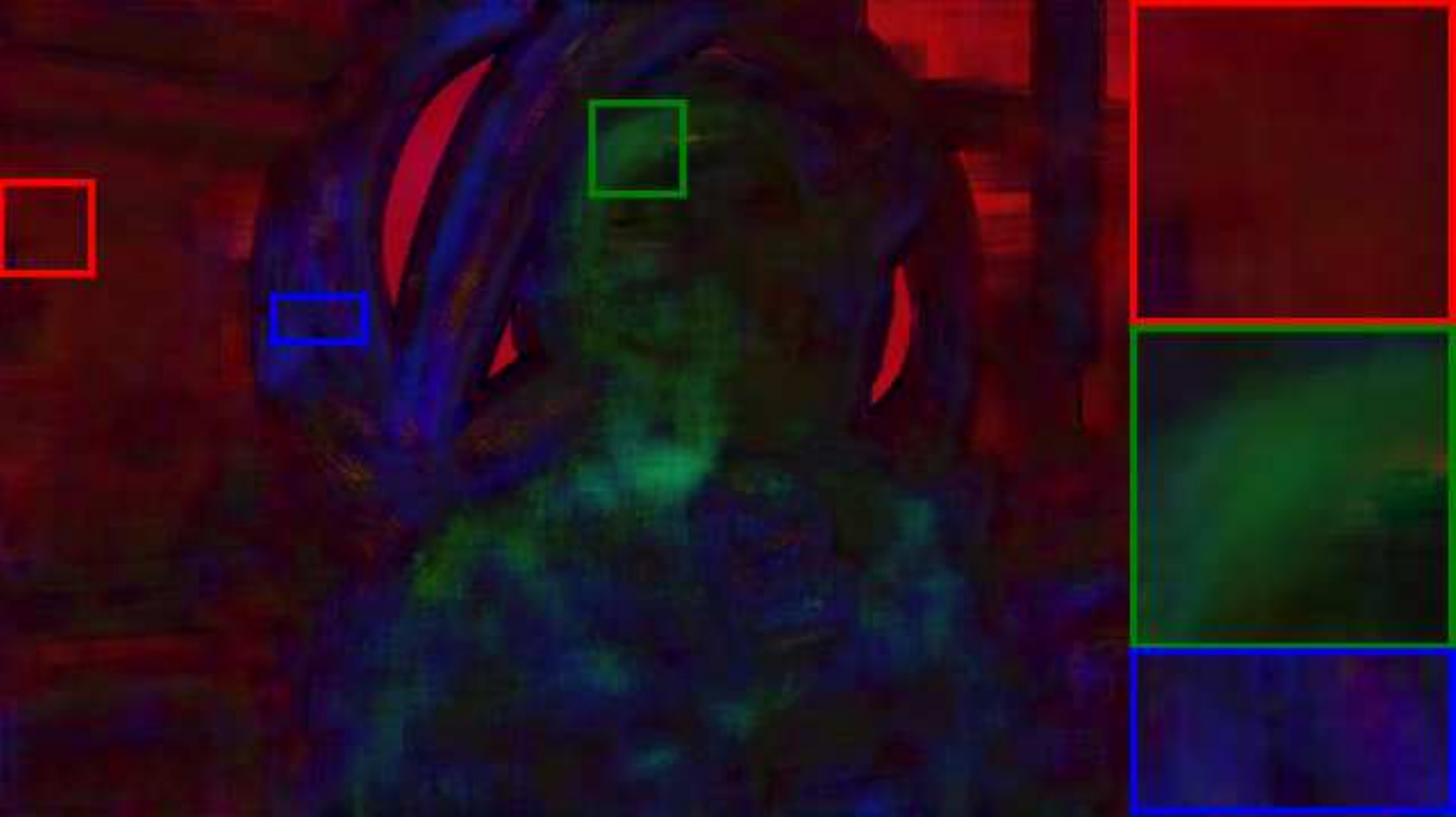}
    \label{shi_7b}
    }
  \subfloat[Restored]{%
       \includegraphics[width=0.32\linewidth]{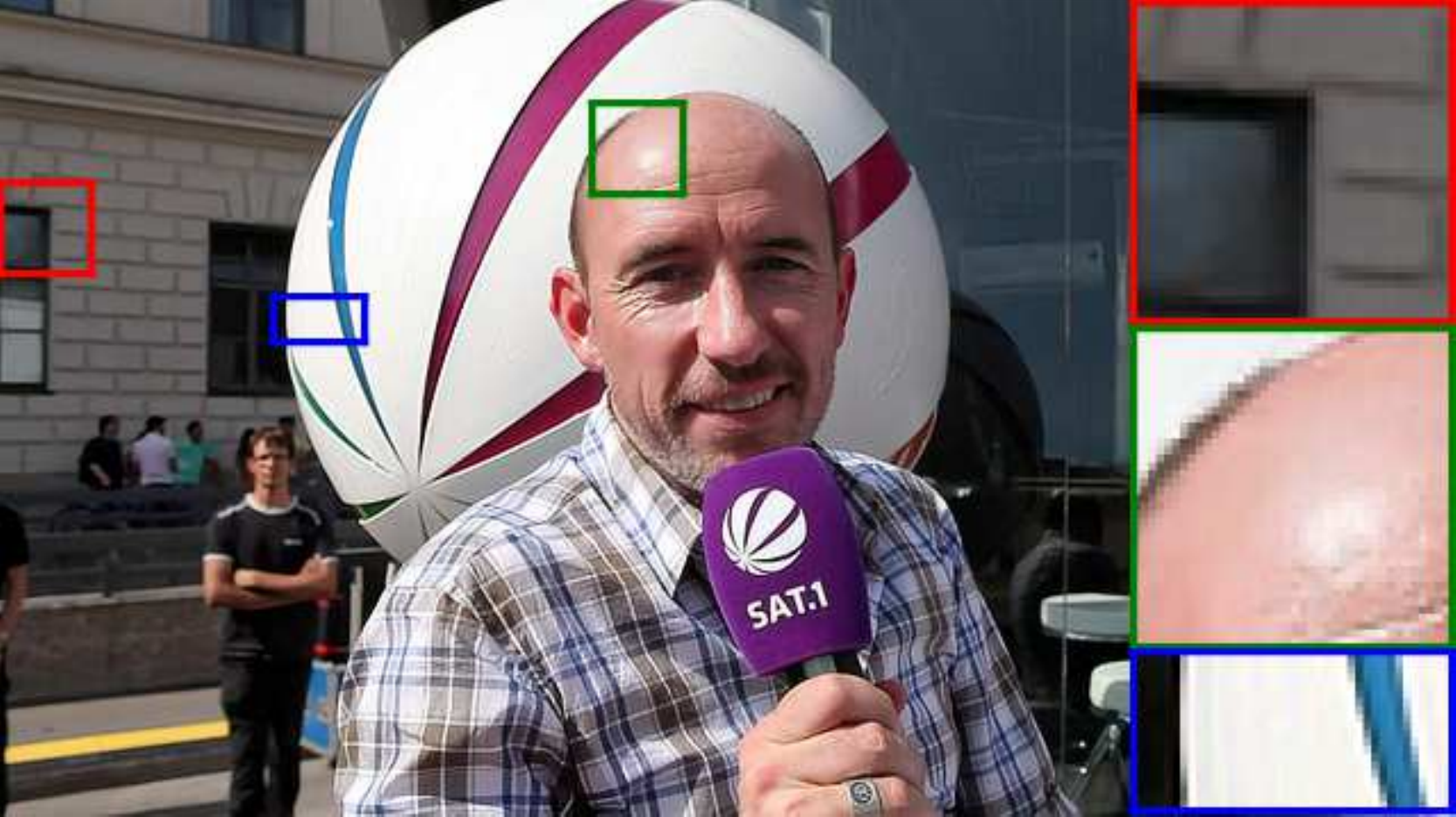}
    \label{shi_7c}\hfill
    }
  \end{center}
    \caption{Another example of restoration from a normal photo: (a) is the original, (b) is the estimated degradations, and (c) is the restored image which is deblurred, denoised and deblocked. }
    \label{fig:shi7}
\end{figure*}

Figure \ref{fig:shi7} shows another example of an actual image.
A region estimated as blurred (Fig. \ref{shi_7b}) is successfully deblurred (Fig. \ref{shi_7c}), whereas a region estimated as AWGN, where a textile pattern appears on a man's shirt, is smoothed (Fig. \ref{shi_7c}).

These samples demonstrate the ability of the model to change the restoration strategy according to the estimated degradation attributes of the input images.

\subsection{Restoration Strategy Control}

As a benefit of the separate architecture of the restoration network, nonblind and interactive restoration can be realized by inputting arbitrary degradation parameter maps instead of estimated degradation properties.
This is particularly useful for human interactive restoration usage by adjusting the strategy and its strength to obtain perceptively better results.

\begin{figure*}[t!]
  \captionsetup{farskip=0pt}
    \begin{center}
    \subfloat[Original]{
      \includegraphics[width=0.30\linewidth]{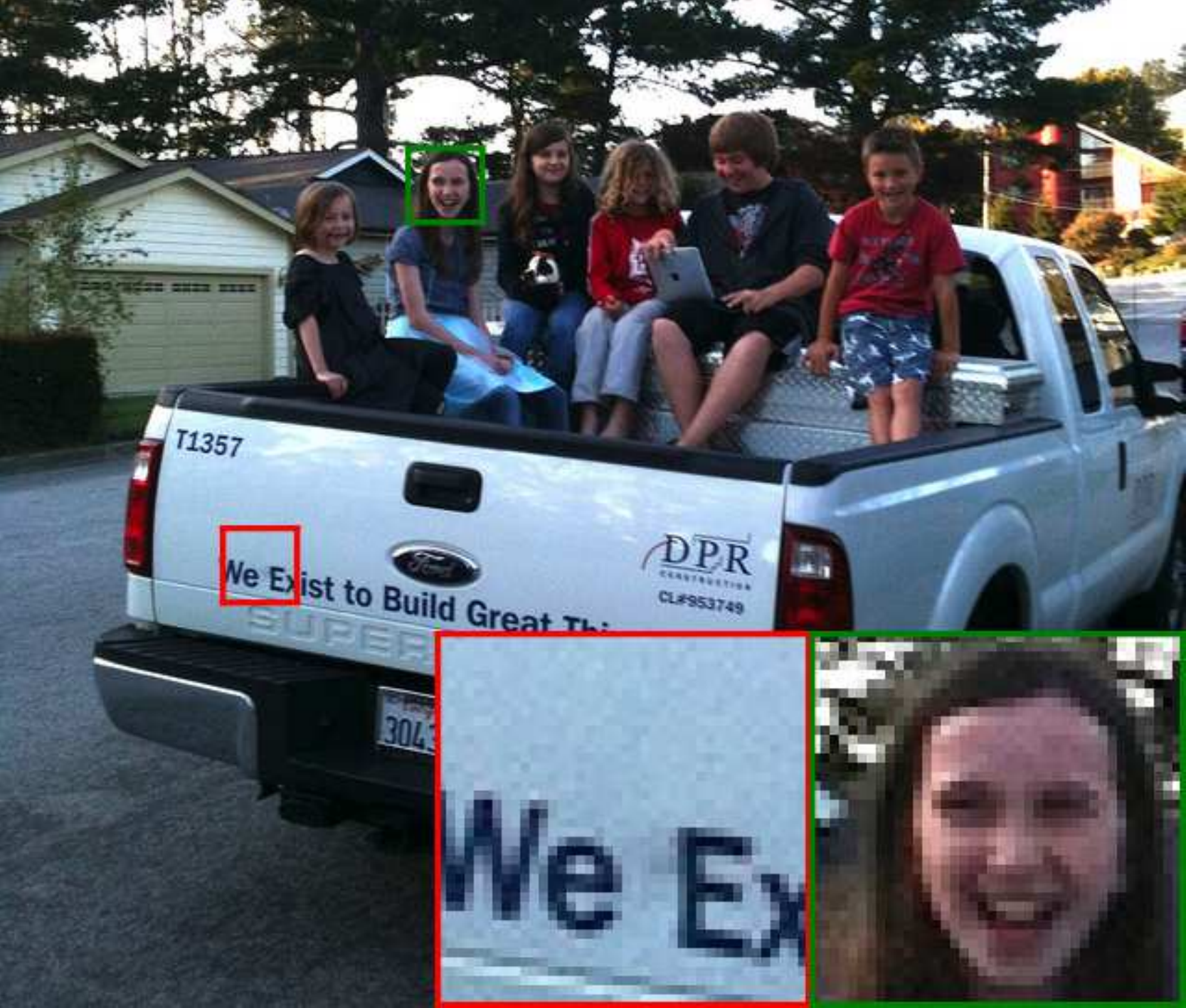}
      \label{fig:coco_original}
      \hfill
    }
    \subfloat[Given Degradation Attribute]{
      \includegraphics[width=0.30\linewidth]{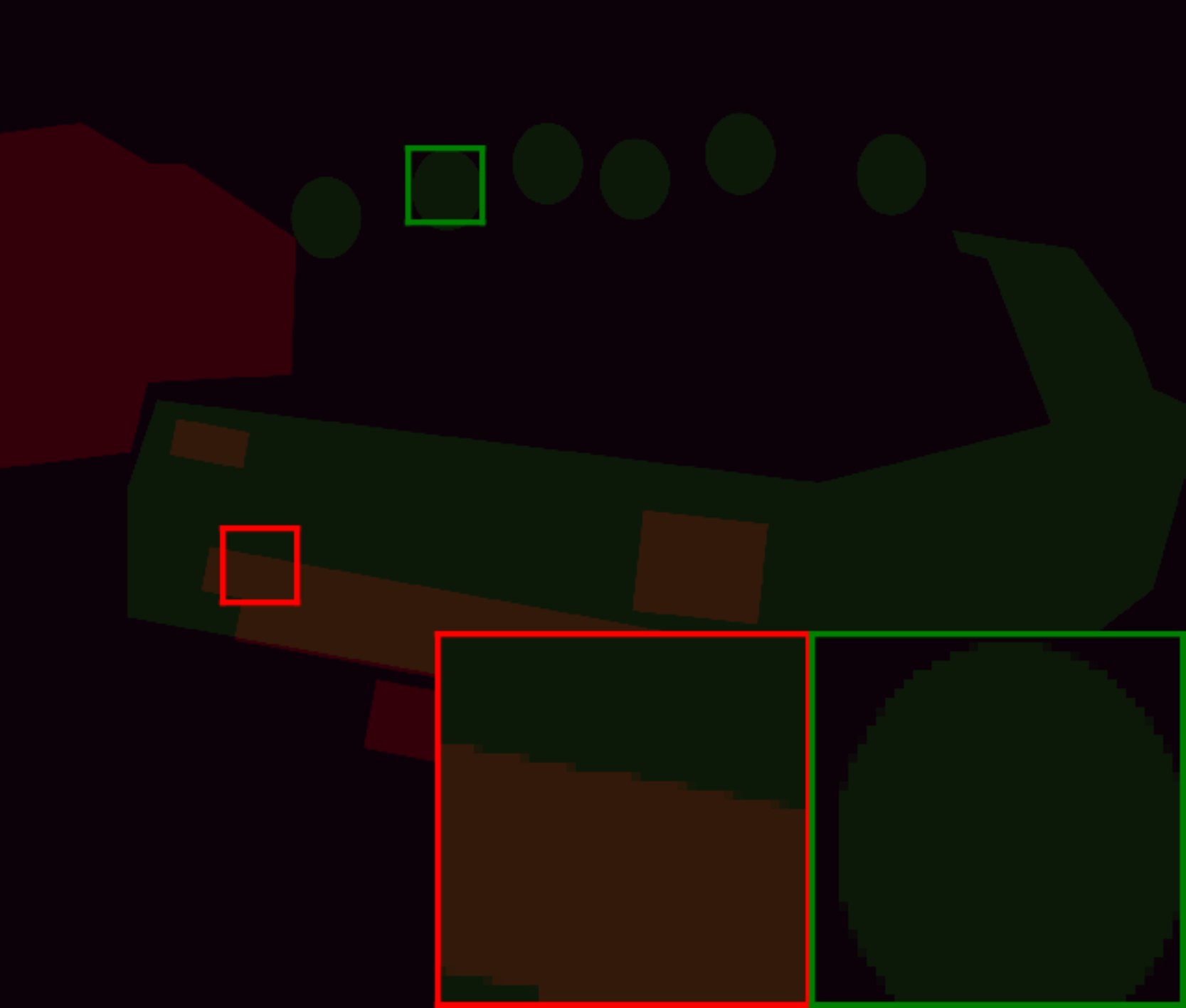}
      \label{fig:coco_attr}
      \hfill
    }
    \subfloat[Restored]{
      \includegraphics[width=0.30\linewidth]{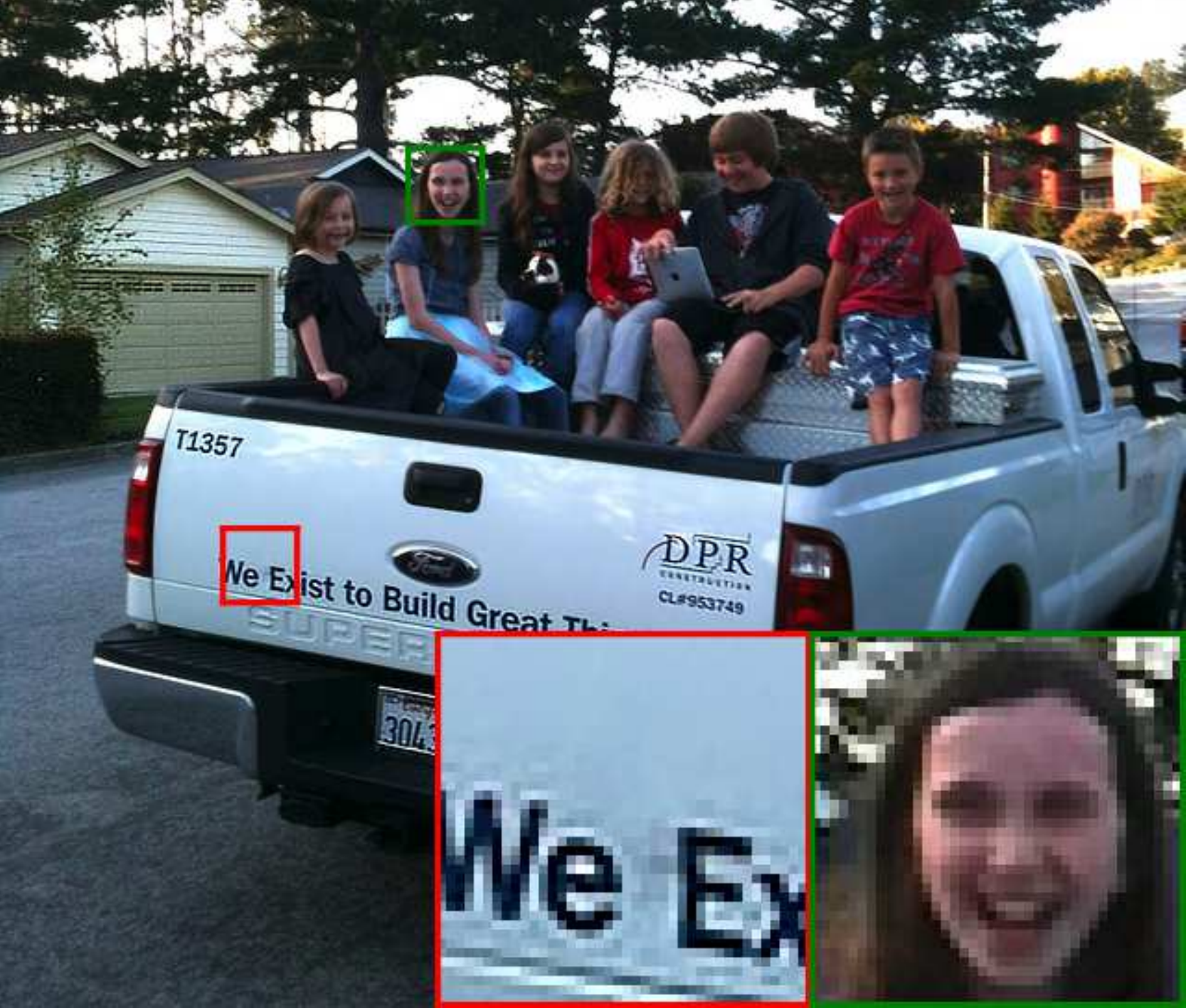}
      \label{fig:coco_restored}
    }
  \end{center}
    \caption{Restoration with a given degradation attribute by region: (a) is the original image, (b) is the given degradation attribute, and (c) is the restored image.}
    \label{fig:coco}
\end{figure*}

Figure \ref{fig:coco} shows an example of controlling the restoration strategy, and Figure \ref{fig:coco_original} shows the original image.
Figure \ref{fig:coco_attr} is handcrafted with the given attribute channels.
Considering that the car body and the faces of the children are noisy, the degradation parameter for AWGN is set high on the spots.
The letters on the car body seem blurred; thus, the degradation parameter for blur is set high on the spots.
Figure \ref{fig:coco_restored} shows the output image.
As instructed by the degradation parameter maps, the car body and the faces are denoised, and the letters are deblurred and sharpened.

\section{Conclusion}

\label{sec:conclusion}

We proposed a CNN model for degradation estimation and restoration from compositional degradation.
Compositional degradation is common in the image capturing process by an image sensor.
However, most image restoration methods target single degradation types. We addressed the degradation estimation and restoration problem of compositional degradation and proposed a CNN-based model to infer degradation properties and restore degraded images.

The proposed model contains two subnetworks: a degradation estimation network and a restoration network.
The degradation estimation network infers the degradation types and levels of an input compositional degradation image.
The restoration network restores a degraded image by using the estimated degradation properties.
The separated network enables both blind and nonblind restorations.

Experimental results show that the proposed model can successfully estimate the degradation parameters and restore images better than the combinational algorithm of existing restoration methods.
Pixel-wise degradation estimation and restoration realize the control on restoration strategy and its strength region by region.

\ifCLASSOPTIONcaptionsoff
  \newpage
\fi



\bibliographystyle{IEEEtran}
\bibliography{IEEEabrv,report}
%



%

\begin{IEEEbiography}{Kazutaka Uchida}
  received bachelor's and master's degrees in control and systems engineering from Tokyo Institute of Technology, Tokyo, Japan, in 2000 and 2003, respectively. He joined Sony Corporation in 2003 and developed next-generation audio and visual applications.
  In 2008, he cofounded a startup company called Kadinche Corporation, where he currently works on the development of immersive virtual reality applications. He is currently pursuing his PhD.
\end{IEEEbiography}

\begin{IEEEbiography}{Masayuki Tanaka}
received his bachelor's and master's degrees in control engineering and Ph.D. degree from Tokyo Institute of Technology in 1998, 2000, and 2003. He joined Agilent Technology in 2003. He was a Research Scientist at Tokyo Institute of Technology since 2004 to 2008. Since 2008, He has been an Associated Professor at the Graduate School of Science and Engineering, Tokyo Institute of Technology. He was a Visiting Scholar with Department of Psychology, Stanford University, CA, USA.
\end{IEEEbiography}


\begin{IEEEbiography}{Masatoshi Okutomi}
  received a B.Eng. degree from the Department of Mathematical Engineering and Information Physics, the University of Tokyo, Japan, in 1981 and an M.Eng. degree from the Department of Control Engineering, Tokyo Institute of Technology, Japan, in 1983. He joined Canon Research Center, Canon Inc., Tokyo, Japan, in 1983. From 1987 to 1990, he was a visiting research scientist in the School of Computer Science at Carnegie Mellon University, USA. In 1993, he received a D.Eng. degree for his research on stereo vision from Tokyo Institute of Technology. Since 1994, he has been with Tokyo Institute of Technology, where he is currently a professor in the Department of Systems and Control Engineering,
  the School of Engineering.
\end{IEEEbiography}




\end{document}